\documentclass[10pt, a4paper]{article}

\usepackage{stackengine}
\usepackage{amsmath}
\usepackage{amscd}
\usepackage{amsthm}
\usepackage{amssymb}
\usepackage{amsfonts}
\usepackage{algorithm}
\usepackage{algpseudocode}
\usepackage{enumerate}
\usepackage{bm}
\usepackage{cite}
\usepackage{mathtools}
\usepackage{mathdots}
\usepackage[T1]{fontenc}
\usepackage{framed, color}
\usepackage{dsfont}
\usepackage{graphicx}
\usepackage{hyperref}
\usepackage{latexsym}
\usepackage{mathrsfs}
\usepackage{verbatim}
\usepackage{booktabs} 
\usepackage{placeins}
\usepackage{thmtools, thm-restate}
\usepackage{multirow}
\usepackage{subcaption}
\usepackage{url}
\usepackage{hyperref}

\usepackage[english]{babel}  
\addto\extrasenglish{

}

\usepackage{xargs}   
\usepackage[pdftex,dvipsnames]{xcolor}  
\usepackage[colorinlistoftodos,prependcaption,textsize=small]{todonotes}
\newcommandx{\unsure}[2][1=]{\todo[linecolor=red,backgroundcolor=red!25,bordercolor=red,#1]{#2}}
\newcommandx{\change}[2][1=]{\todo[linecolor=blue,backgroundcolor=blue!25,bordercolor=blue,#1]{#2}}
\newcommandx{\info}[2][1=]{\todo[linecolor=OliveGreen,backgroundcolor=OliveGreen!25,bordercolor=OliveGreen,#1]{#2}}
\newcommandx{\improvement}[2][1=]{\todo[linecolor=Plum,backgroundcolor=Plum!25,bordercolor=Plum,#1]{#2}}

\usepackage[parfill]{parskip}

\usepackage[subrefformat=parens,labelformat=parens]{subcaption}

\theoremstyle{plain}
\newtheorem{theorem}{Theorem}[section]

\newtheorem{lemma}[theorem]{Lemma}

\newtheorem{remark}[theorem]{Remark}

\theoremstyle{definition}
\newtheorem{definition}[theorem]{Definition}
\newtheorem{example}[theorem]{Example}

\newcommand{\RR}{\mathbb R}
\newcommand{\ZZ}{\mathbb Z}
\newcommand{\CC}{\mathbb C}
\newcommand{\NN}{\mathbb N}

\newcommand{\XX}{\mathcal{X}}
\newcommand{\PP}{\mathbb{P}}
\newcommand{\E}{\mathbb{E}}

\newcommand{\Ltwo}{L^{2}(\RR)}

\newcommand{\elltwo}{\ell^{2}(\ZZ)}
\newcommand{\M}{\mathcal{M}}
\newcommand{\N}{\mathcal{N}}
\newcommand{\G}{\mathcal{G}}
\newcommand{\B}{\mathcal{B}}
\newcommand{\dimN}{p-q}

\newcommand{\dt}{\ \mbox{d}t}

\newcommand{\DD}{\mathcal{D}}
\newcommand{\DFT}{\textbf{DFT}}

\newcommand{\tensor}{\otimes}

\renewcommand{\thefootnote}{\arabic{footnote}}
\author{ Eric Marcus$^{* \, 1,2}$ 
	     \ Ray Sheombarsing$^{* \, 1}$ 
	     \ Jan-Jakob Sonke$^{1,2}$ 
	     \ Jonas Teuwen$^{1,2}$ 
	     \\[2ex]      
	      \emph{\normalsize{$^1$The Netherlands Cancer Institute}} \\
       	       \emph{\normalsize{Amsterdam, Plesmanlaan $121$, $1066$ CX, The Netherlands}} \\[2ex]
       		\emph{\normalsize{$^2$University of Amsterdam}} \\
       		\emph{\normalsize{Amsterdam, Science Park $900$, $1012$ WX, The Netherlands}	  }   
}
\date{}
\title{\bf Constrained Empirical Risk Minimization:\\[4pt] Theory and Practice}

\begin{document}

\maketitle

\def\thefootnote{*}\footnotetext{These authors contributed equally to this work.}

\begin{abstract}
	Deep Neural Networks (DNNs) are widely used for their ability to effectively approximate large classes of functions. This flexibility, however, makes the strict enforcement of constraints 
    on DNNs an open  problem. Here we present a framework that, under mild assumptions, allows the exact enforcement of constraints on parameterized sets of functions such as DNNs. 
    Instead of imposing ``soft'' constraints via additional terms in the loss, we restrict (a subset of) the DNN parameters to a submanifold on which the constraints are satisfied exactly 
    throughout the entire training procedure. We focus on constraints that are outside the scope of equivariant networks used in Geometric Deep Learning. As a major example of the 	
	framework, we restrict filters of a Convolutional Neural Network (CNN) to be wavelets, and apply these wavelet networks to the task of 
	contour prediction in the medical domain.
\end{abstract}

\vfill

\vspace{1 ex}

\thispagestyle{empty}
\newpage

\setcounter{tocdepth}{2}  
\tableofcontents

\section{Introduction}
Empirical risk minimization (ERM) is currently the most prevalent framework for supervised learning.  
In this framework, observations and outcomes are interpreted as realizations 
of random variables. The goal is to find a function
to map inputs to associated targets for all representative (potentially unseen) observations. To find
such a mapping, one introduces a loss function to quantify the discrepancy between observed and predicted
targets. An optimal map is then found by minimizing the expected loss. In 
large-scale settings, such as deep learning, the resulting minimization problem is solved using Stochastic 
Gradient Descent (SGD) and various variants thereof, see
\cite{Adam, loshchilov2017decoupled, loshchilov2018fixing, You2020Large} for instance. 

As deep learning applications become more specialized, domain-specific needs become increasingly vital. These are
often formulated in terms of constraints on the permissible mappings. For instance, the constraint for translation-equivariance led to the development and success of convolution 
neural networks (CNNs) \cite{lecun1995convolutional}. In general, however, it is a highly non-trivial task to construct network architectures that satisfy a set of constraints, if they exist at all. It is therefore common practice to incorporate constraints 
directly into the loss function by including additional terms, usually referred to as ``soft'' constraints. This setup, however, has the 
drawback that the constraints are only approximately (on average) satisfied due to the formulation of ERM. Moreover, incorporating many different objectives in a loss function may lead to suboptimal results for the individual objectives. Another common strategy, used for many types of constraints, e.g., ``flip'' invariance, is based on data augmentation.
However, not every constraint can be achieved using data augmentation, and it also leads at best to constraints 
being approximately satisfied. 

Recently, approaches that circumvent the loss-based soft constraints have been proposed, see \cite{POLICE, input_convex_neural_networks,neural_conservation_laws, miyato2018spectral, pathak2015constrained} for example. 
Furthermore, the field of Geometric Deep Learning (GDL) is engaged in ways to precisely embody symmetries on the domain into the networks themselves, see \cite{bronstein2021geometric, weiler2021coordinate, bronstein2017geometric} and the references therein. In GDL, one considers a very specific but powerful type of constraint, namely that network layers are equivariant with respect to some group action. Such constraints can, in principle, be posed as a set of equations on the network parameters, which 
is the setup of this paper. The GDL approach, however, allows for a more direct modification of the network architecture.

Not all constraints arise as equivariance principles. A large class of examples comes from highly specialized requirements on the output of a neural network,
e.g., that the output is a divergence-free vector field, a contour, or perhaps a surface.
For example, in medical image segmentation, a natural requirement is that the output of the segmentation network corresponds to a continuous (closed) curve. 
This can be enforced by imposing constraints on the parameters (filters) of the network, e.g., by requiring that they correspond to a suitable set of basis functions. 
In this paper, we provide a major example of such a constraint, where the filters of a CNN are restricted to be so-called wavelets, which excel in 
multiresolution signal analysis.

The ubiquitous presence of constraints in the field of deep learning, then, asks for a general framework for incorporating constraints into the optimization procedure. An earlier attempt at incorporating constraints 
is described in \cite{donti2021dc}. However, this method is only able to deal with linear constraints. Non-linear constraints are only approximately satisfied 
using soft constraints. Other works include \cite{leimkuhler2021better, leimkuhler2020constraint, marquez2017imposing}, which are related to the method of Lagrange multipliers and have their optimization and training 
dynamics largely determined by variants of Newton's method. We discuss the differences between our SGD-compatible method and that of Lagrange multipliers in more detail in \autoref{sec:lagrange_multipliers}.

In this paper, we present a general method for incorporating constraints directly into the ERM framework. More precisely, we consider 
a parametric family of admissible mappings $\mathcal{G}$, e.g., neural networks, and consider constraints that can be formulated
as a finite-dimensional system of equations imposed on (a subset of) the tunable parameters. Under mild conditions, 
the solution set of this system is guaranteed to be a smooth finite dimensional Riemannian manifold $\mathcal{M}$. We directly 
formulate and solve the constrained ERM problem on this manifold thereby ensuring that the desired constraints are satisfied \emph{exactly}
up to numerical precision. In particular, we explain in detail how to perform SGD on Riemannian manifolds arising from a finite-dimensional system 
of equations. Performing SGD on Riemannian manifolds has been studied 
before, e.g., \cite{roy2018geometry, bonnabel2013stochastic, NIPS2016_98e6f172, kasai2019riemannian, sato2019riemannian}. Our method, in particular,
heavily relies on the Implicit Function Theorem, which is used to construct explicit charts amenable to numerical computations. This allows for efficient evaluation 
of the (induced) Riemannian metric and gradients, which are vital for performing SGD on Riemannian manifolds. We will make our code publicly available.

\paragraph{Overview} The contributions of this paper are ordered as follows. In \autoref{sec:cerm} we introduce the theory and mathematical details of our proposed Constrained Empirical Risk 
Minimization framework. We end the section with examples of constraints that can be embedded into the framework. In \autoref{sec:wavelet} we dive deeper into the practical side; we consider a
highly non-trivial example in which we constrain filters of a CNN to be so-called wavelets. We will use the resulting wavelet networks in \autoref{sec:contours} to find data-driven wavelets 
for contour prediction. Specifically, we use wavelet networks to perform contour prediction in the medical domain, where we outperform strong baselines.

\section{Constrained Empirical Risk Minimization}
\label{sec:cerm}
In this section, we introduce a general framework for performing ERM with constraints, which we will refer to as Constrained Empirical Risk Minimization (CERM). We start with a brief review of the 
traditional ERM setup \cite{vapnik1991principles, vapnik1999nature} introducing the necessary terminology, notation, and assumptions. Next, we explain how to incorporate
constraints into the ERM framework in the form of a system of equations. We provide sufficient conditions on the system of equations to guarantee that the solution set is a Riemannian manifold $\M$. 
Finally, we explain how the (induced) Riemannian metric and associated geometric quantities can be numerically evaluated, 
which in turn enables us to directly perform SGD on the Riemannian manifold $\M$. 

\subsection{Mathematical setup}
The central notion in supervised learning is ``data'', which consists of input-target pairs. In the ERM framework, 
data is modeled as realizations of a pair of random variables. Therefore, in order to formally argue about data, we 
first introduce the necessary probabilistic notation and terminology. 

\paragraph{Probabilistic setup}
Let $\left( \Omega, \Sigma, \PP \right)$ be a probability space and $X : \Omega \rightarrow \mathcal{X}$ a random variable
whose realizations are interpreted as ``input''. Here $\mathcal{X}$ is a measurable space typically chosen to be a vector
space. In a supervised setting, the random variable $X$ is paired with a random variable $Y : \Omega \rightarrow \mathcal{Y}$, 
whose realizations correspond to ``targets'' associated with the input. Here $\mathcal{Y}$ is also a measurable space. 
As a side note, self-supervised settings fall into this framework as well, in which case the target $Y$ is created on the fly as a function of $X$.
Realizations $(x,y) \in \mathcal{X} \times \mathcal{Y}$ of $(X,Y)$ are together interpreted as input-target pairs. 
For example, $X$ could correspond to discretized images and $Y$ to associated contours describing the boundaries of
(simply connected) regions of interest. In this case, one may choose $\mathcal{X} = [0,1]^{n_{1} \times n_{2}}$, where
$n_{1}, n_{2} \in \NN$ are the dimensions of the images, and $\mathcal{Y} = C_{\text{per}}^{1}([0,1]; \RR^{2})$, both equipped 
with the Borel $\sigma$-algebra. 

\paragraph{Empirical risk minimization}
The goal of the ERM framework is to find a measurable map $G: \mathcal{X} \rightarrow \mathcal{Y}$ such that 
$G(x) \approx y$ for ``most'' realizations of $(X,Y)$. To precisely describe in what sense this approximation 
should hold, one quantifies the discrepancy between predicted and observed targets, $G(x)$ and $y$, respectively,
using a loss function $L: \mathcal{Y} \times \mathcal{Y} \rightarrow [0, \infty)$. We assume without loss of generality
that $L$ assumes positive values only, and that $L$ decreases as the accuracy of predictions increases.
In this setting, zero corresponds to ``perfect'' predictions, i.e., $G(x) = y$.  
The main objective of ERM is then to find an optimal map $G^{\ast}:  \mathcal{X} \rightarrow \mathcal{Y} $, which 
solves the minimization problem
\begin{align}
    \label{eq:ERM}
	\min_{G \in \G} \int_{\mathcal{X} \times \mathcal{Y}} L(G(x), y) \ \mbox{d}\PP_{(X,Y)}. 
\end{align}
Here $\PP_{(X,Y)}$ is the push-forward measure of $\PP$ on the sample space $\mathcal{X} \times \mathcal{Y}$ 
and $\mathcal{G}$ is a suitable subset of measurable functions $G: \mathcal{X} \rightarrow \mathcal{Y}$.
Note that the existence of a minimum is a key assumption in this framework.

In all our applications, we assume that $X$ and $Y$ are random vectors with sample spaces 
$\mathcal{X} = \RR^{n}$ and $\mathcal{Y} = \RR^{m}$, where $m, n \in \NN$. Furthermore, we assume
$\G$ is a parametric set that consists of mappings $\G = \{ G(\cdot, \xi): \xi \in \RR^{p} \}$,
where $G: \RR^{n} \times \RR^{p} \rightarrow \RR^{m}$ is a continuously differentiable map. 
Here $p \in \NN$ denotes the number of free parameters. With these assumptions in place, the minimization problem 
in \eqref{eq:ERM} is equivalent to $\min_{\xi \in \RR^{p}} \E \left( L(G(X, \xi), Y) \right)$.

\begin{remark}
In practice, we only have a finite set of observations $\mathcal{D}_{n_{s}}:= \{ (x^{i}, y^{i}): 1 \leq i \leq n_{s} \}$
at our disposal, where $(x^{i}, y^{i})$ are i.i.d. samples drawn from $(X, Y)$, and $n_{s} \in \NN$ is the number 
of samples. For a sufficiently large sample size $n_{s}$, the expected loss can be accurately approximated with an
arithmetic average by the Strong Law of Large Numbers. For this reason, we replace \eqref{eq:ERM} 
with the approximate problem 
\begin{align*}
        \min_{\xi \in \RR^{p}} \frac{1}{n_{s}} \sum_{j=1}^{n_{s}} L \left( G(x^{j},  \xi), y^{j} \right).
    \end{align*}
\end{remark}

\subsection{Imposing constraints}
\begin{figure}[!t]
	\centering
	{{\includegraphics[width=0.95\textwidth]{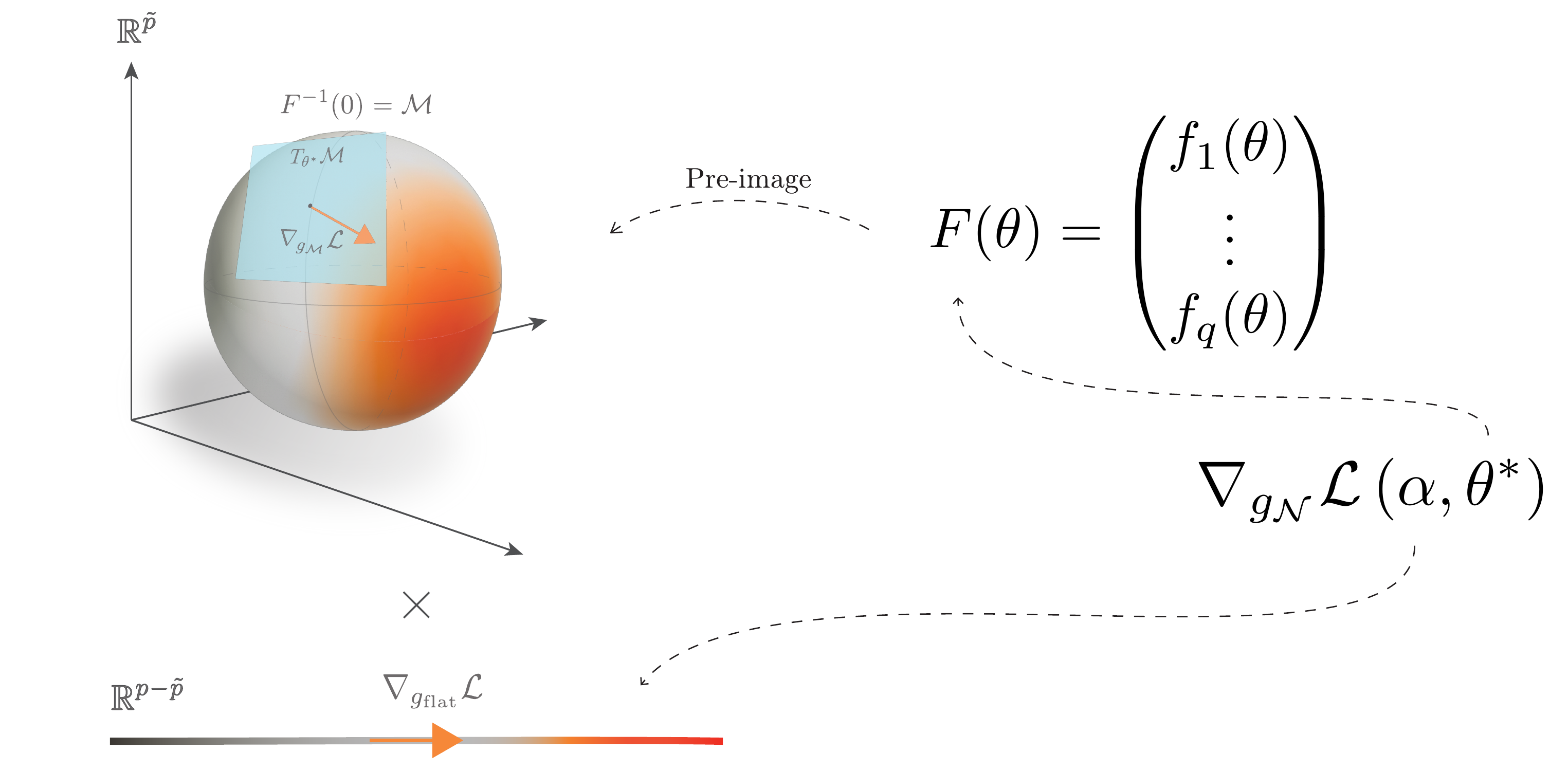}}} \\[1ex]
	\caption{An overview of the CERM framework: on the top-left side the full gradient of the loss $\mathcal{L}$ is shown, where $\alpha$ are the unconstrained parameters, and $\theta$ the constrained ones. The 
     constraints can be written in the form $F(\theta)=0$; the solution set $F^{-1}(0)$ is an embedded submanifold $\mathcal{M}$ of $\RR^{\tilde p}$. The constrained parameters are updated by following a path
     on $\mathcal{M}$ in the direction of the negative gradient $-\nabla_{g_{\mathcal{\M}}} \mathcal{L}( \alpha, \theta^{\ast})$. This ``constrained'' part of the full gradient is contained in the tangent space 
     $T_{\theta^{\ast}}\M$ of the embedded submanifold. The color of the manifold indicates the value of the loss function. By restricting the relevant components 
     of our descent trajectories to the embedded submanifold $\mathcal{M}$, we 
     \emph{always} satisfy the constraints imposed by $F$. 
     The gradient $\nabla_{g_\text{flat}}\mathcal{L}(\alpha, \theta^{\ast})$ and parameter updates for the unconstrained parameters $\alpha$ are computed as usual 
     using standard SGD for flat space (depicted on the bottom-left side). 
    \label{fig:gradients_overview}
	}
\end{figure}
In this section, we explain how to incorporate constraints on a subset of the parameters $\xi$ directly into the ERM framework. 
In addition, we provide explicit examples of simple constraints, which may be used for instance to encode equivariance in 
Multi Perceptration Layers (MLPs). 

\paragraph{Constraints}
We consider constraints given in the form of a system of equations. More explicitly, let
$F: \RR^{\tilde p} \rightarrow \RR^{q}$ be a twice continuously differentiable map, where $\tilde p \in \NN$
denotes the number of constrained parameters and $q \in \NN$ is the number of equations. 
We assume that $q < \tilde p \leq p$. For notational convenience, we decompose $\RR^{p} = \RR^{p - \tilde p} \oplus \RR^{\tilde p}$,
where the first and second subspaces correspond to the unconstrained and constrained parameters, respectively. 
We take $\pi_{p - \tilde p}: \RR^{p} \rightarrow \RR^{p - \tilde p}$ and $\pi_{\tilde p} : \RR^{p} \rightarrow \RR^{\tilde p}$ to be 
the projections onto the unconstrained and constrained parameter subspace, respectively. We will denote the
unconstrained and constrained parameters by $\alpha \in \RR^{p - \tilde p}$ and $\theta \in \RR^{\tilde p}$, respectively, 
i.e., $\alpha = \pi_{p - \tilde p}( \xi)$ and $\theta = \pi_{\tilde p}(\xi)$. The \emph{constrained} ERM problem is
defined below.
\begin{definition}[CERM]
	\label{def:CERM}
	Let $G: \RR^{n} \times \RR^{p} \rightarrow \RR^{m}$ be a continuously differentiable parameterization
	of admissible mappings and $F: \RR^{\tilde p} \rightarrow \RR^{q}$ a twice continuously differentiable
	constraint, where $q < \tilde p \leq p$. Suppose $L: \RR^{m} \times \RR^{m} \rightarrow [0, \infty)$ is a
	continuously differentiable loss function. The \emph{constrained} 
	ERM problem for $(X,Y)$ with respect to $(G, F, L)$ is defined by
	\begin{align}
   		\label{eq:CERM}
    		\begin{cases}
       			\min\limits_{\xi \in \RR^{p}} \E \left( L(G(X, \xi), Y) \right), \\
       			 \text{s.t.} \ F(\pi_{\tilde p}(\xi)) = 0.
   		\end{cases}
	\end{align}	
\end{definition}

Note the generality of the admissible mappings $G$ in Definition \ref{def:CERM}. Although we will focus on neural networks 
from now on, the proposed framework applies to any parametric model, e.g., logistic or polynomial regression models. 
Next, we show that the CERM problem in \eqref{eq:CERM} can be reformulated as an
ordinary ERM problem on a Riemannian manifold $(\mathcal{N}, g_{\mathcal{N}})$,
provided that the system of equations satisfies a mild non-degeneracy condition. 
This result allows us to consider the admissible parameters as a geometric object 
in its own right, whose intrinsic geometry we exploit to solve \eqref{eq:CERM}. 
\begin{theorem}
	\label{thm:CERM}
	If zero is a regular value of $F$, then the CERM problem in \eqref{eq:CERM} is equivalent to 
	solving an ordinary ERM problem on a Riemannian manifold $(\mathcal{N}, g_{\mathcal{N}})$
	of dimension $p-q$. Here $\mathcal{N} = \RR^{p - \tilde p} \times \M$ is an embedded $C^{2}$-submanifold of $\RR^{p}$
	and $\M := F^{-1}(0)$. The equivalent minimization problem is given by 
	\begin{align}
   		\label{eq:CERM_riemann}
        		\min\limits_{(\alpha, \theta) \in \mathcal{N}} \E \left( L \left(G \left (X, \alpha \oplus \iota(\theta) \right), Y \right) \right),
	\end{align}
	where $\iota: \M \rightarrow \RR^{\tilde p}$ is the inclusion map. 
	\begin{proof}
 		The solution set $\M := F^{-1}(0)$ is an embedded $C^{2}$-submanifold of $\RR^{\tilde p}$ 
		of dimension $\tilde p - q$ by the Implicit Function Theorem, since zero is a regular value of $F$. A detailed
		review of this statement is provided in Theorem \ref{theorem:preimage}. Since $\M$ is naturally 
		embedded in $\RR^{\tilde p}$, we may endow it with the pull-back metric $g_{\M}$, turning it into a Riemannian manifold 
		$(\M, g_{\M})$. Here $g_{\M} := \iota^{\ast}g_{\text{flat}} $, where 
		$g_{\text{flat}}$ is the standard Euclidean metric on $\RR^{\tilde p}$.
		The constrained ERM problem can now be reformulated as an ordinary ERM problem on the product manifold 
		$(\mathcal{N}, g_{\mathcal{N}}) := \left(\RR^{p - \tilde p} \times \M, \  g_{\text{flat}} \oplus g_{\M} \right)$. Note that
		$\text{dim}(\mathcal{N}) = p -q$. 
		Here $g_{\mathcal{N}} =g_{\text{flat}} \oplus g_{\M}$ is the product metric and $g_{\text{flat}}$ corresponds\footnote{Formally, we should incorporate the dimension $\tilde p$ into the notation for 
		the flat metric on $\RR^{\tilde p}$. However, to avoid clutter in the notation, 
		we will denote the standard Euclidean metric on any finite-dimensional 
		vector space in the same way.} to the standard Euclidean metric on $\RR^{p - \tilde p}$. 
		Altogether, having these geometric structures in place, the CERM problem in \eqref{eq:CERM} is equivalent to 
		\eqref{eq:CERM_riemann}, which proves the statement.
	\end{proof}
\end{theorem}

For convenience, we shall from now on refer to the objective $\mathcal{L}: \mathcal{N} \rightarrow [0, \infty)$ 
in \eqref{eq:CERM_riemann} as simply the loss. We end this section with two simple examples of
constraints that fit into our framework.

\begin{example}[Equivariance]
    For our first example, we show how to constrain layers in MLPs to be equivariant with respect to a given 
	family of commuting operators $\mathcal{A} \subset \RR^{n \times n}$. A well-known example is the case
	when $\mathcal{A}$ consists of circular shifts on $\RR^{n}$, which corresponds to translation equivariance. 
	To illustrate the technique, we consider a fully connected (pre-activated) layer $\eta: \RR^{n} \times \RR^{p} \rightarrow \RR^{n}$
	without bias, i.e., $\eta(x) = Wx$, for some weight matrix $W \in \RR^{n \times n}$. 
	In this setup, there are no unconstrained parameters and
	$p = \tilde p = n^{2}$. 
	We require that $A \eta(x) = \eta(Ax)$ for all $x \in \RR^{n}$ and $A \in \mathcal{A}$. This is equivalent 
	to $[A, W]=0$ for all $A \in \mathcal{A}$. 
	
	To set up constraints, we assume there exists an operator $A_{0} \in \mathcal{A}$ 
	which has $n$ simple eigenvalues. In this case, $A_{0}$ has $n$ linearly independent, 
	possibly complex-valued, eigenvectors $v_{1}, \ldots, v_{n} \in \CC^{n}$. For such an operator, it is straightforward to show  that $A_{0}$ commutes with $W$ if and only if there exists 
	a change of basis in which both operators are diagonal. More precisely, the commutator
	$[A_{0}, W] = 0$ if and only if $V^{-1} W V$ is diagonal, where 
    $V = \begin{bmatrix} v_{1} & \ldots & v_{n} \end{bmatrix}$ are
	eigenvectors of $A_{0}$. 
	This implies in particular that $V^{-1}AV$ is diagonal for all $A \in \mathcal{A}$. 
	
	The latter observation provides a convenient method for imposing 
	the desired constraint; we simply need to ensure that $V^{-1}WV$ is diagonal. 
	We consider the case that $V \in \CC^{n \times n}$ is complex-valued. 
	The real-valued case is dealt with similarly.
	Define $F: \CC^{n \times n} \rightarrow \CC^{q}$, where $q = n(n-1)$, by 
	\begin{align*}
		[F( \tilde W)]_{kl} = [V^{-1} \tilde W V]_{kl}, \quad
		1 \leq k, l \leq n, \quad k \not = l.
	\end{align*}
	Then $F(\tilde W) = 0$ if and only if $V^{-1}\tilde WV$ is diagonal. 
	Furthermore, if zero is a regular value of $F$, then $F^{-1}(0)$ is a 
	\emph{complex analytic} manifold of dimension $n$. This manifold can
	be identified with a \emph{real-analytic} manifold of dimension $2n$, which
	directly fits into our framework. In particular, since we seek 
	real-valued operators, we set $W$ equal to either the real or imaginary part of $\tilde W$, 
	which both commute with all operators in $\mathcal{A}$.
	
	As a concrete instantiation of this method, we consider the case 
	of translation equivariance again, where $\mathcal{A}$ is the set of 
	circular shifts on $\RR^{n}$. We choose $A_{0}$ to be the left-shift operator, which has
	$n$ simple eigenvalues; the $n$-th roots of unity. An associated collection of 
	eigenvectors is given by 
	\begin{align*}
		v_{j} = 
		\begin{bmatrix}
			1 & \omega_{n}^{j-1} & \ldots & \omega_{n}^{(j-1)(n-1)}
		\end{bmatrix}^{T}, 
		\quad \omega_{n} := e^{i \frac{2 \pi}{n}},
		\quad 1 \leq j \leq n. 
	\end{align*}
	Of course, for translation equivariance, we can solve the equation $F(\tilde W)=0$
	by hand, and show that $W$ needs to be a so-called Toeplitz or circular matrix. 
	This is equivalent to the statement that $\eta$ needs to be a convolutional layer. 
\end{example}

\begin{example}[Orthogonal filters]
	In the next example, we consider the work of \cite{Hu2020Provable}, where filters of a CNN were initialized to be orthogonal. 
	While not the intention of their paper, we can use the CERM framework to extend the orthogonality beyond initialization. 
	The constraints are relatively easy to set up, complementing the more involved constraints considered
	in our main application. This example may be interpreted as a warm-up towards our MRA example in \autoref{sec:wavelet}. 
	 
	Consider a convolutional layer with filters of size
	$M \times M$, where $M \geq 2$. We require that the filters are orthonormal throughout the entire training process.
    To be more precise, consider the case of one filter $h \in \RR^{M \times M}$.
	We require that 
	$
		hh^{T} = \bm{I}_{M \times M}.
	$
	This is equivalent to the following system of equations:
	\begin{align*}
		h_{\cdot l}^{T}h_{\cdot k} = \delta_{kl}, \quad l \leq k \leq M,
	\end{align*}
	for each $1 \leq l \leq M$. Motivated by this observation, we define $f_{l}: \RR^{M \times M} \rightarrow \RR^{M - l + 1}$
	by 
	\begin{align*}
		[f_{l}(h)]_{k-l+1} := h_{\cdot l}^{T}h_{\cdot k} - \delta_{kl}, \quad l \leq k \leq M,
	\end{align*}
	and $F: \RR^{M \times M} \rightarrow \RR^{\frac{1}{2}M(M+1)}$ by 
	$F := (f_{1}, \ldots, f_{M})$. Then zeros of $F$ correspond to orthonormal filters. 
	In this example, there are no unconstrained parameters, i.e., $p = \tilde p$. Furthermore, 
	$\tilde p = M^{2}$ and $q = \frac{1}{2}M(M+1)$. 
	The pre-image $\M = F^{-1}(0)$ is a smooth manifold of dimension $\frac{1}{2}M(M-1)$, 
	referred to as the orthogonal group $O(M)$. 
\end{example}

\subsection{Relation to Lagrange Multipliers}
\label{sec:lagrange_multipliers}
We briefly compare our strategy with a related alternative, namely the method of Lagrange Multipliers.
Lagrange Multipliers can be understood from a geometric perspective by essentially writing down the necessary conditions for 
stationarity in a special local chart, namely one in which $\M$ is embedded into $\RR^{\tilde p}$ as the graph of the inverse chart.  
The resulting necessary conditions for a point $\xi^{\ast} \in \RR^{p}$ to solve \eqref{eq:CERM} is the existence of a so-called 
Lagrange multiplier $\mu^{\ast} \in \RR^{q}$ so that
\begin{align}
	\label{eq:KKT}
	\begin{cases}
		\nabla_{g_{\text{flat}}} H(\xi^{\ast}) + \sum_{j=1}^{q} \mu^{\ast}_{j} \pi_{\tilde p}^{T} \nabla_{g_{\text{flat}}} F_{j} ( \pi_{\tilde p} (\xi^{\ast}) ) = 0, \\[1ex]
		F( \pi_{\tilde p}(\xi^{\ast}) ) = 0.
	\end{cases}
\end{align}
Here we have defined $H: \RR^{p} \rightarrow \RR$ by $H(\xi) := \E \left( L(G(X, \xi), Y) \right)$. 

The system of equations in \eqref{eq:KKT} is referred to as the Karush–Kuhn–Tucker (KKT) conditions. 
For general nonlinear problems, the KKT-conditions constitute a highly nonlinear system of equations
and are difficult to solve directly. Many techniques for solving the constrained problem in \eqref{eq:CERM} are based
on adaptations of Newton's method for \eqref{eq:KKT}, e.g., Sequential Quadratic Programming (SQP)
or Interior Point methods to name a few, see \cite{bertsekas2014constrained} for more. The dynamics of such algorithms, i.e., 
the behavior of the generated sequence of points, takes place in a higher dimensional space 
$\RR^{p} \times \RR^{q}$ than what we started with and is largely determined by Newton's method for solving \eqref{eq:KKT}. 

Our approach is 
fundamentally different from such methods in the following sense. Firstly, the dynamics of our optimization
scheme takes place on a \emph{lower} dimensional submanifold $\mathcal{N}$ defined by the constraints. Once 
we have initialized \emph{any} initial point on $\mathcal{N}$, we use the intrinsic geometry of the manifold to find a next point
by following descent trajectories \emph{confined to the manifold}, e.g., geodesics. We therefore satisfy the desired 
constraints \emph{throughout the entire} optimization procedure thereby exploring the space of feasible 
parameters directly. Finally, the dynamics of our algorithm is completely determined by the (negative) gradient flow 
of the objective, and not by Newton's method for \eqref{eq:KKT}.

\subsection{Graph coordinates on $\M$}
\label{sec:graph_chart}
In this section we explain how to construct a special (local) coordinate system, a so-called graph chart, 
on $\mathcal{M}$ around a point $\theta^{\ast} \in \M$. This chart will be used extensively to perform 
numerical computations, e.g., to evaluate the Riemannian metric $g_{\M}$. The existence of this special chart is guaranteed by the Implicit Function Theorem
and naturally comes up in the proof of the so-called Pre-Image Theorem \cite{lee2013smooth}, which provides sufficient
conditions for $\M = F^{-1}(0)$ to be an embedded submanifold of $\RR^{\tilde p}$. Below, we will essentially repeat the proof 
of this theorem, in a somewhat simplified setting, see \cite{lee2013smooth} for the slightly more general case dealing 
with smooth maps between general manifolds. The reason for including an explicit proof is that the 
computational steps form the backbone of our method.

\begin{theorem}[Pre-image theorem]
	\label{theorem:preimage}
	Let $F: \RR^{\tilde p} \rightarrow \RR^{q}$ be a map of class $C^{k}$, where $k \geq 2$.    
	If zero is a regular value of $F$, then $F^{-1}(0)$ is an embedded $C^{k}$-submanifold of 
	$\RR^{\tilde p}$ of dimension $\tilde p - q$. 
	\begin{proof}
		Assume zero is a regular value of $F$ and let $\theta^{\ast} \in F^{-1}(0)$ be arbitrary.
		Then $DF(\theta^{\ast})$ must have $q$ linearly independent columns. For the sake
		of concreteness, assume 
		\begin{align}
			\label{eq:DF_sub_invertible}
			\begin{bmatrix}
				\dfrac{\partial F}{ \partial \theta_{j_{1}} }(\theta^{\ast}) & \ldots & \dfrac{\partial F}{ \partial \theta_{j_{q}} }(\theta^{\ast})
			\end{bmatrix}
		\end{align} 
		is an isomorphism on $\RR^q$, where $j_{1} < \ldots < j_{q}$ and $1 \leq j_{k} \leq \tilde p$. 
		This gives rise to the decomposition $\RR^{\tilde p} = \RR^{q} \oplus \RR^{\tilde p -q}$, 
		where the first subspace corresponds to the coordinates with multi-index $(j_{1}, \ldots, j_{q})$,
		and the second subspace contains the remaining coordinates. Let $\pi_{q}: \RR^{\tilde p} \rightarrow \RR^{q}$ and 
		$\pi_{\tilde p - q} : \RR^{\tilde p} \rightarrow \RR^{\tilde p - q}$ denote the projections onto the first, 
		and second subspace, respectively, and write $v := \pi_{q}(\theta)$ and 
		$\beta := \pi_{\tilde p - q}(\theta)$ for the corresponding coordinates. We may then view
		$F$ as a function of $(v, \beta)$. More formally, we define a new map $\tilde F:  \RR^{q} \oplus \RR^{\tilde p -q} \rightarrow \RR^{q}$
		by $\tilde F(v, \beta) := F(\nu(v, \beta))$, where $\nu : \RR^{q} \oplus \RR^{\tilde p - q} \rightarrow \RR^{\tilde p}$ is a permutation 
		which puts the coordinates $(v, \beta)$ back in the original ordering. 
		
		Next, write $v^{\ast} = \pi_{q}(\theta^{\ast})$, $\beta^{\ast} = \pi_{\tilde p - q}(\theta^{\ast})$ and observe that 
		$D_{v}\tilde F(v^{\ast}, \beta^{\ast})$ is an isomorphism on $\RR^{q}$ by construction. Therefore, by the Implicit 
		Function Theorem, there exists a unique $C^{k}$-map $\tilde \zeta: B \subset \RR^{\tilde p - q} \rightarrow \RR^{q}$, 
		where $B$ is an open neighborhood of $\beta^{\ast}$, such that  $\tilde \zeta( \beta^{\ast} ) = v^{\ast}$ and 
       		$\tilde F\left( \tilde \zeta( \beta), \beta \right) = 0$ for all $\beta \in B$. Altogether, this shows that the map 
		$\zeta : B \rightarrow F^{-1}(0)$ defined by $\zeta(\beta) := \nu(\tilde \zeta(\beta), \beta)$ is a local parameterization 
		of $F^{-1}(0)$, i.e., its inverse $\Lambda := \zeta^{-1}$ is a local chart on $U := \zeta(B) \subset F^{-1}(0)$. 
		Therefore, since $\theta^{\ast} \in F^{-1}(0)$ is arbitrary, it follows from this observation 
		that $F^{-1}(0)$ is an embedded $C^{k}$-submanifold of dimension $\tilde p - q$. 
	\end{proof}
\end{theorem}

\begin{remark}[Relaxation]
	\label{remark:preimage_relaxation}
	Strictly speaking, one still needs to show that $U$ is open in $F^{-1}(0)$, and that there is 
	a chart in the ambient manifold $\RR^{\tilde p}$ in which $F^{-1}(0)$ is locally described
	by setting the first $q$ coordinates to zero. We omitted the details because they follow in a
	straightforward manner from our arguments.
	In particular, the proof of Theorem \ref{theorem:preimage} also shows that we may relax the condition 
	that zero is a regular value of $F$. Specifically, let $\mathcal{R} \subset F^{-1}(0)$ be the set of
	regular points of $F$. If $\mathcal{R} \not = \emptyset$, then $\mathcal{R}$ is an embedded 
    $C^{k}$-submanifold of $\RR^{\tilde p}$ of dimension $\tilde p - q$. 
\end{remark}
\begin{remark}[Graph coordinates and Lagrange Multipliers]
	The coordinates associated with the chart $\Lambda$ are commonly referred to as graph coordinates 
	since $\M$ is locally parameterized by the graph of $\tilde \zeta$. The existence of 
	Lagrange Multipliers can be proven by writing the necessary conditions for stationarity 
	of the objective in \eqref{eq:CERM} in this chart. 
\end{remark}
\begin{remark}[Regularity]
	If $F$ is $C^{\infty}$ or analytic, then the manifold inherits the same regularity.
\end{remark}

Throughout this section, we assume that zero is a regular value of $F$, which guarantees that $(\M, g_{\M})$
is (an embedded) $C^{2}$ Riemannian manifold. In the discussion below, we will consider a point $\theta^{\ast} \in \M$,
and explain how to explicitly evaluate the Riemannian metric at this point relative to the chart $\Lambda$. In turn, 
this will enable us to compute gradients. To avoid clutter in the notation, we henceforth 
assume without loss of generality, that the first $q$ components of $DF(\theta^{\ast})$ are linearly independent, i.e., 
$(j_{1}, \ldots, j_{q}) = (1, \ldots, q)$, and hence $F = \tilde F$. Note that this assumption will hold on an entire
open neighborhood of $\theta^{\ast}$. For points outside this neighborhood, one needs to choose another set
of components that constitute a linearly independent system, thereby obtaining a different chart $\Lambda$.

In practice, we do not have an explicit formula for the chart $\Lambda$ constructed 
in Theorem \ref{theorem:preimage}. Nonetheless, we can compute with it implicitly as explained below. For the sake 
of illustration, however, we will first consider a toy example before we proceed, in which explicit computations and 
formulae are available. We will continue this example throughout this section to complement the otherwise abstract 
numerical recipes. 

\begin{example}[The unit sphere $\mathbb{S}^{2}$]
	\label{example:unit_sphere}
	Consider the map $F: \RR^{3} \rightarrow \RR$ defined by 
	$F(\theta) := \theta_{1}^{2} + \theta_{2}^{2} + \theta_{3}^{2} - 1$. Clearly, $\M = F^{-1}(0)$ corresponds to the unit 
	sphere $\mathbb{S}^{2}$. We will use Theorem \ref{theorem:preimage} to prove that $\mathbb{S}^{2}$
	is a $C^{\infty}$ two-dimensional embedded submanifold of $\RR^{3}$. While one can easily prove this by constructing
	explicit charts, e.g., using stereographic projection or polar coordinates, our goal is to demonstrate how to use
	Theorem \ref{theorem:preimage} and explicitly construct the chart $\Lambda$.
	
	First observe that $DF(\theta) = 2 \begin{bmatrix} \theta_{1} & \theta_{2} & \theta_{3} \end{bmatrix}$. Further note that
	for any $\theta \in F^{-1}(0)$ at least one of the components $\theta_{j}$ must be nonzero. Therefore, $DF(\theta)$
	is surjective for all $\theta \in F^{-1}(0)$, i.e., zero is a regular value of $F$. Consequently, $\mathbb{S}^{2} = F^{-1}(0)$ is a $2$-dimensional embedded
	submanifold of $\RR^{3}$ by Theorem \ref{theorem:preimage}. Moreover, without explicitly constructing charts, we 
	immediately see that $\mathbb{S}^{2}$ is a $C^{\infty}$-manifold (analytic even), since $F$ is a $C^{\infty}$-map. 
	The chart $\Lambda$ from the proof is easily constructed in this case. To see this, suppose $\theta_{1} >0$,
	then $\beta = (\theta_{2}, \theta_{3})$, 
	$\zeta(\beta_{1}, \beta_{2}) = \left( \sqrt{ 1 - \beta_{1}^{2} - \beta_{2}^{2}}, \beta_{1}, \beta_{2} \right)$
	and $\Lambda(\theta) = (\theta_{2}, \theta_{3})$. The (maximal) domain of this chart is
        $U = \{ \theta \in \mathbb{S}^{2}: \ \theta_{1} > 0 \}$.
\end{example}

\subsection{Riemannian metric on $\mathcal{N}$}
\label{sec:riemann_metric}
In this section we express the product metric on $\mathcal{N}$ in local coordinates with respect to the
chart $\Phi := (\text{id}_{\RR^{p - \tilde p}}, \Lambda)$. Here $\text{id}_{\RR^{p - \tilde p}}$ denotes
the identity map on $\RR^{p - \tilde p}$. We start by deriving a representation of $g_{\M}$ relative to $\Lambda$. 
For this purpose, denote the coordinates associated to $\Lambda$ by $(\lambda^{1}, \ldots, \lambda^{\tilde p -q})$,
and the standard coordinates on $\RR^{\tilde p - q}$ by $\left (\beta^{1}, \ldots, \beta^{\tilde p -q} \right)$.
Recall that the pullback metric on $\M$ is given by $g_{\M} = \iota^{\ast} \langle \cdot, \cdot \rangle$. 
Therefore, in local coordinates, we have 
$
	g_{\M} = (g_{\M})_{ij} \ d\lambda^{i} \tensor d\lambda^{j},
$ 
where $(g_{\M})_{ij} : U \rightarrow \RR$ is given by 
\begin{align*}
	(g_{\M})_{ij}(\theta)  &= 
	\left \langle
		\iota_{\ast, \theta} \left( \frac{ \partial }{ \partial \lambda^{i} } \bigg \vert_{\theta} \right), \ 
		\iota_{\ast, \theta} \left( \frac{ \partial }{ \partial \lambda^{j} } \bigg \vert_{\theta} \right) 
	\right \rangle
	\\[2ex]
	&=
	\left \langle
		\frac{ \partial \zeta }{ \partial \beta^{i} } (  \Lambda (\theta) ), \
		\frac{ \partial \zeta }{ \partial \beta^{j} } (  \Lambda (\theta) )
	\right \rangle, \quad 1 \leq i, j \leq \tilde p - q \ ,
\end{align*}
where we recall that $\zeta=(\tilde \zeta (\beta), \beta))$ is a local parameterization of the manifold.
In practice, we are only interested in a specific choice for $\theta$, namely $\theta = \theta^{\ast}$. 
For this choice, the chart $\Lambda:=\zeta^{-1}$ is explicitly known:
$\Lambda(\theta^{\ast}) = \beta^{\ast}$. Hence, to evaluate the metric 
at $\theta^{\ast}$, we need to explicitly compute $D \zeta( \beta^{\ast})$. 

To evaluate $D \zeta\left( \beta^{\ast} \right)$, first observe that 
$D \zeta ( \beta) = \begin{bmatrix} D \tilde \zeta( \beta)^{T} & \bm{I}_{(\tilde p - q) \times (\tilde p - q)} \end{bmatrix}^{T}$  
for any $\beta \in B$. Here $\bm{I}_{(\tilde p - q) \times (\tilde p - q)}$ denotes the $(\tilde p - q) \times (\tilde p - q) $ identity matrix.
Furthermore, we can compute the derivative of $\tilde \zeta$ by using its defining property (see the proof of \autoref{theorem:preimage})
\begin{equation*}
    F\left( \tilde \zeta( \beta), \beta \right) = 0, \quad \beta \in B.
\end{equation*}
More precisely, differentiating both sides of this equation and evaluating at $\beta^{\ast}$ yields
\begin{align}
	\label{eq:Dzeta}
	D_{v} F ( \theta^{\ast} ) D \tilde \zeta ( \beta^{\ast} ) = - D_{\beta}F( \theta^{\ast} ). 
\end{align}
Both $D_{v} F ( \theta^{\ast} )$ and  $D_{\beta}F( \theta^{\ast} )$ can be explicitly evaluated. Moreover, 
$D_{v} F ( \theta^{\ast} )$ is a non-singular $q \times q$ matrix. Hence we can compute $D \tilde \zeta ( \beta^{\ast} )$ 
by solving the linear system of equations in \eqref{eq:Dzeta}. Subsequently, we can explicitly
evaluate the components of the Riemannian metric at $\theta^{\ast}$:
\begin{align}
	\label{eq:metric_eval}
	(g_{\M})_{ij}(\theta^{\ast}) &= 
	\left \langle
		\frac{ \partial \zeta }{ \partial \beta^{i} } (  \beta^{\ast} ), \
		\frac{ \partial \zeta }{ \partial \beta^{j} } (   \beta^{\ast} )
	\right \rangle, \quad 1 \leq i, j \leq \tilde p - q. 
\end{align}

Finally, we evaluate the product metric $g_{\mathcal{N}} = g_{\text{flat}} \oplus g_{\M}$ on $\mathcal{N}$
relative to $(\text{id}_{\RR^{p - \tilde p}}, \Lambda)$ at $\theta^{\ast}$:
\begin{align}
	\label{eq:product_metric}
	g_{\mathcal{N}} (\alpha, \theta^{\ast}) \simeq 
	[g_{\mathcal{N}}(\alpha, \theta^{\ast}) ]_{\Lambda} := 
	\begin{bmatrix}
		\bm{I}_{(p - \tilde p) \times (p - \tilde p)} & \bm{0}_{(p - \tilde p) \times (\tilde p - q)} \\[1ex]
		\bm{0}_{(\tilde p -q) \times (p - \tilde p) } & [g_{\M}(\theta^{\ast})]_{\Lambda}
	\end{bmatrix}, \quad 
	\alpha \in \RR^{p - \tilde p},
\end{align}
where $[g_{\M}(\theta^{\ast})]_{\Lambda} \in \text{GL}(\tilde p -q, \RR)$ is the symmetric matrix whose 
$(i,j)^{\text{th}}$ component is given by $(g_{\M})_{ij}(\theta^{\ast})$.

\begin{example}[The unit sphere $\mathbb{S}^{2}$ - continued]
	We end this section by continuing Example \ref{example:unit_sphere} 
	and computing the components of the Riemannian metric $g_{\mathbb{S}^{2}}$ 
	relative to $\Lambda$. This computation is only included to provide a concrete 
	application of the abstract theory above. In practice, the computations, e.g., 
	solving the equation in \eqref{eq:Dzeta}, are implemented numerically. 
	Now, a straightforward computation shows that 
	\begin{align*}
		D\zeta(\beta) = 
		\begin{bmatrix}
			- \dfrac{\beta_{1}}{ \sqrt{1 - \beta_{1}^{2} - \beta_{2}^{2}} } & - \dfrac{\beta_{2}}{ \sqrt{1 - \beta_{1}^{2} - \beta_{2}^{2}} } \\
			1 & 0 \\
			0 & 1
		\end{bmatrix}. 
	\end{align*}
	Therefore, the components of the Riemannian-metric relative to $\Lambda$ are given by
	\begin{align*}
		[g_{\mathbb{S}^{2}}(\theta)]_{\Lambda} = 
		\dfrac{1}{1 - \theta_{2}^{2} - \theta_{3}^{2} }\begin{bmatrix}
			1 - \theta_{3}^{2} & \theta_{2} \theta_{3} \\
			\theta_{2} \theta_{3} & 1 - \theta_{2}^{2}
		\end{bmatrix}. 
	\end{align*}
\end{example}

\subsection{Computing gradients on $\mathcal{N}$}

In this section we explain how to compute the gradient of a smooth map $\mathcal{L} : \mathcal{N} \rightarrow \RR$
relative to $\Phi = (\text{id}_{\RR^{p - \tilde p}}, \Lambda)$. For notational convenience, we denote the coordinates associated to 
$(\text{id}_{\RR^{p - \tilde p}}, \Lambda)$ by $(u^{1}, \ldots, u^{p - q})$, where $\left(u^{1}, \ldots, u^{p - \tilde p}\right) = 
\left(\alpha^{1}, \ldots, \alpha^{p - \tilde p} \right)$ and $( u^{p - \tilde p + 1}, \ldots, u^{p - q} ) = \left (\lambda^{1}, \ldots, \lambda^{\tilde p -q} \right)$
are the coordinates associated to $\text{id}_{\RR^{p - \tilde p}}$ and $\Lambda$, respectively. In the next section, we will use these computations
to find a minimizer of $\mathcal{L}$ using SGD. We remind the reader that our specific use case is the 
constrained ERM problem in \eqref{eq:CERM}, which corresponds to finding a minimum of 
\begin{align*}
    \mathcal{L}(\alpha, \theta) = \E \left( L \left(G \left (X, \alpha \oplus \iota(\theta) \right), Y \right) \right).
\end{align*}

The gradient of $\mathcal{L}$  on $\mathcal{N}$ with respect to $g_{\mathcal{N}}$ is the unique vector field  
$\nabla_{g_{\mathcal{N}}} \mathcal{L} \in \mathfrak{X}(\mathcal{N})$ satisfying $d\mathcal{L} = g_{\mathcal{N}}(\cdot, \nabla _{g_{\mathcal{N}}}\mathcal{L})$.
Such a vector field must exist since $g_{\mathcal{N}}$ is non-degenerate. In local coordinates, 
\begin{align*}
	d\mathcal{L} = \dfrac{ \partial \mathcal{L} }{ \partial u^{j}} du^{j}, \quad 
	\nabla_{g_{\mathcal{N}}} \mathcal{L}  = c^{j} \dfrac{ \partial }{ \partial u^{j}},
\end{align*} 
where $c^{1}, \ldots c^{p - q}: \mathcal{N} \rightarrow \RR$ are smooth (uniquely determined) functions. 
We can easily determine these functions by plugging them into the defining equation for the gradient and evaluating both sides
at $\dfrac{ \partial }{ \partial u^{i}}$. This yields the following linear system of equations: 
\begin{align*}
	c^{j} (g_{\mathcal{N}})_{ij} = \dfrac{ \partial \mathcal{L} }{ \partial u^{i}},  \quad 1 \leq i \leq p - q. 
\end{align*}
Here $(g_{\mathcal{N}})_{ij}: \RR^{p - \tilde p} \times U \rightarrow \RR$ are the components of $g_{\mathcal{N}}$ relative to $\Phi$. 
Similar as before, we define $[g_{\N}(\alpha, \theta)]_{\Phi} \in \text{GL}(p -q, \RR)$ to be the symmetric matrix whose 
$(i,j)^{\text{th}}$ component is given by $(g_{\mathcal{N}})_{ij}(\alpha, \theta)$. Then
\begin{align*}
	\nabla_{g_{\mathcal{N}}} \mathcal{L} = g^{ij}_{\mathcal{N}} \dfrac{ \partial \mathcal{L}}{ \partial u^{j}} \dfrac{ \partial}{ \partial u^{i} },
\end{align*} 
where $g_{\mathcal{N}}^{ij}(\alpha, \theta)$ are the components of the inverse of $[g_{\mathcal{N}}(\alpha, \theta)]_{\Phi}$. 

In practice, of course, we will not invert the matrix $[g_{\mathcal{N}}(\alpha, \theta^{\ast})]_{\Phi}$. Instead,
we numerically solve the system of equations at our point of interest $(\alpha, \theta^{\ast})$ for the unknown-coefficients 
$\left( c^{j}( \alpha, \theta^{\ast} ) \right)_{j=1}^{p - q}$ by exploiting the block structure of the metric, see \eqref{eq:product_metric}. 
In particular, we immediately see that the first $p - \tilde p$ components of $\nabla_{g_{\mathcal{N}}} \mathcal{L}(\alpha, \theta^{\ast})$ are given by
$c^{j}(\alpha, \theta^{\ast}) = \dfrac{ \partial \mathcal{L} }{ \partial \alpha^{j}}(\alpha, \theta^{\ast})$, where $1 \leq j \leq p - \tilde p$. 
In other words, since the metric on $\RR^{p - \tilde p}$ is flat, the associated components of the gradient 
reduce to the usual ones. On the other hand, for the coordinates on $\M$, we have  
\begin{align*}
	 \sum_{j=1}^{p - q} c^{j}(\alpha, \theta^{\ast}) (g_{\mathcal{N}})_{ij}(\alpha, \theta^{\ast}) = 
	 \sum_{j=p - \tilde p + 1}^{p - q} c^{j}(\alpha, \theta^{\ast}) \left( [g_{\M}(\theta^{\ast}) ]_{\Lambda} \right)_{(i + \tilde p -p, j + \tilde p -p)}, 
	\quad p - \tilde p + 1 \leq i \leq p - q
\end{align*}
by \eqref{eq:product_metric}. Therefore, the last $\tilde p - q$ components 
$\left( c^{j}(\alpha, \theta^{\ast}) \right)_{j=p - \tilde p + 1}^{p-q}$ 
of $\nabla_{g_{\mathcal{N}}} \mathcal{L}(\alpha, \theta^{\ast})$ can be obtained by solving the linear (square) system 
\begin{align}
	\label{eq:gradient_linear_system}
	 [g_{\M}(\theta^{\ast}) ]_{\Lambda}  \begin{pmatrix}
	 	c^{p - \tilde p + 1}(\alpha, \theta^{\ast}) \\
		\vdots \\
		c^{p - q}(\alpha, \theta^{\ast})
	 \end{pmatrix} = 
	 \begin{pmatrix}
	 	 \dfrac{ \partial \mathcal{L} }{ \partial \lambda^{1}}(\alpha, \theta^{\ast}) \\
		 \vdots \\
		  \dfrac{ \partial \mathcal{L} }{ \partial \lambda^{\tilde p - q }}(\alpha, \theta^{\ast})
	 \end{pmatrix}.
\end{align}

\paragraph{Computing partial derivatives}
We need one final ingredient to compute the gradient of $\mathcal{L}$. Namely, we need to 
evaluate its partial derivatives with respect to the coordinate system defined by $\Phi = (\text{id}_{\RR^{p - \tilde p}}, \Lambda)$.
Clearly there is no difficulty in computing $\dfrac{ \partial \mathcal{L}}{ \partial \alpha^{i} }( \alpha,  \theta^{\ast})$, since 
$( \alpha^{1}, \ldots, \alpha^{p - \tilde p} )$ are the standard coordinates on $\RR^{p - \tilde p}$, and 
thus correspond to the ``usual'' partial derivatives one encounters in calculus on vector spaces. For 
the partial derivatives with respect to $(\lambda^{1}, \ldots, \lambda^{\tilde p - q})$, however, we have to be
more careful, and compute from the perspective of the (non-trivial) chart:
\begin{align}
	\label{eq:partial_derivatives}
	\frac{ \partial \mathcal{L}}{ \partial \lambda^{i} }( \alpha, \theta^{\ast}) &= 
	\frac{ \partial (\mathcal{L} \circ \Phi^{-1}) }{ \partial \beta^{i}}( \Phi(\alpha, \theta^{\ast}) ) \nonumber \\[2ex] &= 
	\frac{ \partial }{ \partial \beta^{i}} \bigg \vert_{\beta^{\ast}} 
	( \beta \mapsto \mathcal{L}( \alpha, \zeta(\beta) ) \nonumber \\[2ex] &= 
	D_{\theta} \mathcal{L} (\alpha, \theta^{\ast}) \frac{ \partial \zeta}{ \partial \beta^{i}}( \beta^{\ast}), \quad 1 \leq i \leq \tilde p -q,
\end{align}
since $\Phi^{-1} = (\text{id}_{\RR^{p - \tilde p}}, \zeta)$ and $\zeta( \beta^{\ast} ) = \theta^{\ast}$. In the last line
we assumed that $\mathcal{L}(\alpha, \cdot)$ has a smooth extension to some open neighborhood $V \subset \RR^{\tilde p}$
of $\M$ for all $\alpha \in \RR^{p - \tilde p}$. This is the case for all our applications, where $\mathcal{L}$ comes from 
the constrained minimization problem in \eqref{eq:CERM}. 
 
Altogether, we now have all the ingredients to numerically evaluate the gradient of a smooth map 
$\mathcal{L} : \mathcal{N} \rightarrow \RR$ relative to the chart $(\text{id}_{\RR^{p - \tilde p}}, \Lambda)$. 
The steps are summarized in Algorithm \ref{algorithm:gradient}. 
\begin{algorithm}[tb]
	\caption{\label{algorithm:gradient} Compute $\nabla_{g_{\mathcal{N}}} \mathcal{L}(\alpha, \theta^{\ast})$ 
		     relative to $\Phi$ given $(\alpha, \theta^{\ast}) \in \mathcal{N}$.}
	\begin{algorithmic}[1]
	\State Compute $DF(\theta^{\ast})$. 
	\State Compute $D \zeta (\beta^{\ast}) = \begin{bmatrix} D \tilde \zeta(\beta^{\ast})^{T} & \bm{I}_{(\tilde p - q) \times (\tilde p - q)} \end{bmatrix}^{T}$   by solving \eqref{eq:Dzeta}. 
	\State Compute $[g_{\N}(\alpha, \theta^{\ast})]_{\Phi}$ by evaluating \eqref{eq:product_metric}. 
	\State Compute the components of $\nabla_{g_{\text{flat}}}\mathcal{L}( \alpha, \theta^{\ast})$ by evaluating $D_{\alpha} \mathcal{L}(\alpha, \theta^{\ast})$. 
	\State Compute the partial derivatives $\frac{ \partial \mathcal{L}}{ \partial \lambda^{i} }( \alpha, \theta^{\ast})$ for $1 \leq i \leq \tilde p -q$ using \eqref{eq:partial_derivatives}.
	\State Compute the components of $\nabla_{g_{\M}} \mathcal{L}(\alpha, \theta^{\ast})$  by solving \eqref{eq:gradient_linear_system}.
	\end{algorithmic}
\end{algorithm}

\begin{example}[The unit sphere $\mathbb{S}^{2}$ - continued]
	We continue our example of the unit sphere and explain how to 
	compute the gradient of a smooth map $\mathcal{L}: \mathbb{S}^{2} \rightarrow \RR$. 
	We assume that $\mathcal{L}$ can be smoothly extended to an open neighborhood
	of $\mathbb{S}^{2}$ in $\RR^{3}$. To compute the gradient relative to $\Lambda$, 
	we need to solve the system in \eqref{eq:gradient_linear_system}. For this purpose, 
	we first explicitly compute the inverse of $[g_{\mathbb{S}^{2}}(\theta)]$:
	\begin{align*}
		\left( [g_{\mathbb{S}^{2}}(\theta)]_{\Lambda} \right)^{-1} = 
		\begin{bmatrix}
			1 - \theta_{2}^{2} & - \theta_{2} \theta_{3} \\
			- \theta_{2} \theta_{3} & 1 - \theta_{3}^{2}
		\end{bmatrix}.
	\end{align*} 
	Again, we stress that in practice, we do not invert this matrix, but solve
	the system of equations numerically instead. Next, we compute the partial derivatives of 
	$\mathcal{L}$ relative to $\Lambda = (\lambda^{1}, \lambda^{2})$ using \eqref{eq:partial_derivatives}:
	\begin{align*}
		\dfrac{ \partial \mathcal{L} }{ \partial \lambda^{1} }(\theta) = 
		\dfrac{ \partial \mathcal{L} }{ \partial \theta_{2} }(\theta) - \dfrac{\theta_{2} }{ \theta_1}  \dfrac{ \partial \mathcal{L} }{ \partial \theta_{1} }(\theta), \quad
		\dfrac{ \partial \mathcal{L} }{ \partial \lambda^{2} }(\theta) = 
		\dfrac{ \partial \mathcal{L} }{ \partial \theta_{3} }(\theta)  - \dfrac{\theta_{3} }{\theta_1} \dfrac{ \partial \mathcal{L} }{ \partial \theta_{1} }(\theta).
	\end{align*} 
	Here $\left( \dfrac{ \partial L}{ \partial \theta_{j} } \right)_{j=1}^{3}$ denote the partial derivatives
	with respect to the standard coordinates on $\RR^{3}$, i.e., these are the ``usual'' partial derivatives from calculus on vector spaces. 
	Hence 
	\begin{align*}
		\nabla_{g_{\mathbb{S}^{2}}} \mathcal{L} (\theta) = 
		c_{1}(\theta) \frac{ \partial }{ \partial \lambda^{1}} \biggl \vert_{\theta} + c_{2}(\theta) \frac{ \partial }{ \partial \lambda^{2}} \biggl \vert_{\theta}
		\simeq 
		\begin{bmatrix} 
			c_{1}(\theta) \\
			c_{2}(\theta) \\
		\end{bmatrix},
	\end{align*}
	where
	\begin{align*}
		c_{1}(\theta) &= \dfrac{ \partial \mathcal{L} }{ \partial \theta_{2} }(\theta) - 
		\theta_{2} \left( 
			\theta_{1} \dfrac{ \partial \mathcal{L} }{ \partial \theta_{1} }(\theta) 
			+ \theta_{2} \dfrac{ \partial \mathcal{L} }{ \partial \theta_{2} }(\theta) 
			+ \theta_{3} \dfrac{ \partial \mathcal{L} }{ \partial \theta_{3} }(\theta)
		\right), \\
		c_{2}(\theta) &= \dfrac{ \partial \mathcal{L} }{ \partial \theta_{3} }(\theta) - 
		\theta_{3} \left( 
			\theta_{1} \dfrac{ \partial \mathcal{L} }{ \partial \theta_{1} }(\theta) 
			+ \theta_{2} \dfrac{ \partial \mathcal{L} }{ \partial \theta_{2} }(\theta) 
			+ \theta_{3} \dfrac{ \partial \mathcal{L} }{ \partial \theta_{3} }(\theta)
		\right).
	\end{align*}
\end{example}

\subsection{Stochastic Gradient Descent}
\label{sec:appendix_cerm_sgd}
In this section we explain how to perform SGD on Riemannian manifolds using graph coordinates. For previous work on SGD on Riemannian manifolds, we
refer the reader to \cite{roy2018geometry, bonnabel2013stochastic, NIPS2016_98e6f172, kasai2019riemannian, sato2019riemannian}. 
The presented technique is completely intrinsic to the manifold $\mathcal{N}$ and 
involves following (approximate) geodesics in the direction of the (negative) gradient of $\mathcal{L}$. To explain this idea in more detail, we first briefly recall
the notion of geodesics and refer the reader to \cite{lee2006riemannian, lee2013smooth} for a more comprehensive introduction to differential geometry. 

\subsubsection{Geodesics and parallel transport}
The analog of a gradient descent step on a Riemannian manifold $(\N, g_{\N})$ is to follow ``a straight line'',
confined to the manifold, in the direction of the negative gradient. In order to make sense of this, one first needs to generalize the notion of a
straight line to arbitrary Riemannian manifolds. On Euclidean vector spaces, one can define a straight line as a curve
 whose velocity is constant. This notion makes sense on a vector space, since different tangent spaces can be related to 
 one another, but does not make sense on a general manifold. An equivalent notion, which \emph{can} be generalized to a Riemannian
 manifold, is to define a straight line as a curve whose acceleration is zero. The key idea here is that the notion of acceleration can
 be made sense of on any Riemannian manifold. More precisely, one can define a so-called affine connection or covariant derivative $\nabla$, 
 not to be confused with the notation for a gradient, which allows
 one to measure the change of one vector field in the direction of another. Formally, a connection is a differential operator 
 $\nabla: \mathfrak{X}(\N) \times \mathfrak{X}(\N) \rightarrow \mathfrak{X}(\N)$, which is $C^{\infty}(\N)$-linear in the
 first variable, $\RR$-linear in the second, and satisfies the Leibniz rule. Given two vector fields $V, W \in \mathfrak{X}(\N)$, 
one typically writes $\nabla_{V}W$ and interprets this new vector field as measuring the change of $W$ in the direction of $V$. 
 
A connection is a so-called \emph{local} operator in the sense that $\nabla_{V}W(u)$ is completely determined by $V(u) \in T_{u} \N$
and the behavior of $W$ in a neighborhood around $u \in \mathcal{N}$. We may therefore write $\nabla_{V}W(u) =\nabla_{V(u)}W(u)$.
This local property can in turn be used to measure the change of a vector field in the direction of a curve. More precisely, given a
curve $\gamma$, there exists a unique (differential) operator $D_{t}$ associated to $\gamma$ and $\nabla$, which enables one to 
differentiate vector fields $V \in \Gamma(\gamma)$ in the direction of $\gamma$. This operator is uniquely determined by three properties: it is
$\RR$-linear, satisfies the Leibniz rule, and if $V \in \Gamma(\gamma)$ can be extended to a vector field $\tilde V$ defined on 
an open neighborhood of $\gamma(t)$, then $D_{t}V(t) = \nabla_{\dot \gamma(t)} \tilde V(\gamma(t))$. 
One can now make sense of acceleration by defining it as the derivative of the velocity field $\dot \gamma$ in the direction of $\gamma$ itself, 
i.e., acceleration is defined by $D_{t} \dot \gamma$.  A ``straight line'' or geodesic is then simply defined as a curve whose acceleration field is zero. 
The existence of geodesics is guaranteed, at least locally, by the existence and uniqueness theorem for ODEs, see the discussion below. 

\begin{figure}[tb]
	\centering
	{{\includegraphics[width=0.75\textwidth]{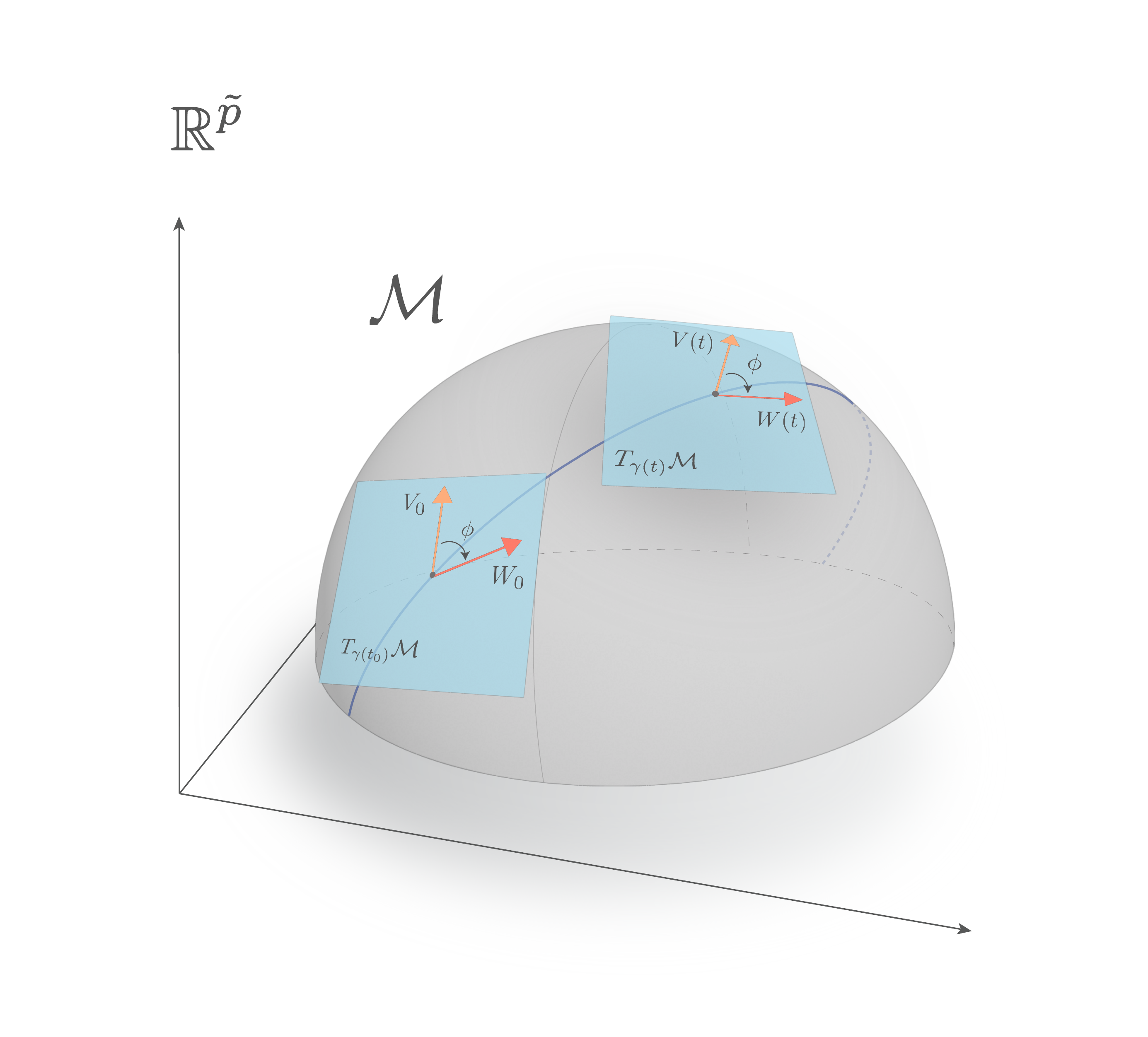}}} \\[1ex]
	\caption{In this figure we depict a curve $\gamma: [0, T] \rightarrow \mathcal{M}$ (in blue) on which we have drawn two points, 
		     $\gamma(t_{0})$ and $\gamma(t)$, for some $t, t_{0} \in (0, T)$. In addition, we have drawn the tangent spaces associated
		     to these points.
		     The tangent vectors $V_{0}, W_{0} \in T_{\gamma(t_{0})} \mathcal{M}$
		     are ``parallel transported'' along $\gamma$ resulting in vector fields $V, W \in \Gamma(\gamma)$. 
		     The Levi-Civita connection is the unique torsion free connection for which the angle between any two vectors 
		     $V_{0}, W_{0} \in T_{\gamma(t_{0})} \mathcal{M}$ and their parallel extensions remains constant.
		     \label{fig:metric_compatibility}
	}
\end{figure}
A covariant derivative $\nabla$ allows one to generalize many more familiar concepts from
Euclidean vector spaces to Riemannian manifolds. For instance, given a curve  $\gamma: [0, T] \rightarrow \N$ and tangent vector 
$V_{0} \in T_{\gamma(t_{0})} \N$, one may extend $V_{0}$ to a vector field $V \in \Gamma(\gamma)$ which ``is parallel'' to $V_{0}$ everywhere, 
see Figure \ref{fig:metric_compatibility}. This extension $V$ is referred to as the parallel transport of $V_{0}$ along $\gamma$. 
The notions of geodesics and parallel transport, however, heavily depend on the choice of connection.
In general, there exist infinitely many connections on a Riemannian manifold. There exists exactly one connection, 
however, the so-called Levi-Civita connection, which in a sense is ``naturally aligned'' with the Riemannian metric. This specific connection may 
be summarized in a geometric way by the following two conditions, which are usually taken for granted on Euclidean spaces.
First, if $\gamma: [0, T] \rightarrow \N$ is a curve and $V_{0}, W_{0} \in T_{\gamma(t_{0})}\N$ are tangent vectors with angle $\phi$ 
between them, then the parallel extensions $V, W \in \Gamma(\gamma)$ must have angle $\phi$ between them as well at any point on $\gamma$
(metric compatibility), see Figure \ref{fig:metric_compatibility}. Secondly, for any coordinate chart on $\N$, the rate of change
of one coordinate direction in the direction of another must not change if we swap directions (torsion free). 
In this paper we always use the Levi-Civita connection. 

Finally, we provide a local description of a geodesic $\gamma$. Let $t_{0} \in (0,T)$ and assume $(U, u^{1}, \ldots, u^{\dimN})$ is 
any chart containing $\gamma(t_{0})$, then there exists a $\delta >0$ such that $\gamma( (t_{0} - \delta, t_{0} + \delta )) \subset \N$. 
Write $\partial_{l} = \frac{ \partial }{ \partial u^{l} }$ and observe that for each $1 \leq i, j \leq \dimN$, there exist smooth functions 
$\Gamma^{k}_{ij}: U \rightarrow \RR$ such that $\nabla_{\partial_{i}} \partial_{j} = \Gamma^{k}_{ij} \partial_{k}$, since 
$(\partial_{l})_{l=1}^{\dimN}$ is a frame on $U$. The coefficients $\left \{ \Gamma^{k}_{ij}: 1 \leq i, j, k \leq \dimN \right \}$ are called 
the \emph{Christoffel symbols} of $\nabla$ on $U$. They completely characterize the connection on $U$. The equation for a 
geodesic starting at an initial point $u_{0}$ with initial velocity $V_{0}$ is given by
\begin{align}
	\label{eq:geodesic_equation}
	\begin{cases}
		\ddot \gamma^{k}(t) + \dot \gamma^{i}(t) \dot \gamma^{j}(t) \Gamma^{k}_{ij}( \gamma(t) ) = 0, & 1 \leq k \leq \dimN, \\[2ex]
		\dot \gamma^{k}(t_{0}) = V^{k}_{0}, & 1 \leq k \leq \dimN, \\[2ex]
		\gamma(t_{0}) = u_{0},
	\end{cases}	
\end{align}
see \cite{lee2006riemannian}. Here we have expressed $\gamma$ and the components of its velocity in local coordinates:
\begin{align*}
	\dot \gamma(t) = \dot \gamma^{i}(t) \partial_{i} \bigl \vert_{\gamma(t)}, \quad \gamma^{i} := u^{i} \circ \gamma.
\end{align*} 

This is a second-order ordinary differential equation for the unknown curve (geodesic) $\gamma$. In general, this 
equation is \emph{nonlinear}. The existence and uniqueness theorem for ODEs only guarantees the existence of a 
\emph{local} solution. The solution may be extended outside of $U$ by considering other charts. However, due to the
nonlinearity, there may be obstructions to extending the solution beyond a certain point. 
In general, there is no guarantee that a geodesic can be extended and defined for all $t \in \RR$. A manifold with the 
property that geodesics exist for all time is called \emph{complete}. In particular, any compact manifold is 
complete \cite{lee2006riemannian}. We remark that for the purpose of SGD local existence is sufficient, since 
we need to take sufficiently small steps on the manifold to guarantee descent of the objective. 

\subsubsection{Gradient descent steps}
\label{sec:gradient_descent}
We will now explain how to define a gradient descent step on our manifold of interest 
$(\N, g_{\N})= \left( \RR^{p - \tilde p} \times \M, g_{\text{flat}} \oplus g_{\M} \right)$ by computing
approximate solutions of the geodesic equation \eqref{eq:geodesic_equation}. The main idea is to follow
the geodesic starting at our current point $(\alpha, \theta^{\ast})$ in the direction of the negative gradient $-\nabla_{g_{\N}} \mathcal{L}(\alpha$, $\theta^{\ast})$ 
for a small amount of time. While there exist many efficient techniques to compute high order approximate solutions of ODEs, e.g., Runge-Kutta solvers, 
they typically rely on evaluating the associated vector field on a neighborhood of the initial condition. In our set up, this would correspond
to evaluating the Christoffel symbols at different points on the manifold. While it would be possible to explore nearby points in our chart 
$\Phi = (\text{id}_{\RR^{p - \tilde p}}, \Lambda)$, e.g, by computing a second or higher order Taylor-expansion of $\zeta$, our objective
is not to just simply explore $\N$. Instead, we are only interested in following paths on $\N$ which lead to a decrease in $\mathcal{L}$. 
In particular, we are limited to choosing sufficiently small step-sizes, since we wish to stay on descent directions for 
$\mathcal{L}$. For this reason, since we only need to integrate the geodesic equation for small amounts of time, 
we use a first or second order Taylor-expansion to approximate the solution of \eqref{eq:geodesic_equation}. 

More precisely, let 
$\bm{\gamma} : = \begin{bmatrix} \gamma^{1} & \ldots & \gamma^{p-q} \end{bmatrix}^{T}$ denote the curve in local coordinates, then 
\begin{align*}
	\bm{\gamma}(t_{0} + h) = \Phi(u_{0}) + [V_{0}]_{\Phi} h - \dfrac{1}{2} h^{2} V^{i}_{0} V^{j}_{0} \bm{\Gamma}_{ij}( u_{0} ) + o(h^{2}), \quad
	\bm{\Gamma}_{ij}( u_{0} ) := \begin{bmatrix} \Gamma_{ij}^{1}(u_{0}) \\ \vdots \\ \Gamma_{ij}^{p-q}(u_{0}) \end{bmatrix}
\end{align*}
as $h \rightarrow 0$. For our particular case, we set 
\begin{align*}
	u_{0} = (\alpha, \theta^{\ast}), \quad V_{0} = -\nabla_{g_{\N}} \mathcal{L}\left( \alpha, \theta^{\ast} \right) \simeq 
	- \bm c(\alpha, \theta^{\ast}), \quad  \bm c(\alpha, \theta^{\ast}) := \begin{bmatrix} c^{1}(\alpha, \theta^{\ast}) \\ \ldots  \\ c^{p - q}(\alpha, \theta^{\ast}) \end{bmatrix}, 
\end{align*} 
where $\bm c(\alpha, \theta^{\ast})$ are the components of the gradient relative to $\Phi$.
We define \emph{the second order gradient descent step} with step-size $h$ based at $(\alpha, \theta^{\ast})$ 
for $\mathcal{L}$ by 
\begin{align*}
	\begin{bmatrix}
		\tilde \alpha \\
		\tilde \beta
	\end{bmatrix} = 
	\begin{bmatrix}
		\alpha \\
		\beta^{\ast}
	\end{bmatrix} - 
	\bm{c}(\alpha, \theta^{\ast})h - \dfrac{1}{2} h^{2} c^{i}(\alpha, \theta^{\ast}) c^{j}(\alpha, \theta^{\ast}) \bm{\Gamma}_{ij}(\alpha, \theta^{\ast}). 
\end{align*}
Here $\Phi(\alpha, \theta^{\ast}) = (\alpha, \beta^{\ast})$ is the coordinate representation of $(\alpha, \theta^{\ast})$. 
Similarly, we define \emph{the first order gradient descent step} with step-size $h$ based at $(\alpha, \theta^{\ast})$ by
\begin{align*}
	\begin{bmatrix}
		\tilde \alpha \\
		\tilde \beta
	\end{bmatrix} = 
	\begin{bmatrix}
		\alpha \\
		\beta^{\ast}
	\end{bmatrix} -
	\bm{c}(\alpha, \theta^{\ast})h.
\end{align*}
Note very carefully that the gradient descent steps are taken \emph{in the local coordinate system}. For sufficiently small $h$, we are guaranteed 
that the new point $(\tilde \alpha, \tilde \beta)$ is contained in the current chart for both the first and second order steps. However, to get back to the
manifold, we have to evaluate $\Phi^{-1}(\tilde \alpha, \tilde \beta) = ( \tilde \alpha, \zeta( \tilde \beta) )$. In addition, we also have to 
explicitly evaluate the Christoffel symbols. The computational details are given below.  

\subsubsection{Evaluating the inverse chart}
\begin{figure}[tb]
	\centering
	{{\includegraphics[width=0.75\textwidth]{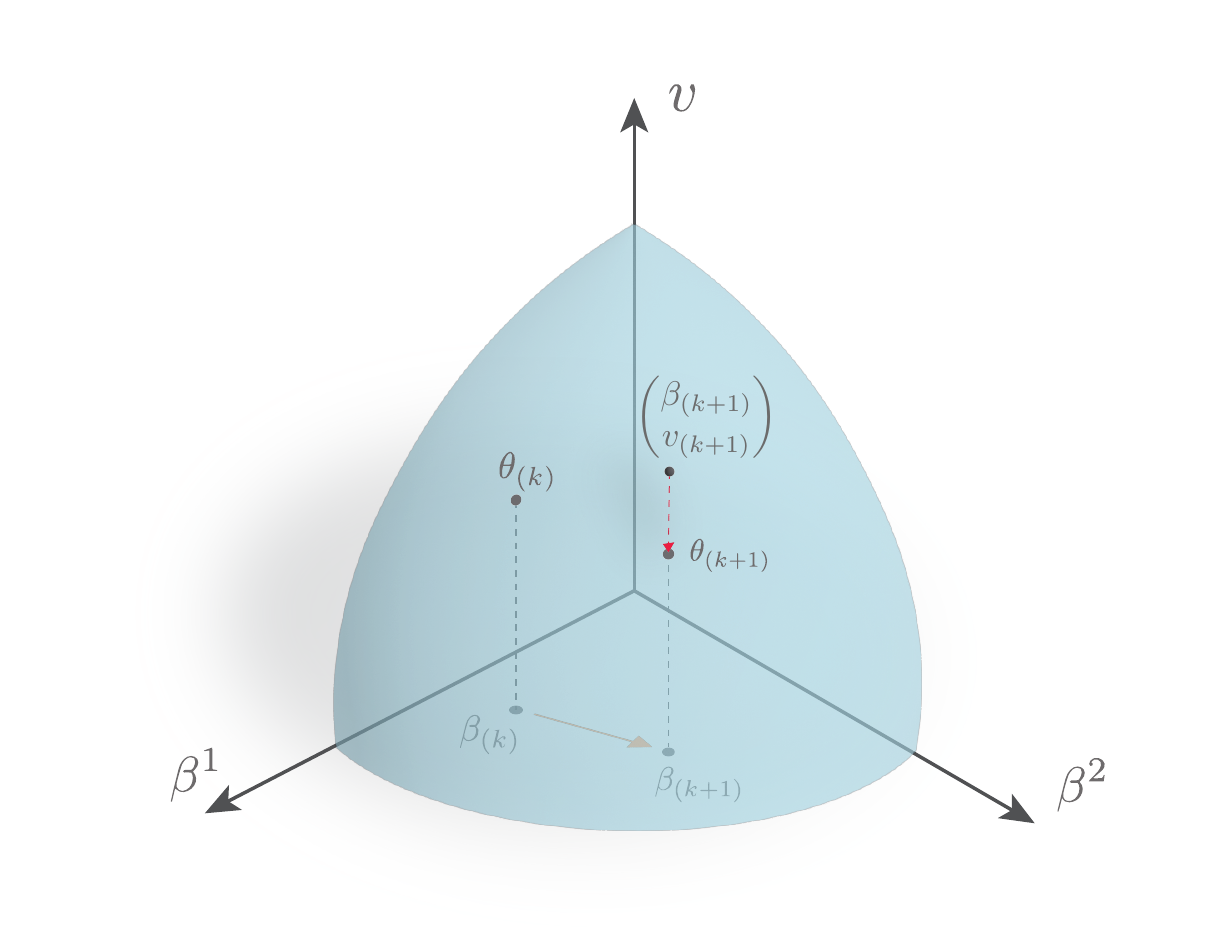}}} \\[1ex]
	\caption{In this figure we visualize the computational steps for performing SGD on $\mathcal{N}$. 
		     We assume for the sake of clarity that there are no unconstrained parameters, i.e., $\N = \M$. We start at a previously
		      computed point $\theta_{(k)} \in \M$ with associated coordinates $\beta_{(k)}$ relative to $\Lambda$. 
		      We remind the reader that the inverse of $\Lambda$ embeds a patch of $\M$ into $\RR^{\tilde p}$ as the graph of $\tilde \zeta$. 
		      Next, we perform a gradient descent step by following 
		      the first or second order Taylor expansion of the geodesic (depicted in orange) starting at $\beta_{(k)}$ in the direction of 
		      $- \nabla_{g_{\M}} \mathcal{L} \left ( \theta_{(k)} \right)$ for a small amount of time. This yields the next point $\beta_{(k+1)}$,
              which is still contained in the chart. 
		      Finally, we evaluate the inverse chart $\zeta$ at the new point in two steps. First, we approximate 
		      $\tilde \zeta\left( \beta_{(k+1)} \right) \approx v_{(k+1)}$ using a first or second order Taylor expansion of $\tilde \zeta$, see
		       \eqref{eq:initial_guess_newton}. 
		      We then use Newton's method to refine this approximation and compute $\theta_{(k+1)} =  \zeta\left( \beta_{(k+1)} \right)$. 
		     \label{fig:newton_geodesic}
	}
\end{figure}
We will use a Taylor expansion to evaluate the inverse chart $\zeta$ on $\M$ at $\tilde \beta$ . 
Subsequently, we use Newton's method to refine the approximation. The resulting point that we find 
must necessarily correspond to $\zeta (\tilde \beta)$, and is thus completely determined by $\tilde \beta$, since 
$\zeta$ is locally unique as explained in Theorem \ref{theorem:preimage}. This justifies the claim made
in \autoref{sec:lagrange_multipliers} that the search dynamics of our algorithm is completely determined 
by the negative gradient flow of $\mathcal{L}$, since $\tilde \beta$ is. 

Below we provide the computational details for the case of a second order Taylor expansion; the first order case is 
obtained by ignoring the second order terms. To avoid clutter in the notation, we will henceforth (interchangeably) write 
\begin{align*}
	\begin{bmatrix}
		\alpha_{(k+1)} \\
		\beta_{(k+1)}
	\end{bmatrix} =	
	\begin{bmatrix}
		\tilde \alpha \\
		\tilde \beta
	\end{bmatrix}, \quad
	\begin{bmatrix}
		\alpha_{(k)} \\
		\beta_{(k)}
	\end{bmatrix} =	
	\begin{bmatrix}
		\alpha \\
		\beta^{\ast}
	\end{bmatrix},
	\quad 
	\theta_{(k+1)} = \zeta \left( \beta_{(k+1)} \right), \quad
	\theta_{(k)} = \zeta \left( \beta_{(k)} \right).
\end{align*}
This notation also emphasizes that we move from a given point at step $k \in \NN_{0}$ to a next point.

The second order Taylor expansion of $\tilde \zeta$ around $\beta_{(k)}$ is given by 
\begin{align*}
	\tilde \zeta \left ( \beta_{(k+1)} \right) = 
	\tilde \zeta \left( \beta_{(k)} \right) 
	+ D \tilde \zeta \left( \beta_{(k)} \right) d_{k}
	+ \frac{1}{2} D^{2} \tilde \zeta \left( \beta_{(k)} \right) [d_{k}, d_{k}] + 
	o \left( \left \Vert d_{k} \right \Vert^{2}_{2} \right), \quad
	d_{k} := \beta_{(k+1)} - \beta_{(k)}
\end{align*}
as $\beta_{(k+1)} \rightarrow \beta_{(k)}$. We have explained in Section \ref{sec:riemann_metric} how to 
explicitly compute $D \tilde \zeta \left( \beta_{(k)} \right)$, which was needed to evaluate the Riemannian metric.  
Here we employ the same strategy to compute the second derivative 
$D^{2} \tilde \zeta \left(\beta_{(k)} \right) \in \B^{2}( \RR^{\tilde p - q}, \RR^{q} )$, where $\B^{2}( \RR^{\tilde p - q}, \RR^{q} )$ 
denotes the space of $\RR^{q}$-valued $2 \choose 0$-tensors on $\RR^{\tilde p - q}$. We start by 
rewriting \eqref{eq:Dzeta} as
\begin{align*}
	DF \left( \tilde \zeta(\beta), \beta  \right)  \begin{bmatrix} D \tilde \zeta( \beta) \\ \bm{I}_{\RR^{\tilde p -q}} \end{bmatrix} = 0, \quad \beta \in B.
\end{align*}
Next, we differentiate both sides with respect to $\beta$ and evaluate at $\beta_{(k)}$. This yields 
\begin{align}
	\label{eq:D^2F}
	D_{v}F \left( \theta_{(k)} \right) D^{2} \tilde \zeta \left( \beta_{(k)} \right)[s_{1}, s_{2}] = 
	- D^{2}F \left( \theta_{(k)}  \right) \left[ 
		 \begin{pmatrix} D \tilde \zeta \left( \beta_{(k)} \right)s_{1} \\ s_{1} \end{pmatrix}, 
		  \begin{pmatrix} D \tilde \zeta \left( \beta_{(k)} \right)s_{2} \\ s_{2} \end{pmatrix}
		 \right]
\end{align}
for all $s_{1}, s_{2} \in \RR^{\tilde p -q}$. To compute the $(i,j)^{\text{th}}$ component of
$D^{2} \tilde \zeta \left( \beta_{(k)} \right)$ with respect to the standard basis, i.e., in order to compute 
$\frac{ \partial ^{2} \tilde \zeta }{ \partial \beta^{i} \partial \beta^{j} } \left(\beta_{(k)} \right)$, we evaluate both 
sides of \eqref{eq:D^2F} at $(s_{1}, s_{2}) = (e_{i}, e_{j})$ and solve the equation
\begin{align}
	\label{eq:D^2zeta}
	D_{v}F \left( \theta_{(k)} \right) \frac{ \partial ^{2} \tilde \zeta }{ \partial \beta^{i} \partial \beta^{j} }\left(\beta_{(k)} \right) = 
	- D^{2}F \left( \theta_{(k)}  \right) \left[ 
		 \begin{pmatrix} \dfrac{ \partial \tilde \zeta }{ \partial \beta^{i} } \left(\beta_{(k)} \right) \\ e_{i} \end{pmatrix}, 
		  \begin{pmatrix} \dfrac{ \partial \tilde \zeta }{ \partial \beta^{j} } \left(\beta_{(k)} \right) \\ e_{j} \end{pmatrix} 
		 \right], 
\end{align}
for each $1 \leq i, j \leq \tilde p -q$. This equation admits a unique solution, since $D_{v}F \left( \theta_{(k)} \right)$ 
is an isomorphism on $\RR^{q}$. 

Finally, we approximate $\tilde \zeta \left( \beta_{(k+1)} \right)$ using its second (or first) order Taylor expansion and then
use Newton's method to evaluate
\begin{align*}
	\Phi^{-1} \left( \alpha_{(k+1)}, \beta_{(k+1)} \right) = 
	\left (\alpha_{(k+1)}, \zeta \left( \beta_{(k+1)} \right) \right).
\end{align*}
More precisely, we first approximate $\zeta \left ( \beta_{(k+1)} \right)$ by
\begin{align}
	\label{eq:initial_guess_newton}
	\zeta \left ( \beta_{(k+1)} \right) \approx 
	\begin{bmatrix}
		v_{(k+1)} \\
		\beta_{(k+1)}
	\end{bmatrix}, 
	\quad 
	v_{(k+1)} := \tilde \zeta \left( \beta_{(k)} \right) 
	+ D \tilde \zeta \left( \beta_{(k)} \right) d_{k}
	+ \frac{1}{2} D^{2} \tilde \zeta \left( \beta_{(k)} \right) [d_{k}, d_{k}]. 	
\end{align}
We then refine this approximation by finding a zero of the map $v \mapsto F \left (v, \beta_{(k+1)} \right)$ using Newton's method
and $v_{(k+1)}$ as initial guess. 
In particular, we solve the equation for $v$, while $\beta_{(k+1)}$ \emph{remains fixed}. 
The zero that we find must necessarily correspond to $\zeta \left ( \beta_{(k+1)} \right)$, since $\zeta$ is locally unique as
explained in Theorem \ref{theorem:preimage}. Altogether, this yields the desired point $\left( \alpha_{(k+1)}, \theta_{(k+1)} \right) \in \N$.
See Figure \ref{fig:newton_geodesic} for a visualization of the steps described in this section. 

\subsubsection{Evaluating the Christoffel symbols} 
We end this section by explaining how to explicitly evaluate the Christoffel symbols $\Gamma^{k}_{ij}$
at $(\alpha, \theta^{\ast})$. Recall that a connection is locally completely characterized by the Christoffel
symbols. The constraints that uniquely determine the Levi-Civita connection, i.e., metric compatibility 
and torsion-freeness, therefore also impose 
constraints on the Christoffel symbols. In fact, the standard proof for the existence of the Levi-Civita 
connection is constructive and establishes an explicit relationship between the Christoffel symbols and
the Riemannian metric:
\begin{align*}
	\Gamma^{k}_{ij} = \frac{1}{2} (g_{\N})^{kl} 
	\left( 
		\frac{ \partial (g_{\N})_{jl} }{ \partial u^{i} } + 
		\frac{ \partial (g_{\N})_{il} }{ \partial u^{j} } - 
		\frac{ \partial (g_{\N})_{ij} }{ \partial u^{l} }
	\right), 
	\quad 1 \leq i, j, k \leq p - q,
\end{align*}
see \cite{lee2006riemannian, lee2013smooth} for instance. We will use this expression to numerically
evaluate the Christoffel symbols. 

It follows immediately from the block structure of the metric $g_{\N}$ 
in \eqref{eq:product_metric} that
 \begin{align*}
 	 \Gamma^{k}_{ij}(\alpha, \theta^{\ast}) &= 0, \quad 1 \leq i \leq p - \tilde p, \ 1 \leq j \leq p - q, \\[2ex]
	 \Gamma^{k}_{ij}(\alpha, \theta^{\ast}) &= 0, \quad p - \tilde p + 1 \leq i \leq p - q, \ 1 \leq j \leq p - \tilde p,
\end{align*}
for all $1 \leq k \leq p - q$. The reason why these coefficients are zero is because there is no interplay
between the submanifolds $\RR^{p - \tilde p}$ and $\M$, which together make up $\N$, and because
the metric on $\RR^{p - \tilde p}$ is flat. In particular, this shows that the component in $\RR^{p - \tilde p}$
of a geodesic on $\N$ is just a straight line as expected. 

It remains to consider the case $p - \tilde p + 1 \leq i, j \leq p - q$, which is associated to the non-trivial 
metric $g_{\M}$ on $\M$. We use the expression in \eqref{eq:metric_eval} to compute the partial derivatives 
of the relevant components of $g_{\M}$. More precisely, observe that 
\begin{align*}
	\dfrac{ \partial \left( g_{\M} \right)_{ij} }{ \partial \lambda^{l} }( \theta^{\ast} ) 
	 &= \frac{ \partial }{ \partial \beta^{l} } \bigg \vert_{ \beta^{\ast} }
	 \left( \beta \mapsto \left \langle \frac{ \partial \zeta }{ \partial \beta^{i} }( \beta),  \frac{ \partial \zeta }{ \partial \beta^{j} }( \beta) \right \rangle \right)
	 \\[2ex] 
	 & = 
	 \left \langle \frac{ \partial \tilde \zeta }{ \partial \beta^{i} }( \beta^{\ast} ), \frac{ \partial^{2} \tilde \zeta }{ \partial \beta^{l} \partial \beta^{j}} (\beta^{\ast}) \right \rangle + 
	  \left \langle \frac{ \partial \tilde \zeta }{ \partial \beta^{j} }( \beta^{\ast} ), \frac{ \partial^{2} \tilde \zeta }{ \partial \beta^{l} \partial \beta^{i}} (\beta^{\ast}) \right \rangle
\end{align*}
for $1 \leq i, j, l \leq \tilde p - q$. We can evaluate this expression numerically, since we can explicitly evaluate
$D \tilde \zeta( \beta^{\ast})$ and $D^{2} \tilde \zeta( \beta^{\ast})$. Finally, to compute the relevant
Christoffel symbols, we define vectors $\bm{w}_{ij}(\beta^{\ast}) \in \RR^{\tilde p -q}$
for each $1 \leq i, j \leq \tilde p - q$ by 
\begin{align*}
	[\bm{w}_{ij}(\beta^{\ast})]_{l} := \frac{1}{2} \left( 
		\frac{ \partial (g_{\M})_{jl} }{ \partial \lambda^{i} }(\beta^{\ast}) + 
		\frac{ \partial (g_{\M})_{il} }{ \partial \lambda^{j} }(\beta^{\ast}) - 
		\frac{ \partial (g_{\M})_{ij} }{ \partial \lambda^{l} }(\beta^{\ast})
	\right), \quad 1 \leq l \leq \tilde p - q.
\end{align*}
The remaining (non-zero) Christoffel symbols associated to $\M$ can now be computed by solving the following linear system of equations: 
\begin{align*}
	[g_{\M}(\theta^{\ast})]_{\Lambda} [\Gamma^{k}_{\tilde i \tilde j}(\alpha, \theta^{\ast})]_{k=1}^{\tilde p -q} =  \bm{w}_{ij}(\beta^{\ast}), \quad
	\tilde i = i + p - \tilde p, \ \tilde j = j + p - \tilde p. 
\end{align*}

\section{Multiresolution Analysis and CERM}\label{sec:wavelet}
In this section we present a non-trivial application of the CERM framework to learn optimal wavelet bases for a given task. 
Specifically, we explain how to set up a system of equations (constraints) whose solution set corresponds to wavelets. 
To set up appropriate constraints, we first review the needed theory from Multiresolution Analysis (MRA) \cite{Mallat,HarmonicAnalysis,WaveletsTheory}.
Multiresolution analysis provides a natural framework for defining and analyzing wavelets. Moreover, it can be used to 
characterize a large class of finitely supported wavelets as solutions of a finite system of equations. We review in detail how 
to derive these equations and how to efficiently compute wavelet decompositions using Mallat's Pyramid Algorithm \cite{Mallat}, 
which together form the backbone of our main example in Section \ref{sec:contours}, where we train networks for predicting wavelet 
decompositions of contours in the medical domain. 
Before we continue, however, we briefly discuss examples of tasks where wavelets arise naturally. 

\paragraph{Applications of MRAs} 
There are several tasks at which one expects wavelet-based neural networks to excel. 
Wavelet decompositions naturally lend themselves to representing continuous objects 
such as curves, images, vector fields, or other higher-dimensional objects. Hence any task 
where the object of interest can be identified with a smooth or continuous function is  well-suited for wavelet-based neural networks. There is an abundance of 
such examples to be found in computer vision, e.g., boundary prediction, image registration, and so forth.
Another family of interesting applications can be found in signal analysis, e.g., in compression and denoising, where wavelets 
are long-standing tools that have proven to be extremely efficient \cite{Mallat}. The main idea in these areas is to extract information 
about noise, smoothness, and even singularities, through analysis of the wavelet coefficients. Subsequently, by modifying a subset of the coefficients, 
e.g., through thresholding, the signal can be ``cleaned up'' or denoised.

In this paper, we consider one-dimensional wavelets only, which will be applied to boundary prediction of simply-connected two-dimensional domains
in \autoref{sec:contours}. The wavelet framework, however, is easily adapted to higher-dimensional domains, such as images, by using tensor products of 
the one-dimensional bases.

\subsection{Multiresolution Analysis}
In this section we briefly review what Multiresolution Analyses (MRA) are, how wavelets come into play, 
and why they are useful. We closely follow the exposition in  \cite{WaveletsTheory, HarmonicAnalysis} and
refer the reader to these references for a more comprehensive introduction. 

\begin{figure}[!tb]
	\centering
	\subfloat[\centering Father wavelet \label{fig:scaling_function}]{{\includegraphics[width=0.4\textwidth]{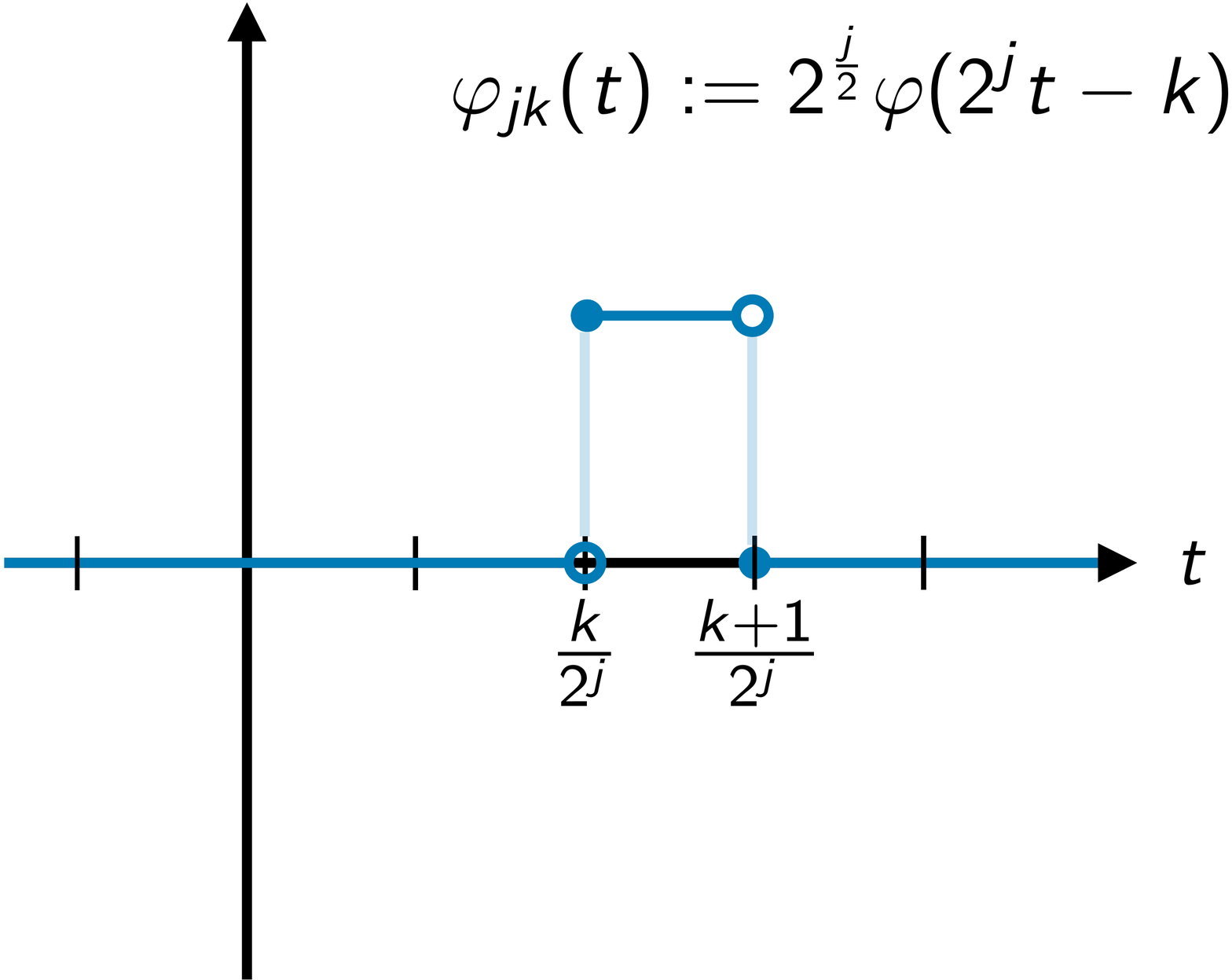}}} \quad
	\subfloat[\centering $V_{j}$ \label{fig:approx_subspace}]{{\includegraphics[width=0.4\textwidth]{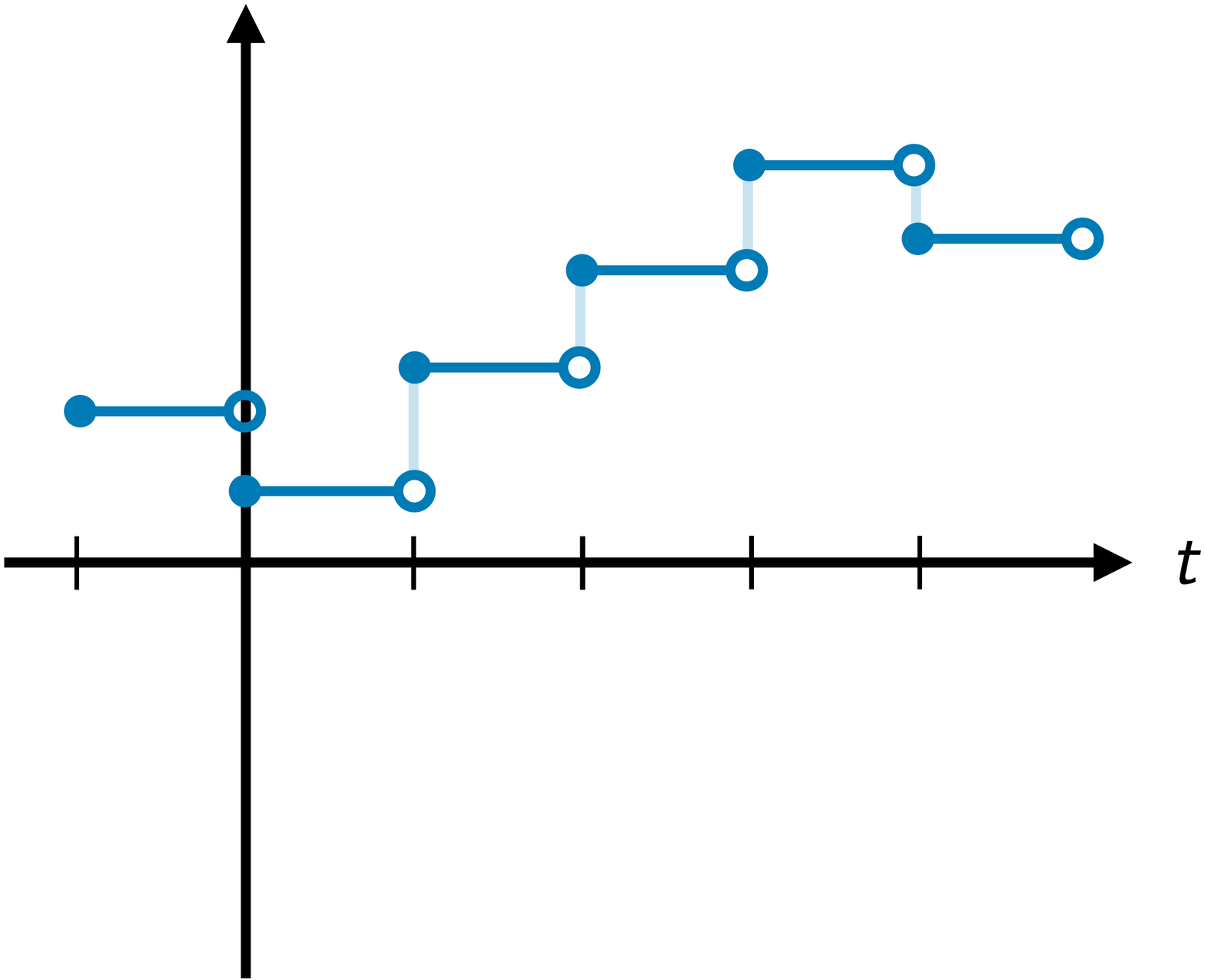}}} \quad \\[2ex]
	\subfloat[\centering Mother wavelet \label{fig:mother_wavelet}]{{\includegraphics[width=0.4\textwidth]{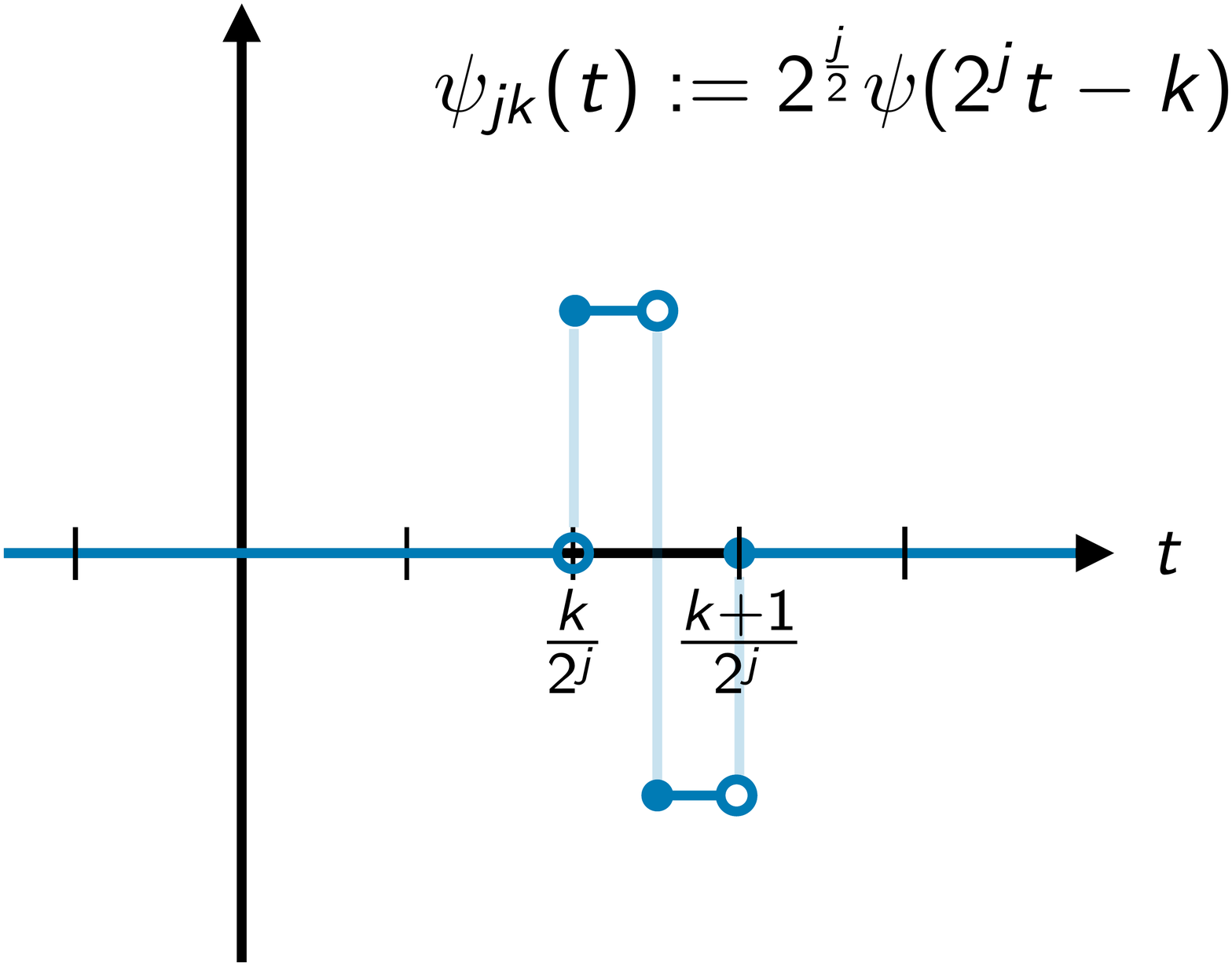}}} \quad
	\subfloat[\centering $W_{j}$ \label{fig: detail_subspace}]{{\includegraphics[width=0.4\textwidth]{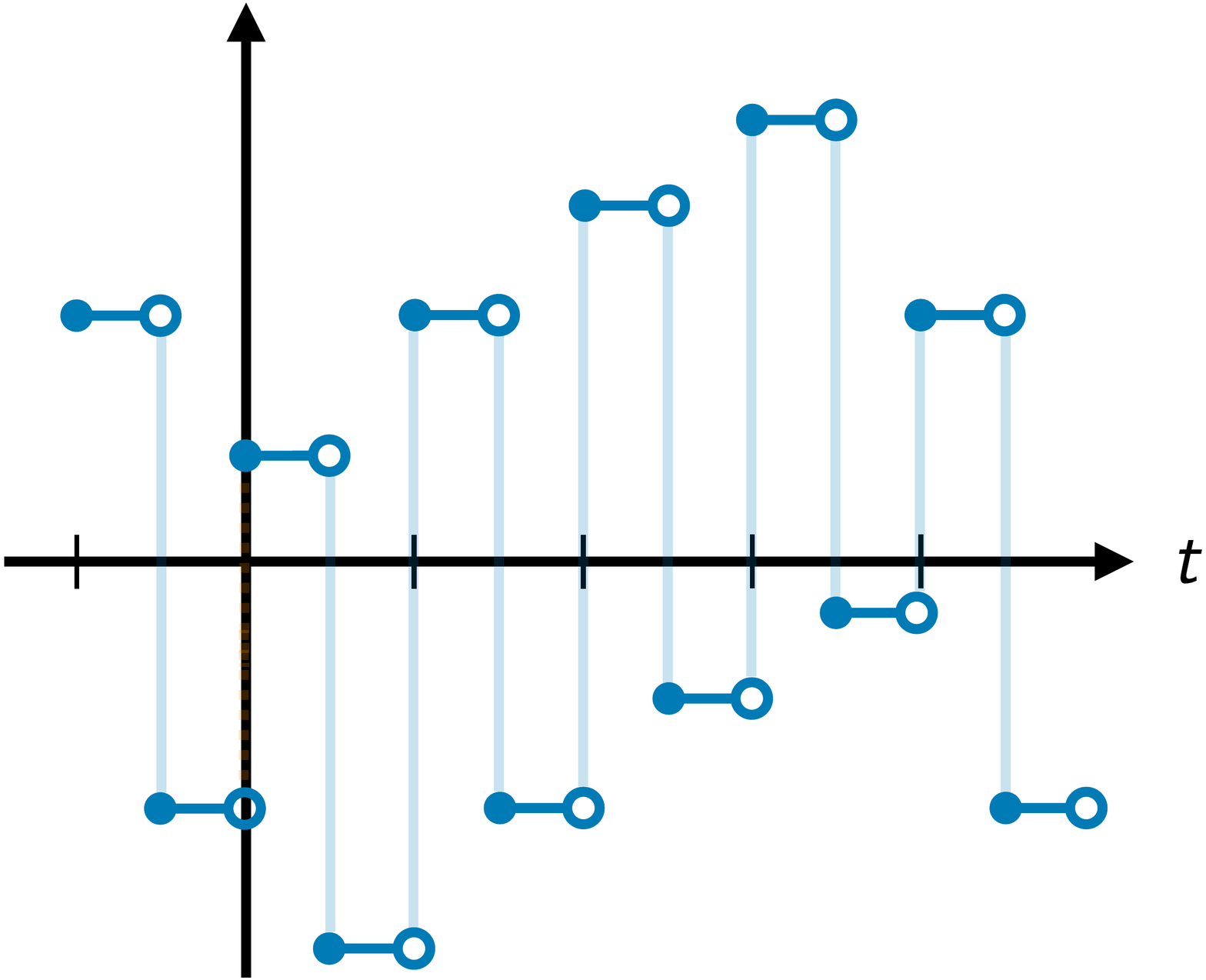}}} 
	\caption{Example of the Haar MRA: ~\protect \subref{fig:scaling_function} Dilated translation of the Haar scaling map $\varphi = \bm{1}_{[0, 1)}$.
			~\protect \subref{fig:approx_subspace} The approximation subspace at level $j$ consists of all step-functions with step-size $2^{-j}$.
			~\protect \subref{fig:mother_wavelet} Dilated translation of the mother wavelet
			$\psi = \bm{1}_{[0, \frac{1}{2})} - \bm{1}_{[\frac{1}{2}, 1)} $. 
			~\protect \subref{fig: detail_subspace} Example of a function in the detail subspace at level $j$.
			}
	\label{fig:MRA}
\end{figure}

The uncertainty principle in Fourier analysis states that a signal $\gamma \in \Ltwo$ cannot be simultaneously localized 
in the time and frequency domain. Multiresolution analysis aims to address this shortcoming by decomposing
a signal on different \emph{discrete} resolution levels. The idea is to construct subspaces 
$V_{j} \subset \Ltwo$, associated to various resolution levels $j \in \ZZ$, spanned by integer 
shifts of a localized mapping $\varphi_{j}$. The level of localization associated to $V_{j}$ is
determined by taking an appropriate dilation of a prescribed map $\varphi$; the so-called
\emph{scaling function}. In the MRA framework the dilation factors are chosen to be powers of
two. Formally, we require that $(\varphi_{jk})_{k \in \ZZ}$ is an orthonormal basis for $V_{j}$, 
where $\varphi_{jk}(t) := 2^{\frac{j}{2}}\varphi(2^{j}t - k)$, see Figures \ref{fig:scaling_function} 
and \ref{fig:approx_subspace}. Altogether, this yields an increasing sequence of closed subspaces 
$V_{j} \subset V_{j+1} \subset \Ltwo$ dense in $\Ltwo$, where $V_{j+1}$ 
is the next level up in resolution after $V_{j}$.  For the sake of completeness, we provide the formal 
definition of a MRA below. 

\begin{definition}[Formal definition MRA \cite{Mallat}]
\label{def:MRA}
Let $T_{k}: \Ltwo \rightarrow \Ltwo$ and $\DD_{j}: \Ltwo \rightarrow \Ltwo$ 
denote the translation and normalized dilation operator, respectively, defined by 
$T_{k}\gamma (t) = \gamma(t - k)$ and $\DD_{j} \gamma (t) = 2^{\frac{j}{2}} \gamma( 2^{j} t )$
for $\gamma \in \Ltwo \cap C^{\infty}_{0}(\RR)$ and $j, k \in \ZZ$. 
A multiresolution analysis of $\Ltwo$ is an increasing sequence of subspaces $(V_{j})_{j \in \ZZ}$, such that 
\begin{enumerate}[\itshape(i)]
	\item $\bigcap_{j \in \ZZ} V_{j} = \{ 0 \}$,
	\item $\bigcup_{j \in \ZZ} V_{j}$ is dense in $\Ltwo$,
	\item $\gamma \in V_{j}$ if and only if $\DD_{1} \gamma \in V_{j+1}$,
	\item $V_{0}$ is invariant under translations,
	\item $\exists \varphi \in \Ltwo$ such that $\{ T_{k}\varphi \}_{k \in \ZZ}$ is an orthonormal basis for $V_{0}$. 
\end{enumerate}
\end{definition}
Condition $(ii)$ formalizes the idea that any signal in $\Ltwo$ can be arbitrarily well approximated 
using an appropriate resolution level. Condition $(iii)$ encapsulates the idea that $V_{j+1}$ is the next resolution 
level with respect to our choice of dilation operators $\DD_{j}$, i.e., there are no other resolution levels between $V_{j}$ and $V_{j+1}$.  Combined 
with $(iv)$ it implies that each subspace $V_{j}$ is invariant under integer shifts. Finally, condition $(v)$ formalizes 
the idea that the subspaces are spanned by translations and dilations of the map $\varphi$; the so-called 
\emph{scaling function} or \emph{father wavelet}. Indeed, it is straightforward to show that $\{ \varphi_{jk} : k \in \ZZ \}$ 
is an orthonormal basis for $V_{j}$, where $\varphi_{jk} := \DD_{j}T_{k} \varphi$.

\paragraph{Decomposing a signal}
Next, we explain how the MRA framework can be used to analyze a signal $\gamma \in \Ltwo$. The main idea is to 
approximate $\gamma$ at different resolution levels by projecting it onto the subspaces $V_{j}$. 
More precisely, we define the approximation of $\gamma$ at resolution level $j$ by $\gamma_{j} := P_{j} (\gamma)$, 
where $P_{j}: \Ltwo \rightarrow V_{j}$ is the orthogonal projection onto $V_{j}$ (see Figure \ref{fig:approx_subspace}). 
The coefficients of $\gamma_{j}$ with respect to the basis $(\varphi_{jk})_{k \in \ZZ}$ for $V_{j}$, 
denoted by $a_{j}(\gamma) = (a_{jk}(\gamma) )_{k \in \ZZ} \in \elltwo$, are called the \emph{approximation 
coefficients} of $\gamma$ at level $j$.

To study the information that is lost when a signal in $V_{j+1}$ is projected onto $V_{j}$, we consider the 
operator $Q_{j}: = P_{j+1} - P_{j}$.
The range of $Q_j$ is denoted by $W_j$ and referred to as the the \emph{detail} subspace at level $j$
(see Figure \ref{fig: detail_subspace}). The subspace $W_j$ is the orthogonal complement of 
$V_{j}$ in $V_{j+1}$. 
The detail subspaces $( W_{j})_{j \in \ZZ}$ are mutually disjoint and orthogonal by construction. Furthermore, since $V_{j} = V_{j-1} \oplus W_{j-1}$ for any $j \in \ZZ$, it follows that
\begin{equation}
	\label{eq:direct_sum_details}
    	V_{j} = V_{j_{0}} \oplus \bigoplus_{l=j_{0}}^{j-1} W_{l} \ , \qquad \forall j > j_0.
\end{equation}
This decomposition shows that a signal on resolution level $j$ can be reconstructed
from any lower level $j_{0}$ if all the details in between are known.

A fundamental result, known as Mallat's Theorem, states that the subspaces $W_{j}$ can also be spanned by 
dilating and shifting a single map. More precisely, there exists a map $\psi \in W_{0}$, the so-called 
\emph{mother wavelet}, such that $( \psi_{jk} )_{k \in \ZZ} $ is an orthonormal basis for $W_{j}$, 
see \cite{HarmonicAnalysis}. Here we have used the notation $\psi_{jk} := \DD_{j}T_{k}\psi$ as before. 
The coefficients of $Q_{j} (\gamma)$ with respect to the basis for $W_{j}$, denoted by 
$d_{j}(\gamma) := (d_{jk}(\gamma))_{k \in \ZZ} \in \elltwo$, are referred to as the \emph{detail coefficients} of $\gamma$ at resolution level $j$.
The detail coefficients store the information needed to go back one level up in resolution. 
\begin{remark}
	We will frequently omit the dependence of the approximation and detail coefficients 
	on the underlying signal $\gamma$, i.e., write $a_{j}(\gamma) = a_{j}$ and 
	$d_{j}(\gamma) = d_{j}$, whenever there is no chance of confusion. 
\end{remark}

In general, given approximation
coefficients $a_{j_{0}} \in \ell^{2}(\ZZ)$ at level $j_{0}$ and detail coefficients $d_{l} \in \ell^{2}(\ZZ)$ at
levels $j_{0} \leq l \leq j-1$, we can reconstruct the approximation at level $j$ using \eqref{eq:direct_sum_details}:
\begin{equation*}
    	\gamma_{j} = \sum_{k \in \ZZ} a_{j_{0}k} \varphi_{j_{0}k} + \sum_{j_{0} \leq l \leq j - 1} \sum_{k \in \ZZ} d_{lk} \psi_{lk}. 
\end{equation*}
Altogether, these observations give rise to the following terminology: 
\begin{definition}[Multiresolution decomposition of a signal]
    Let $j_{0} < j_{1}$ be resolution levels. A finite $(j_{0}, j_{1})$-multiresolution decomposition 
    of a signal $\gamma \in \Ltwo$ is the sequence 
	\begin{align*}
		\left( a_{j_{0}}(\gamma), d_{j_{0}}(\gamma), \ldots, d_{j_{1}-1}(\gamma) \right).
	\end{align*}
\end{definition}

\subsection{The scaling equation}
\label{sec: scaling_equation}
In this section we review the so-called scaling equation, which is key for understanding 
many fundamental aspects of MRAs, both theoretical and computational. We will heavily
rely on it in the subsequent sections to set up the desired constraints and to efficiently 
compute with wavelets. The key observation is that since $V_{0} \subset V_{1}$, there exists a 
unique sequence $h \in \elltwo$ such that 
\begin{equation}
	\label{eq:scaling_eqn}
	\varphi = \sum_{k \in \ZZ} h_{k} \varphi_{1k}  .
\end{equation}
This equation is referred to as the \emph{scaling equation}; one of the fundamental properties of a scaling function. 

\paragraph{Low and high pass filters}
The sequence $h$ is called the low-pass filter of the MRA. It completely characterizes the scaling function and therefore
also the corresponding MRA. We will often refer to $h$ as simply a \emph{wavelet filter}. 
Similarly, since $\psi \in W_{0} \subset V_{1}$, 
there exists a unique sequence $g \in \elltwo$, the so-called \emph{high-pass filter} associated to $h$, 
such that 
\begin{equation}
	\label{eq:scaling_mother_wavelet}
    \psi = \sum_{k \in \ZZ} g_{k} \varphi_{1k}. 
\end{equation}
For Mallat's mother wavelet, we have $g_{k} = (-1)^{k-1} h_{1-k}$. In practice, to define a MRA, one 
only needs to specify an ``appropriate'' low-pass filter $h$. In \autoref{sec:wavelets_constraints}
we derive a finite set of equations whose solutions correspond to low-pass filters, provided a mild
non-degeneracy condition is satisfied, and characterize a finite-dimensional family of compactly 
supported wavelets. 

\begin{example}[Haar MRA]
	A simple example of a MRA is the so-called Haar MRA; the father wavelet is given by $\varphi=\bm{1}_{[0, 1)}$, 
	and the mother wavelet by $\psi = \bm{1}_{[0, \frac{1}{2})} - \bm{1}_{[\frac{1}{2}, 1)}$. They are visualized  in 
	Figure \ref{fig:scaling_function} and Figure \ref{fig:mother_wavelet}, respectively. The associated low and high
	pass filters are given by 
	\begin{align*}
		h_{k} = 
		\begin{cases}
			\dfrac{1}{\sqrt{2}}, & k \in \{0, 1\} \\[2ex]
			0, & \mbox{otherwise}, 
		\end{cases}, \quad
		g_{k} = 
		\begin{cases}
			\dfrac{(-1)^{k-1}}{\sqrt{2}}, & k \in \{0, 1\}, \\[2ex]
			0, & \mbox{otherwise},
		\end{cases}		
	\end{align*}
	respectively. 
\end{example}

\paragraph{The refinement mask}
An important observation follows from taking the Fourier Transform of the scaling equation, which yields 
\begin{align}
	\label{eq:scaling_equation_fourier}
	\hat \varphi(\xi) = H\left(\frac{\xi}{2} \right) \hat \varphi \left( \frac{\xi}{2} \right), \quad
	H(\xi) :=  \frac{1}{\sqrt{2}} \sum_{k \in \ZZ} h_{k} e^{-2 \pi i \xi k}.
\end{align}
Here $H: [0,1] \rightarrow \CC$ is a 1-period map typically referred to as the \emph{refinement mask}.
Throughout this paper, we shall abuse terminology and frequently refer to both $H$ and $h$ as the low-pass 
filter associated to $\varphi$. Both the low-pass filter and refinement mask completely characterize
the scaling function. The relation in \eqref{eq:scaling_equation_fourier} 
will be used extensively in \autoref{sec:wavelets_constraints} to derive constraints on admissible filters.

\paragraph{Existence and uniqueness of MRAs}
The scaling equation plays a seminal role in establishing the existence and uniqueness of an MRA given a candidate 
$h$ for a low-pass filter. While there is no need to explicitly construct $\varphi$, we do briefly discuss its existence here
to justify the claim that we are learning wavelets. In addition, the discussion will reveal a necessary condition on $H$. 
The idea for proving the existence of a scaling map $\varphi$, given a low-pass filter $h$, is to ``reconstruct'' its Fourier transform 
$\hat \varphi$ using the scaling equation. To see how, suppose we start with a scaling map $\varphi$. Then repeated application of \eqref{eq:scaling_equation_fourier} yields
\begin{align*}
	\hat \varphi(\xi) = \hat \varphi \left( \frac{\xi}{2^k} \right) \prod_{j=1}^{k} H \left(  \frac{\xi}{2^j} \right), \quad \xi \in \RR.
\end{align*}
Assuming that $\hat \varphi$ is continuous at $\xi=0$, we may consider the limit as $k \rightarrow \infty$, which yields
\begin{equation}
	\label{eq:infinite_product_scaling}
    	\hat \varphi(\xi) = \hat \varphi(0) \prod_{j=1}^{\infty}  H \left(  \frac{\xi}{2^j} \right),
\end{equation}
provided the latter product exists. Since $\hat \varphi$ is not identically zero, we must have that $\hat \varphi(0) \not = 0$. 
This imposes a constraint on $H$, namely $H(0) = 1$. Without loss of generality, we may further assume that $\hat \varphi(0) = 1$.

Conversely, if we start with a sequence $h$ instead of a scaling map $\varphi$, we may try to use the right-hand side of 
\eqref{eq:infinite_product_scaling} to define a candidate for $\hat \varphi$. More precisely, if the infinite product converges to a map in $\Ltwo$, one 
may use the inverse Fourier transform to define a corresponding candidate for $\varphi$. As it turns out, if $h$ decays sufficiently fast to zero,
and we assume that $H(0) = 1$, where we now \emph{define} $H$ via \eqref{eq:scaling_equation_fourier}, then 
$\xi \mapsto \prod_{j=1}^{\infty}  H \left(  \frac{\xi}{2^j} \right)$ is in $\Ltwo$, continuous at $\xi = 0$, and satisfies \eqref{eq:infinite_product_scaling}. 
For a more precise statement, we refer the reader to \cite{frazier2001introduction, WaveletsTheory}. In this paper, we exclusively deal with finite sequences $h$, 
for which these assumptions are always (trivially) satisfied. Hence we may use \eqref{eq:infinite_product_scaling} to define a 
\emph{candidate} for a scaling map $\varphi$. However, we still need to impose additional constraints on $h$, to ensure that the translates of
$\varphi$ are orthogonal, see \autoref{sec:wavelets_constraints}. 

\subsection{The Discrete Wavelet Transform}\label{sec:discrete_wavelet_transform}
\begin{figure}[tb!]
	\centering
	\subfloat[\centering Decomposition \label{fig:pyramid_down}]{{\includegraphics[width=0.30\textwidth]{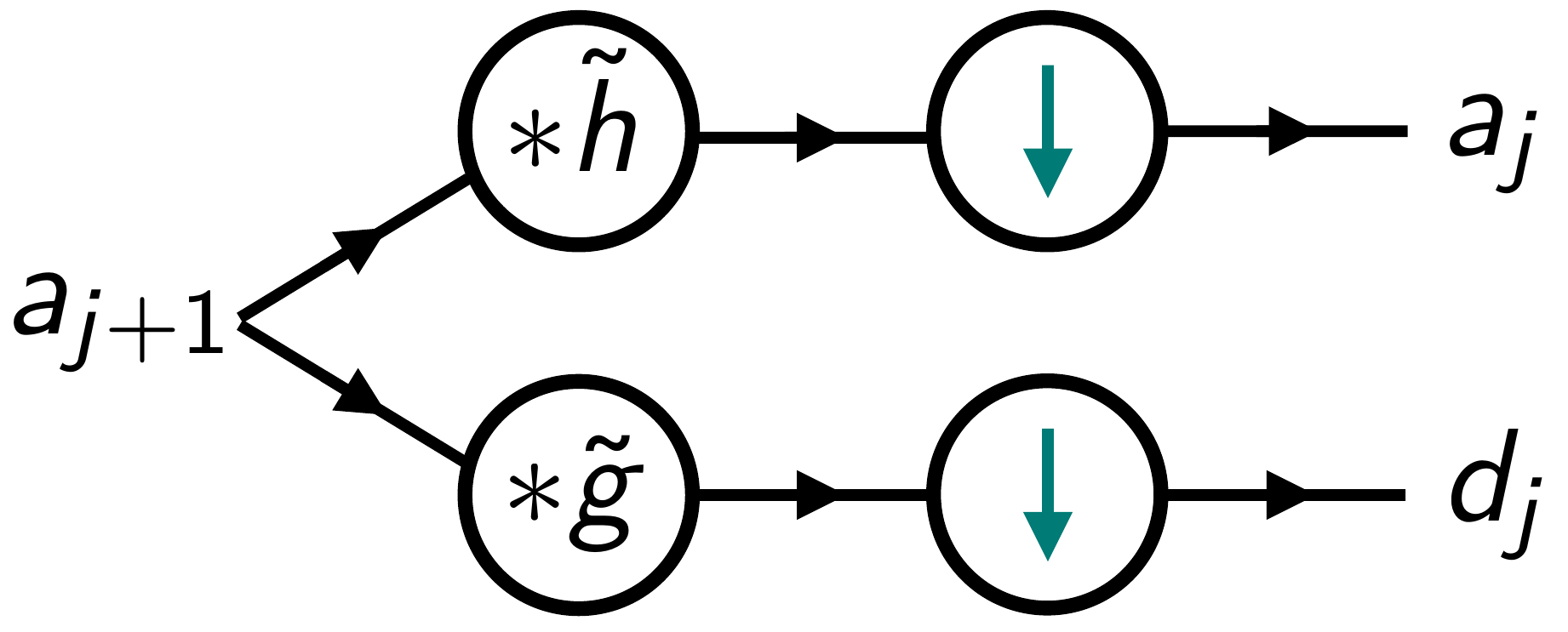}}} \qquad 
	\subfloat[\centering Reconstruction \label{fig:pyramid_up}]{{\includegraphics[width=0.41\textwidth]{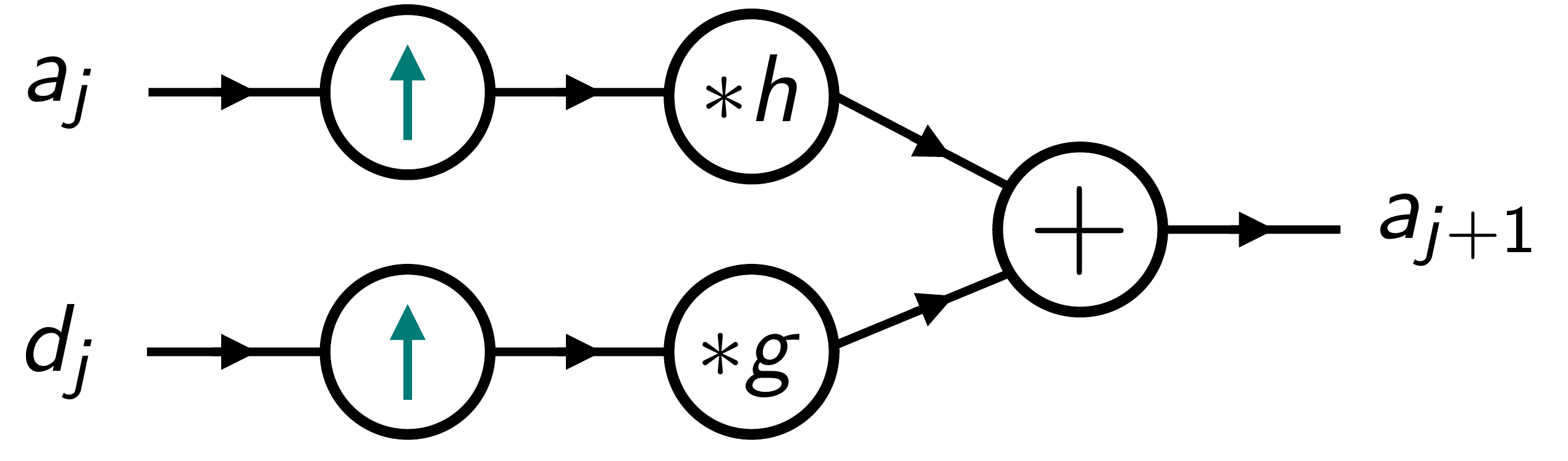}}} \\[1ex]
	\caption{~\protect \subref{fig:pyramid_down} Decomposing approximation coefficients at level $j+1$ into approximation 
	and detail coefficients at level $j$. Here $\tilde h$ and $\tilde g$ are defined in \eqref{eq:decomp_approx} and \eqref{eq:decomp_detail}, respectively, and $\ast$ is the two-sided discrete convolution.
	The symbol $\downarrow$ corresponds to operator $S^{\downarrow}$, which downsamples a sequence by discarding all terms with odd index. 
	~\protect \subref{fig:pyramid_up} Reconstruction of the approximation coefficients at level $j+1$ from the approximation and detail
	coefficients at level $j$. The symbol $\uparrow$ corresponds to operator $S^{\uparrow}$, which upsamples
	a sequence by putting zeros in between every term.}
	\label{fig:pyramid}
\end{figure}

The scaling equation \eqref{eq:scaling_eqn} can be used to derive an efficient scheme for computing a (finite) multiresolution decomposition of a signal $\gamma$. 
More precisely, given initial approximation coefficients $a_{j+1}$ at level $j+1$, the scaling equation can be used to compute the 
approximation and detail coefficients at level $j$. Conversely, the orthogonal decomposition $V_{j+1} = V_{j} \oplus W_{j}$ can be used to 
reconstruct $a_{j+1}$ given the approximation and detail coefficients $a_{j}$ and $d_{j}$, respectively, at resolution level $j$. The mapping
associated to these operations is called the ($1$-level) Discrete Wavelet Transform (DWT). It provides an efficient way to obtain a multiresolution
decomposition of a signal. The associated algorithm, which iteratively applies the $1$-level DWT, is known as the so-called Pyramid Algorithm
\cite{Mallat}. 

\paragraph{Decomposition} 
Let $a_{j+1} \in \elltwo$ be approximation coefficients at an initial resolution level $j+1$, where $j \in \ZZ$. To obtain  
the approximation and detail coefficients at level $j$, we first note that
\begin{equation}
	\label{eq:scaling_equation_father_wavelet}
	\varphi_{jk} = \sum_{l \in \ZZ} h_{l - 2k} \varphi_{j+1, l}, \quad k \in \ZZ. 
\end{equation}
This relation between $\varphi_{j+1}$ and $\varphi_{j}$ can be easily derived by substituting the
right hand side of the scaling equation \eqref{eq:scaling_eqn} into the definition of $\varphi_{jk}$. Consequently, 
\begin{align}
	a_{jk} = \langle \gamma_{j+1}, \varphi_{jk} \rangle = 
	\left( S^{\downarrow} \left( a_{j+1} \ast \tilde h \right) \right)_{k}, 
	\quad \tilde h_{k} := h_{-k},
	 \label{eq:decomp_approx}
\end{align}
where $\ast : \ell^{2}(\ZZ) \times \ell^{2}(\ZZ)  \rightarrow \ell^{2}(\ZZ)$ denotes the two-sided discrete convolution
and $S_{h}^{\downarrow} : \ell^{2}(\ZZ) \rightarrow \ell^{2}(\ZZ)$ is defined by $(S^{\downarrow}(c))_{k} := c_{2k}$. 
The resulting map $a_{j+1} \mapsto S^{\downarrow} \left( a_{j+1} \ast \tilde h \right)$ is typically referred to as the DWT at 
level $j$. An analogous computation for the detail coefficients shows that
\begin{equation}
	\label{eq:decomp_detail}
   	 d_{j} = S^{\downarrow} \left( a_{j+1} \ast \tilde g \right), \quad \tilde g_{k} := g_{-k}.
\end{equation}
The decomposition of the approximation coefficients at level $j+1$ into approximation and detail coefficients 
at level $j$ is illustrated in Figure \ref{fig:pyramid_down}. 

\paragraph{Reconstruction}
The inverse DWT can be derived in a similar fashion using the decomposition $V_{j+1} = V_{j} \oplus W_{j}$. 
To make the computation explicit, we use \eqref{eq:scaling_mother_wavelet} and the scaling equation again to write 
\begin{align*}
	\psi_{jk} = \sum_{l \in \ZZ} g_{l-2k} \varphi_{j+1, l}, \quad k \in \ZZ. 
\end{align*}
Consequently, since $V_{j+1} = V_{j} \oplus W_{j}$, 
\begin{align*}
	\gamma_{j+1} &= \sum_{k \in \ZZ} a_{jk} \varphi_{jk} +  \sum_{k \in \ZZ} d_{jk} \psi_{jk} 
	 =
	\sum_{k, l \in \ZZ} \left( a_{jk}h_{l-2k} + d_{jk}g_{l-2k} \right) \varphi_{j+1, l} 
	\\[2ex]&= 
	\sum_{k \in \ZZ} \left( S^{\uparrow}(a_{j}) \ast h +  S^{\uparrow}(d_{j}) \ast g \right)_{k} \varphi_{j+1, k},
\end{align*}
where $S^{\uparrow}: \elltwo \rightarrow \elltwo$ is defined by 
\begin{align*}
	(S^{\uparrow}c)_{k} :=
	\begin{cases}
	 c_{\frac{k}{2}}, & k \equiv 0 \mod 2, \\
	 0, & k \equiv 1 \mod 2.
	 \end{cases}
\end{align*} 
This shows that the approximations coefficients at level $j+1$ are given by
\begin{equation*}
    a_{j+1} =  S^{\uparrow}(a_{j}) \ast h +  S^{\uparrow}(d_{j}) \ast g.     
\end{equation*}
The reconstruction procedure is schematically shown in Figure \ref{fig:pyramid_up}.

\begin{remark}[Numerical implementation DWT]
	The convolutions appearing in the decomposition and reconstruction formulae can
	be efficiently computed using the Fast Fourier Transform (FFT), see Appendix \ref{appendix:dft}. 
\end{remark}

\subsection{Setting up constraints for wavelet filters}
\label{sec:wavelets_constraints}
In this section we set up a finite system of equations whose zeros, under a mild non-degeneracy condition, correspond to wavelet filters. 
Recall that a wavelet filter is a sequence $h \in \elltwo$ that characterizes a scaling function $\varphi$. 
We reformulate the key requirements on $\varphi$, namely that its translates are orthogonal and $H(0) = 1$, 
in terms of its low-pass filter $h$. In turn, this imposes constraints on admissible filters $h$ in the form of a system of equations. 
Solutions of this system are commonly referred to as Quadratic Mirror Filters (QMFs), see Definition \ref{def:QMF}. We remark that these equations 
and conditions are well-known and refer the reader to \cite{WaveletsTheory, HarmonicAnalysis} for a more comprehensive treatment. 

To reformulate the orthogonality conditions into a system of equations for $h$, we first rewrite the system
$\langle \varphi_{0k}, \varphi_{0l} \rangle =\delta_{kl}$ in frequency space. The recurrence relation for the Fourier 
transform of $\varphi$ in \eqref{eq:scaling_equation_fourier} may then be used to derive a necessary condition on the refinement
mask $H$. Subsequently, we can reformulate this necessary condition as an equivalent condition on $h$.
The details can be found in \cite{HarmonicAnalysis}. Here we only state the relevant results. 
\begin{lemma}[Orthogonality refinement mask]
	\label{lemma:orthogonality_refinement_mask}
	Suppose $\varphi \in \Ltwo$ satisfies the dilation equation for a refinement mask $H$ with
	Fourier coefficients $h \in \elltwo$.
	If the family $(\varphi_{0k})_{k \in \ZZ}$ is orthonormal, then 
	\begin{align}	
		\label{eq:orthogonality_refinement_mask}
		\vert H(\xi/2)\vert^2  + \vert H (\xi /2 + 1/2) \vert^2  = 1,
	\end{align}
	for a.e. $\xi \in \RR^{2}$. 
	\begin{proof}
		See \cite{HarmonicAnalysis}.
	\end{proof}
\end{lemma}
\begin{remark}
	The condition in \eqref{eq:orthogonality_refinement_mask} is often referred to as the Quadratic Mirror Filter condition. 
\end{remark}
\begin{definition}[Quadratic Mirror Filter]
	\label{def:QMF}
	A Quadratic Mirror Filter (QMF) is a sequence $h \in \elltwo$ which satisfies \eqref{eq:orthogonality_refinement_mask} and $H(0) = 1$. 
\end{definition}
The reason for introducing this terminology is that QMFs correspond to wavelet filters under an additional non-degeneracy condition. 
Here we only state the result for finite filters. 
\begin{theorem}
	Suppose $h$ is a finite QMF. If $\inf_{0 \leq \xi \leq \frac{1}{4}} \vert H(\xi) \vert > 0$, then 
	\begin{align*}
		\varphi := \mathcal{F}^{-1} \left( \xi \mapsto \prod_{j=1}^{\infty}  H \left(  \frac{\xi}{2^j} \right) \right)
	\end{align*}
	is a scaling function and defines an MRA of $\Ltwo$. Here $\mathcal{F} : \Ltwo \rightarrow \Ltwo$ denotes the Fourier transform. 
	\begin{proof}
		See \cite{walnut2002introduction} Theorem $8.35$.
	\end{proof}
\end{theorem}
\begin{remark}
	\label{remark:nondegeneracy_condition}
	One may expect that any finite filter $h$ satisfying \eqref{eq:orthogonality_refinement_mask} will define a scaling function
	whose translates are orthogonal. However, this is unfortunately not the case, and the additional requirement that 
	$\inf_{0 \leq \xi \leq \frac{1}{4}} \vert H(\xi) \vert > 0$ is needed to avoid degenerate cases. 
\end{remark}

Next, we derive a system of equations for $h$ that is equivalent to \eqref{eq:orthogonality_refinement_mask}. 
To formulate this system of equations, we define operators $M, R : \elltwo \rightarrow \elltwo$ by 
$(Mc)_{k} := (-1)^{k}c_{k}$ and $(Rc)_{k} := c_{-k}$. 
For brevity, we will frequently write $\tilde c := R(c)$ as before. Even though we are dealing with real-valued filters $h$ in practice, 
below we state the results for general complex-valued sequences.

\begin{lemma}[Orthogonality low-pass filter]
	\label{lemma:orthogonality_low_pass_filter}
	Suppose $H$ is a refinement mask with Fourier coefficients $h \in \elltwo$. Then the orthonormality constraint 
	in \eqref{eq:orthogonality_refinement_mask} is equivalent to the following system of equations:
	\begin{align}
		\label{eq:orthogonality_low_pass_filter}
		\begin{cases}
			\displaystyle \sum_{l \in \ZZ} \vert h_{l} \vert^{2} = 1, & k=0, \\[4ex]
			\displaystyle \sum_{l \in \ZZ} h_{l -2 k} \overline{h_{l}} = 0, & k \in \NN.
		\end{cases}
	\end{align}
	\begin{proof}
		We start by computing the Fourier coefficients $c(h) \in \ell^{1}(\ZZ)$ of the lefthand-side of \eqref{eq:orthogonality_refinement_mask}. 
		To this end, observe that the 2-periodic map $\xi \mapsto H \left( \frac{\xi}{2}\right)$
		and its conjugate have Fourier coefficients $\frac{1}{\sqrt{2}}\tilde h$ and $\frac{1}{\sqrt{2}} \overline h$, respectively. Therefore,
		since $H \in L^{2}([0, 1])$, the product $\xi \mapsto \left \vert H \left( \frac{\xi}{2} \right) \right \vert^{2}$
		is $L^{1}$ with Fourier coefficients $\frac{1}{2} \tilde h \ast \overline h$. Similarly, the Fourier coefficients of 
		$\xi \mapsto \left \vert H \left( \frac{\xi+1}{2} \right) \right \vert^{2}$ are given by  $\frac{1}{2} M ( \tilde h ) \ast M (\overline h)$.
		Hence 
		\begin{align*}
			2c(h) = 
			\tilde h \ast \overline{h} 
			+ M(\tilde h) \ast M(\overline h).
		\end{align*}
		Unfolding the definitions yields
		\begin{align*}
			(c(h))_{k} = \frac{1}{2} \sum_{l \in \ZZ} \left(1 +  (-1)^{k}  \right) h_{l - k} \overline{h}_{l}, \quad k \in \ZZ. 
		\end{align*}
		Note that $(c(h))_{k} = 0$ whenever $k$ is odd, since 
		\begin{align}
			\label{eq:odd_terms_zero}
			\left( (-1)^{k}+ 1 \right) =
			\begin{cases}
			2, & k \equiv 0 \mod 2, \\
			0, & \mbox{otherwise}.
			\end{cases}
		\end{align}		
		
		The equation in \eqref{eq:orthogonality_refinement_mask} is equivalent to the statement that $(c(h))_{k} = \delta_{0k}$ 
		for $k \in \ZZ$, since the Fourier coefficients of a $L^{1}$-function are unique. 
		Hence \eqref{eq:orthogonality_refinement_mask} is equivalent to $(c(h))_{2k} = \delta_{0, 2k}$ for $k \in \ZZ$ by 
		the observation in \eqref{eq:odd_terms_zero}. 
		Finally, the latter statement is equivalent to $(c(h))_{2k} = \delta_{0, 2k}$ for $k \in \NN_{0}$, since 
		\begin{equation*}
		    			\overline{ \sum_{l \in \ZZ} h_{l -2 k} \overline{h_{l}}}  = 
			 \sum_{l \in \ZZ} h_{l +2 k} \overline{h_{l}} 
		\end{equation*}
		for any $k \in \ZZ$.
		The two cases in \eqref{eq:orthogonality_low_pass_filter} show the demands for $k=0$ and positive even indices, respectively. 
		This establishes the result.
		\end{proof}
\end{lemma}
\begin{remark}
	A more direct way to arrive at \eqref{eq:orthogonality_low_pass_filter} is to plug in the dilation relation into
	$\langle T_{k} \varphi, \varphi \rangle$ and use the orthogonality of $(\varphi_{1k})_{k \in \ZZ}$. The equivalence 
	with \eqref{eq:orthogonality_refinement_mask} can then be established in a similar (but slightly different) way.
\end{remark}

\paragraph{QMF conditions}
We are now ready to set up the desired constraints. In general, the QMF conditions are not sufficient to guarantee that $h$ 
is the low pass filter of a scaling function, see the discussion in Remark \ref{remark:nondegeneracy_condition}. However, 
in numerical experiments, we never seem to violate the non-degeneracy condition when only imposing the QMF conditions. 
For this reason, the only constraints 
that we impose are the QMF conditions. We do provide an option to include the non-degeneracy condition in
Remark \ref{remark:H_no_zeros}. 

To properly write down the QMF conditions as constraints on a sequence $h$, 
we introduce some additional notation. Let $\mathcal{A}_M(\RR)$ denote the space of one-dimensional 
$\RR$-valued two-sided sequences of order $M$, i.e., 
\begin{align*}
	\mathcal{A}_{M}(\RR) : = \left \{ a \ \biggr \vert \ a : \{ 1 - M, \ldots, M -1 \} \rightarrow \RR \right \}. 
\end{align*}
Note that $\mathcal{A}_{M}(\RR)$ is a vector space over $\RR$ of dimension $2M - 1$. In particular, $\mathcal{A}_{M} \simeq \RR^{2M-1}$. 
The reason for introducing this notation is to explicitly keep track of the two-sided ordering of sequences. 
We are now ready to gather all the demands that we have derived, and place them into the general framework of
Section \ref{sec:cerm}.
\begin{definition}
	\label{def:QMF_map}
	Let $M \in \NN_{\geq 3}$ be a prescribed order. The QMF-map is the function $F_{M} : \mathcal{A}_{M}(\RR) \rightarrow \RR^{M}$ 
	defined by 
	\begin{align*}
		(F_{M}(h))_{k} := 
		\begin{cases}
			(h^{-} \ast h)_{0} - 1, & k = 0, \\[2ex]
			(h^{-} \ast h)_{2k}, & 1 \leq k \leq M-1, \\[2ex]
		 	- \sqrt{2}  + \displaystyle \sum_{\vert l \vert \leq M - 1} h_{l}, & k = M. 
		\end{cases}
	\end{align*}
\end{definition}
The first $M$ equations correspond to the orthonormality constraints. Note that we only have to impose $(h^{-} \ast h)_{2k} = 0$
for $1 \leq k \leq M-1$, since $(h^{-} \ast h)_{2k} = 0$ for $k \geq M$. The last equation corresponds to the condition that $H(0)=1$.  
The set of regular points in $F_{M}^{-1}(0)$ is a real-analytic $(M-2)$-dimensional submanifold of $\RR^{2M-1}$ by Remark 
\ref{remark:preimage_relaxation}. In particular, we can get as many degrees of freedom as desired by choosing a sufficiently 
large order $M$.  

We summarize the interpretation and importance of the constraints in a theorem.
\begin{theorem}
	\label{thm:zeros_FM}
	If $F_{M}(h) = 0$ and $\inf_{0 \leq \xi \leq \frac{1}{4}} \vert H(\xi) \vert > 0$, then $h$ is the low-pass filter 
	of a scaling map $\varphi$.
\end{theorem}

\begin{remark}[Imposing the non-degeneracy condition]
	\label{remark:H_no_zeros}
	 The additional non-degeneracy condition $\inf_{0 \leq \xi \leq \frac{1}{4}} \vert H(\xi) \vert > 0$ can be imposed, for instance, by 
	requiring that $H$ has no zeros in $[0, \frac{1}{4}]$. Since we consider finite filters only, the refinement mask $H$ is 
	analytic (entire even). Hence the latter condition may be imposed by requiring that
	\begin{align}
		\label{eq:contour_integral}
		\oint_{\partial \mathcal{E}_{r}} \frac{ H'(z) }{ H(z) } \ \mbox{d}z = 0, 
	\end{align}
	where $\mathcal{E}_{r} \subset \CC$ is an ellipse with foci $0$ and $\frac{1}{4}$ and $r >0$ is a free parameter which controls
	the sum of the major and minor axis. We remind the reader that the above integral counts the zeros of $H$ (up to a scaling factor) 
        in $\mathcal{E}_{r}$, provided $H$ has no zeros on $\partial \mathcal{E}_{r}$. For any parameterization of $\partial \mathcal{E}_{r}$, we
        can numerically evaluate the integrand of \eqref{eq:contour_integral} on an associated uniform grid by using the Fourier expansion of $H$. 
        We may therefore numerically compute a Fourier expansion of the integrand, which in turn allows numerical approximation of 
        the contour integral.
\end{remark}

\section{Contour Prediction using MRA}
\label{sec:contours}
In this section we present a non-trivial application of the CERM framework to learn
optimal wavelet bases for contour prediction in medical images. Wavelets have, as
discussed in the previous section, 
the ability to represent signals at multiple resolution levels, allowing for both detailed analysis of local features and a 
broad overview of the overall signal. This ability makes wavelets an ideal tool for contour prediction in two-dimensional 
images, such as slices of CT or MRI scans. 

In the context of contour prediction, wavelets can be used to represent the boundary of a region in an image using a simple 
closed curve. While a Fourier basis appears to be a natural candidate for this task, its global nature impedes
accurate predictions of curves that exhibit highly localized behavior, requiring accurate estimates of small noisy
high-frequency modes. For this reason, we have chosen to represent contours using MRA and wavelets. 
More precisely, we consider two-dimensional gray-valued images $x \in \mathcal{X} := \left[0, 1 \right]^{n \times n}$, 
e.g., slices of MRI or CT scans of size $n \times n $. We assume that each image $x$ contains 
a (uniquely identifiable) simply connected region $R(x) \subset \RR^{2}$, e.g., an organ, with boundary $\partial R(x)$. 
It is assumed that $\partial R(x)$ can be parameterized by a simple closed piecewise $C^{2}$-curve $\gamma(x)$. 
We will develop a deep learning framework for computing such parameterizations
by learning a multiresolution decomposition of $\gamma(x)$ using the methods developed in Sections \ref{sec:cerm}
and \ref{sec:wavelet}. 

This section is organized as follows. In \autoref{sec:periodic_wavelet} 
we explain how to represent periodic curves using wavelets. In \autoref{sec:data} we provide
details about the data, e.g., how ground truth curves are constructed, what preprocessing steps
are taken, etc.. In \autoref{sec:model} and \autoref{sec:training}, we present the full details of our network architecture and
training schedule. Finally, in \autoref{sec:results}, we examine the performance of our auto-contouring models for 
the spleen and prostate central gland. In addition, we visualize the task-optimized wavelets. 

\subsection{Wavelet Representations of periodic curves}
\label{sec:periodic_wavelet}
We start by explaining how to compute a multiresolution decomposition of a scalar-valued \emph{periodic} signal 
$\gamma$ with period $\tau >0$. First, we address the issue that periodic signals are not contained in $\Ltwo$ by 
considering the cut-off $\tilde \gamma(t) := \gamma(t) \mathbf{1}_{[-\tau, \tau]}(t)$, which is contained in $\Ltwo$. 
In general, such a cut-off will introduce discontinuities at the boundary points $-\tau$ and $\tau$. These artifacts do not
present an issue for us, however, since (by periodicity) we can restrict our analysis to a strict subset 
$[I_{0}, I_{1}] \subset [-\tau, \tau]$ of length $\tau$.

To compute a multiresolution decomposition of $\tilde \gamma$ using the DWT, we need to 
compute the approximation coefficients $a_{j_{1}}(\tilde \gamma) \in \elltwo$ of $\tilde \gamma$ 
at some initial resolution level $j_{1} \in \NN$.  To explain how such an initial approximation can be obtained in the first place, 
we derive an explicit formula for the approximation coefficients $a_{jk}(\tilde \gamma) = \langle \tilde \gamma, \varphi_{jk} \rangle$. 
While we will not directly use this formula, we do remark it can be efficiently implemented and provides an alternative 
method to initialize wavelet coefficients thereby addressing the so-called wavelet crime \cite{WaveletCrime, Liu1997OnTI}. For our purposes, 
this expression will be key for identifying which coefficients to consider, i.e., which spatial locations $k \in \ZZ$ associated 
to $a_{jk}(\tilde \gamma)$ are relevant for representing $\tilde \gamma$. 

\begin{lemma}[Initialization approximation coefficients]
	\label{lemma:init_approx_coeffs}
	Let $\varphi \in \Ltwo$ be the scaling map of an MRA with low-pass filter $h \in \elltwo$ and associated refinement mask $H$. 
	Assume $h$ is nonzero for only a finite number of indices $k \in \ZZ$ so that $\mbox{supp}(\varphi) \subset [-r_{1}, r_{2}]$
	for some $r_{1}, r_{2} > 0$. If $\gamma \in C_{\text{per}}^{2}([0, \tau])$ is a $\tau$-periodic map with Fourier coefficients 
	$\left( \gamma_{m} \right)_{m \in \ZZ}$, then 
	\begin{align}
		\label{eq:approx_coeffs}
		\left \langle \tilde \gamma, \varphi_{jk} \right \rangle = 
		2^{-\frac{j}{2}} \sum_{m \in \ZZ} \gamma_{m} e^{i \omega(\tau) m \frac{k}{2^{j}} } \prod_{n=1}^{\infty}  H \left( -\frac{m}{ \tau 2^{j+n}} \right),
	\end{align}
	for any $j \in \ZZ$ and $k \in \{\lceil r_{1} -2^{j}\tau \rceil, \ldots, \lfloor 2^{j} \tau - r_{2} \rfloor \}$, where
	$\omega(\tau) := \frac{2 \pi}{\tau}$ is the angular frequency of $\gamma$. 
	\begin{proof}
		Let $j \in \ZZ$ and $k \in \{\lceil r_{1} -2^{j}\tau \rceil, \ldots, \lfloor 2^{j} \tau - r_{2} \rfloor \}$ be arbitrary. 
		A change of variables shows that
		\begin{align*}
			\left \langle \tilde \gamma, \varphi_{jk} \right \rangle = 
			2^{-\frac{j}{2}} \int_{[r_{1}, r_{2}]} \tilde \gamma \left( 2^{-j}(t+k) \right) \varphi(t) \dt, 
		\end{align*}
		since  $\mbox{supp}(\varphi)\subset [-r_{1}, r_{2}]$. Note that the latter holds for all $k \in \ZZ$. 
		For $k \in \{\lceil r_{1} -2^{j}\tau \rceil, \ldots, \lfloor 2^{j}\tau - r_{2} \rfloor \}$ in particular, we have that $2^{-j}(t+k) \in [-\tau, \tau]$ for 
		all $t \in [-r_{1}, r_{2}]$. Therefore, for such $k$, we may plug in the Fourier expansion for $\tilde \gamma$ and compute
		\begin{align*}
			\int_{[-r_{1}, r_{2}]} \tilde \gamma \left( 2^{-j}(t+k) \right) \varphi(t) \dt = 
			 \int_{[-r_{1}, r_{2}]} \sum_{m \in \ZZ} \gamma_{m} e^{i \omega(\tau) m \frac{t + k }{2^{j}}} \varphi(t) \dt.
		\end{align*}
		Next, note that that series inside the integral converges pointwise to $\gamma\left(2^{-j}(t+k)\right) \varphi(t)$ on $[-r_{1}, r_{2}]$.
		Furthermore, the partial sums can be bounded from above on $[-r_{1}, r_{2}]$ by a constant, since $\gamma \in C_{\text{per}}^{2}([0, \tau])$ and $\varphi$ is bounded. 
		Therefore, we may interchange the order of summation and integration by the Dominated Convergence Theorem:
		\begin{align*}
			\int_{[-r_{1}, r_{2}]} \sum_{m \in \ZZ} \gamma_{m} e^{i \omega(\tau) m \frac{t + k }{2^{j}}} \varphi(t) \dt &= 
			\sum_{m \in \ZZ} \gamma_{m} e^{i \omega(\tau) m \frac{k}{2^{j}} } \int_{[-r_{1}, r_{2}]} e^{i \omega(\tau) m \frac{t}{2^{j}} } \varphi(t) \dt.
		\end{align*}
		Finally, changing the domain of integration to $\RR$ again, we see that
		\begin{align*}
			\sum_{m \in \ZZ} \gamma_{m} e^{i \omega(\tau) m \frac{k}{2^{j}} } \int_{[-r_{1}, r_{2}]} e^{i \omega(\tau) m \frac{t}{2^{j}} } \varphi(t) \dt = 
			\sum_{m \in \ZZ} \gamma_{m} e^{i \omega(\tau) m \frac{k}{2^{j}} } \hat \varphi \left( - \frac{m}{\tau 2^{j}} \right).
		\end{align*}
		The stated result now follows from the observation that $\hat \varphi(\xi) = \prod_{l=1}^{\infty} H( \frac{\xi}{2^{l}})$ holds pointwise
		for any $\xi \in \RR$, since $h$ is nonzero for only a finite number of indices, see \cite{WaveletsTheory} Theorem $8.34$.
	\end{proof}
\end{lemma}
\begin{remark}
	It is straightforward to show that the partial sums converge uniformly on $[-r_{1}, r_{2}]$. It is therefore not needed to resort to 
	the Dominated Convergence Theorem. 
\end{remark} 
\begin{remark}
The bounds $\lceil r_{1} -2^{j}\tau \rceil$ and $\lfloor 2^{j} \tau - r_{2} \rfloor$ are the smallest and largest integer, respectively, for which
the Fourier series for $\gamma$ can be plugged into $\left \langle \tilde \gamma, \varphi_{jk} \right \rangle$.
The bounds are somewhat artificial, however, since the argument may be repeated for any cut-off of $\gamma$ on $[-s\tau, s\tau]$, 
where $s \in \NN_{\geq 2}$. The choice for $s$ is ultimately irrelevant, however, since we are interested in the minimal number 
of approximation coefficients needed to cover $\gamma$; see the discussion below. 
\end{remark}

\begin{figure}[tb]
	\centering
	{{\includegraphics[width=0.85\textwidth]{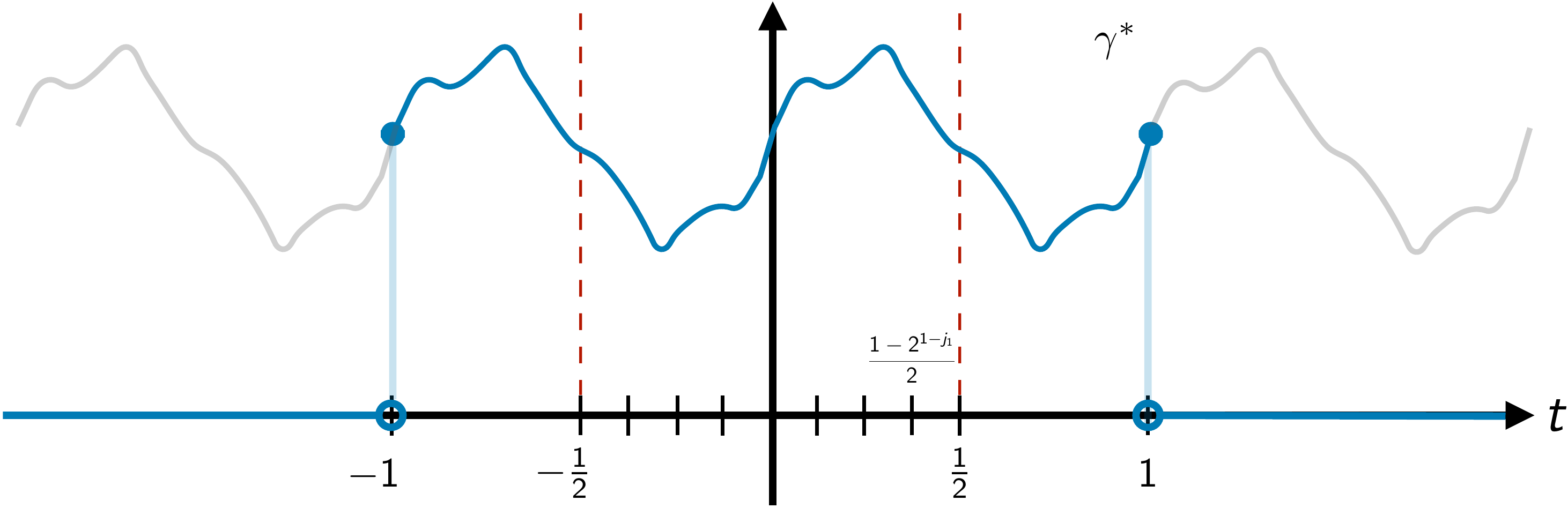}}} \\[1ex]
	\caption{ 
 The re-parameterized cut-off signal $\gamma^{\ast}(t)= \gamma(\tau t) \mathbf{1}_{[-1, 1]}(t)$ depicted in blue. We only need to compute
		     approximation coefficients associated to the smaller region $[-\frac{1}{2}, \frac{1}{2}]$. 	
		     \label{fig:cutoff}
	}
\end{figure}
Lemma \ref{lemma:init_approx_coeffs} provides a convenient way to initialize approximation coefficients. To explain how, we first 
re-parameterize $\gamma$ to have period $1$ and consider the cut-off $\gamma^{\ast} (t) := \gamma(\tau t) \mathbf{1}_{[-1, 1]}(t)$.
The motivation for this re-parameterization is that we can now conveniently relate specific approximation coefficients to sample values 
of $\gamma$. To be more precise, recall that $\hat \varphi$ is continuous at zero and $H(0) = 1$. 
Therefore, if the initial resolution level $j_{1}$ is sufficiently large, the infinite product in \eqref{eq:approx_coeffs} 
will be close to $1$ (for small $m$). Furthermore, in practice, we have a finite number of Fourier coefficients, i.e., 
$\gamma_{m} = 0$ for $\vert m \vert \geq N$. Therefore, if $j_{1}$ is
sufficiently large relative to $N$, then 
\begin{align}
	\label{eq:approx_approx_coeffs}
	a_{j_{1}k}( \gamma^{\ast}) \approx 2^{-\frac{j_{1}}{2}} \gamma^{\ast}( k 2^{-j_{1}} ), 
	\quad \lceil r_{1} -2^{j_{1}} \rceil \leq k \leq \lfloor 2^{j_{1}} - r_{2} \rfloor.
\end{align} 
That is, on sufficiently high-resolution levels 
the approximation coefficients  are close to the (scaled) sample values of the underlying signal; a well-known general fact of MRAs. 
Consequently, the approximation coefficients needed to cover $[-1, 1]$ (approximately) are 
$(a_{j_{1}k}(\gamma^{\ast}))_{k= \lceil r_{1} -2^{j_{1}} \rceil}^{\lfloor 2^{j_{1}} - r_{2} \rfloor}$. 
Motivated by this observation, and the fact that we only need $\gamma^{\ast}$ on $[-\frac{1}{2}, \frac{1}{2}]$, we
use the scaled sample values in \eqref{eq:approx_approx_coeffs} to initialize the coefficients 
$(a_{j_{1}k}(\gamma^{\ast}))_{k= -2^{j_{1} -1 }}^{2^{j_{1} - 1} - 1}$, which cover
$[-\frac{1}{2}, \frac{1 - 2^{1 - j_{1}}}{2}]$ approximately, see Figure \ref{fig:cutoff}. 

We stress that in order for the above approximations to be accurate, the initial resolution level $j_{1}$ needs to be 
sufficiently large.  Furthermore, to ensure that $ -2^{j_{1} -1 } > \lceil r_{1} -2^{j_{1}} \rceil$ and 
$2^{j_{1} -1} -1 < \lfloor 2^{j_{1}} - r_{2} \rfloor$, we require that
\begin{equation*}
    j_{1} \geq \max \left \{ \left \lceil \frac{ \log{(r_{1} + 1)} }{ \log(2) } + 1 \right \rceil,
    		 \left \lceil \frac{ \log{(r_{2} - 1)} }{ \log(2) } + 1 \right \rceil \
    		 \right \}.
\end{equation*}
One can explicitly express the support of $\varphi$ in terms of the order $M$ of the wavelet. 
Specifically, the scaling relation can be used to shown that $\text{supp} \ \varphi \subset [1-M, M-1]$,
thus providing explicit values for $r_{1}$ and $r_{2}$. A rigorous proof is out
of the scope of this paper and we refer the reader to \cite{walnut2002introduction} Theorem $8.38$.

Finally, we remark that after the initial approximation coefficients are initialized, the periodicity of $\gamma^{\ast}$
has to be taken into account in the implementation of the DWT, see Appendix \ref{appendix:periodic_conv}
for the details. 

\subsection{Data and preprocessing}
\label{sec:data}
We have used public datasets from the Medical Decathlon Challenge \cite{antonelli2022medical}
to illustrate the effectiveness of the CERM methodology. 
The selected data consists of CT scans of the spleen of size $512 \times 512$ and T2-weighted MRI 
images of the prostate central gland, henceforth abbreviated as just the prostate, of size $320 \times 320$. The scans 
were cropped to size $224 \times 224$ and  $192 \times 192$, respectively. Furthermore, the images were resampled 
to the median sample spacing, which resulted in $(5.00 \ \text{mm}, 0.793 \ \text{mm}, 0.793 \ \text{mm})$ and 
$(3.6 \ \text{mm}, 0.625 \ \text{mm}, 0.625 \ \text{mm})$ spacings for the spleen and prostate, respectively.

\subsubsection{Construction ground truth}
\label{subsec:groundtruth}
Let $(x, y) \in \XX \times \RR^{n_{s} \times n_{p}}$ be an image (slice) - contour pair, where $x$ is a slice of the CT or MRI scan, 
$y$ is a sequence of $n_{p} \in \NN$ points approximating the boundary of a simply connected region $R = R(x)$, 
and $n_{s}=2$ is the number of spatial components. Since we only have access to binary masks, and not to the raw annotations themselves, we extract $y$ 
using \textsc{opencv}. We remark that $y$ is not constrained to an integer-valued grid. 

\paragraph{Approximation coefficients}
The ground truth consists of the approximation coefficients of $\gamma^{\ast}$ at an initial resolution level $j_{2} \in \NN$. 
Here $\gamma^{\ast}$ is the re-parameterized cut-off of an initial parameterization $\gamma$ of $\partial R$ as explained in 
the previous section. We approximate the approximation coefficients using \eqref{eq:approx_approx_coeffs}, which requires evaluating 
$\gamma^{\ast}$ on a dyadic grid. To accomplish this, we compute a Fourier expansion for $\gamma$. To be more precise,
we first parameterize $\partial R$ by arc length resulting in a curve $\gamma$. The arc length $\tau$ is approximated by summing up the Euclidian 
distances between subsequent points on $y$. We re-parameterize $\gamma$ to have period $1$, as explained in 
\autoref{sec:periodic_wavelet}, and additionally ``center'' it using the average midpoint of the contours in the training set.  
The Fourier coefficients of the resulting contour are then computed by evaluating it on an equispaced grid of $[0, 1]$ of size $2N-1$, where $N \in \NN$, 
using linear interpolation and the Discrete Fourier Transform. Since the contours are real-valued, we only store the Fourier coefficients 
$(\tilde \gamma_{m})_{m=0}^{N-1} \in (\CC^{n_{s}})^{N}$.  Fourier coefficients that are too small, i.e., have no relevant contribution, 
are set to zero; see Appendix \ref{appendix:truncation} for the details. Finally, we use the approximation in 
\eqref{eq:approx_approx_coeffs} to initialize the approximation coefficients $a_{j_{2}}$. 

\paragraph{Consistency}
To have consistent parameterizations for all slices, we ensure that $\partial R$ is always traversed anti-clockwise (using \texttt{opencv}).
Furthermore, since parameterizations are only determined up to a translation in time, we need to pick out a specific one. 
We choose the unique parameterization such that $\gamma^{\ast}$ starts at angle zero at time zero relative to the midpoint 
$c = (c_{1}, c_{2} ) \in \RR^{2}$ of $R$. The implementation details are provided in Appendix \ref{appendix:midpoint}. 

The resulting dataset $\mathcal{D}$ thus consists of tuples $( x, a_{j_{2}})$. Before feeding the images $x$ into the model, 
we linearly rescale the image intensities at each instance to $[0,1]$. Furthermore, we use extensive data augmentation: we use random shifts,
random rotations, random scaling, elastic deformations and horizontal shearing. A custom (random) split of the
available data was made to construct a train-validation-test split. The sizes of the datasets are reported in Table \ref{table:datasets}. 

\begin{table}[tb]
	\centering
	\scalebox{1.0}{
	\begin{tabular}[htb]{cccc}
		\toprule
		ROI & $\vert \mathcal{D}_{\text{train}} \vert$ &	$\vert \mathcal{D}_{\text{val}} \vert$  & $\vert \mathcal{D}_{\text{test}} \vert$ \\
		\midrule
		Spleen & $2509$ & $386$ & $371$ \\
		Prostate & $454$ & $77$ & $63$ \\
		\midrule
		\bottomrule
	\end{tabular}}
	\caption{\label{table:datasets}The number of samples (slices) in the train-val-test splits for the prostate and spleen. This count includes empty slices, i.e., 
		      slices which do not contain a contour. The split was made on volume (patient) level.}
\end{table}

\subsection{Model objective and architecture}
\label{sec:model}
In this section we describe the model architecture and its objective. 

\subsubsection{Objective}
In order to define the objective, let $x \in \XX$ be an image containing a simply connected region $R(x)$ with associated boundary $\partial R(x)$. 
Let $\gamma^{\ast}(x)$ be the re-parameterized cut-off of an initial parameterization $\gamma(x)$ of $\partial R(x)$ as explained in Section 
\ref{sec:periodic_wavelet}. 
The objective is to compute the relevant approximation coefficients of $\gamma^{\ast}(x)$. 
More precisely, let $j_{0}, j_{1}, j_{2} \in \NN$ be resolution levels,
where $j_{0} \leq j_{1} \leq j_{2}$. We will construct a convolutional neural network
\begin{equation*}
G: \XX \times \RR^{p} \rightarrow \prod_{j=j_{0}}^{j_{2}} \RR^{2^{j}} \times \RR^{2^{j}} \times \prod_{j=j_{0}}^{j_{1}-1} \RR^{2^{j}} \times \RR^{2^{j}},
\end{equation*}
which predicts the wavelet decomposition of $\gamma^{\ast}(x)$. Here the subspaces $\RR^{2^{j}}$ correspond to
approximation and detail coefficients at level $j$, one for each spatial component. Furthermore, a subset of the parameters
$\xi \in \RR^{p}$ are constrained to be wavelet filters, one wavelet filter per spatial component, using Theorem \ref{thm:zeros_FM}. 

To explain more precisely what the co-range of $G$ represents, we identify $\RR^{2^{j}}$ with truncated
approximation and detail subspaces:
\begin{align}
	\label{eq:approx_truncated_subspace}
	\RR^{2^{j}} &\simeq \mbox{span} \{ \varphi_{jk} : - 2^{j-1} \leq k \leq 2^{j-1} -1 \} \subset V_{j}, \\[2ex] 
	\label{eq:detail_truncated_subspace}
	\RR^{2^{j}} &\simeq \mbox{span} \{ \psi_{jk} : - 2^{j-1} \leq k \leq 2^{j-1} - 1\} \subset W_{j}.
\end{align}
Note very carefully that the identifications in \eqref{eq:approx_truncated_subspace} and \eqref{eq:detail_truncated_subspace} 
explicitly depend on the constrained network parameters, i.e., the wavelet filters which determine the father and
mother wavelets $\varphi$ and $\psi$, respectively. 
The map $G(\cdot, \xi)$ applied to an image $x$ has output
\begin{align*}
    G(x, \xi) = \left(v_{j_{0}}( x, \xi), \ldots, v_{j_{2}}(x, \xi),  w_{j_{0}}( x, \xi), \ldots, w_{j_{1} - 1}(x, \xi) \right).
\end{align*}
Here $v_{j}(x, \xi)$ and $w_{j}(x, \xi)$ are predictions for the approximation and detail coefficients
of $\gamma^{\ast}(x)$ at level $j$, respectively. We only predict detail coefficients up to level $j_{1}$. 
The approximation coefficients at levels $j_{1} < j \leq j_{2}$ are constructed without detail coefficients,
see the next section for motivation. 
Altogether, the goal is to find optimal parameters $\xi \in \mathcal{N}$ such that 
\begin{align*}
	v_{jk}(x, \xi) \approx \left( a_{jk} ( [\gamma^{\ast}(x)]_{1}), a_{jk} ( [\gamma^{\ast}(x)]_{2}) \right), \quad  -2^{j -1} \leq k \leq 2^{j - 1} - 1, \ j_{0} \leq j \leq j_{2}, 
\end{align*}
for ``most'' realizations of $X$.

\begin{figure}[!tb]
	\centering \qquad \qquad
	{{\includegraphics[width=0.85\textwidth]{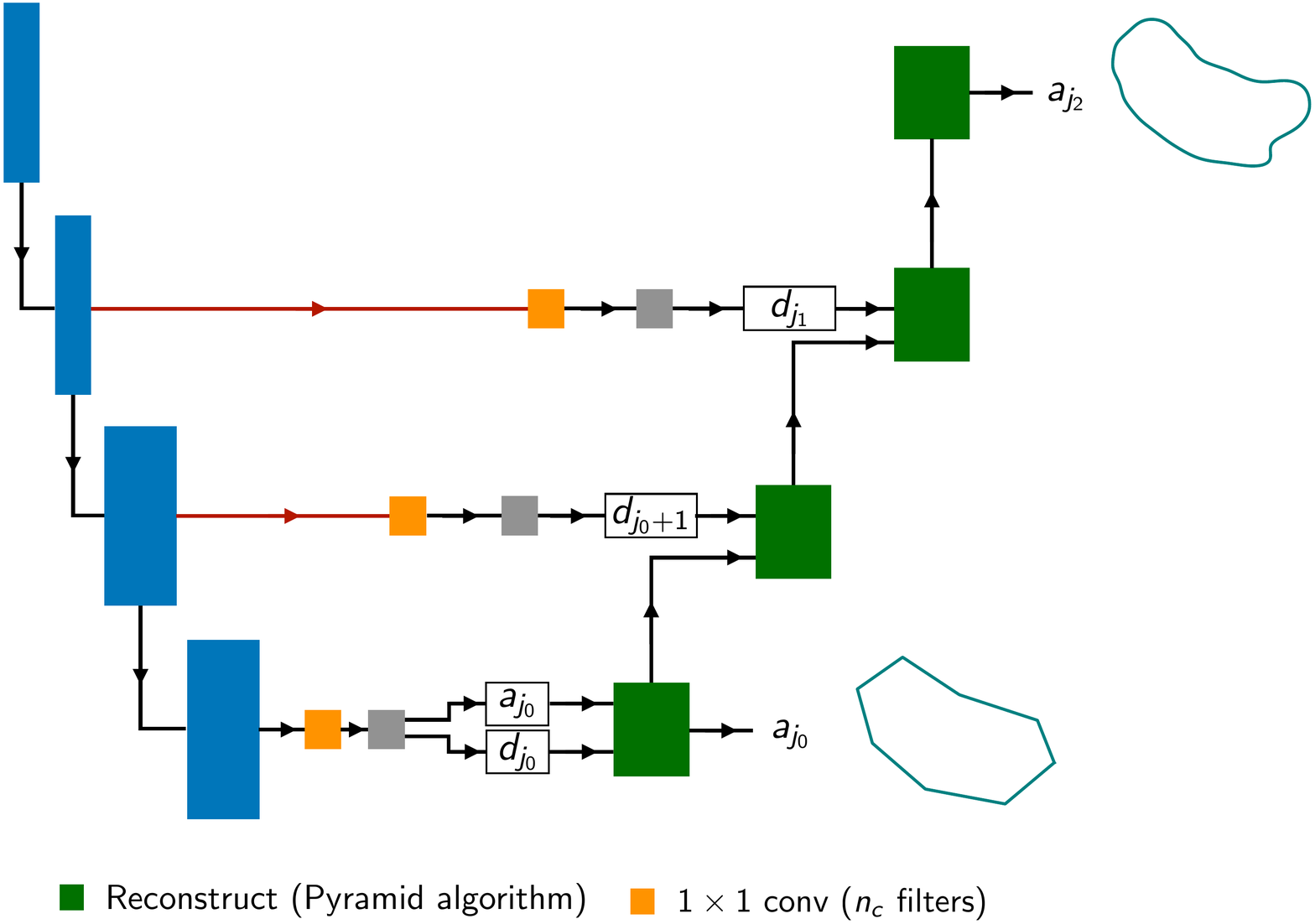}}} \\[1ex]
	\caption{A schematic picture of our network. The encoder consists of residual convolutional blocks depicted in blue.
		      The first residual convolutional block uses $n_{f}$ filters and is doubled after every other residual convolutional block. 
		      Attached to the encoder are fully connected layers to predict approximation and detail coefficients at the lowest resolution level $j_{0}$. 
		      The approximation and detail coefficients are supplied as input to the Pyramid Algorithm (the decoder) to predict a contour on high-resolution level. 
		       Each green block corresponds to a $1$-level-IDWT as depicted in Figure \ref{fig:pyramid_up}.
		      Detail coefficients at higher levels are computed using skip-connections (arrows in red). We only predict detail coefficients up to level $j_{1}$. 
		      No detail coefficients are used at levels $j_{1} + 1 \leq j \leq j_{2}$. In this example, we have set $j_{1} = j_{0} + 2$ and $j_{2} = j_{0} + 3$.
		      In reality, the decoder consists of two upsampling paths, one for each spatial component of the curve. We have only drawn one for notational convenience.
		      During training, only the approximation coefficients at the highest resolution level are supervised. See Figure 
		      \ref{fig:network_components} for more details about the network components. 
		      }
	\label{fig:network}
\end{figure}

\subsubsection{Architecture}
\begin{figure}[!tb]
	\centering
	{{\includegraphics[width=0.92\textwidth]{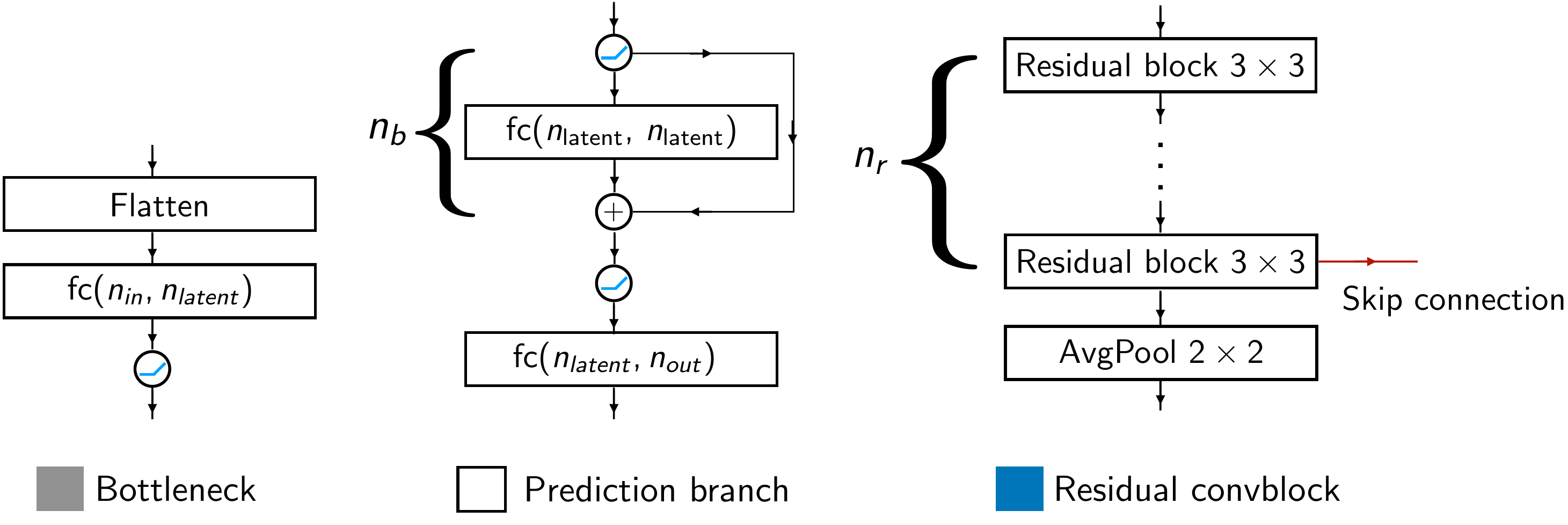}}} \\[1ex]
	\caption{The components of the network: the bottleneck, fully connected prediction layers, 
		      and convolutional block, respectively.
		      }
	\label{fig:network_components}
\end{figure}

Our network is a hybrid analog of the U-Net. It consists of a two-dimensional convolutional encoder, a bottleneck of fully connected layers, 
and a one-dimensional decoder. The encoder and decoder are connected through skip-connections. The approximation and detail 
coefficients at the lowest resolution level $j_{0}$ are predicted in the bottleneck. Afterwards, the Pyramid Algorithm takes over to compute 
approximation coefficients at higher resolution levels (the decoder)  using learnable wavelet filters. The needed detail coefficients at the higher resolution levels
are predicted using the skip-connections. In practice, the detail coefficients are negligible on sufficiently high-resolution levels. For 
this reason, we only predict detail coefficients up to a prescribed level $j_{1}$. The predictions at higher resolution levels $j_{1} < j \leq j_{2}$ 
are computed without detail coefficients. The full architecture is visualized in Figures \ref{fig:network} and \ref{fig:network_components}. 
In addition, we provide a detailed summary below. The specific values for the architecture were determined 
using a hyperparameter search.

\paragraph{Encoder}
The encoder consists of $n_{d} \in \NN$ down-sampling blocks. Each block consists of $n_{r} \in \NN$ (convolutional) residual blocks, 
using GELU-activation and kernels of size $3 \times 3$, followed by an average-pooling layer of size $2 \times 2$. 
The initial number of filters $n_{f} \in \NN$ used in the first block is doubled after each other block. For example, if $n_{d} = 5$ and
the number of kernels at the first block is $n_{f}=32$, then the subsequent blocks have $32$, $64$, $64$, and $128$, kernels, respectively. 

\paragraph{Bottleneck}
The encoder is followed by a bottleneck which consists of a stack of fully connected layers.
The first layer in the bottleneck compresses the feature map from the encoder path to a feature map 
with $n_{c}$ channels using a $1 \times 1$ convolution. Next, this compressed feature map is transformed to a vector
in $\RR^{n_{\text{lat}}}$, where $n_{\text{lat}} \in \NN$ refers to the latent 
dimension of  the MLP. Attached to this layer are four branches to predict the approximation and detail coefficients 
$[v_{j_{0}}(x)]_{s}, [w_{j_{0}}(x)]_{s} \in \RR^{2^{j_{0}}}$, respectively. Here $s \in \{1, 2\}$ corresponds to the spatial component of the contour. 
Each branch consists of $n_{b} \in \NN$ fully-connected layers. The first $n_{b} -1$ layers map 
from $\RR^{n_{\text{lat}}}$ to itself with GELU-activation and residual connections in between. The final layer
transforms the $n_{\text{lat}}$-dimensional output to an element in $\RR^{2^{j_{0}}}$. 

\paragraph{Decoder}
The detail coefficients at levels $j_{0} \leq j < j_{1}$ are predicted using skip-connections. For each
skip-connection, we first compress the feature map from the encoder path  to a feature map with $n_{c}$ 
channels using a $1 \times 1$ convolution. Subsequently, two prediction branches, each having the same
architecture as above,  are used to predict the detail coefficients in $\RR^{2^{j}}$ (one for each spatial component). 
The predicted approximation coefficients at level $j_{0}$ and detail coefficients at levels $j_{0} \leq j \leq j_{1} -1$ 
are used as input to the Pyramid algorithm to reconstruct approximation coefficients up to level $j_{1}$ using learnable wavelet filters. 
The approximation coefficients at levels $j_{1} + 1 \leq j \leq j_{2}$ are reconstructed without detail coefficients.

\paragraph{Hyperparameters} 
The choices for the hyperparameters were based on a hyperparameter search, optimizing the Dice score. For the spleen and prostate
we have set 
\begin{align*}
    (n_{d}, n_{r}, n_{\text{lat}}, n_{b}, n_{c}, j_{2}) = (6, 4, 124, 3, 16, 7), \quad
    (n_{d}, n_{r}, n_{\text{lat}}, n_{b}, n_{c}, j_{2}) = (5, 4, 116, 2, 16, 7),
\end{align*}
respectively, and considered wavelet orders $3 \leq M \leq 8$. 
Furthermore, for each order, we used the lowest possible resolution level $j_{0}$ and $j_{1} = j_{2}$. In particular, 
$j_{0}(M) = 3$ for $M \in \{3, 4\}$ and $j_{0}(M) = 4$ for $M \in \{5, 6, 7, 8 \}$. 

\subsection{Optimization}
\label{sec:training}
In this section, we summarize the training procedure of our contouring-network. 
We use two different optimizers during training: one for the encoder and MLP (unconstrained), 
and one for the decoder (constrained wavelet network). The reason for this is that the needed step sizes on the 
non-trivial submanifold may significantly differ from those on the unconstrained (flat) parameter space. 
We use first order approximations of geodesics to perform SGD on the submanifold of constrained parameters, 
see \autoref{sec:gradient_descent}. 
We remark that the generality of our framework allows for the computation of derivatives with respect to both the constrained 
and unconstrained parameters using normal automatic differentiation, see Algorithm \ref{algorithm:gradient}. 

We use plain SGD for the first eight epochs for both the constrained and unconstrained parameters. During this period the
learning rate for the unconstrained parameters is linearly increased from $10^{-5}$ to $2 \cdot 10^{-4}$. The learning
rate for the constrained parameters (wavelet filters) is linearly increased from $10^{-4}$ to $10^{-2}$. After the initial warmup
stage, we switch to the Adam optimizer for the unconstrained parameters. For both the constrained and unconstrained parameters, 
we use learning rate schedulers and decrease the learning rate by a factor 0.85 if no 
significant improvements in the validation loss are observed during the last ten epochs. We train all models for $250$ epochs using
a batch size of $32$ and use the last epoch for inference. The computations were performed in \textsc{PyTorch} on a Geforce RTX $2080$ Ti.

\subsubsection{Loss}
Next, we set up an appropriate loss to determine network parameters. To measure the discrepancy between the ground truth and the predicted curve, we define
\begin{equation*}
    L(G(x, \xi), a( \gamma^{\ast}(x))) :=
	\Vert [v_{j_{2}}(x, \xi)]_{1} -  a_{j_{2}}( [\gamma^{\ast}(x)]_{1}) \Vert_{2} + 
	\Vert [v_{j_{2}}(x, \xi)]_{2} -  a_{j_{2}}( [\gamma^{\ast}(x)]_{2}) \Vert_{2}.
\end{equation*}
This corresponds to the component-wise $L^{2}$-error between the curves on resolution level $j_{2}$ with approximation coefficients 
$v_{j_{2}}(x, \xi)$ and $a_{j_{2}}(\gamma^{\ast}(x))$.

Notice that $L$ measures the discrepancy between observed and predicted curves on the highest resolution level only. 
We claim that this is sufficient for enforcing the approximation and detail coefficients at intermediate levels to agree as well. Indeed, 
recall the decomposition $V_{j_{2}} = V_{j_{0}} \oplus \bigoplus_{l=j_{0}}^{j_{2}-1} W_{l}$, which shows that any signal in 
$V_{j_{2}}$ can be \emph{uniquely} written as a sum of elements in $V_{j_{0}}, W_{j_{0}}, \ldots, W_{j_{2} - 1}$. Therefore, if 
two signals agree on $V_{j_{2}}$, their associated approximation and detail coefficients on lower levels must agree as well. 

\subsubsection{Performance measures}
We evaluate performance using the two-dimensional dice score, since our models are $2$d and the hyperparameters
were tuned to optimize this metric. We compute the dice score between curves using the implementation in 
\textsc{shapely}. This requires a polygonal approximation of 
the contour, which is directly obtained using the approximation coefficients at level $j_{2}$.

For comparison, we also report the performance of a state-of-the-art baseline $2$d-nnUNet \cite{isensee2021nnu}. We stress, however, that our objective, 
i.e., parameterizing contours, is different from the nnUNet's objective. The binary ground truth matched by the 2d-nnUNet is a fundamentally 
different (often easier) object than the continuous representation of a curve matched by our networks.
Subtle curvature and geometry may be accurately presented using our ground truth curves, e.g., 
by using a sufficiently large number of Fourier coefficients to compute approximation coefficients. 
Binary ground truth masks, however, cannot capture such subtle geometry due to their discrete nature.  

\subsection{Numerical results}
\label{sec:results}
\begin{table}[tb]
	\centering
	\scalebox{1}{
	\begin{tabular}[t!]{lcccr}
		\toprule
		Model &	Dice Spleen & Dice Prostate	\\
		\midrule
		\textsc{nnunet}	&	$0.914$ $(1.74 \cdot 10^{-1})$	&	$0.896$ $(1.27 \cdot 10^{-1})$		 \\
		\textsc{order $3$}	&	$0.911$ $(7.38 \cdot 10^{-2})$	&	$0.929$ $(4.66 \cdot 10^{-2})$	 \\
		\textsc{order $4$}	&	$0.911$ $(7.85 \cdot 10^{-2})$	&	$\bm{0.935}$ $(3.48 \cdot 10^{-2})$		 \\
		\textsc{order $5$}	&	$0.916$ $(6.94 \cdot 10^{-2})$	&	$0.935$ $(4.11 \cdot 10^{-2})$	 \\
		\textsc{order $6$}	&	$0.91$7 $(7.17 \cdot 10^{-2})$	&	$0.934$ $(4.14 \cdot 10^{-2})$ \\
		\textsc{order $7$}	&	$\bm{0.921}$ $(6.91 \cdot 10^{-2})$		&	$0.928$ $(4.03 \cdot 10^{-2})$ \\
		\textsc{order $8$}	&	$0.919$ $(6.64 \cdot 10^{-2})$	&	$0.934$ $(3.62 \cdot 10^{-2})$ \\
		\midrule
		\bottomrule
	\end{tabular}}
	\caption{\label{table:results_test} Mean and standard deviation (in parentheses) of the dice score on the unseen test sets
         for the spleen and prostate. The first row corresponds to the baseline $2$D-nnUNet. 
         The subsequent rows correspond to wavelet networks of different orders $M$.}
\end{table}

\begin{figure}[!t]
	\centering
	\subfloat[\centering Spleen \label{fig:boxplot_spleen}]{{\includegraphics[width=0.83\columnwidth]{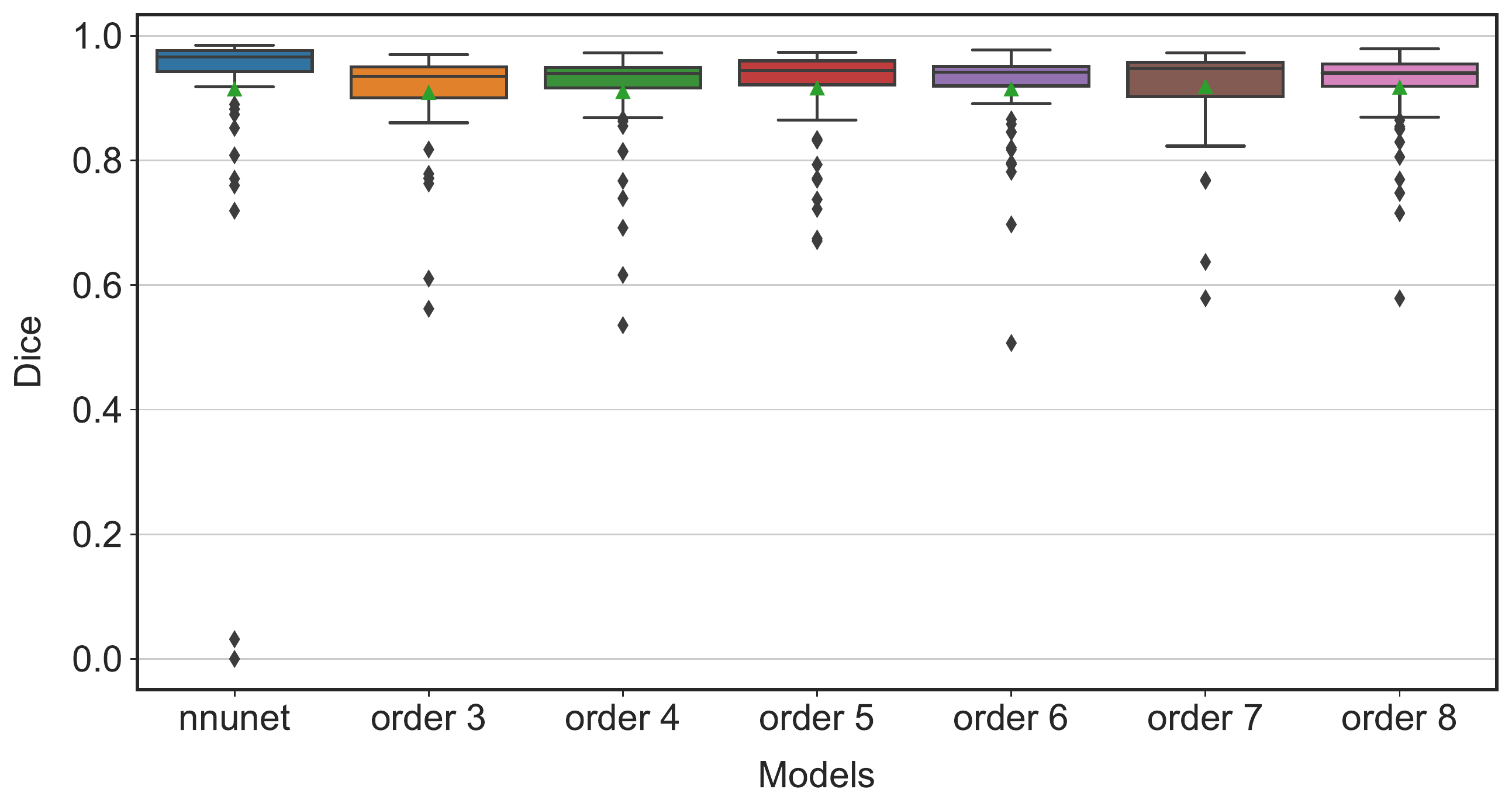}}} \\[2ex]
	\subfloat[\centering Prostate \label{fig:boxplot_prostate}]{{\includegraphics[width=0.83\columnwidth]{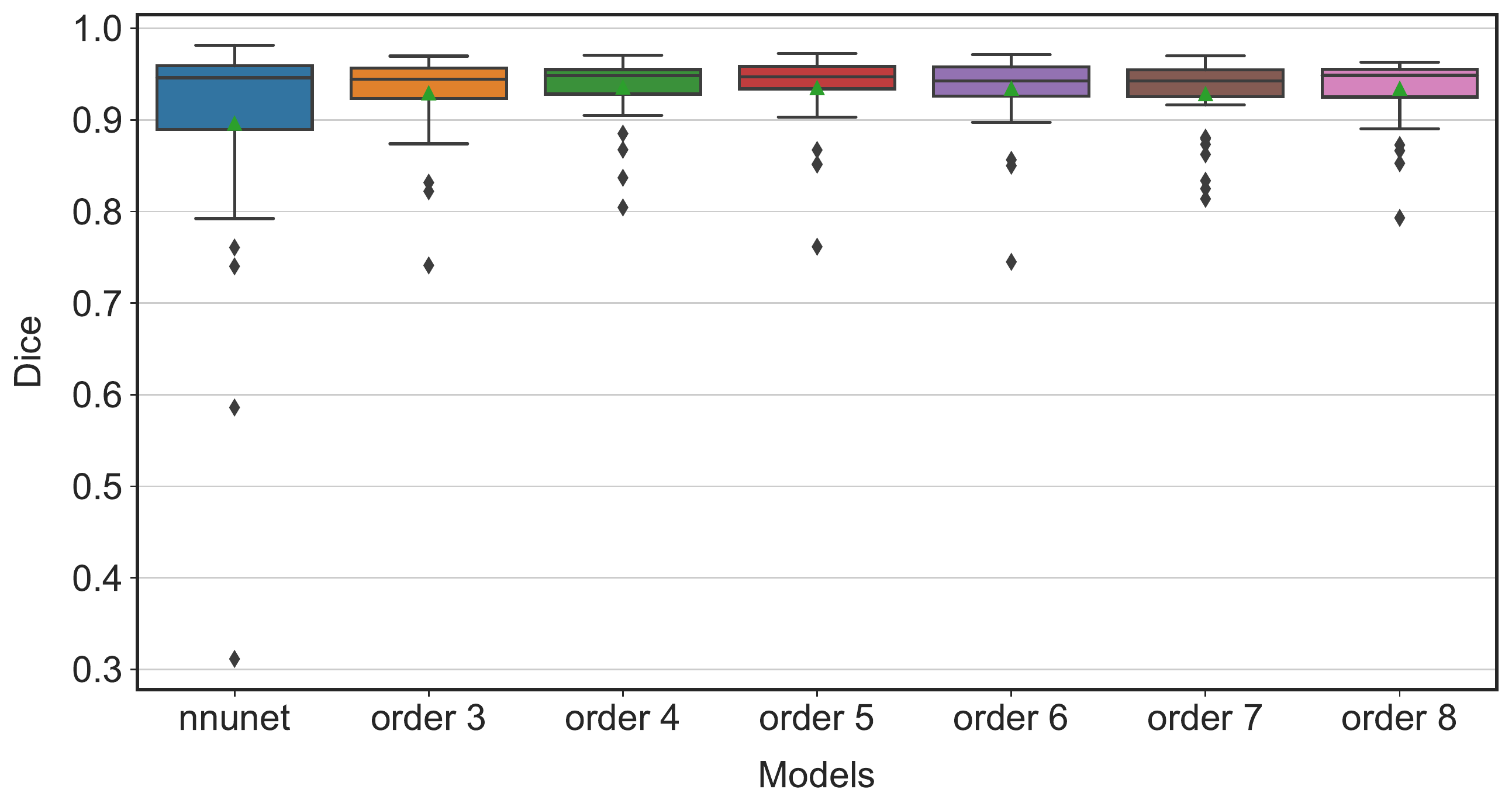}}} \
	\caption{Boxplot of the dice scores on the test set for the spleen and prostate. The green arrow denotes the average over the test set.}
	\label{fig:boxplots}
\end{figure}

\begin{figure}[t]
	\centering
	\subfloat[ \label{fig:138_spleen}]{{\includegraphics[width=0.17\columnwidth]{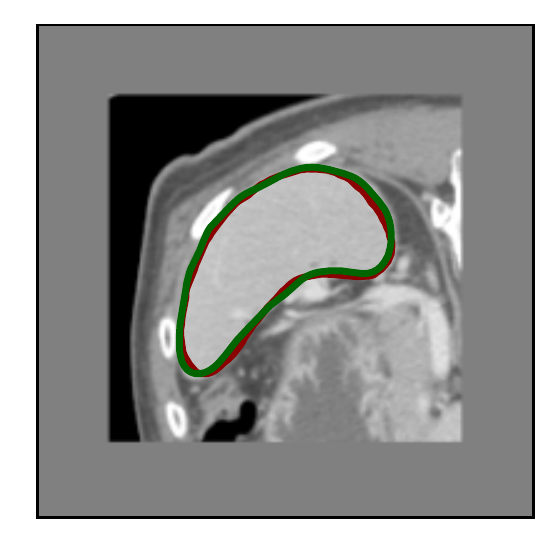}}} \
	\subfloat[\label{fig:144_spleen}]{{\includegraphics[width=0.21\columnwidth]{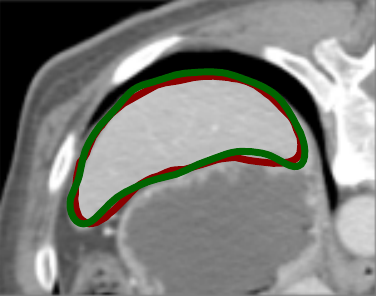}}} \
	\subfloat[ \label{fig:362_spleen}]{{\includegraphics[width=0.15\columnwidth]{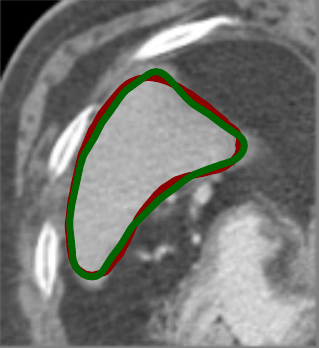}}} \
	\subfloat[ \label{fig:367_spleen}]{{\includegraphics[width=0.17\columnwidth]{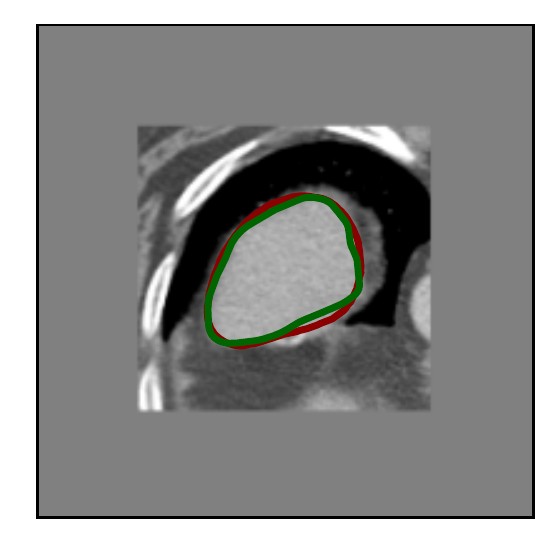}}} \
	\subfloat[ \label{fig:302_spleen}]{{\includegraphics[width=0.24\columnwidth]{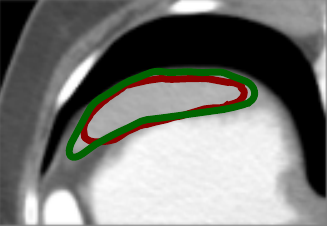}}} \\
	\subfloat[ \label{fig:45_prostate}]{{\includegraphics[width=0.16\columnwidth]{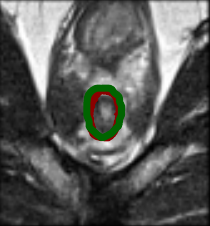}}} \
	\subfloat[ \label{fig:48_prostate}]{{\includegraphics[width=0.2\columnwidth]{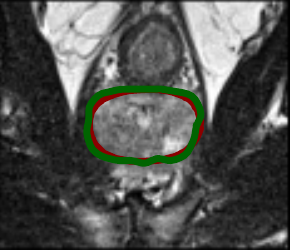}}} \
	\subfloat[ \label{fig:55_prostate}]{{\includegraphics[width=0.2\columnwidth]{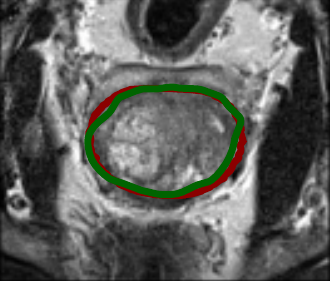}}} \ 
	\subfloat[ \label{fig:13_prostate}]{{\includegraphics[width=0.2\columnwidth]{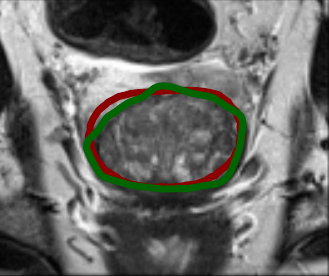}}} \
	\subfloat[ \label{fig:5_prostate}]{{\includegraphics[width=0.18\columnwidth]{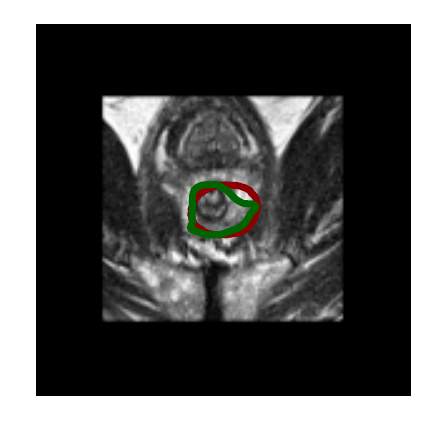}}} \\
	\caption{Examples of predictions on the test set for the spleen and prostate, depicted in the first and second row, respectively,
	for the best performing wavelet models. The green curve corresponds to the ground truth, while the red curve is a 
	prediction made by the wavelet network. The last two columns correspond to typical “hard” examples, where
	our models struggle to predict accurate contours.}
	\label{fig:predictions_test_set}
\end{figure}

We have evaluated the performance of our wavelet networks for different orders on the unseen test data, 
see Table \ref{table:results_test}.  Examples of predictions are depicted in Figure \ref{fig:predictions_test_set}. 
For both the spleen and prostate, we observe that the best wavelet networks outperform the baseline in 
terms of dice score.

\paragraph{Spleen}
The predictions of our wavelet models are accurate and on par with the baseline. The higher order wavelet models
perform slightly better in terms of the mean dice score due to more outliers by the $2$d-nnUNet. 
While the $2$d-nnUNet has cases with higher dice scores, at the same time it has a relatively large 
number of outliers with relatively low dice scores. Our wavelet models, the best performing model in particular, are more robust in 
this sense, which is an especially important property for medical applications. The robustness is 
reflected in smaller standard deviations for the dice score, also see the boxplots in Figure \ref{fig:boxplot_spleen}. 
We have depicted examples of typical predictions in Figures \ref{fig:138_spleen}, \ref{fig:144_spleen} 
and \ref{fig:362_spleen}. These examples also showcase the fact that the ground truth 
curves may describe more subtle (complicated) geometry than binary masks. 
In Figures \ref{fig:367_spleen} and \ref{fig:302_spleen} we have depicted
typical examples of predictions were our wavelet models struggle and the baseline
performs better. 

\paragraph{Prostate}
The predictions of our wavelet models are accurate and outperform the baseline in terms of the dice score.
We have depicted examples of typical predictions in Figures \ref{fig:45_prostate}, \ref{fig:48_prostate}
and \ref{fig:55_prostate}. We observe, as for the spleen, that the wavelet models are more robust than
the baseline, also see the box plots in Figure \ref{fig:boxplot_prostate}. In Figures \ref{fig:13_prostate} and \ref{fig:5_prostate} we 
show typical examples of where the wavelet models struggle to produce accurate predictions. In these examples,
the predicted detail coefficients that correspond to parts of the curve with high curvature were not sufficiently accurate. The main reason for why
we outperform the baseline is that it struggles with predicting ``small'' structures as in Figures \ref{fig:45_prostate}
and \ref{fig:5_prostate}, often only correctly identifying a small number of pixels. It is in these cases where our contour
models have a clear advantage; instead of annotating a few possibly disconnected set of pixels, our models have prior
knowledge about the geometry and always predict a contour.

\paragraph{Task-optimized wavelets}
We observe that the task-optimized wavelets differ significantly from the wavelets randomly initialized at the start of training. 
A comparison of an initial and task-optimized wavelet is depicted in Figure \ref{fig:wavelet_examples}, see Appendix \ref{sec:figures_wavelets}
for more examples. We observed that in most cases, for both the prostate and spleen, the final wavelet appeared to be less ``noisy'' 
exhibiting less variation. In particular, the task-optimized wavelets for the spleen were much less noisy than for the prostate. 
We suspect, however, but did not test, that the wavelets for the prostate models would simplify if we increased the training time. 

In numerical experiments we typically found different wavelets at the end of training for different
initializations. One of the main reasons for this is that there is no unique ``optimal wavelet'' which solves the auto-contouring problem. Finally, as we increased the order
of the wavelet filters, the final wavelets exhibited more oscillatory behavior and the number of zero-crossings increased. Our experiments
did not reveal, however, a clear choice for a ``best'' order for our applications. 

\paragraph{Non-degeneracy condition}
In all numerical experiments the task-optimized filters satisfied the non-degeneracy condition of Theorem \ref{thm:zeros_FM}. We
have illustrated this in Figure \ref{fig:refinement_examples}, where we observe (numerically), that the magnitudes $\vert H(\xi) \vert$ of the final
refinement masks are sufficiently far way from zero on $[0, \frac{1}{4}]$. We have included more examples in Appendix \ref{sec:figures_refinement_masks}. 
We do note, however, that the initial wavelets were in some instances close to ``degenerate'', in the sense $\vert H(\xi)\vert$ came close to
having a zero in $[0, \frac{1}{4}]$, see Figures \ref{fig:refinement_mask_init_0} and \ref{fig:refinement_mask_init_1} for example. In all 
such cases, these ``near-degeneracies'' vanished quickly during the initial stages of training.  

\section{Conclusion}
In this paper, we have introduced the CERM framework for imposing constraints on parametric models such as neural networks. 
The constraints are formulated as a finite system of equations. Under mild smoothness and non-degeneracy conditions, the 
parametric model can be made to obey the constraints \emph{exactly} throughout the entire training procedure by performing SGD 
on a (possibly) curved space. As a major example, we have constructed a convolutional 
network whose filters are constrained to be wavelets. We have applied these wavelet networks to the prediction of boundaries 
of simply connected regions in medical images, where they outperform strong baselines.

\begin{figure}[!t]
	\centering
	\subfloat[\centering Initial father and mother wavelet \label{fig:wavelet_init_0}]{{\includegraphics[width=0.49\columnwidth]{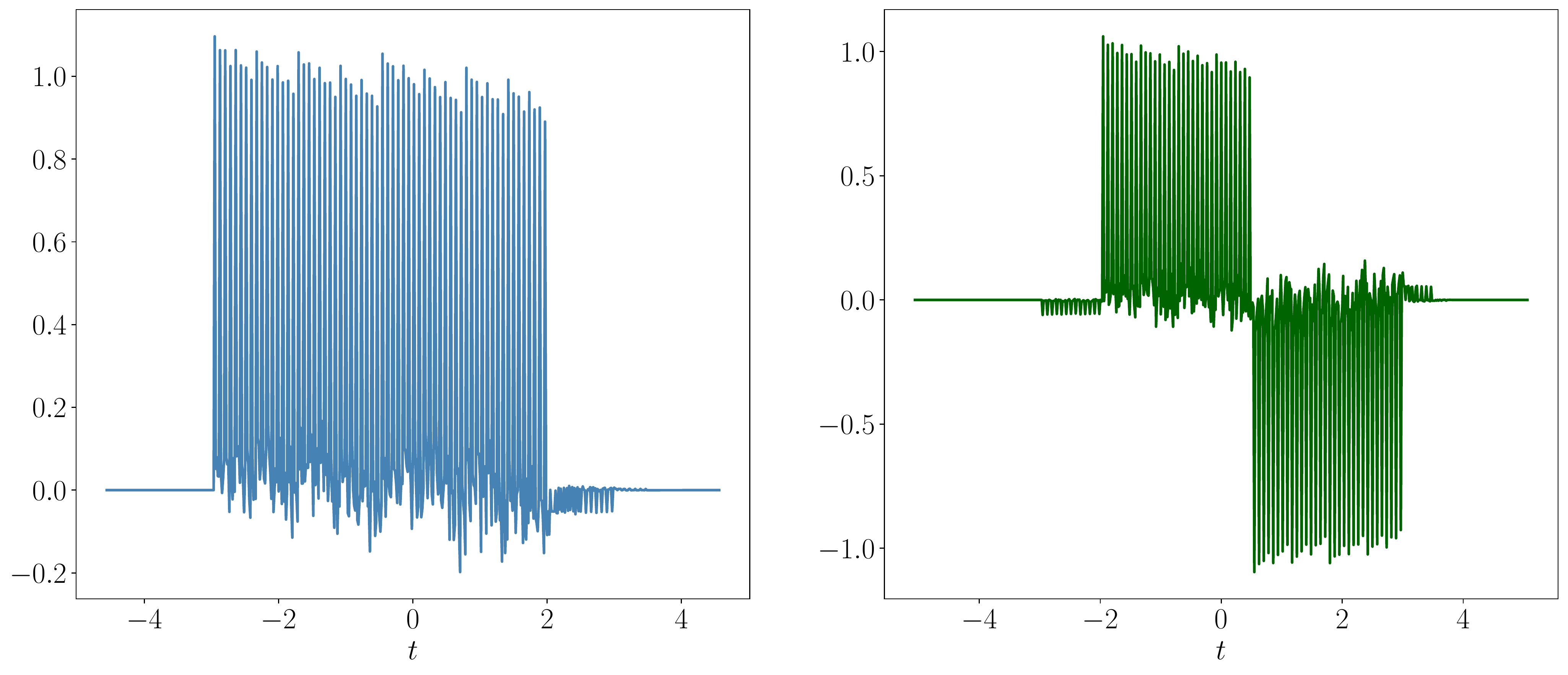}}} \
	\subfloat[\centering Task-optimized father and mother wavelet \label{fig:wavelet_final_0}]{{\includegraphics[width=0.49\columnwidth]{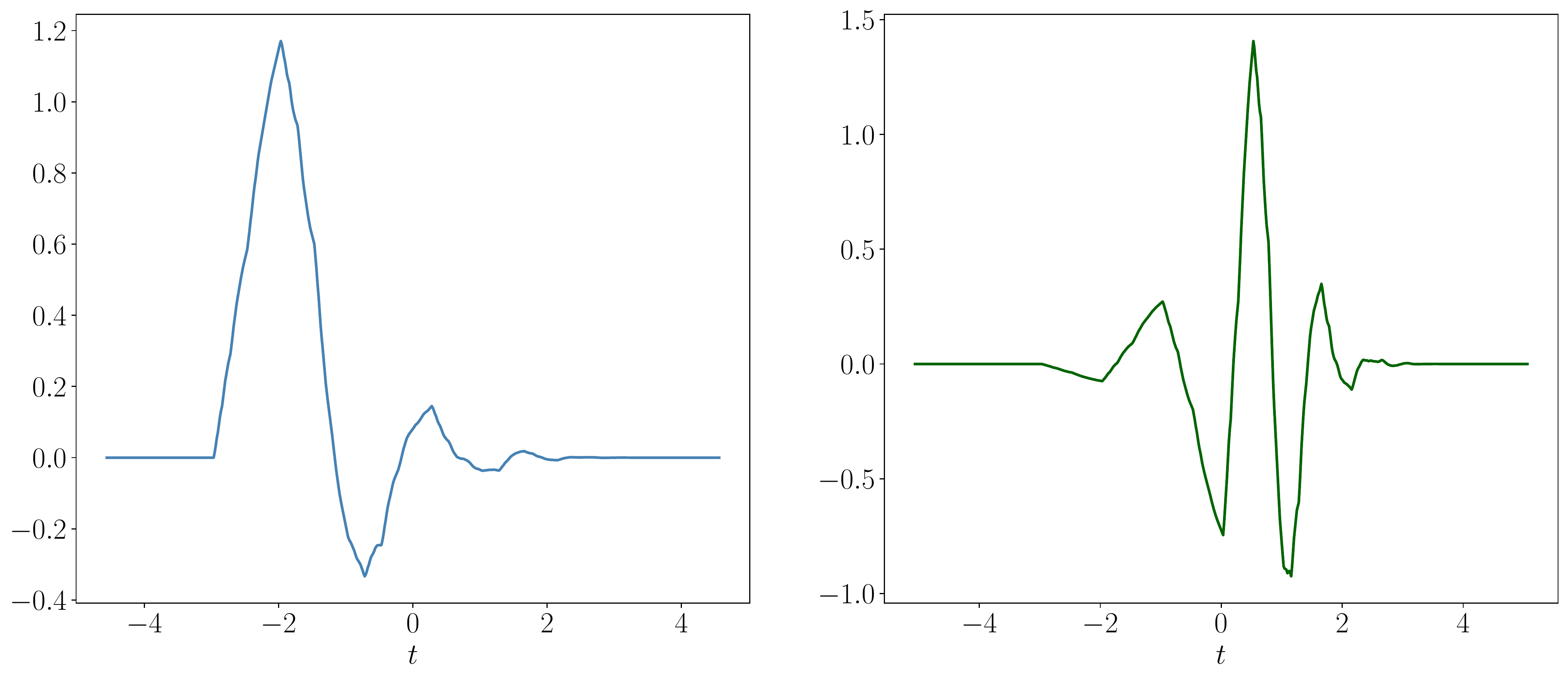}}} \\ 
	\subfloat[\centering Initial father and mother wavelet \label{fig:wavelet_init_1}]{{\includegraphics[width=0.49\columnwidth]{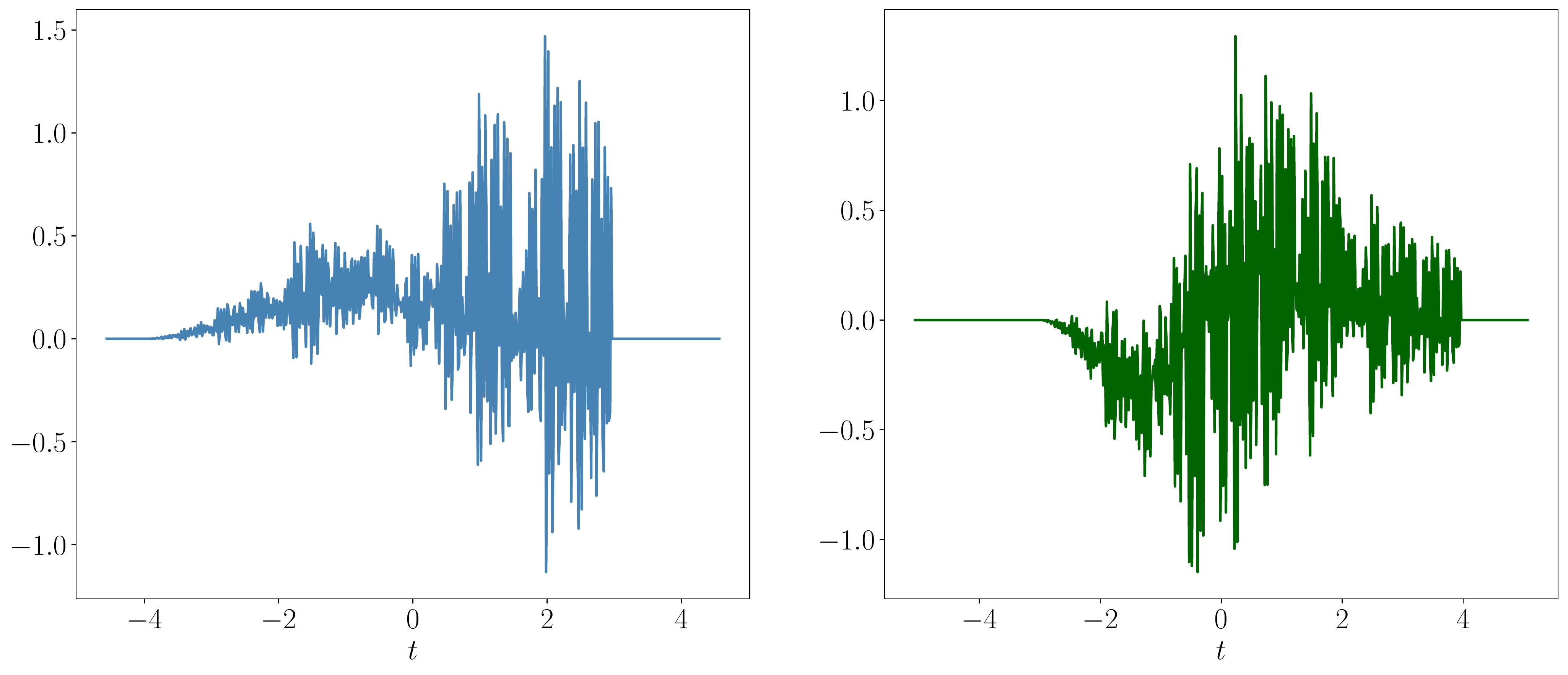}}} \
	\subfloat[\centering Task-optimized father and mother wavelet \label{fig:wavelet_final_1}]{{\includegraphics[width=0.49\columnwidth]{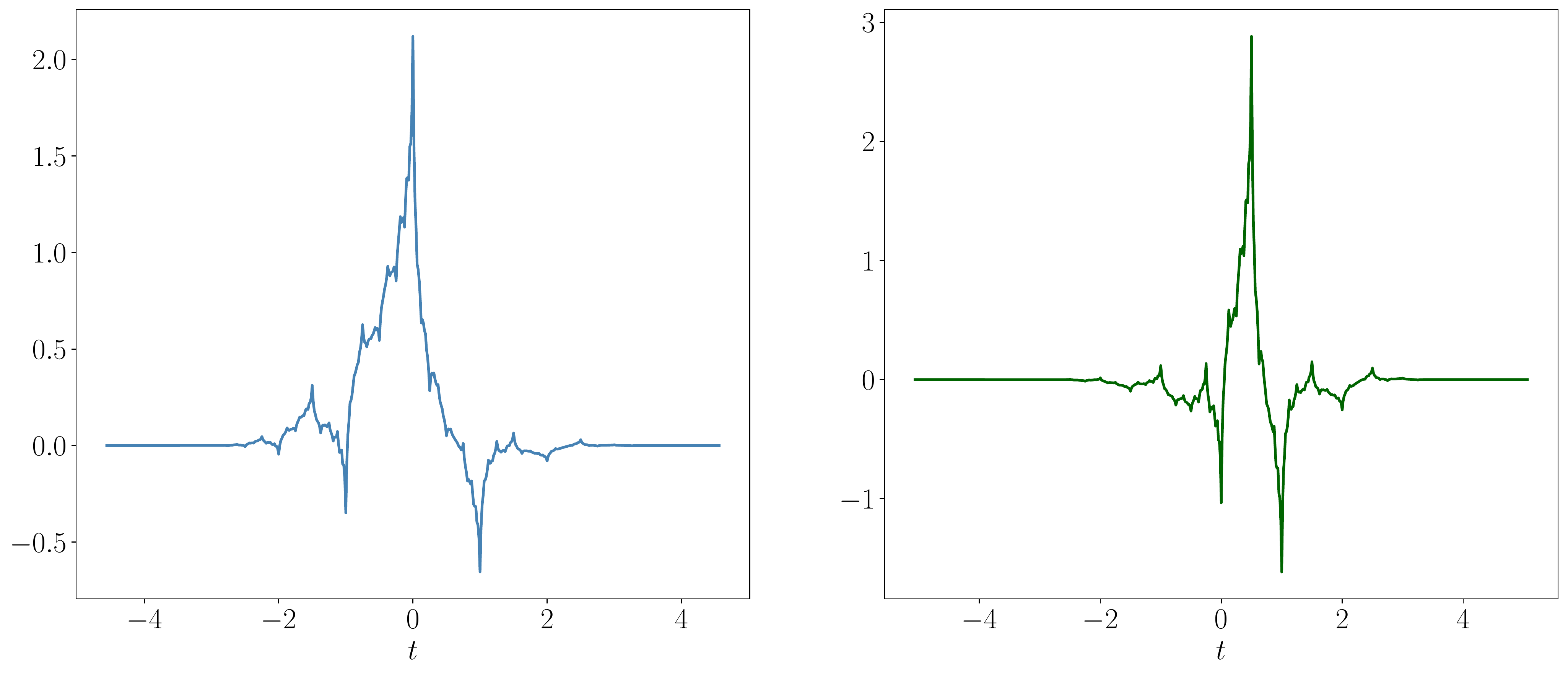}}}
	\caption{Example of wavelets of order $5$ learned during training of the spleen model. We observe
              	     that the task-optimized wavelets are more simple and exhibit less oscillatory behavior.
            ~\protect \subref{fig:wavelet_init_0}, ~\protect \subref{fig:wavelet_final_0} Wavelets associated to the first spatial component.
            ~\protect \subref{fig:wavelet_init_1}, ~\protect \subref{fig:wavelet_final_1} Wavelets associated to the second spatial component. }
	\label{fig:wavelet_examples}
\end{figure}

\begin{figure}[!t]
	\centering
	\subfloat[\centering Initial refinement masks \label{fig:refinement_mask_init_0}]
	{{\includegraphics[width=0.49\columnwidth]{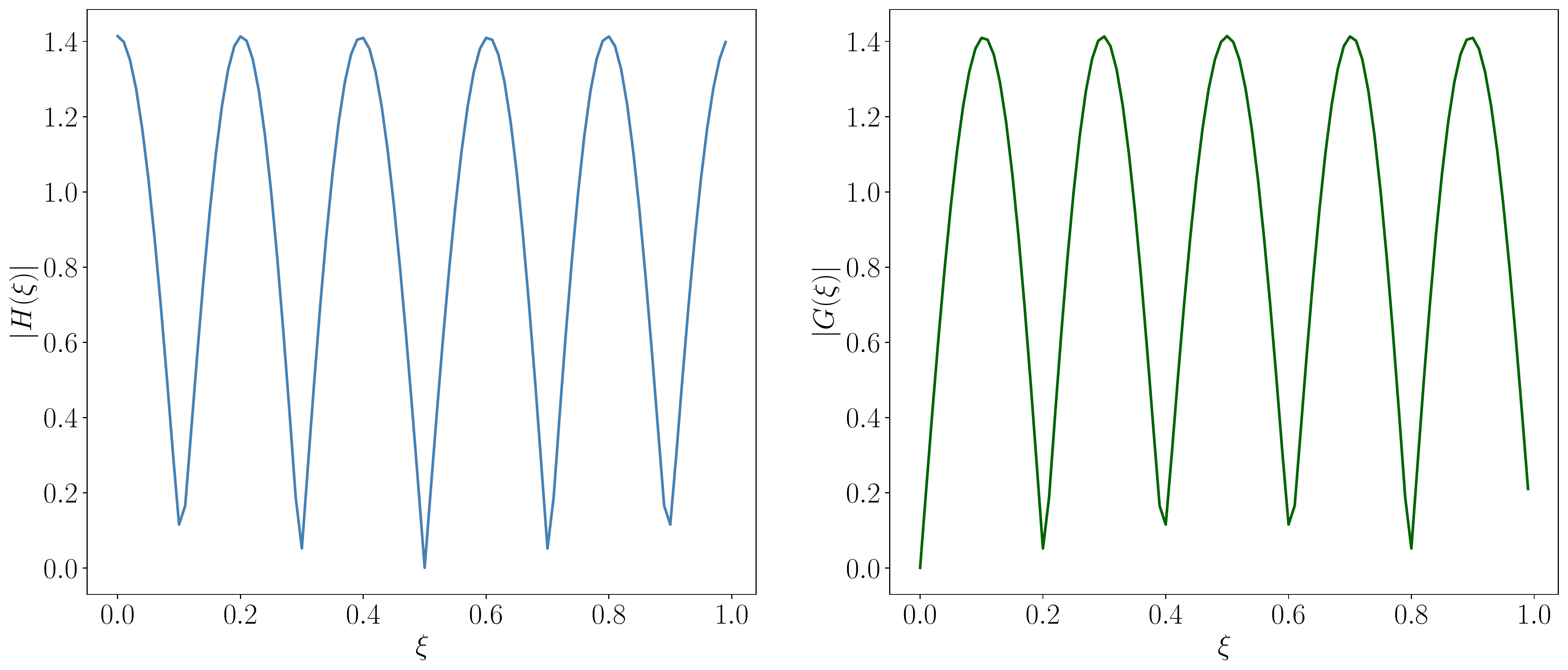} }}
	\subfloat[\centering Final refinement masks \label{fig:refinement_mask_final_0}]
	{{\includegraphics[width=0.49\columnwidth]{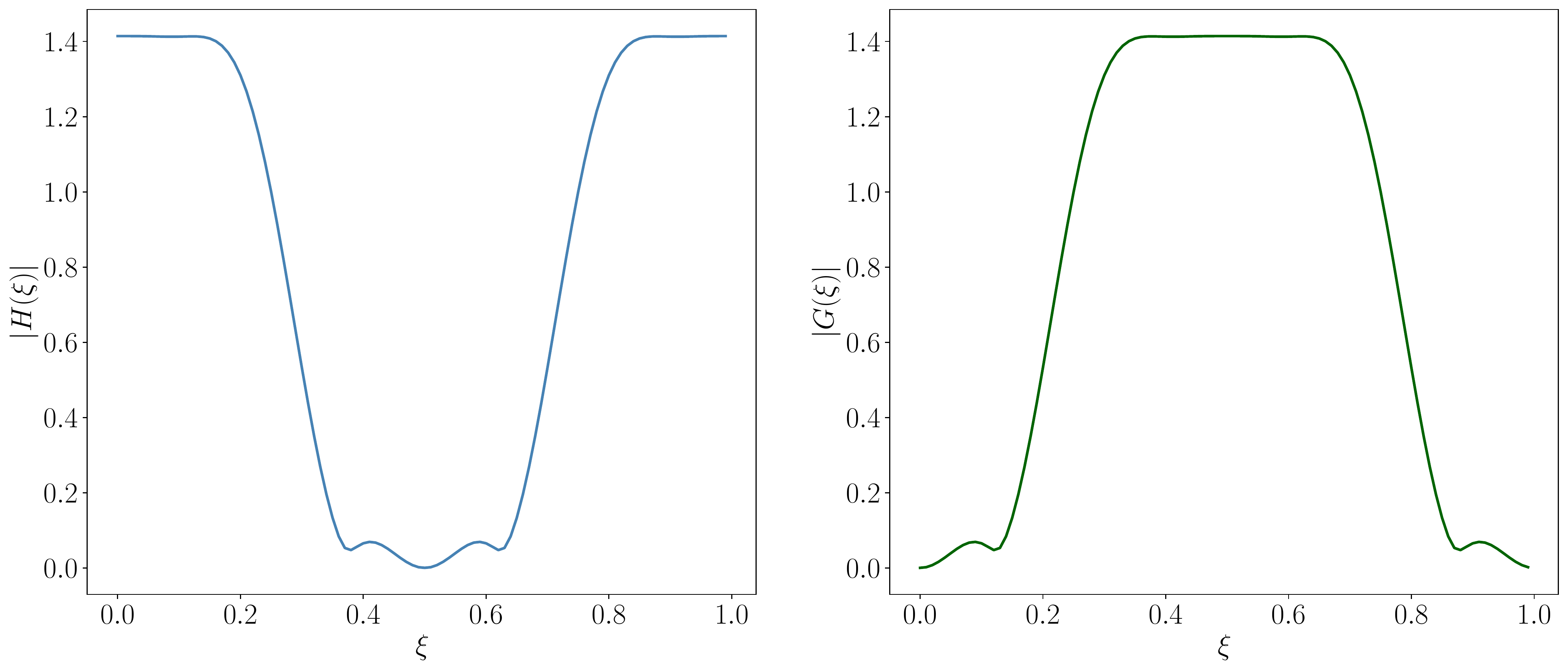} }}
	\\[2ex]    		   	
	\subfloat[\centering Initial refinement masks \label{fig:refinement_mask_init_1} ]
	{{\includegraphics[width=0.49\columnwidth]{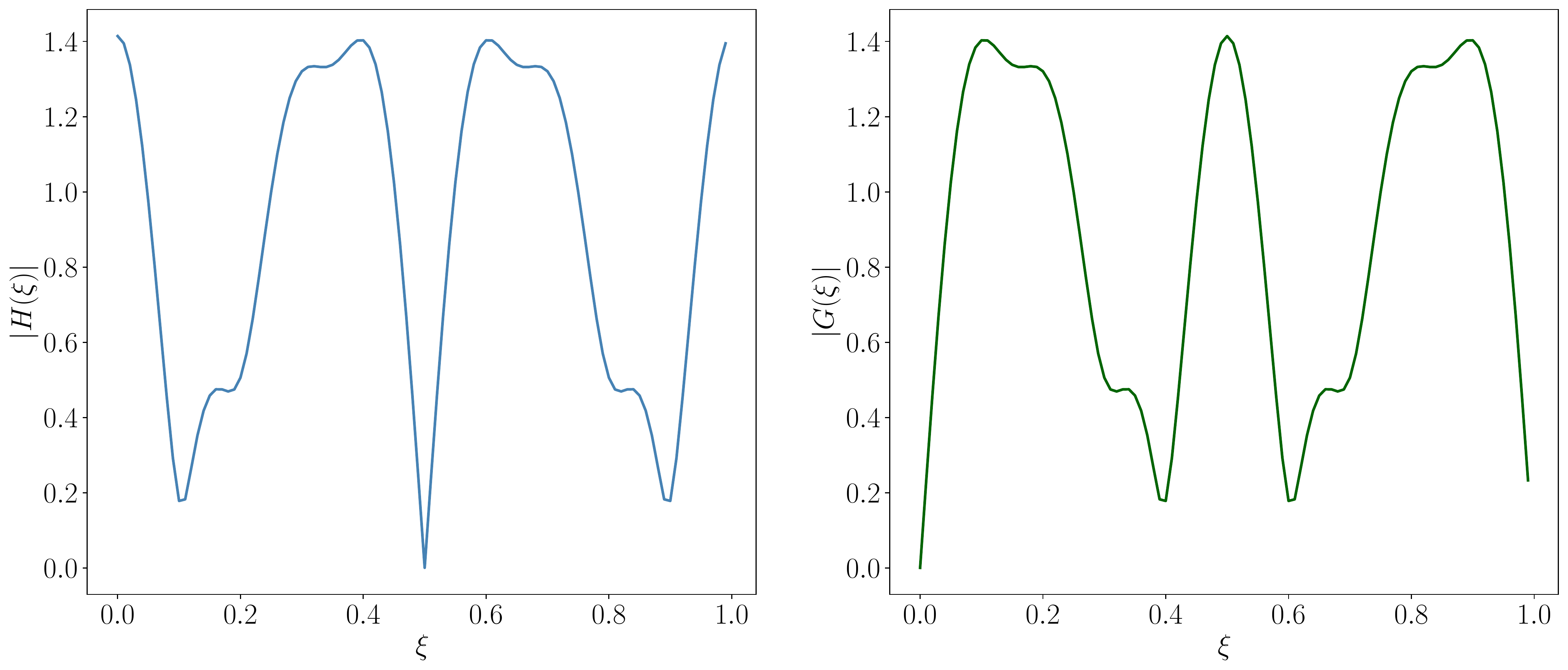} }}
	\subfloat[\centering Final refinement masks \label{fig:refinement_mask_final_1}]
	{{\includegraphics[width=0.49\columnwidth]{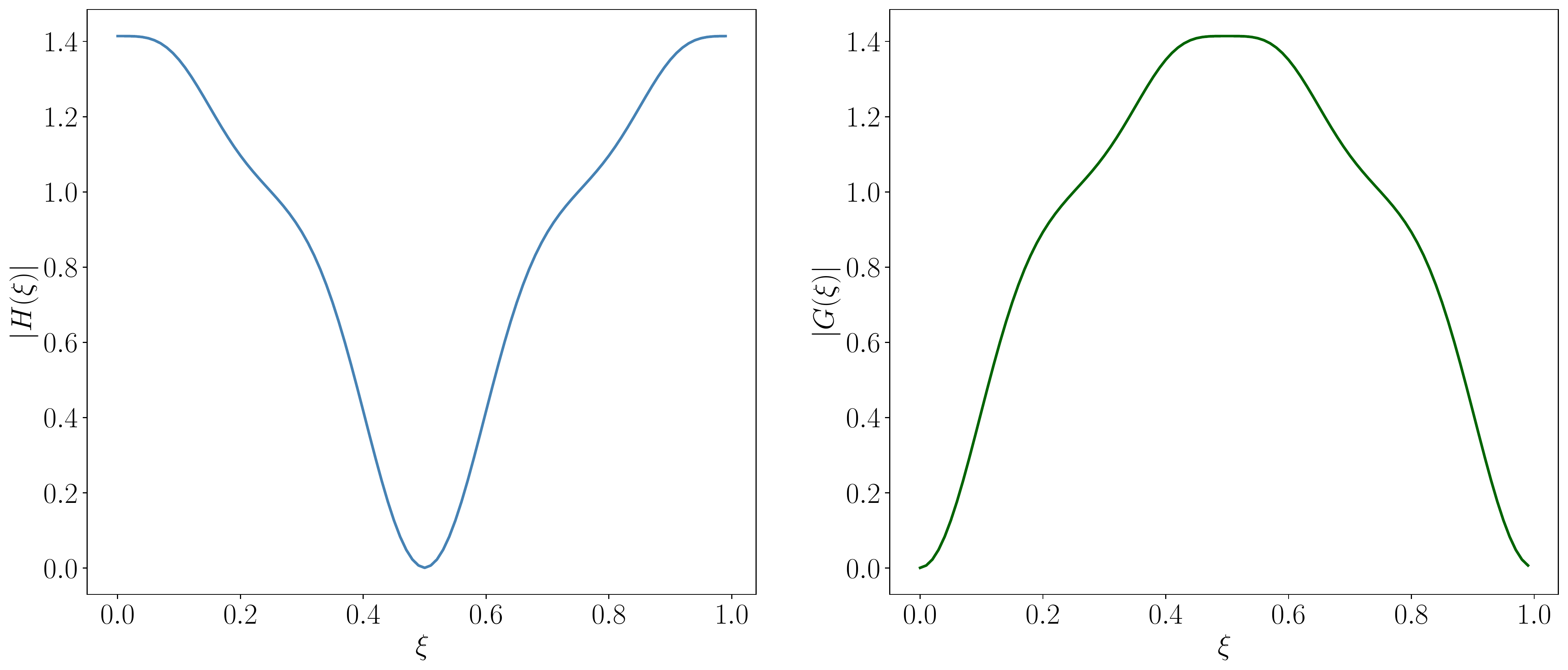} }}
	\caption{Refinement masks $H$ and $G$ associated to the low and high pass filters, respectively, of the wavelets depicted in Figure \ref{fig:wavelet_examples}. 
            ~\protect \subref{fig:refinement_mask_init_0}, ~\protect \subref{fig:refinement_mask_final_0} Refinement masks associated to the first spatial component.
	    ~\protect \subref{fig:refinement_mask_init_1}, ~\protect \subref{fig:refinement_mask_final_1} Refinement masks associated to the second spatial component.}
	\label{fig:refinement_examples}
\end{figure}

\section*{Acknowledgements}
We thank Joren Brunekreef for his helpful feedback and discussions. 

\bibliographystyle{unsrt}
\bibliography{bibliography}

\newpage
\appendix

\section{Computing discrete convolutions using the DFT}
\label{appendix:dft}
In this appendix, we recall how to compute the two-sided convolution using the Discrete Fourier Transform (DFT). For this purpose,
we first set up some terminology. Let $M \in \NN^{d}$ be a $d$-dimensional multi-index and let $\mathcal{A}_{M}$ denote the space
of two-sided $d$-dimensional $\CC$-valued sequences (or arrays) of order $M$, i.e., 
\begin{align*}
	\mathcal{A}_{M} : = \left \{ a \ \biggr \vert \ a : \prod_{i=1}^{d} \{ 1 - M_{i}, \ldots, M_{i} -1 \} \rightarrow \CC \right \}.
\end{align*}
The set $\mathcal{A}_{M}$ is a vector space over $\CC$ of dimension $\prod_{j=1}^{d} (2M_{j} - 1)$. As usual, we will write $a(k) = a_{k}$ 
for $1 - M \leq k \leq M-1$. Throughout this section inequalities involving multi-indices are interpreted component-wise. 
Similarly, we will denote the space of one-sided $d$-dimensional $\CC$-valued sequences or order $M$ by $\mathcal{A}^{+}_{M}$, i.e., 
\begin{align*}
    \mathcal{A}^{+}_{M} := \left \{ a \ \biggr \vert \ a : \prod_{i=1}^{d} \{ 0, \ldots, M_{i} -1 \} \rightarrow \CC \right \}.
\end{align*}

\paragraph{Convolution and multiplication of polynomials}
The two-sided convolution between sequences $a \in \mathcal{A}_{M}$ and $b \in \mathcal{A}_{N}$ is a new 
sequence $a \ast b \in \mathcal{A}_{M + N - 1}$ defined by 
\begin{align*}
	(a \ast b)_{k} := \sum_{ \substack{m + n = k \\ m, n \in \ZZ^{d}} } a_{m} b_{n}.
\end{align*}
Here we have omitted the ranges of $m$ and $n$ in the domain of summation to reduce clutter in the notation. It should be 
clear from the context, however, that $1 - M \leq m \leq M-1$ and $1 - N \leq n \leq N-1$. Strictly speaking, we should 
incorporate $M$ and $N$ into the notation for $\ast$ as well. However, since we may always embed $\mathcal{A}_{M}$ and $\mathcal{A}_{N}$
into $\mathcal{A}_{K}$ by padding with zeros, for any $K \geq M + N -1$, leaving the result of convolution unchanged, we will 
ignore this distinction. 

Convolutions can be efficiently computed using the DFT. To explain how to do so, we interpret $a$ and $b$ as the coefficients of Laurent polynomials 
$\mathcal{T}_{M}(a): \CC^{d} \setminus \{0\} \rightarrow \CC$ and $\mathcal{T}_{N}(b): \CC^{d}  \setminus \{0\}  \rightarrow \CC$, respectively, where
\begin{align*}
	\mathcal{T}_{M}(a)(z) = \sum_{\vert k \vert \leq M -1} a_{k} z^{k}, \quad \mathcal{T}_{N}(b)(z) = \sum_{\vert k \vert \leq N -1} b_{k} z^{k}.
\end{align*}
The motivation for this interpretation is that the product $\mathcal{T}_{M}(a)\mathcal{T}_{N}(b)$ has coefficients $a \ast b$. 
We will exploit this relationship to compute the desired convolution. 
First, note that $\mathcal{T}_{M}(a)$ can be characterized by evaluating it on $\tilde M := \prod_{i=1}^{d} (2M_{i} -1)$ appropriately chosen
points in $\CC^{d} \setminus \{0\}$. Here appropriate means that the evaluation operator mapping $a$ to the corresponding values of $\mathcal{T}_{M}(a)$ 
is an isomorphism on $\mathcal{A}_{M}$. Similarly, $\mathcal{T}_{N}(b)$ can be characterized by evaluation on $\tilde N := \prod_{i=1}^{d} (2N_{i} -1)$ 
appropriate points, and $\mathcal{T}_{M}(a) \mathcal{T}_{N}(b)$ by evaluation on $\tilde K := \prod_{i=1}^{d} (2K_{i} -1)$ appropriate points, where 
$K := M + N - 1$. The key observation here is that if we fix $\tilde K$ appropriately chosen points in $\CC^{d} \setminus \{0\}$, we may go back and forth 
between value and coefficient representations of $\mathcal{T}_{M}(a)\mathcal{T}_{N}(b)$ using the associated evaluation operator. Therefore, if the 
chosen evaluation operator and its inverse are analytically tractable, we can compute $a \ast b$ by evaluating $\mathcal{T}_{M}(a)\mathcal{T}_{N}(b)$.

\paragraph{The Discrete Fourier Transform}
An appropriate choice for evaluation points is the roots of unity. The associated evaluation operator is the 
DFT, which is analytically tractable and computationally efficient. Here we shall consider the DFT from a purely algebraic 
point of view and mostly forget about its relation with Fourier Analysis. The interpretation we adopt is that the DFT 
evaluates (one-sided) multivariate polynomials on an ``uniform discretization'' of the $d$-dimensional 
Torus $\mathbb{T}^{d} := \prod_{j=1}^{d} \mathbb{S}^{1}$. More precisely, for any $n \in \NN$, set $\omega_{n}:= 
e^{ -\frac{ 2 \pi i }{ n} }$ and define 
$
	\bm{\omega}_{M} :=  \left( \omega_{M_{1}}, \ldots, \omega_{M_{d}} \right) 
$
for $M \in \NN^{d}$. We refer to $\{ \bm{\omega}^{j}_{M} : 0 \leq j \leq M - 1 \}$ as the $M$-th order roots of unity. 
The $M$-th order DFT is the map $\DFT_{M}: \mathcal{A}_{M}^{+} \rightarrow \mathcal{A}_{M}^{+}$ defined by
\begin{align*}
	(\DFT_{M}( a ))_{j} := \mathcal{P}_{M}(a)( \bm{\omega}_{M}^{j} ), \quad
	\mathcal{P}_{M}(a)(z) := \sum_{0 \leq k \leq M-1} a_{k} z^{k}. 
\end{align*}
The DFT is an ``appropriate'' evaluation operator, i.e., it is an isomorphism. It characterizes the coefficients 
of a polynomial through evaluation at the roots of unity.  

\paragraph{Evaluating Laurent-polynomials at the roots of unity}
There is a slight difference between our objective, evaluating Laurent-polynomials, and the
choice of evaluation operator (the DFT), which evaluates ordinary (one-sided) polynomials. 
In order to use the DFT for our purposes, we need to relate the evaluation of a Laurent polynomial
at the roots of unity with the evaluation of an ordinary polynomial. This can be accomplished by 
exploiting the symmetry of the roots of unity. 

Let $a \in \mathcal{A}_{M}$ and $1 - M \leq j \leq M-1$ be arbitrary. Evaluation of the associated 
Laurent-polynomial at a root of unity yields
\begin{align}  
    \label{eq:eval_roots_unity}
	\mathcal{T}_{M}(a)( \bm{\omega}_{2M -1}^{j} ) = 
	\sum_{1 - M_{1} \leq k_{1} \leq M_{1} -1} \cdots \sum_{1 - M_{d} \leq k_{d} \leq M_{d} -1}  
	a_{k} \ \omega_{2M_{1} -1}^{j_{1}k_{1}} \ldots \omega_{2M_{d} -1}^{j_{d}k_{d}}. 
\end{align}
The right-hand side of \eqref{eq:eval_roots_unity} can be rewritten as a sum over positive indices only. 
Subsequently, we can exploit the symmetry of the roots of unity and recognize the result as evaluating 
a one-sided polynomial, i.e., as a DFT. More precisely, define $S_{M} : \mathcal{A}_{M} \rightarrow 
\mathcal{A}^{+}_{2M - 1}$ by $(S_{M}(a))_{k} := a_{\hat k}$, where 
\begin{align*}
	\hat k_{l} := 
	\begin{cases}
		k_{l} & 0 \leq k_{l} \leq M_{l} -1, \\
		k_{l} - 2 M_{l} + 1, & M_{l} \leq k_{l} \leq 2(M_{l} -1),
	\end{cases}
	\quad 1 \leq l \leq d. 
\end{align*}
Roughly speaking, the map $S_{M}$ places the components of $a$ with negative indices ``after'' 
the ones with positive indices. In numerical implementations, this operation is commonly referred 
to as a ``fft shift''. This reordering can be used to recognize \eqref{eq:eval_roots_unity} as a DFT:
\begin{align*}
	\mathcal{T}_{M}(a)( \bm{\omega}_{2M -1}^{j} ) &= 
	\sum_{0 \leq k_{1} \leq 2(M_{1} -1)} \cdots \sum_{0 \leq k_{d} \leq 2(M_{d} -1)}  
	(S_{M}(a))_{k} \ \omega_{2M_{1} -1}^{j_{1}k_{1}} \ldots \omega_{2M_{d} -1}^{j_{d}k_{d}} 
	 \\[2ex] &= (\DFT_{M} S_{M}(a))_{j}
\end{align*}
for $0 \leq j \leq 2(M -1)$. This shows that evaluation of $\mathcal{T}_{M}(a)$ at the roots of unity
$\{ \bm{\omega}^{j}_{2M -1} : 0 \leq j \leq 2(M -1) \}$ is equivalent to computing
$(\DFT_{M} \circ S_{M})(a)$. 

\paragraph{Convolution using the DFT}
Finally, we explain how to compute the two-sided convolution $a \ast b$. First, we characterize the
coefficients of the Laurent-polynomial $\mathcal{T}_{M}(a) \mathcal{T}_{N}(b)$ by evaluating it at the roots of unity 
$\{ \bm{\omega}^{j}_{2K -1} : 0 \leq j \leq 2(K -1) \}$. For this purpose, extend $a$ and $b$ to 
sequences in $\mathcal{A}_{K}$ by padding with zeros. More formally, for each $I, J \in \NN^{d}$ such that $J > I$, 
define a padding operator $Z^{J}_{I} : \mathcal{A}_{I} \rightarrow \mathcal{A}_{J}$ by
\begin{align*}
	(Z^{J}_{I}(a))_{k} := 
	\begin{cases}
		a_{k} & 1 - I \leq k \leq I -1, \\
		0 & \text{otherwise}. 
	\end{cases}
\end{align*}
Then evaluation of $\mathcal{T}_{M}(a) \mathcal{T}_{N}(b)$ at the $K$-th order roots of unity corresponds to computing
\begin{align}
    \label{eq:eval_TM_times_TN}
	\left( \DFT_{K} \circ S_{K} \circ Z^{K}_{M} \right)(a)  \odot \left( \DFT_{K} \circ S_{K} \circ Z^{K}_{N} \right)(b),
\end{align} 
where $\odot$ denotes the Hadamard product. Since \eqref{eq:eval_TM_times_TN} is an equivalent representation of 
the Laurent-polynomial $\mathcal{T}_{M}(a) \mathcal{T}_{N}(b)$, which has coefficients $a \ast b$, we conclude that 
\begin{align*}
	a \ast b = S_{k}^{-1} \circ \DFT_{K}^{-1} 
	\biggl( 
	\left( \DFT_{K} \circ S_{K} \circ Z^{K}_{M} \right)(a)  \odot \left( \DFT_{K} \circ S_{K} \circ Z^{K}_{N} \right)(b)
	\biggr).
\end{align*}

\subsection{Periodic Convolutions}
\label{appendix:periodic_conv}
In certain applications, e.g., when dealing with wavelet expansions of periodic signals, one is given an array $a \in \mathcal{A}_{M}$ and
wishes to compute $\tilde a \ast b$, where $\tilde a \in \CC^{\ZZ^{d}}$ is the $(2M-1)$-periodic extension of $a$.  This type of convolution 
is commonly referred to as periodic circular convolution. In particular, note that it suffices to compute $(\tilde a \ast b)_{k}$ for $1 - M \leq k \leq M-1$ only,
since $\tilde a \ast b$ is $(2M-1)$ periodic as well. The periodic convolution can be efficiently computed using the DFT as well provided 
the periodicity has been appropriately taken into account. This is necessary to avoid boundary artifacts, see the explanation below. 

First, observe that the sum 
\begin{align*}
	(\tilde a \ast b)_{k} = \sum_{ 1 - N \leq n \leq N -1 } \tilde a_{k-n} b_{n}, \quad k \in \ZZ^{d},
\end{align*}
contains only a finite number of nonzero terms, since $b \in \mathcal{A}_{N}$ is finite. Furthermore, 
for $1 - M \leq k \leq M -1$, we do not need the full periodic extension of $a$, but only a partial (finite) 
periodic extension $P_{MN}(a)$. More precisely, define $P_{MN}: \mathcal{A}_{M} \rightarrow \mathcal{A}_{M + N -1}$
by $(P_{MN}(a))_{k} = a_{\tilde k}$ for $2 - M - N \leq k \leq M + N - 2$, where 
\begin{align*}
	\tilde k_{j} := 
	\begin{cases}
		k_{j} + 2M_{j} - 1, & 2 - M_{j} - N_{j} \leq k_{j} \leq - M_{j}, \\
		k_{j}, & 1 - M_{j} \leq k_{j} \leq M_{j} -1, \\
		k_{j} + 1 - 2M_{j}, & M_{j} \leq k_{j} \leq M_{j} + N_{j} -2,	
	\end{cases}
	\quad 1 \leq j \leq d.
\end{align*}
Then $(P_{MN}(a) \ast b)_{k}  = (\tilde a \ast b)_{k}$ for $1 - M \leq k \leq M-1$.

Finally, we apply the tools developed in the previous section to compute $(\tilde a \ast b)_{k}$ for $1 - M \leq k \leq M-1$ using the DFT. 
More precisely, set 
\begin{align*}
	\hat a := \DFT_{\tilde K} \circ S_{\tilde K} \circ Z^{\tilde K}_{M + N - 1} \circ P_{MN}(a), \quad
	\hat b :=  \DFT_{\tilde K} \circ S_{\tilde K} \circ Z^{\tilde K}_{N}(b), 
\end{align*}
where $\tilde K: = M + 2(N-1)$, then 
\begin{align*}
	( \tilde a \ast b)_{1 - M \leq k \leq M-1} = \left( S_{K}^{-1} \circ \DFT_{\tilde K}^{-1} \left( \hat a \odot \hat b \right) \right)_{1 - M \leq k \leq M-1}. 
\end{align*}

\section{Preprocessing}
In this section we provide the details of our preprocessing steps. 

\subsection{Truncation Fourier coefficients}
\label{appendix:truncation}
The magnitude of the approximated Fourier coefficients will typically stagnate and stay constant (approximately) beyond some critical order, 
since all computations are performed in finite (single) precision. We locate this critical order $m^{\ast}_{0}(s) \in \NN$ for each component $s \in \{1, 2\}$, 
if present, by iteratively fitting the best line, in the least squares sense, through the points 
\begin{align*}
	\left \{ \left(m,  \left \Vert \left( \left \vert \left[ \tilde \gamma_{\tilde m} \right]_{s} \right \vert \right)_{\tilde m=m_{0}}^{m} \right \Vert_{1} \right) : m_{0} \leq m \leq N -1 \right \}, 
	\quad 1 \leq m_{0} \leq N-1. 
\end{align*}
We iterate this process until the residual is below a prescribed threshold $\delta_{N} >0$. In practice, 
we set $\delta_{N} = 0.1$. The Fourier coefficients with index strictly larger than $m_{0}^{\ast}(s)$ are set to zero. 

\subsection{Consistent parameterizations}
\label{appendix:midpoint}
To have consistent parameterizations we enforce that all contours start at angle zero at time zero relative to the midpoint 
$c = (c_{1}, c_{2} ) \in \RR^{2}$ of the region of interest $R$. This is accomplished by exploiting the Fourier representation of $\gamma$. 
More precisely, let 
\begin{align*}
    \gamma(t) = \sum_{\vert m \vert \leq N-1} \tilde \gamma_{m} e^{i \omega(\tau)m t}, \quad \omega(\tau) = \frac{2 \pi}{\tau}, 
\end{align*}
be the initial contour with Fourier coefficients $\eta :=(\tilde \gamma_{m})_{m = 1-N}^{N-1}$.
The midpoint $c$ of the region enclosed by $\gamma$ is given by 
\begin{align}
	c_{s}  &= \frac{1}{ \lambda(R)} \int_{R} u_{s} \ \mbox{d} \lambda(u_{1}, u_{2}) 
	= (-1)^{s} \frac{ \left(  [\eta]_{1} \ast
	 [\eta]_{2} \ast [\eta']_{s} \right)_{0} }{  \left( [\eta]_{1} \ast [\eta' ]_{2} \right)_{0}},
	\quad s \in \left \{1, 2  \right \}
\end{align}
by Green's Theorem. Here $\lambda$ denotes the Lebesgue measure on $\RR^{2}$ and $[\eta]_{s}$, $[\eta']_{s}$ are the Fourier coefficients of $[\gamma]_{s}$
and its derivative, respectively. 

We can now compute the desired parameterization by determining
$t_{0} \in [0, \tau]$ such that 
\begin{align*}
    \text{arccos} \left( \dfrac{[ \gamma (- t_{0}) - c ]_{1}}{ \Vert \gamma(-t_{0}) - c \Vert_{2}} \right) \approx 0,
\end{align*}
and then use the shifted parameterization $t \mapsto \gamma(t - t_{0})$.  
While $t_{0}$ can be easily found using Newton's method, it suffices in practice to simply re-order $y$ from the start, 
before computing the Fourier coefficients of $\gamma$. More precisely, we first define a shift $\tilde y$ of $y$ by 
\begin{align*}
    \tilde y_{k} := y_{k \ + \ k^{\ast} \ \text{mod} \ n_{p}},  
    \quad k^{\ast} := \text{argmin} \left \{ \text{arccos} \left( \dfrac{[ y_{k} - c ]_{1}}{ \Vert y_{k} - c\Vert_{2}} \right) \right \}_{k = 0}^{n_{p} - 1},
    \quad 0 \leq k \leq n_{p} -1,
\end{align*}
and then compute the Fourier coefficients of the resulting curve. 

\section{Figures}
\label{sec:figures}
\subsection{Wavelets}
\label{sec:figures_wavelets}
In this section we show examples of initialized and task-optimized wavelets.
\newpage
\subsubsection{Spleen - first spatial component}
\begin{figure}[!b]
	\centering
	\subfloat[\centering Order $3$ - initial wavelet ]{{\includegraphics[width=0.44\columnwidth]{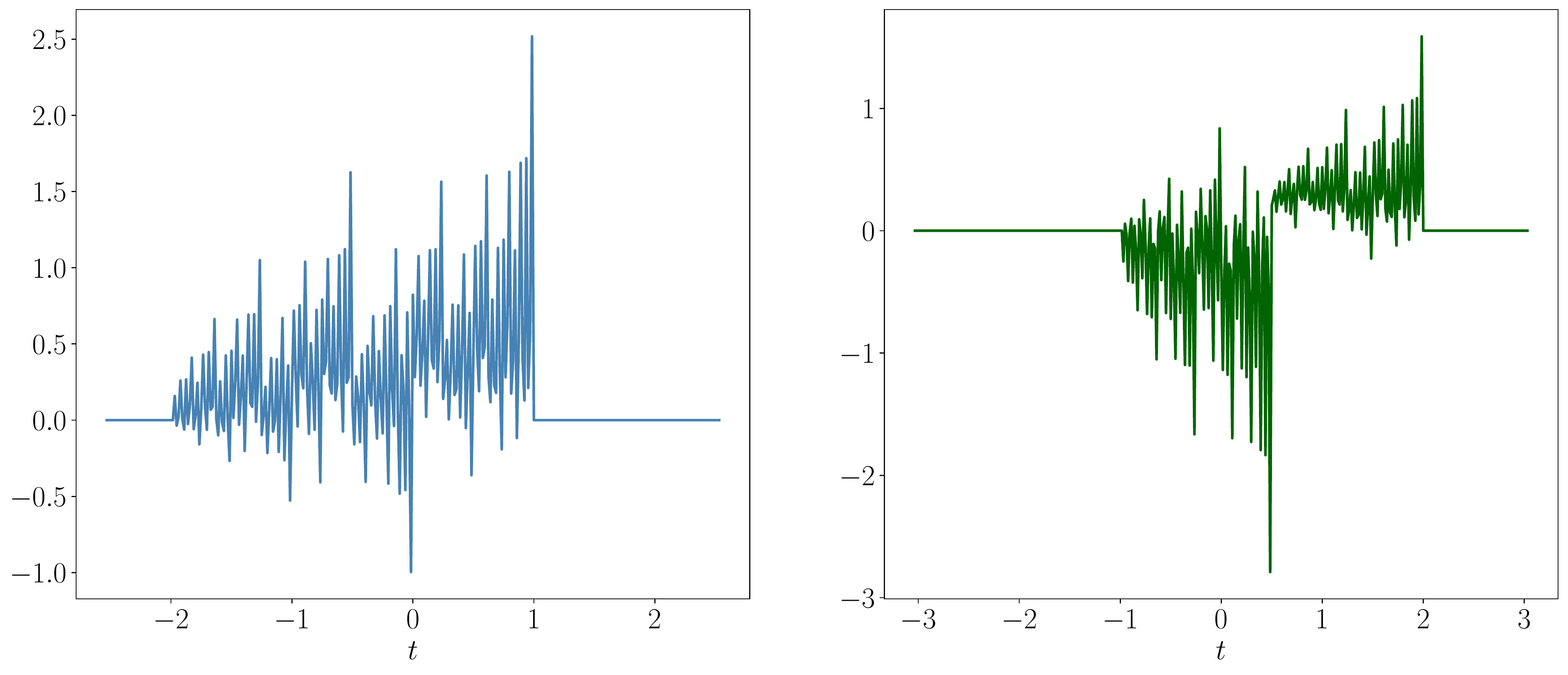} }}
	\subfloat[\centering Order $3$ - task-optimized wavelet ]{{\includegraphics[width=0.44\columnwidth]{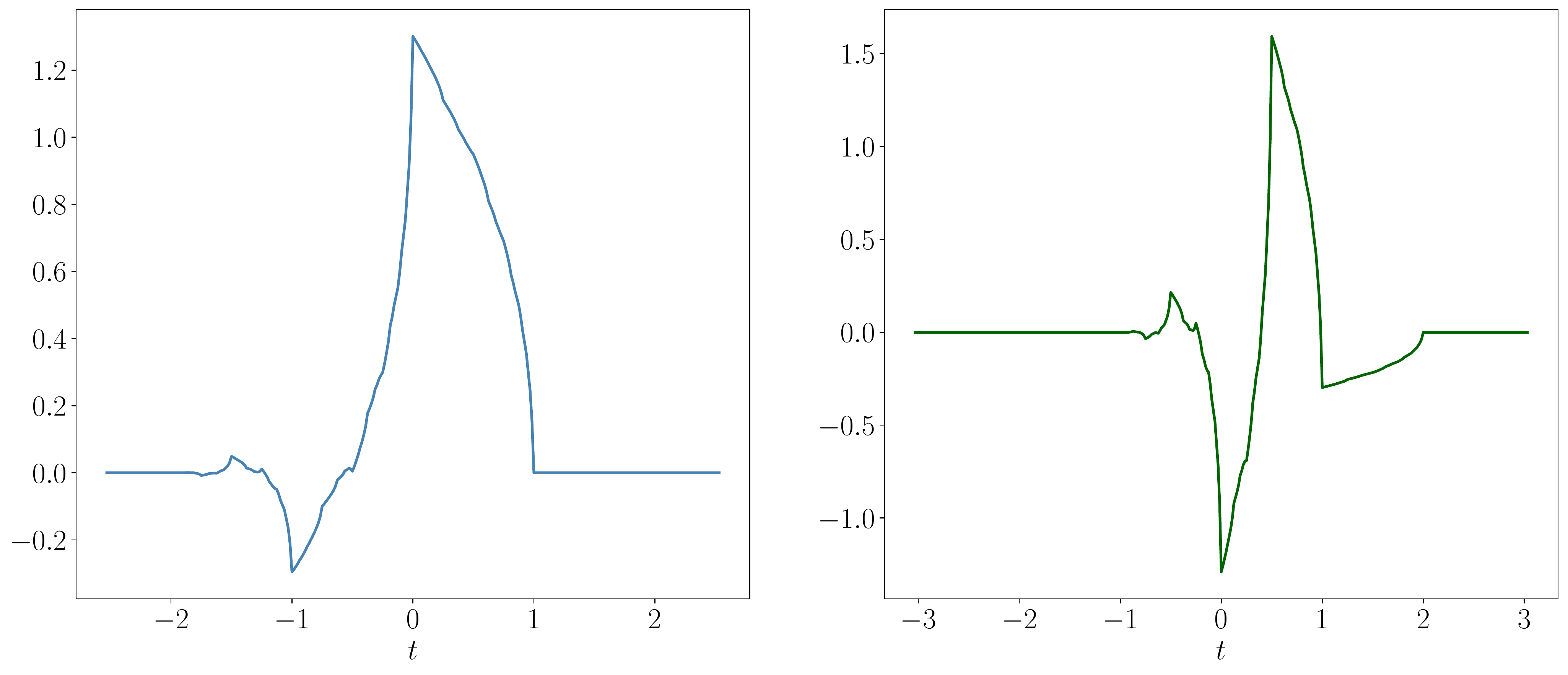} }}
	\\[2ex]    
	\subfloat[\centering Order $4$ - initial wavelet ]{{\includegraphics[width=0.44\columnwidth]{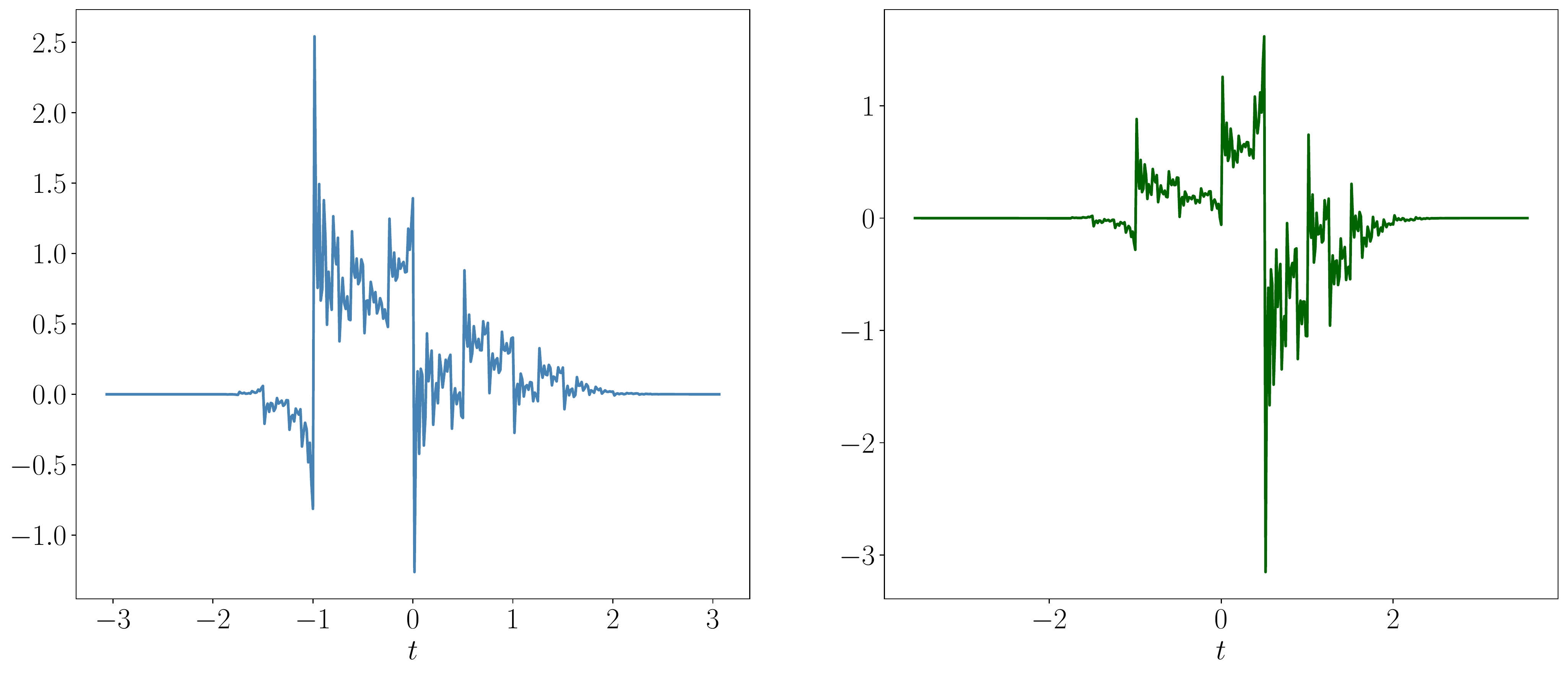} }}
	\subfloat[\centering Order $4$ - task-optimized wavelet ]{{\includegraphics[width=0.44\columnwidth]{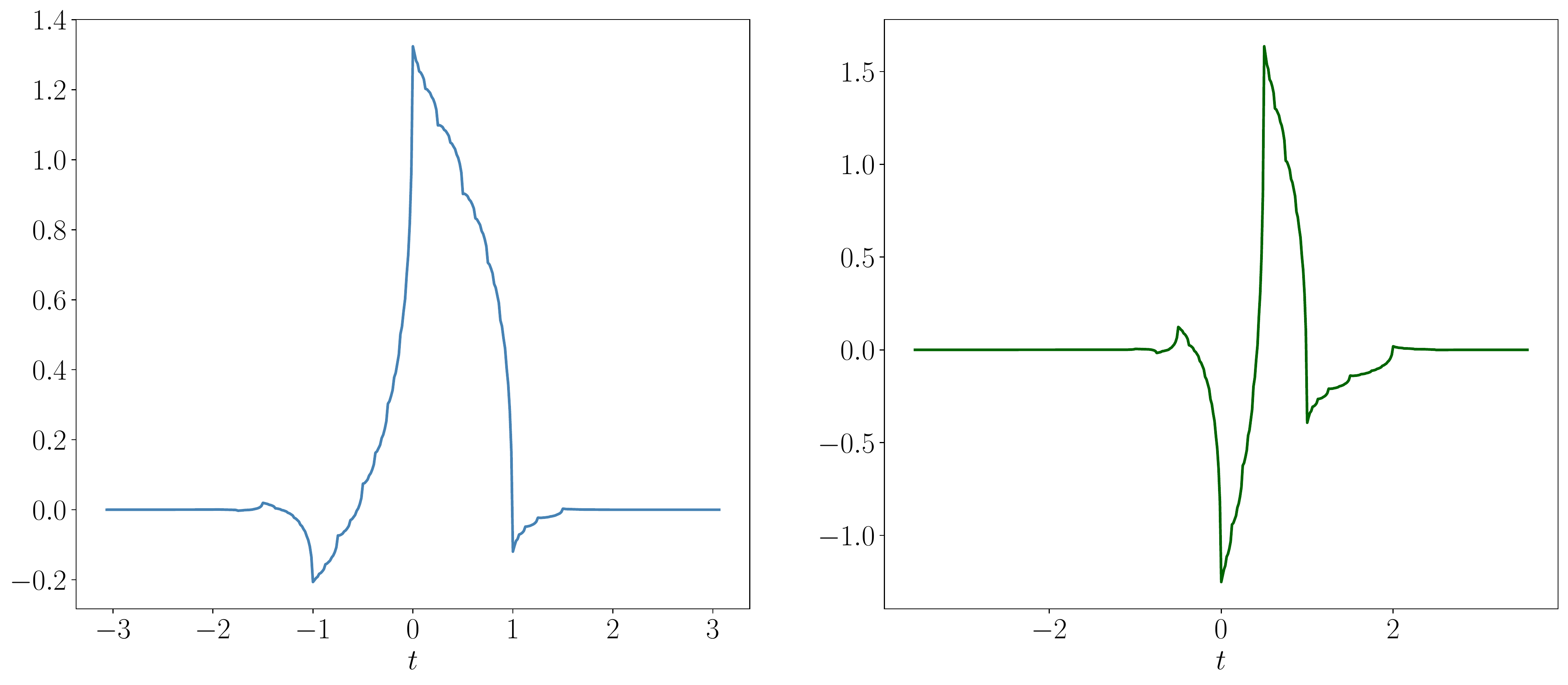} }}
	\\[2ex]    	
	\subfloat[\centering Order $5$ - initial wavelet ]{{\includegraphics[width=0.44\columnwidth]{spleen/order_5_epoch_0_comp_0} }}
	\subfloat[\centering Order $5$ - task-optimized wavelet ]{{\includegraphics[width=0.44\columnwidth]{spleen/order_5_epoch_250_comp_0} }}
	\\[2ex]    	
	\subfloat[\centering Order $6$ - initial wavelet ]{{\includegraphics[width=0.44\columnwidth]{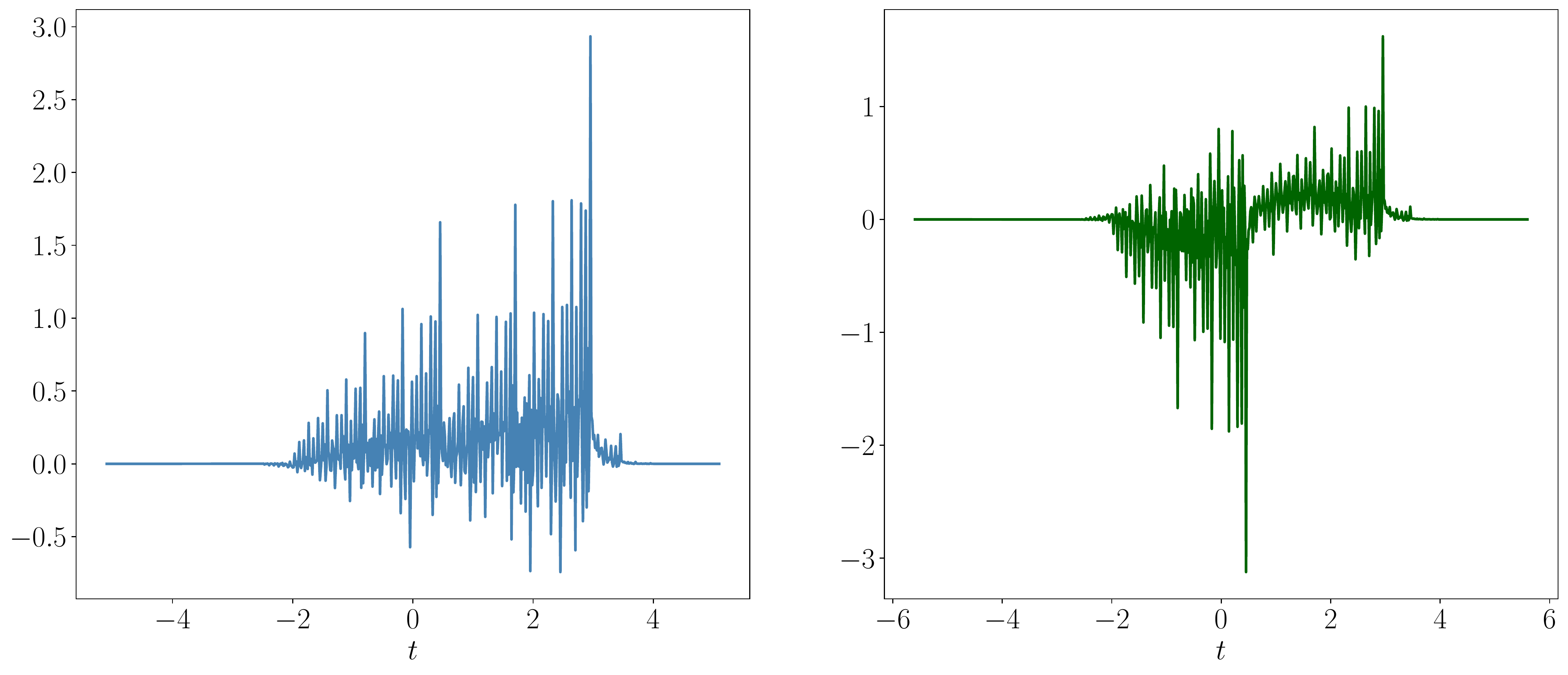} }}
	\subfloat[\centering Order $6$ - task-optimized wavelet ]{{\includegraphics[width=0.44\columnwidth]{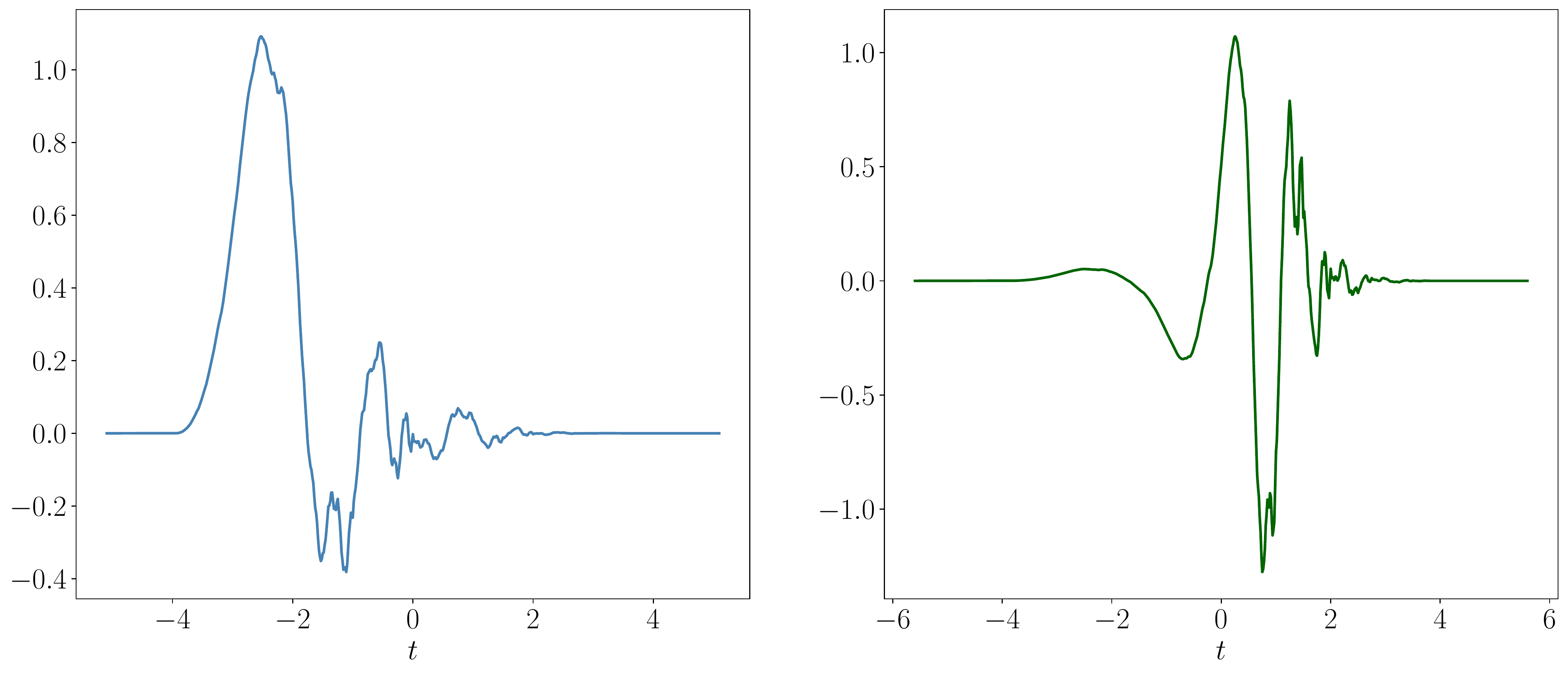} }}
	\\[2ex]    	
	\subfloat[\centering Order $7$ - initial wavelet ]{{\includegraphics[width=0.44\columnwidth]{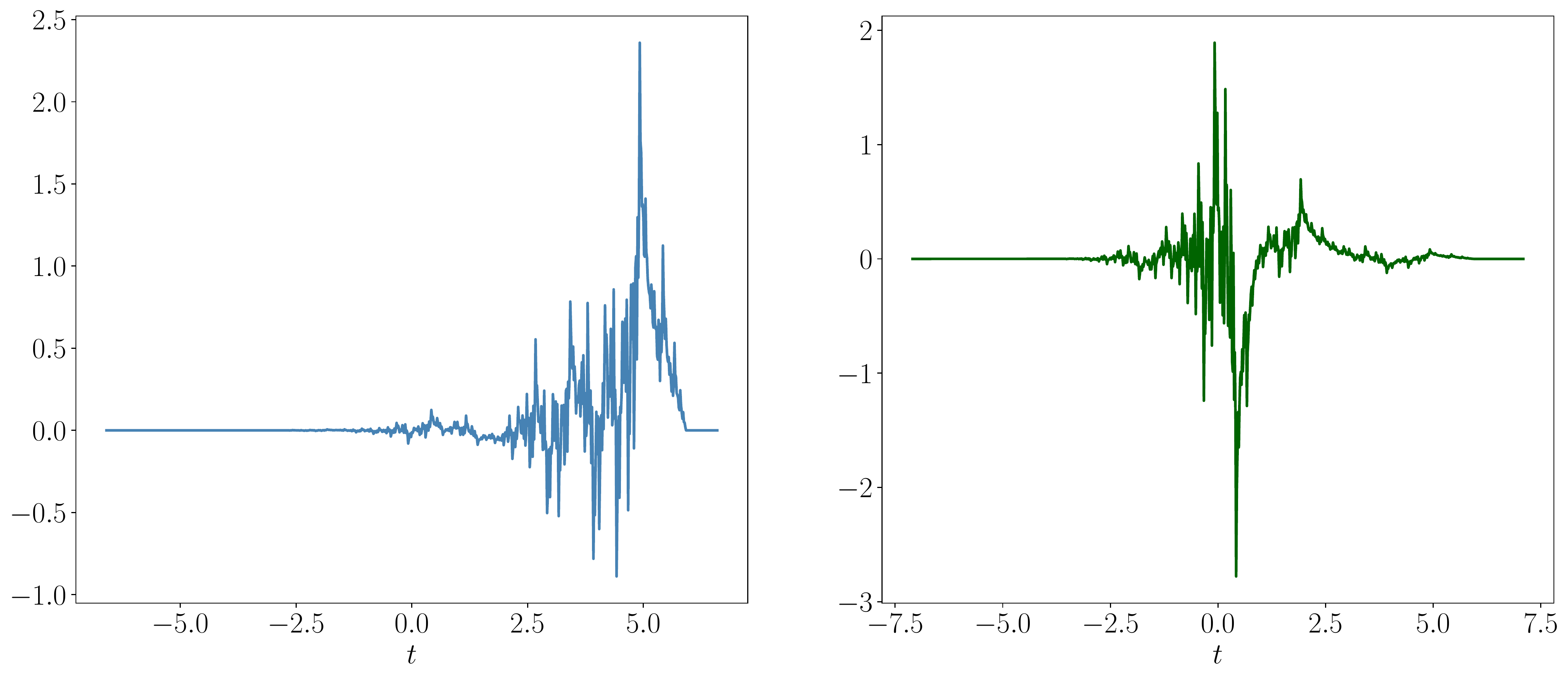} }}
	\subfloat[\centering Order $7$ - task-optimized wavelet ]{{\includegraphics[width=0.44\columnwidth]{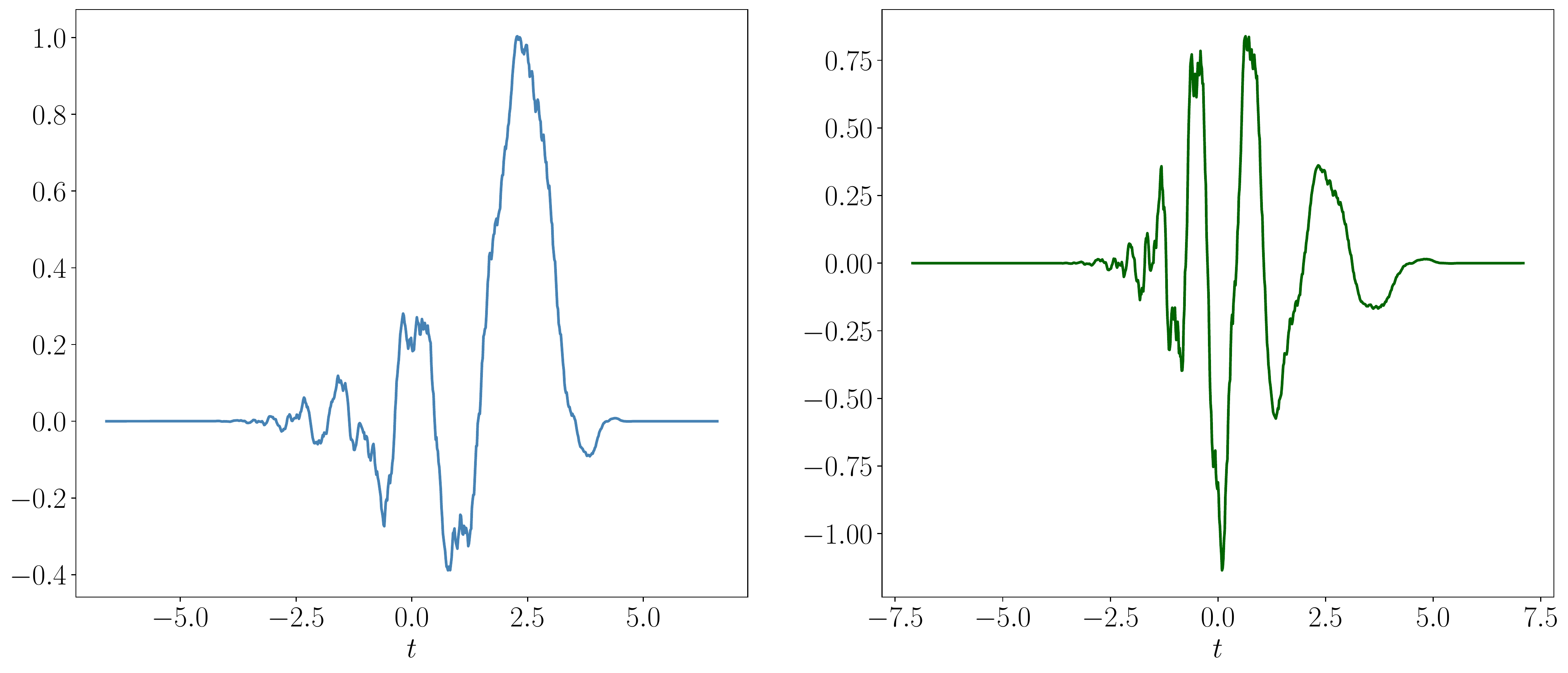} }}
	\\[2ex] 
	\subfloat[\centering Order $8$ - initial wavelet ]{{\includegraphics[width=0.44\columnwidth]{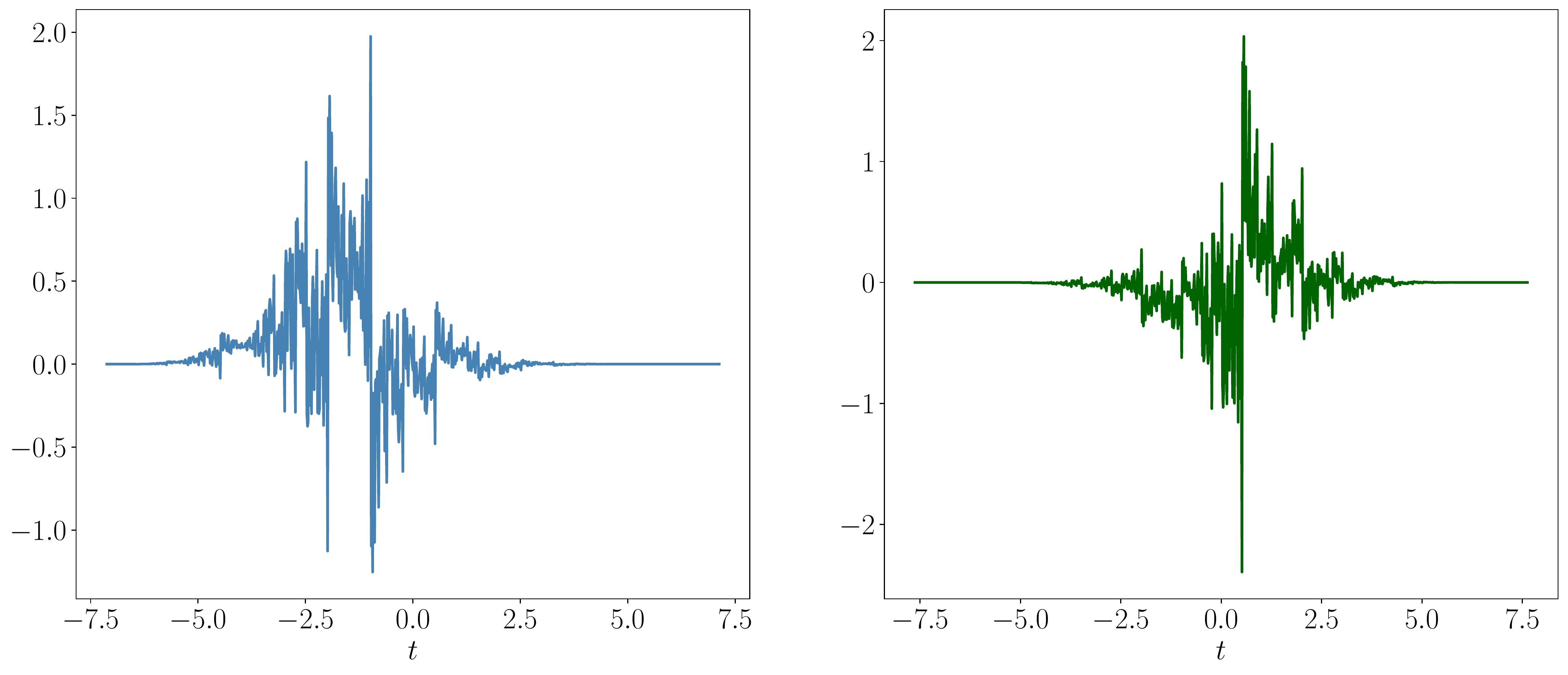} }}
	\subfloat[\centering Order $8$ - task-optimized wavelet ]{{\includegraphics[width=0.44\columnwidth]{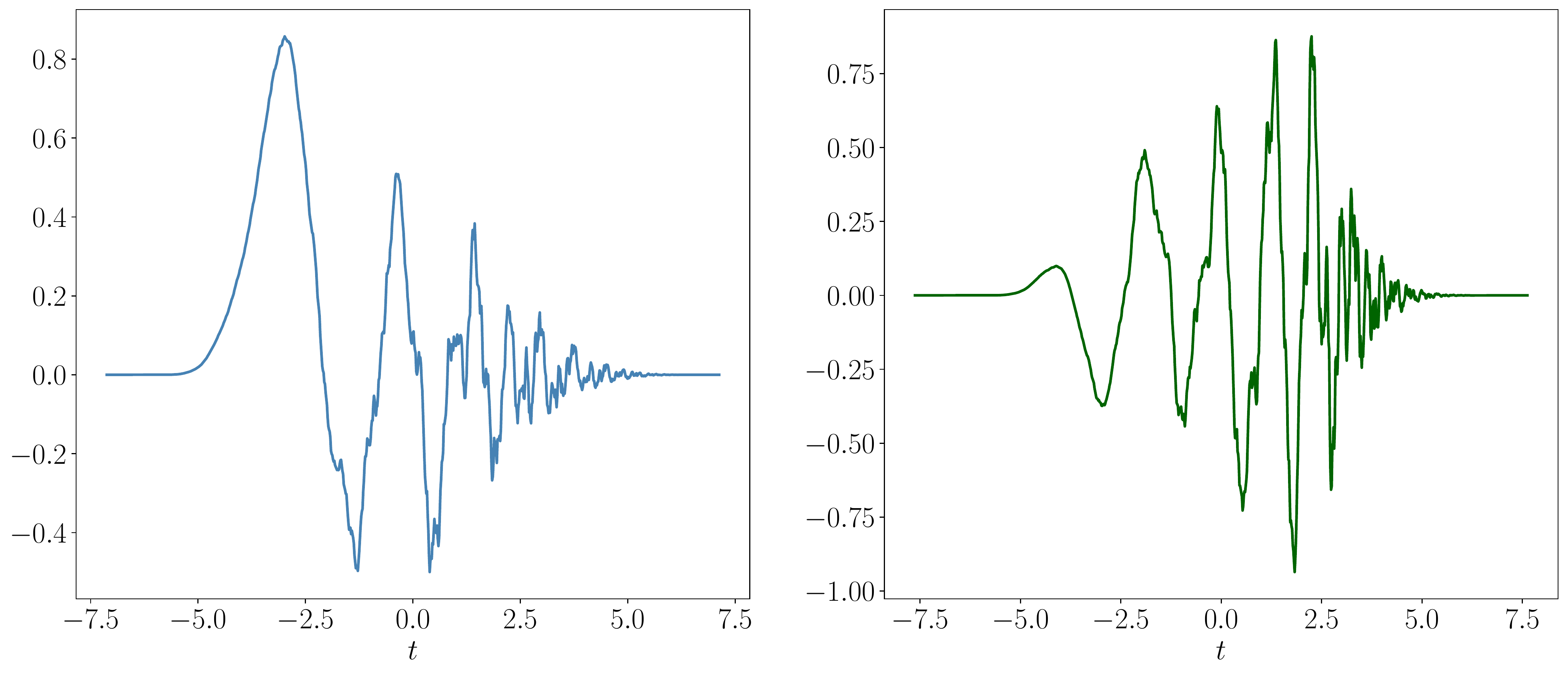} }}
	\\[2ex]    	   	
	\label{fig:spleen_wavelets_comp_0}
\end{figure}
\newpage
\subsubsection{Spleen - second spatial component}
\begin{figure}[!b]
	\centering
	\subfloat[\centering Order $3$ - initial wavelet ]{{\includegraphics[width=0.44\columnwidth]{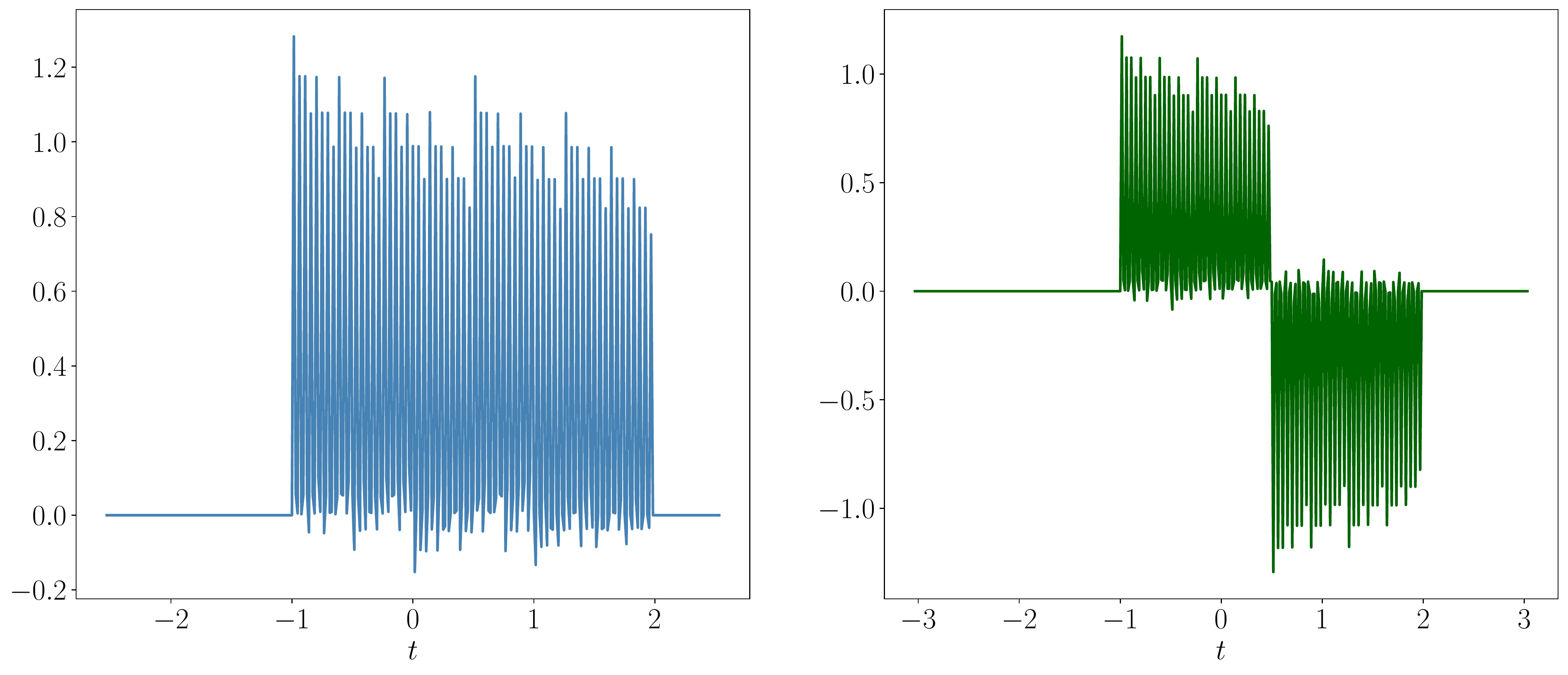} }}
	\subfloat[\centering Order $3$ - task-optimized wavelet ]{{\includegraphics[width=0.44\columnwidth]{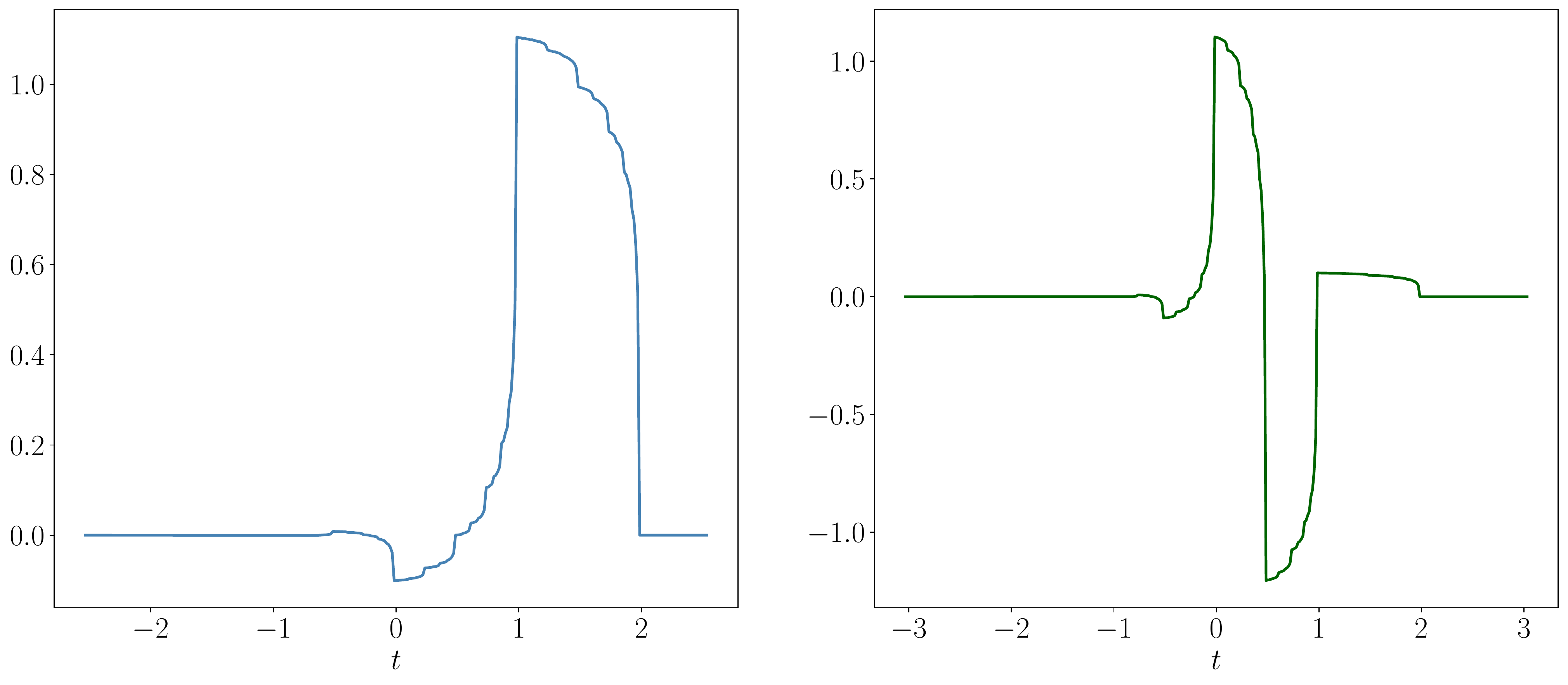} }}
	\\[2ex]    
	\subfloat[\centering Order $4$ - initial wavelet ]{{\includegraphics[width=0.44\columnwidth]{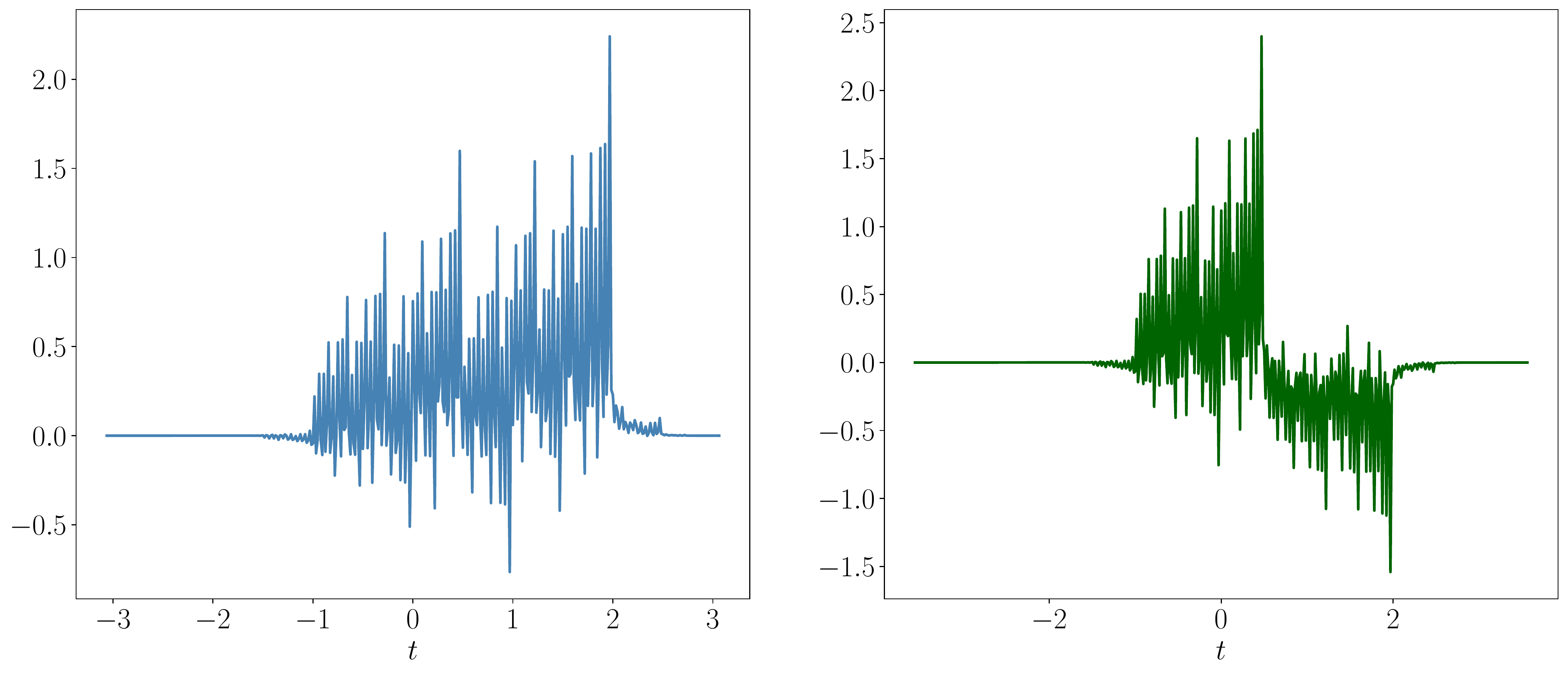} }}
	\subfloat[\centering Order $4$ - task-optimized wavelet ]{{\includegraphics[width=0.44\columnwidth]{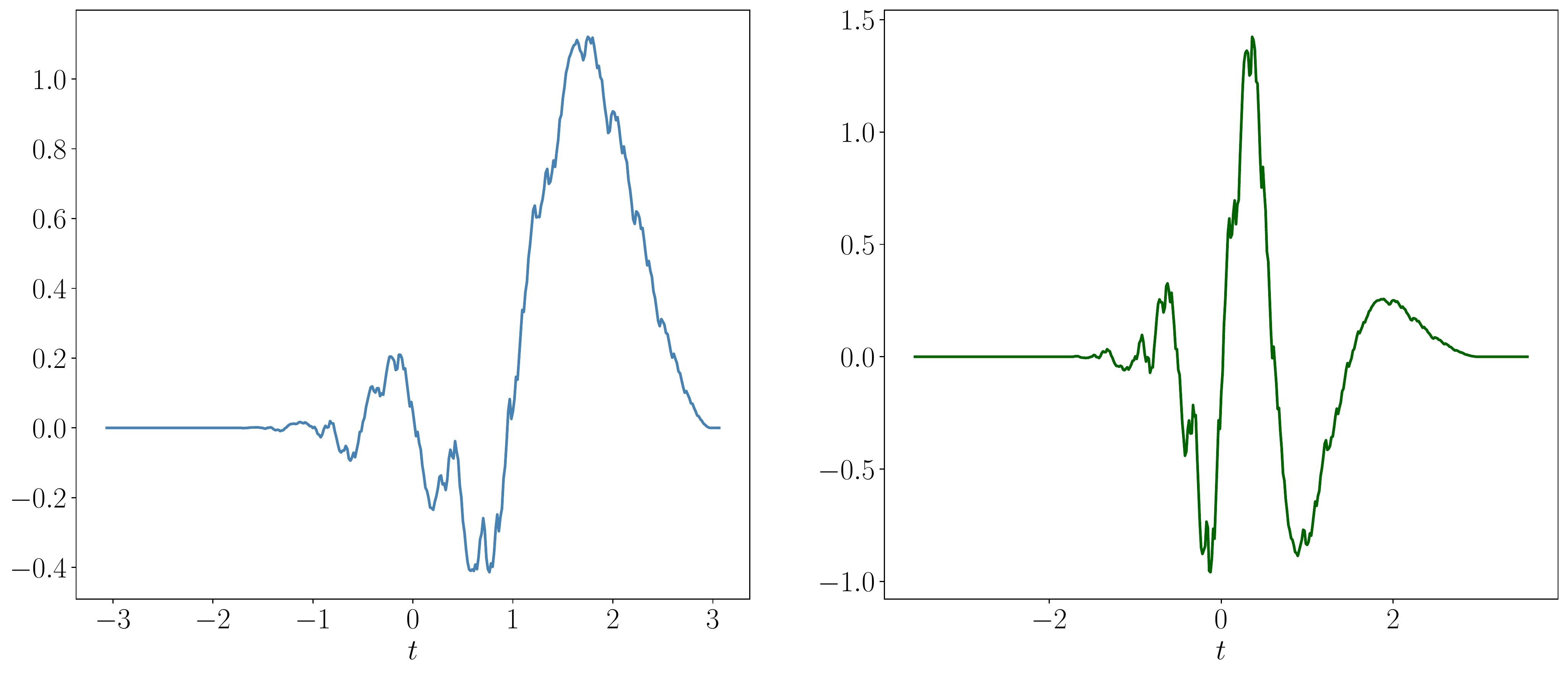} }}
	\\[2ex]    	
	\subfloat[\centering Order $5$ - initial wavelet ]{{\includegraphics[width=0.44\columnwidth]{spleen/order_5_epoch_0_comp_1} }}
	\subfloat[\centering Order $5$ - task-optimized wavelet ]{{\includegraphics[width=0.44\columnwidth]{spleen/order_5_epoch_250_comp_1} }}
	\\[2ex]    	
	\subfloat[\centering Order $6$ - initial wavelet ]{{\includegraphics[width=0.44\columnwidth]{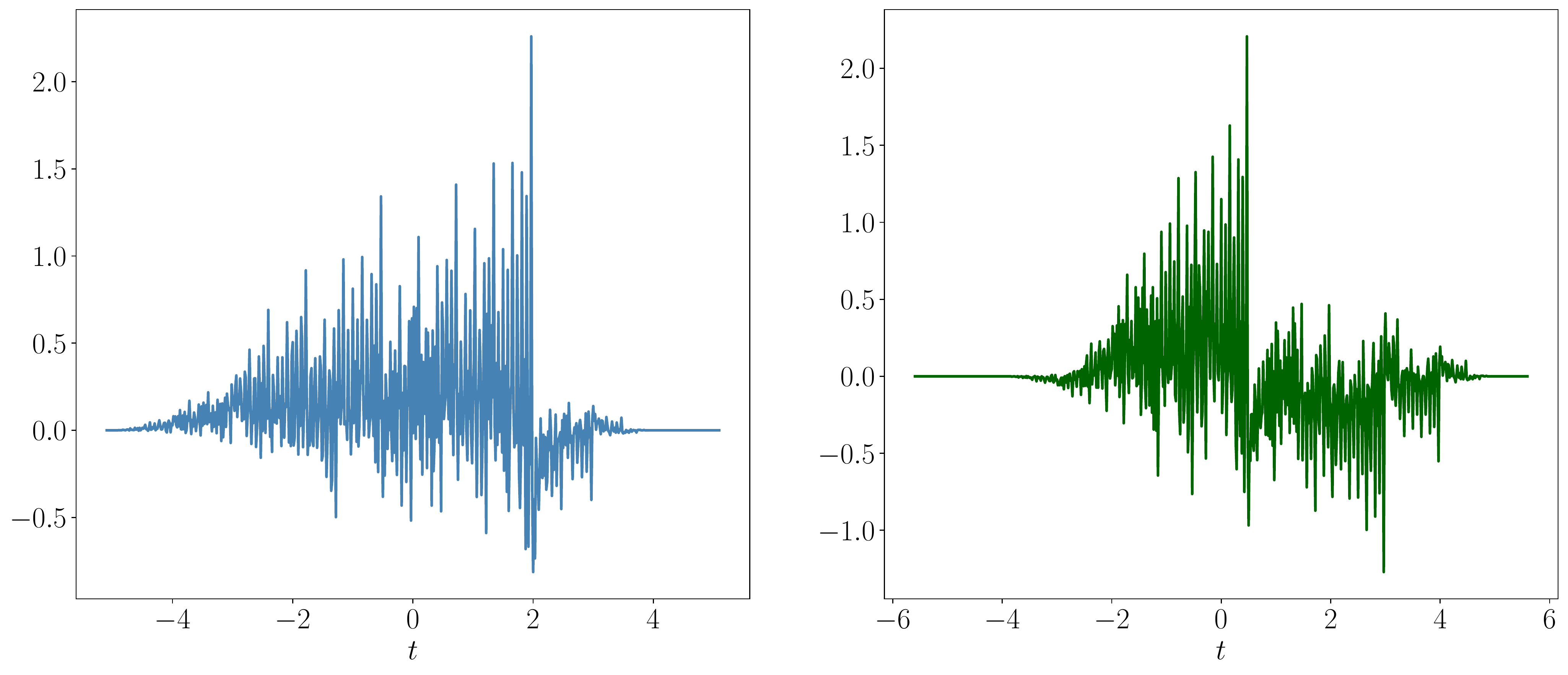} }}
	\subfloat[\centering Order $6$ - task-optimized wavelet ]{{\includegraphics[width=0.44\columnwidth]{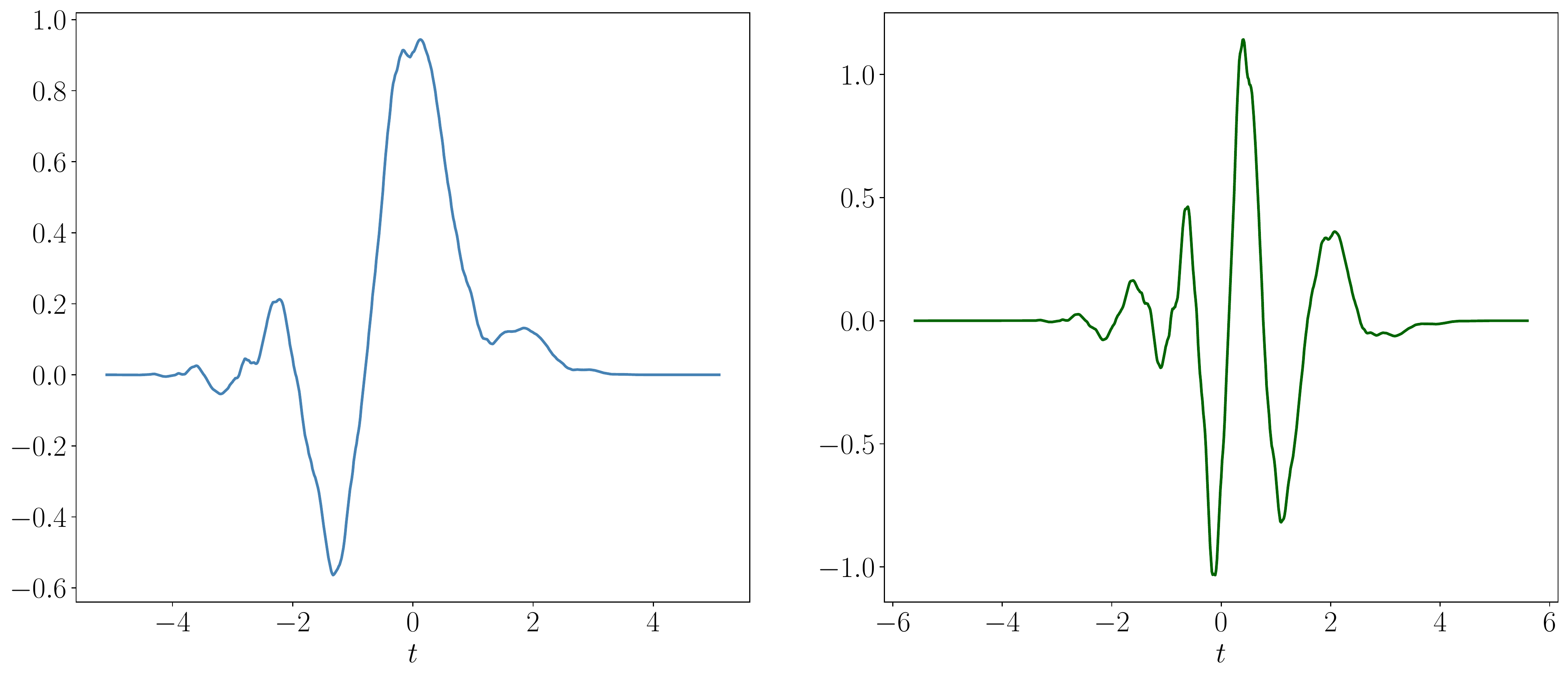} }}
	\\[2ex]    	
	\subfloat[\centering Order $7$ - initial wavelet ]{{\includegraphics[width=0.44\columnwidth]{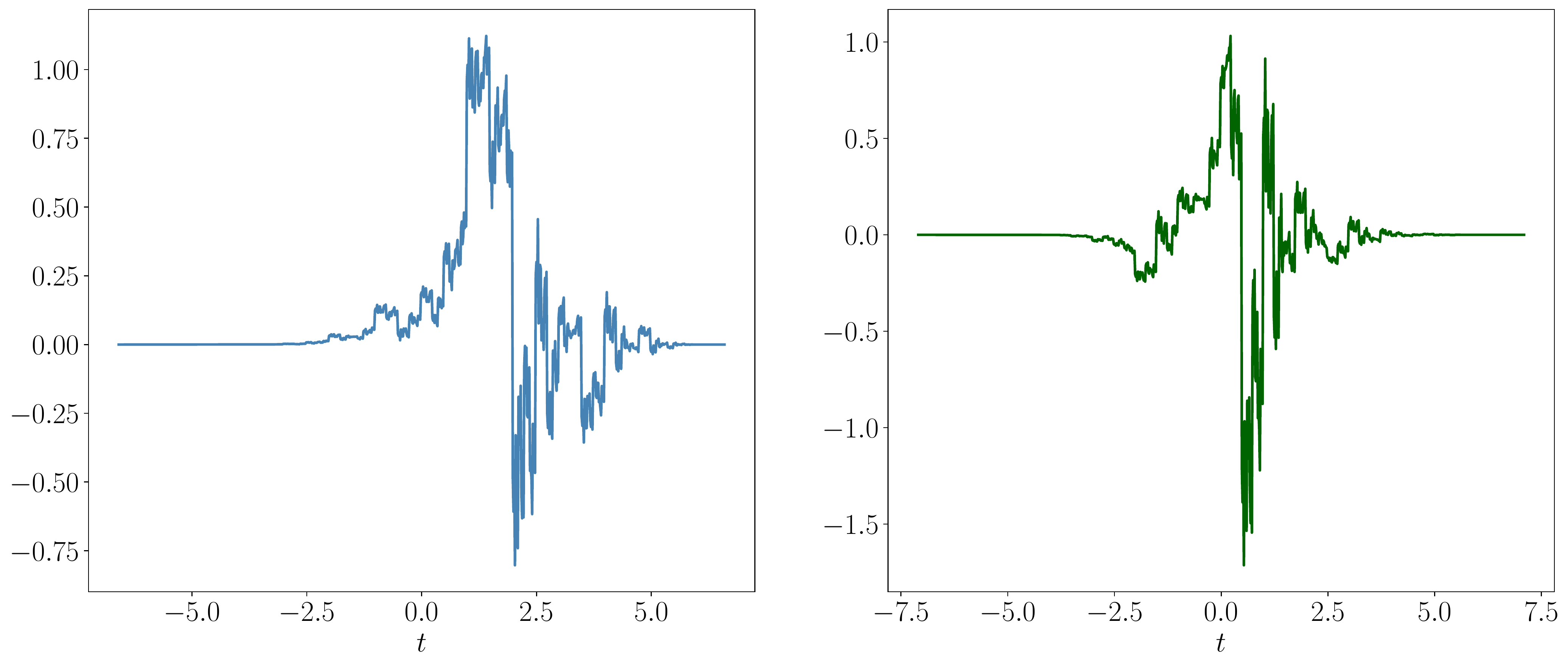} }}
	\subfloat[\centering Order $7$ - task-optimized wavelet ]{{\includegraphics[width=0.44\columnwidth]{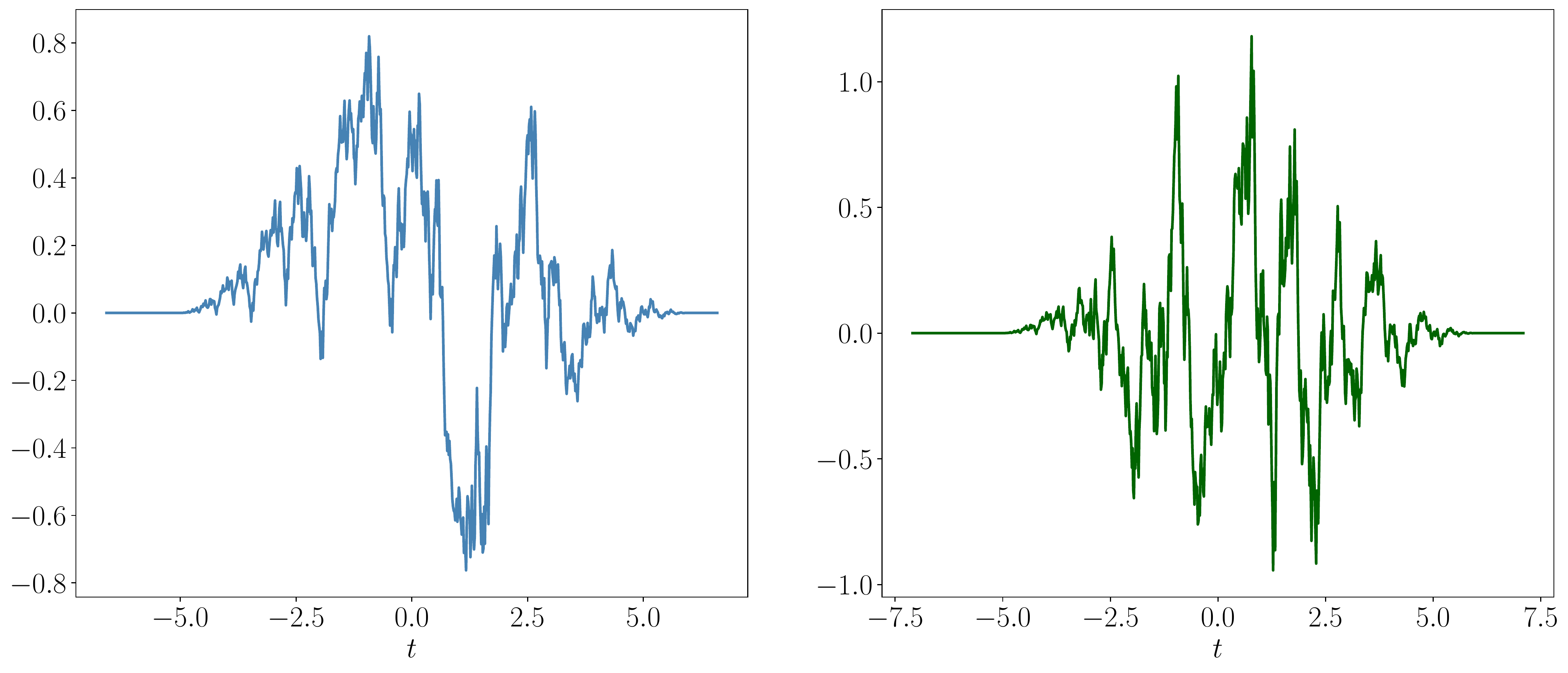} }}
	\\[2ex] 
	\subfloat[\centering Order $8$ - initial wavelet ]{{\includegraphics[width=0.44\columnwidth]{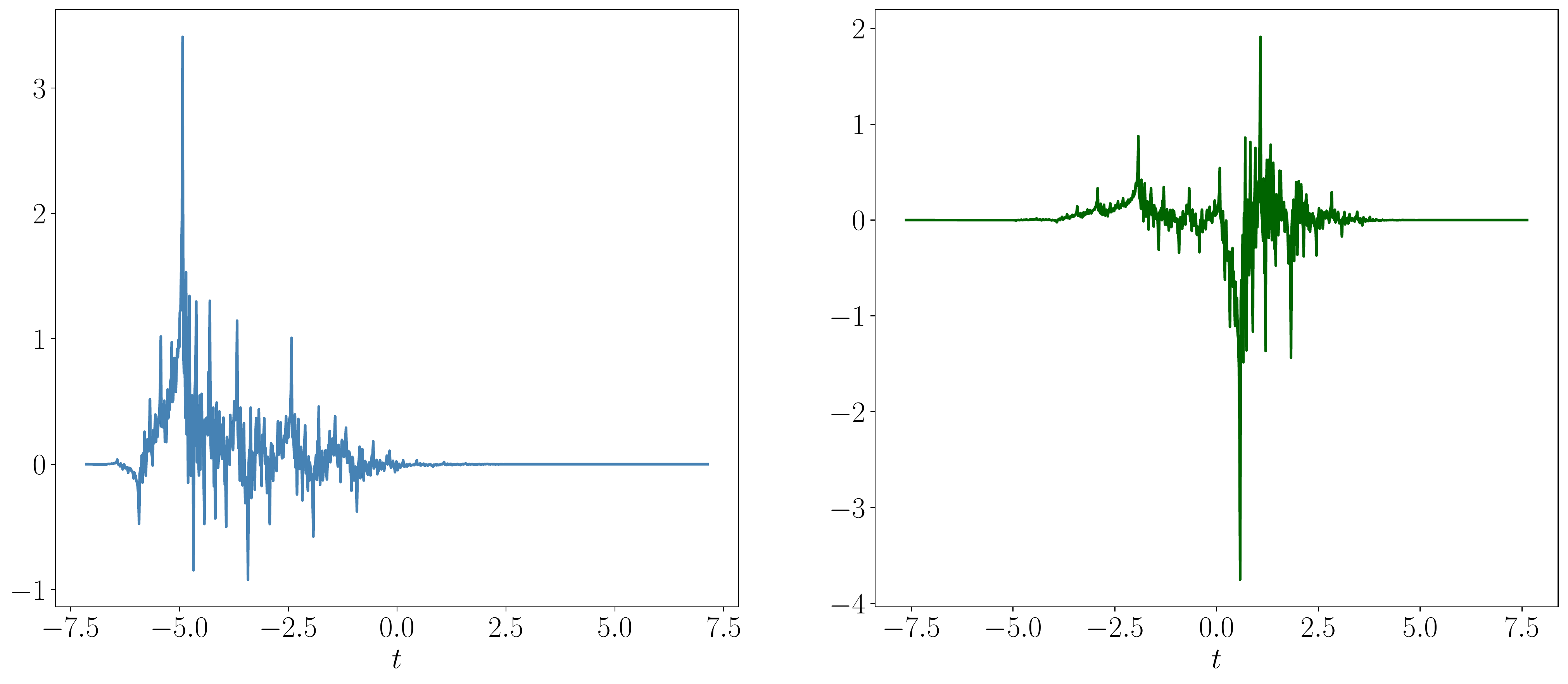} }}
	\subfloat[\centering Order $8$ - task-optimized wavelet ]{{\includegraphics[width=0.44\columnwidth]{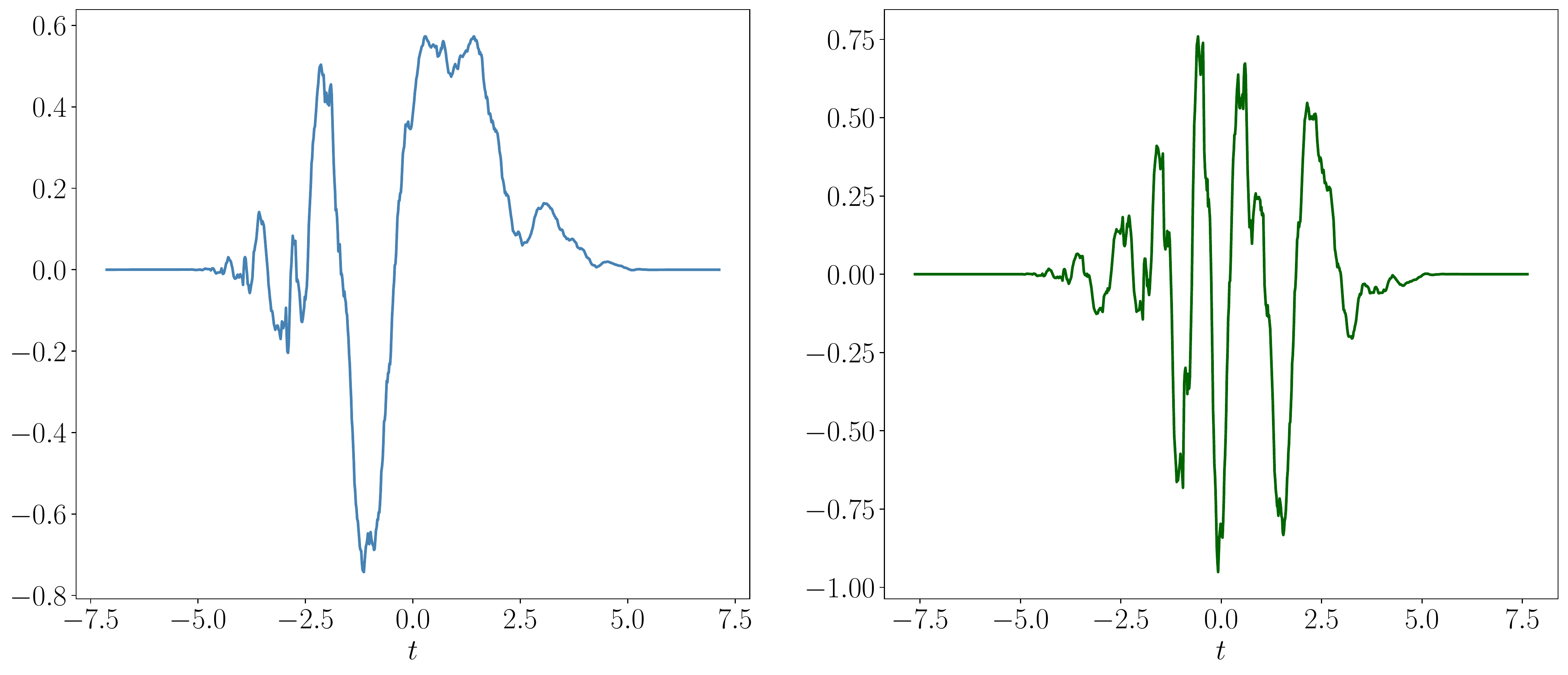} }}
	\\[2ex]    	   	
	\label{fig:spleen_wavelets_comp_1}
\end{figure}
\newpage
\subsubsection{Prostate - first spatial component}
\begin{figure}[!b]
	\centering
	\subfloat[\centering Order $3$ - initial wavelet ]{{\includegraphics[width=0.44\columnwidth]{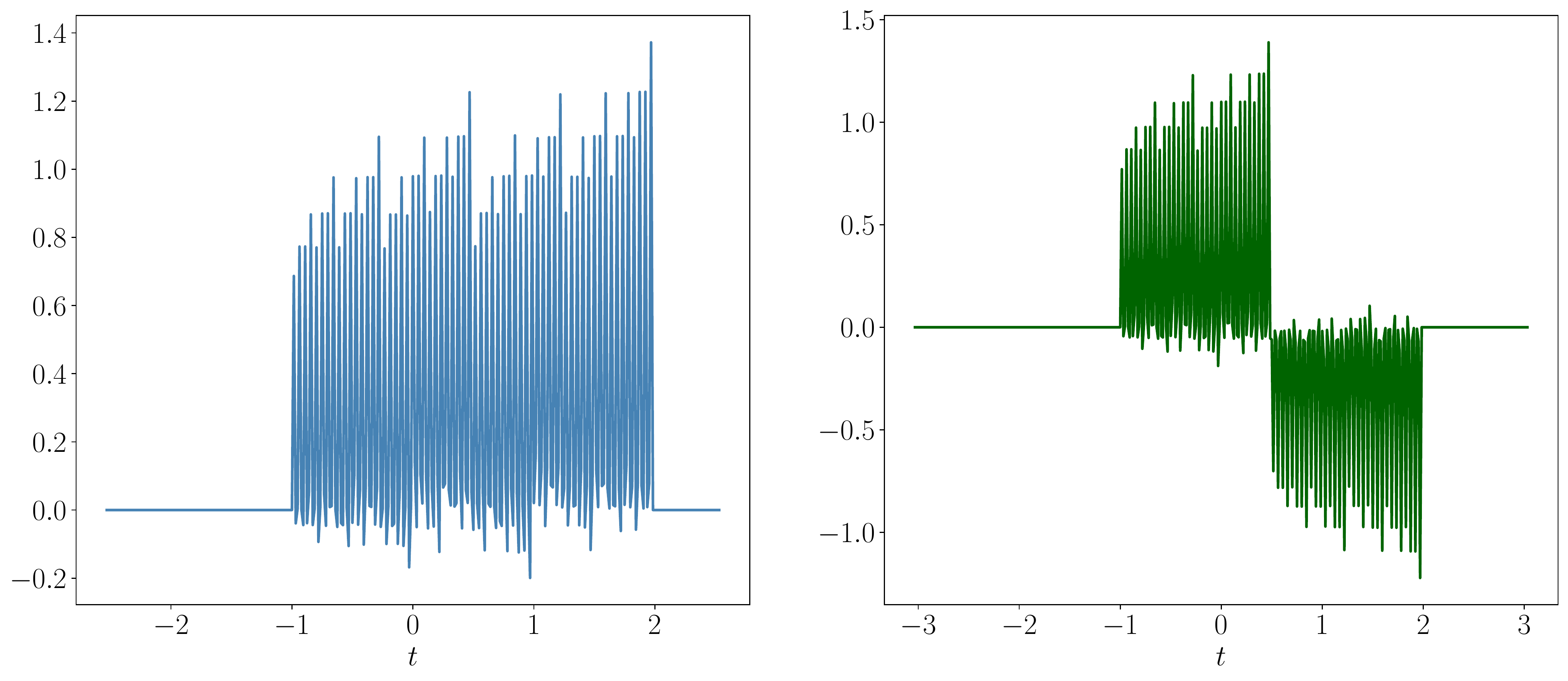} }}
	\subfloat[\centering Order $3$ - task-optimized wavelet ]{{\includegraphics[width=0.44\columnwidth]{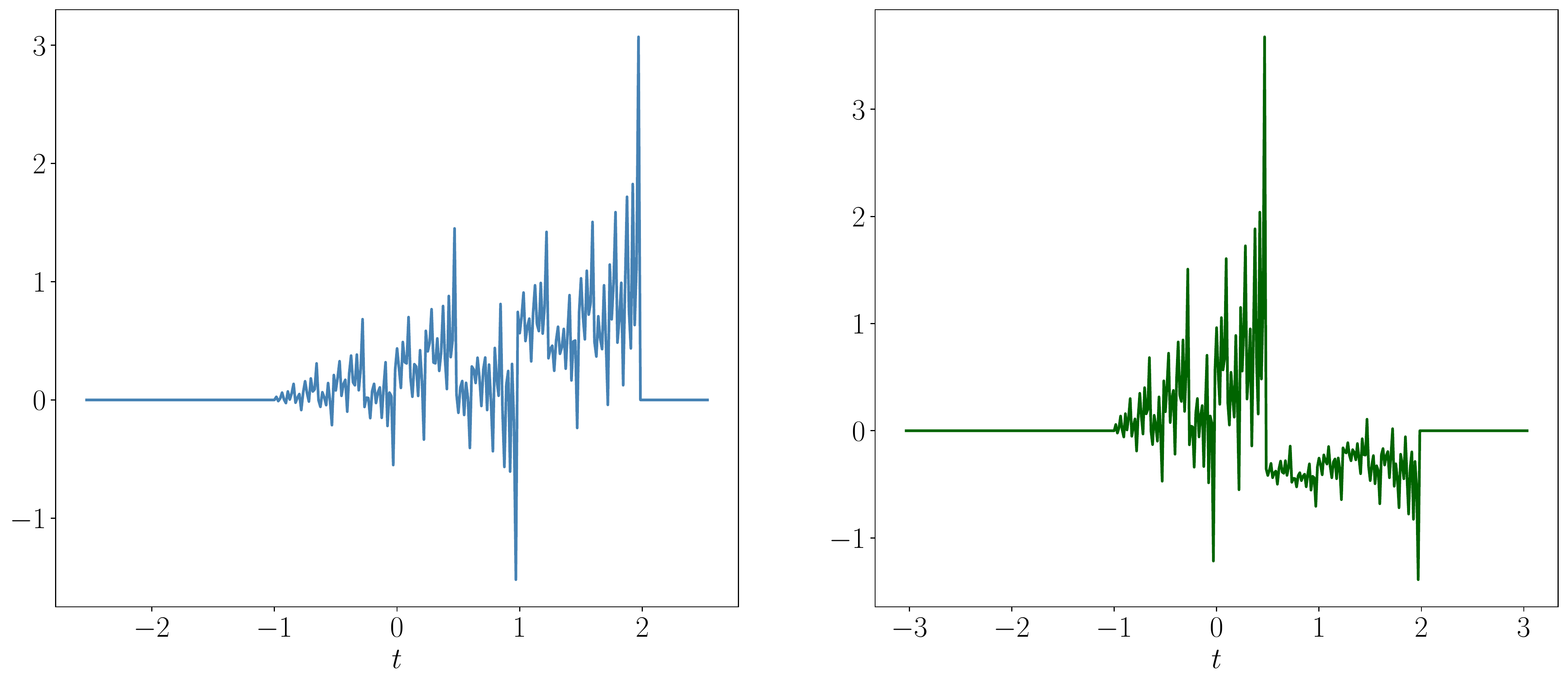} }}
	\\[2ex]    
	\subfloat[\centering Order $4$ - initial wavelet ]{{\includegraphics[width=0.44\columnwidth]{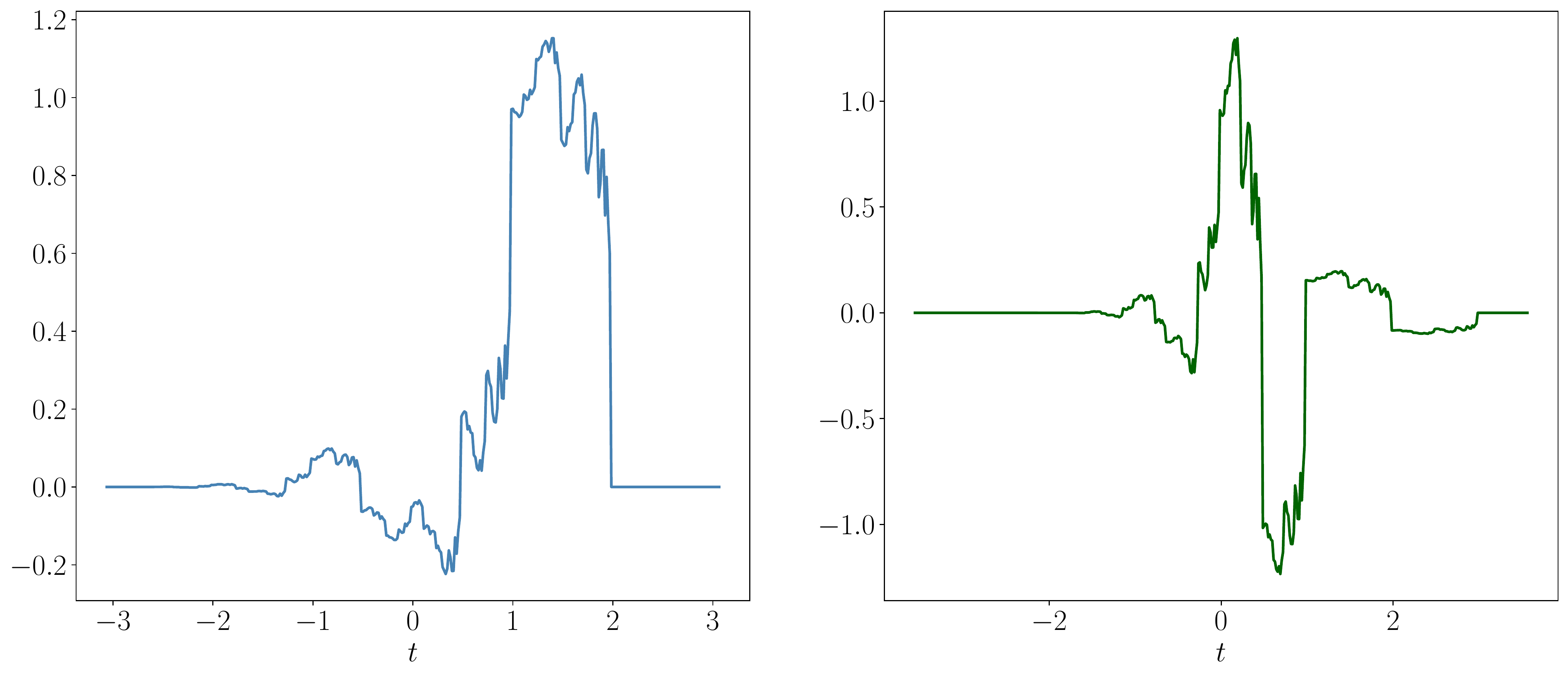} }}
	\subfloat[\centering Order $4$ - task-optimized wavelet ]{{\includegraphics[width=0.44\columnwidth]{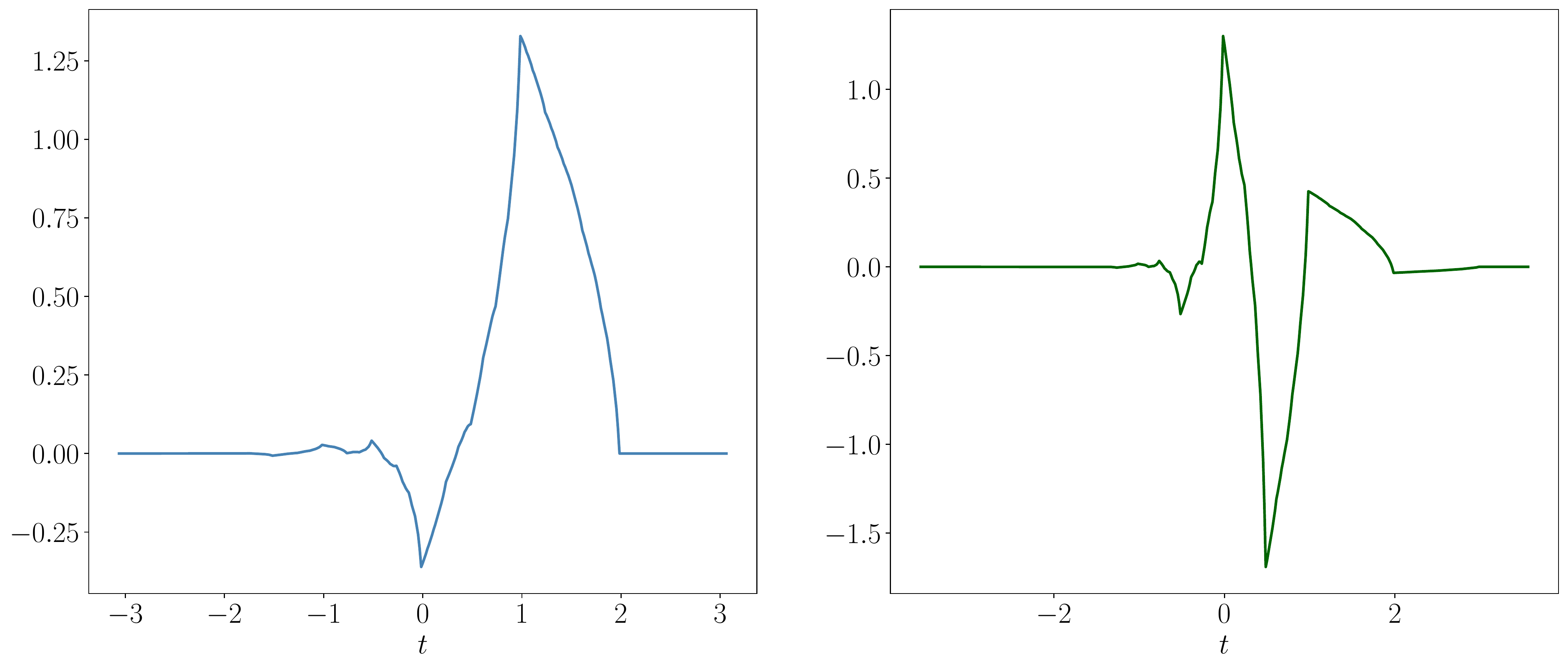} }}
	\\[2ex]    	
	\subfloat[\centering Order $5$ - initial wavelet ]{{\includegraphics[width=0.44\columnwidth]{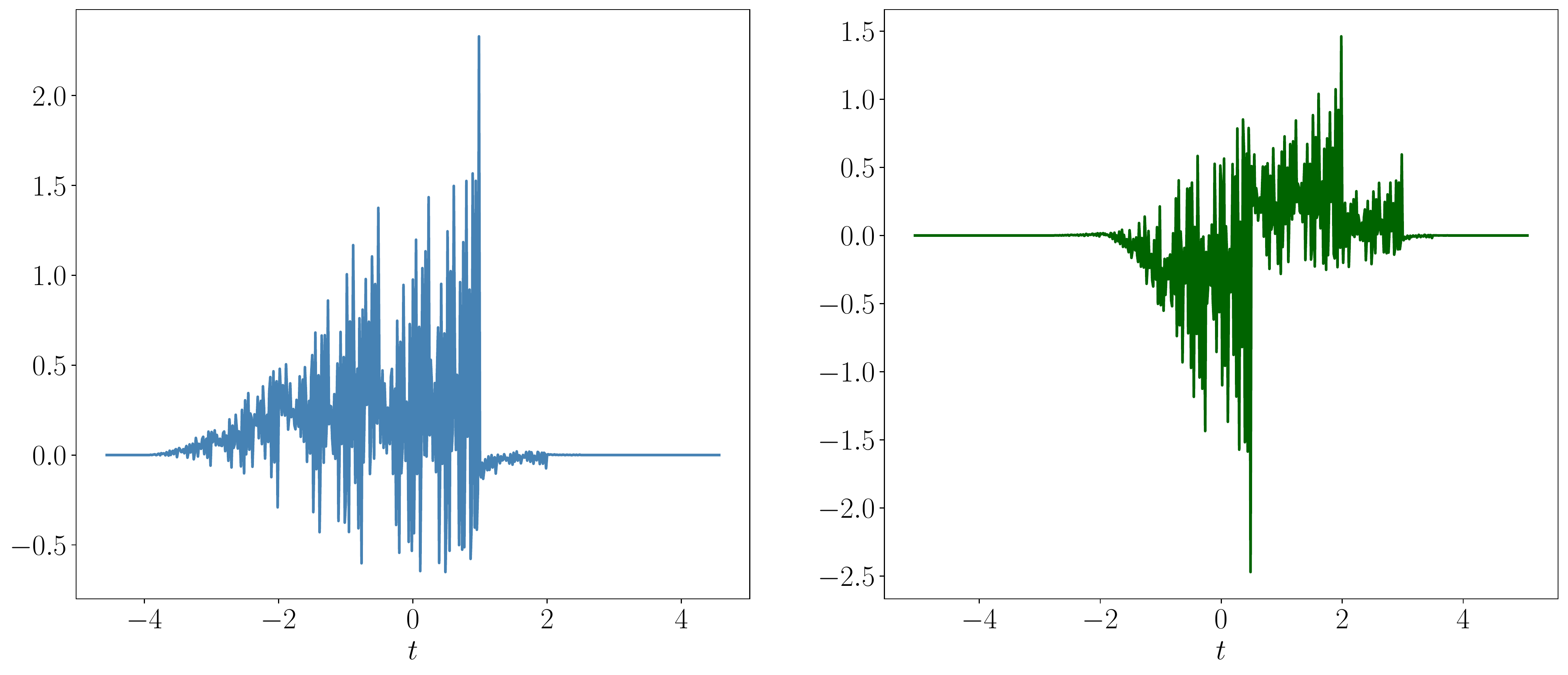} }}
	\subfloat[\centering Order $5$ - task-optimized wavelet ]{{\includegraphics[width=0.44\columnwidth]{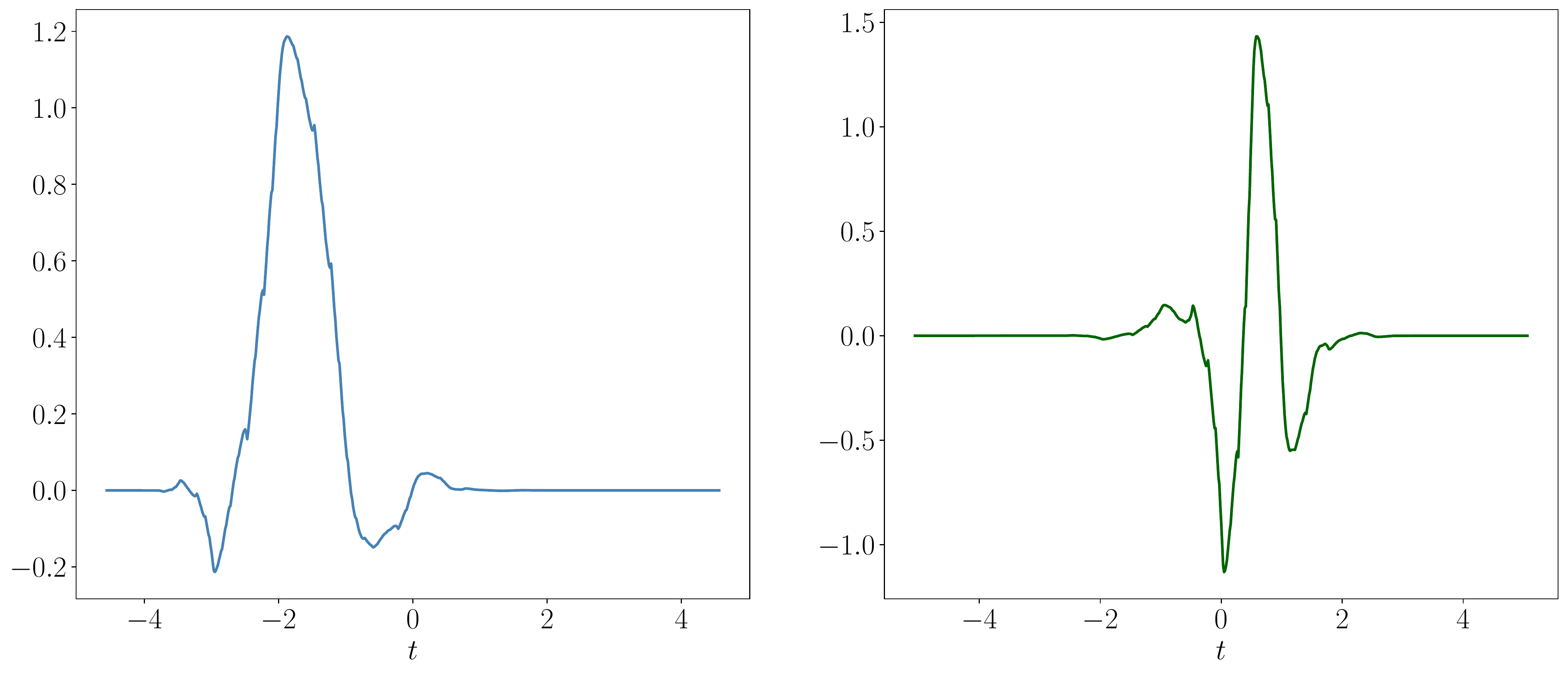} }}
	\\[2ex]    	
	\subfloat[\centering Order $6$ - initial wavelet ]{{\includegraphics[width=0.44\columnwidth]{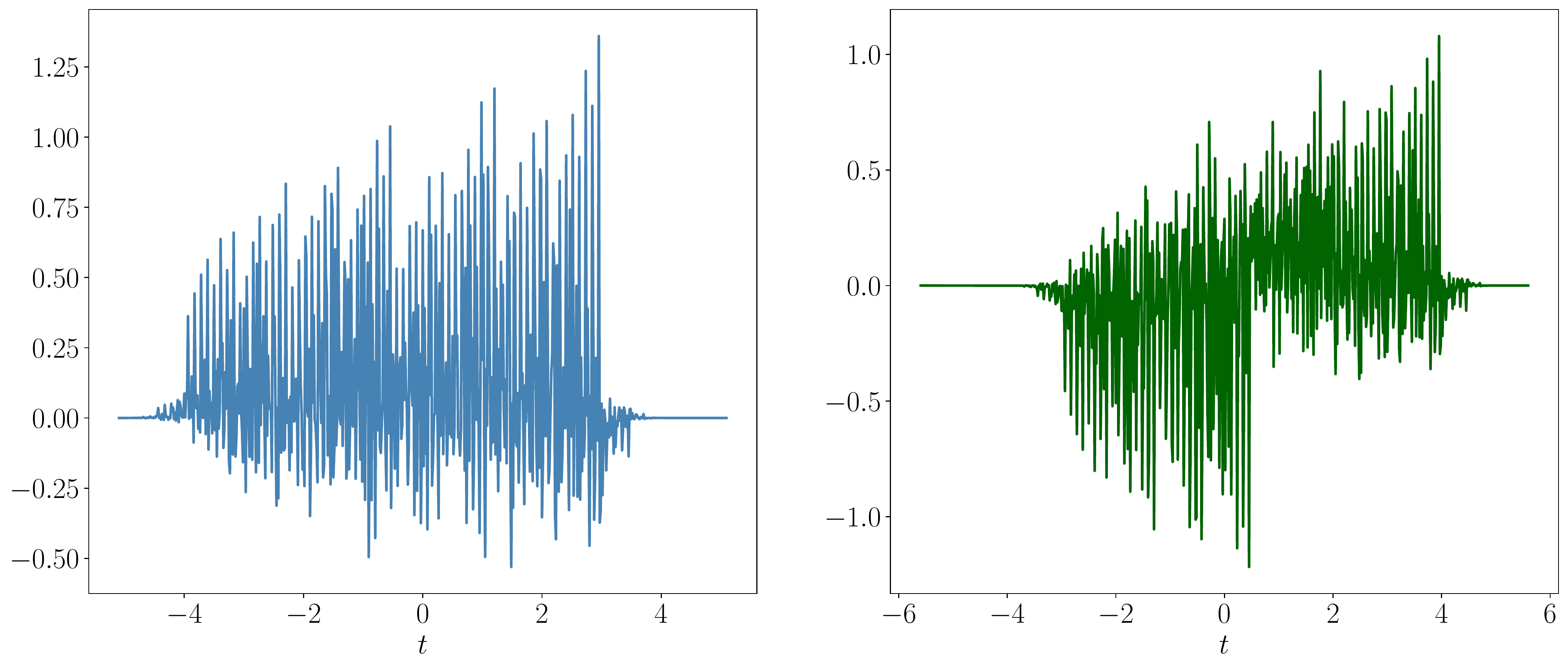} }}
	\subfloat[\centering Order $6$ - task-optimized wavelet ]{{\includegraphics[width=0.44\columnwidth]{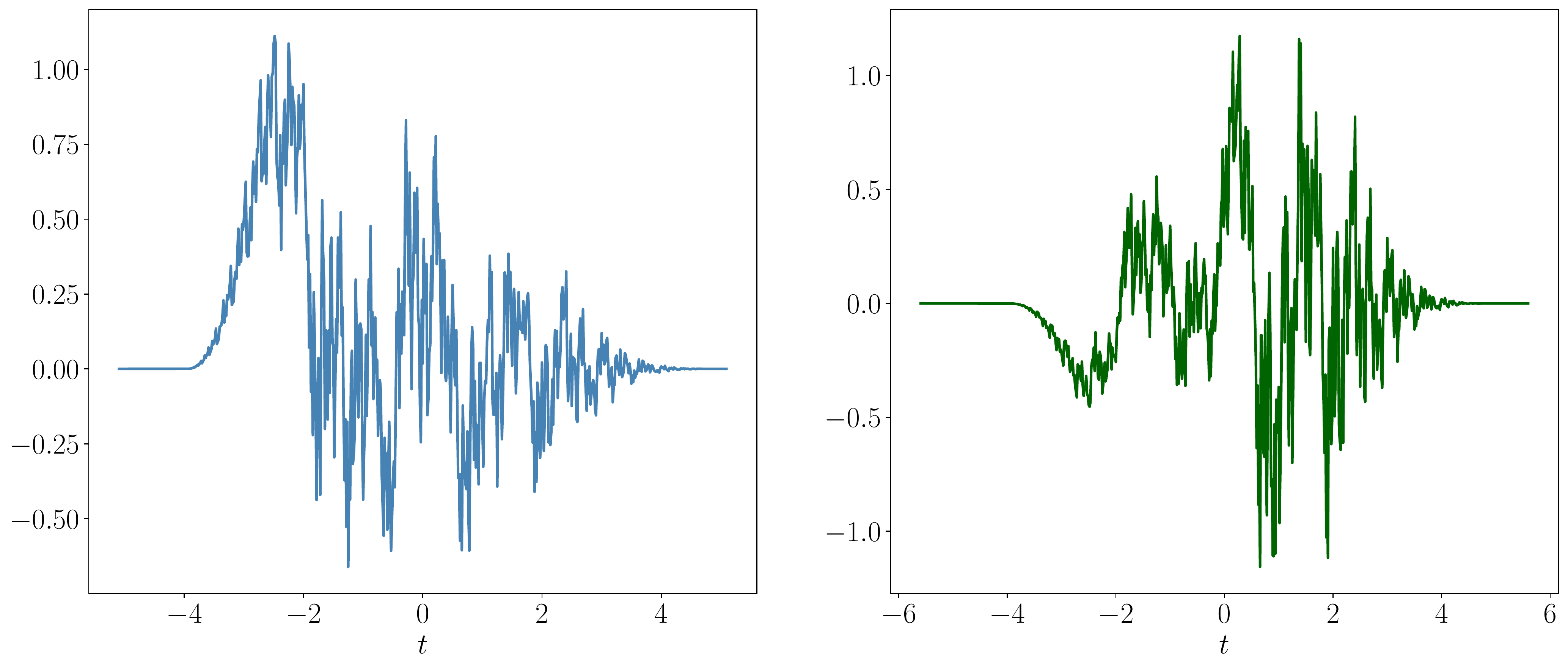} }}
	\\[2ex]    	
	\subfloat[\centering Order $7$ - initial wavelet ]{{\includegraphics[width=0.44\columnwidth]{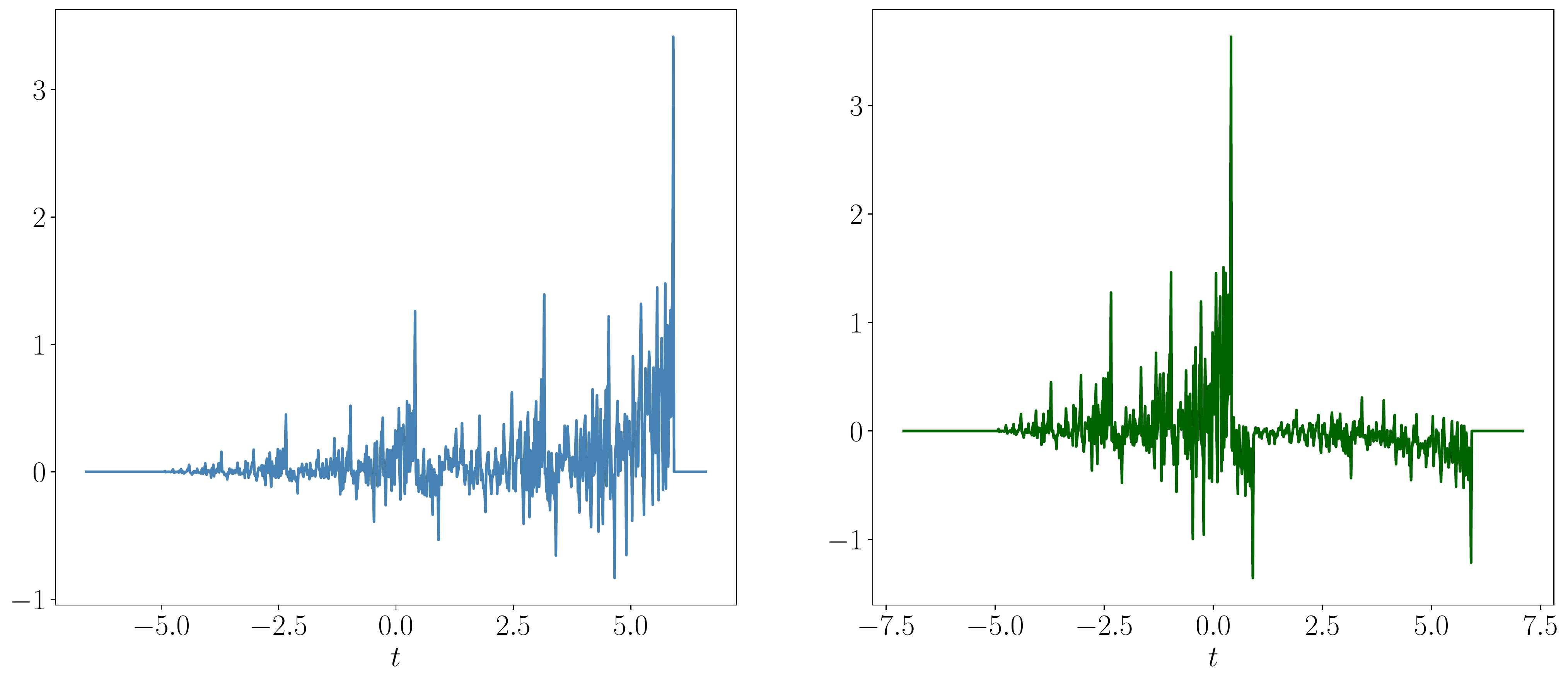} }}
	\subfloat[\centering Order $7$ - task-optimized wavelet ]{{\includegraphics[width=0.44\columnwidth]{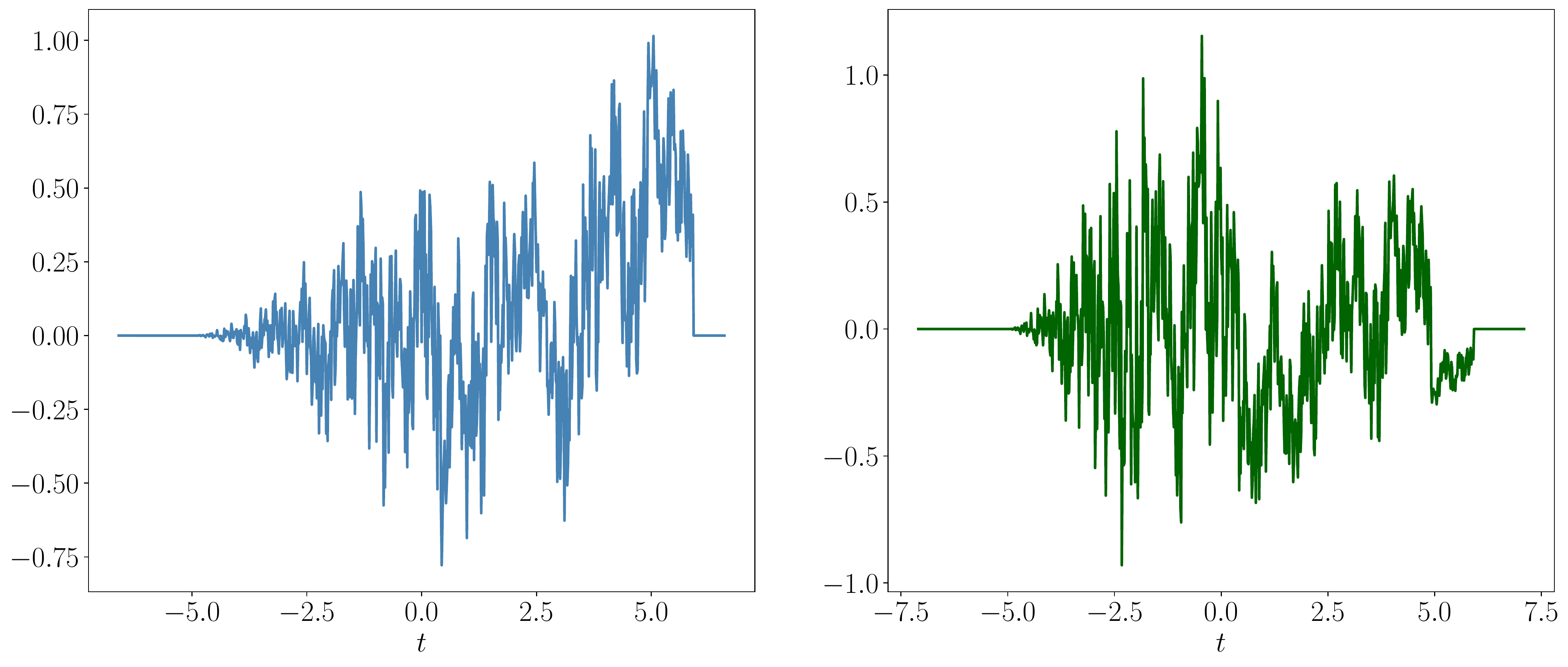} }}
	\\[2ex] 
	\subfloat[\centering Order $8$ - initial wavelet ]{{\includegraphics[width=0.44\columnwidth]{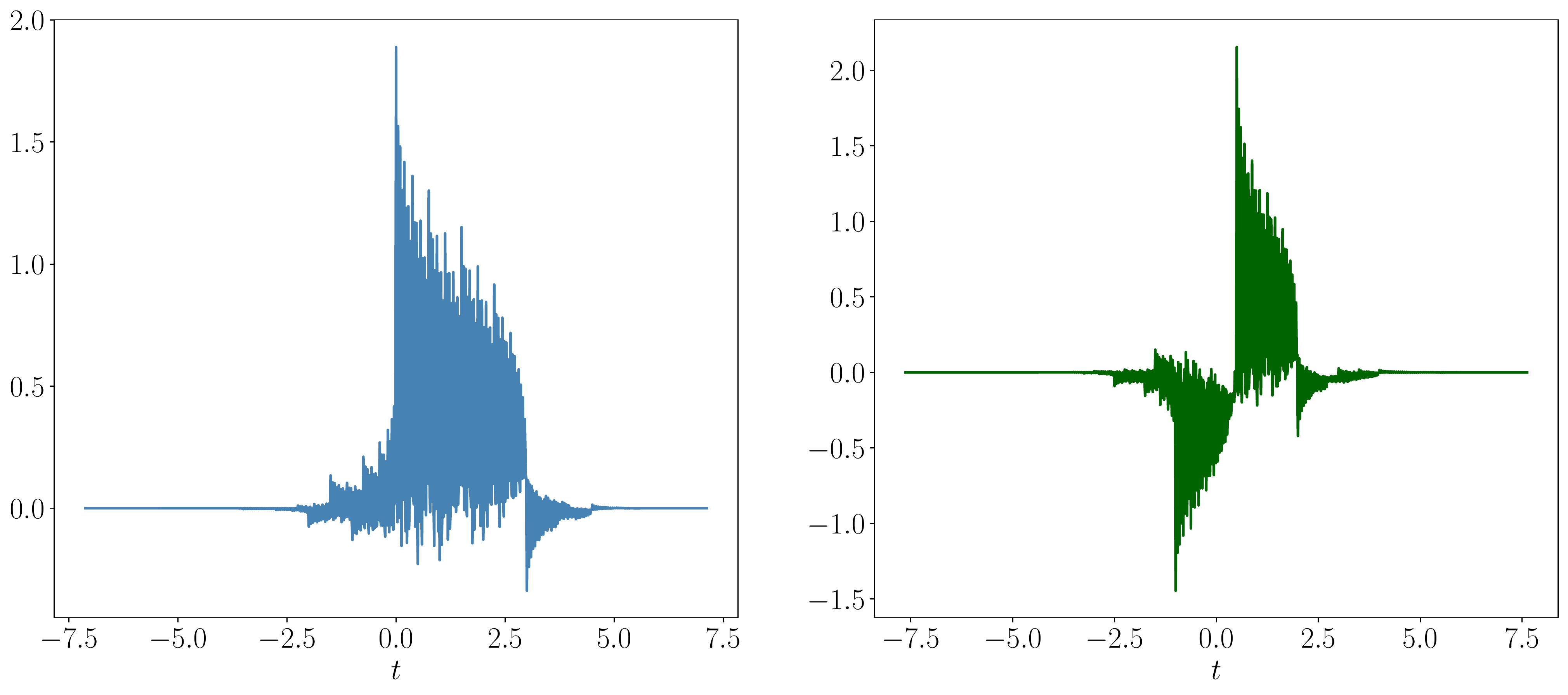} }}
	\subfloat[\centering Order $8$ - task-optimized wavelet ]{{\includegraphics[width=0.44\columnwidth]{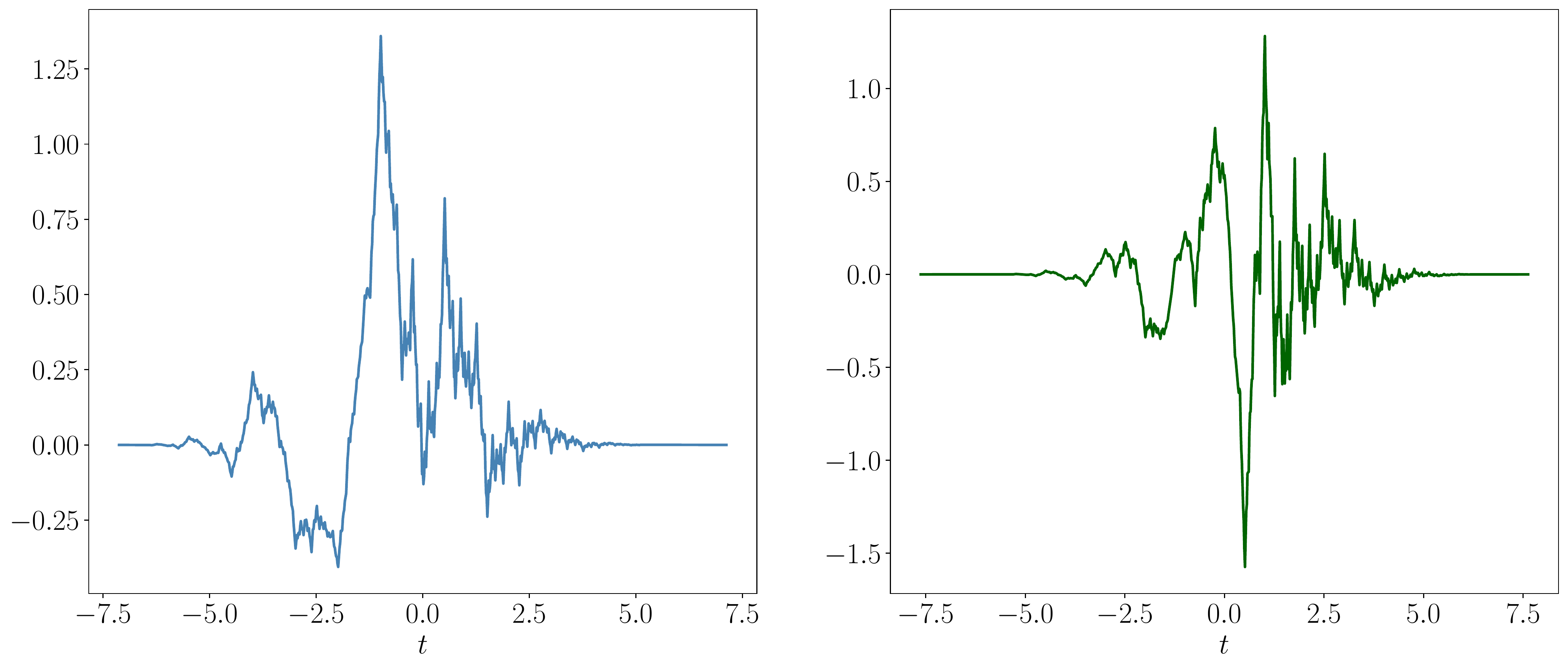} }}
	\\[2ex]    	   	
	\label{fig:prostate_wavelets_comp_0}
\end{figure}
\newpage
\subsubsection{Prostate - second spatial component}
\begin{figure}[!b]
	\centering
	\subfloat[\centering Order $3$ - initial wavelet ]{{\includegraphics[width=0.44\columnwidth]{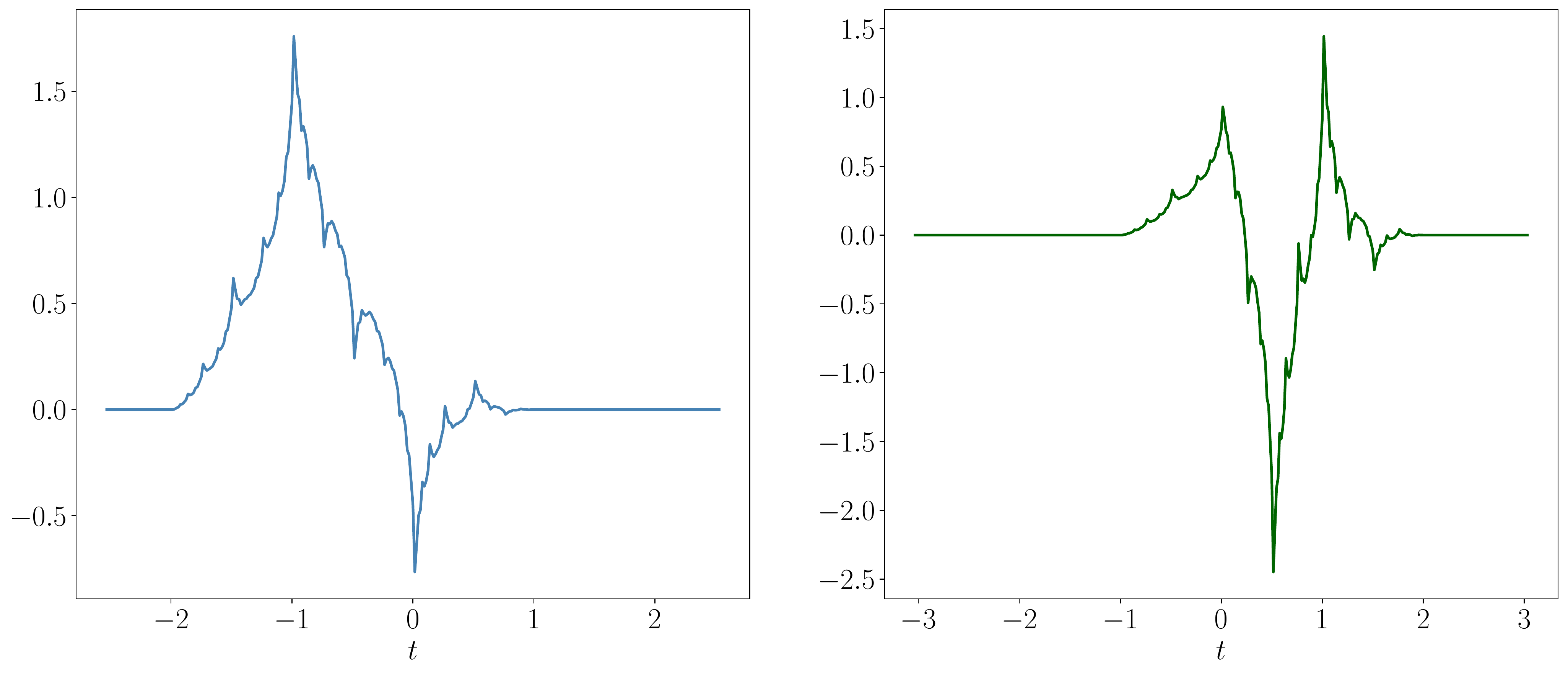} }}
	\subfloat[\centering Order $3$ - task-optimized wavelet ]{{\includegraphics[width=0.44\columnwidth]{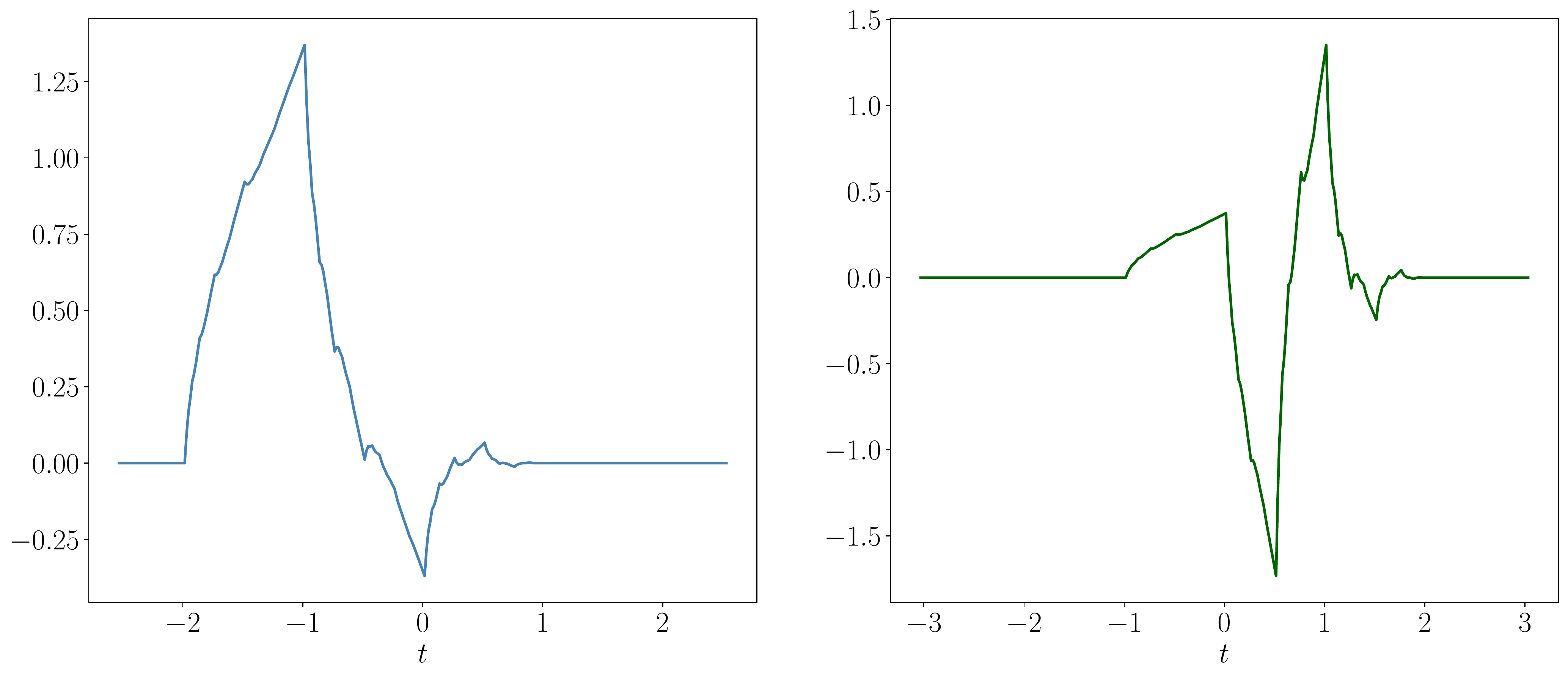} }}
	\\[2ex]    
	\subfloat[\centering Order $4$ - initial wavelet ]{{\includegraphics[width=0.44\columnwidth]{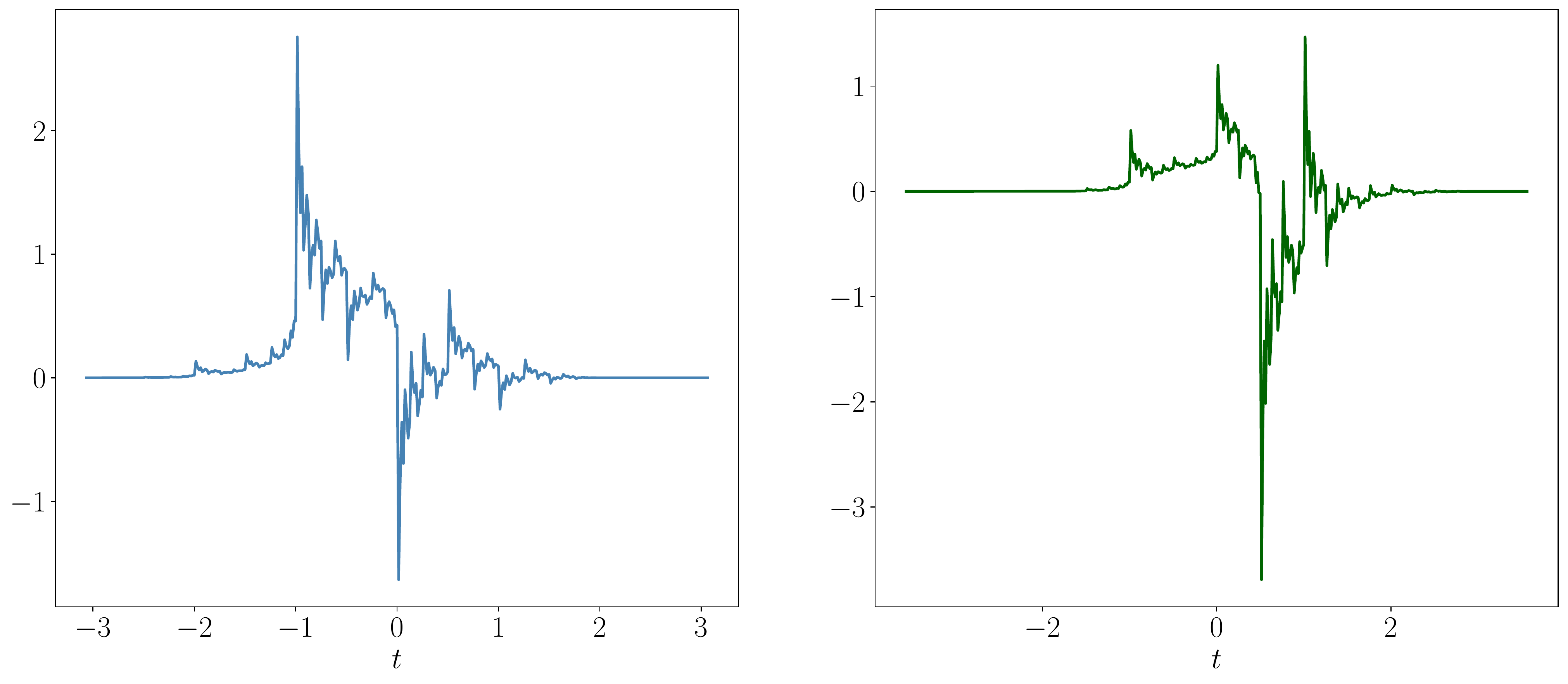} }}
	\subfloat[\centering Order $4$ - task-optimized wavelet ]{{\includegraphics[width=0.44\columnwidth]{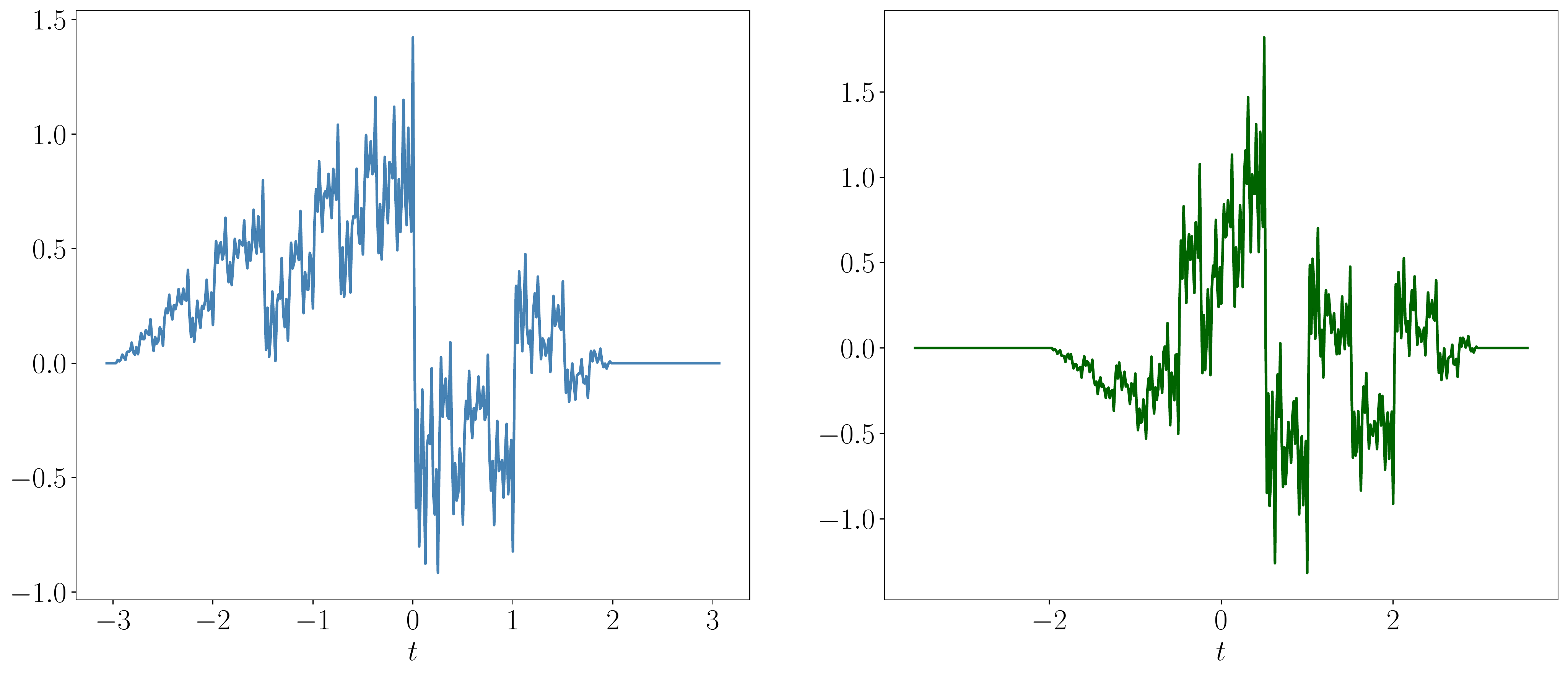} }}
	\\[2ex]    	
	\subfloat[\centering Order $5$ - initial wavelet ]{{\includegraphics[width=0.44\columnwidth]{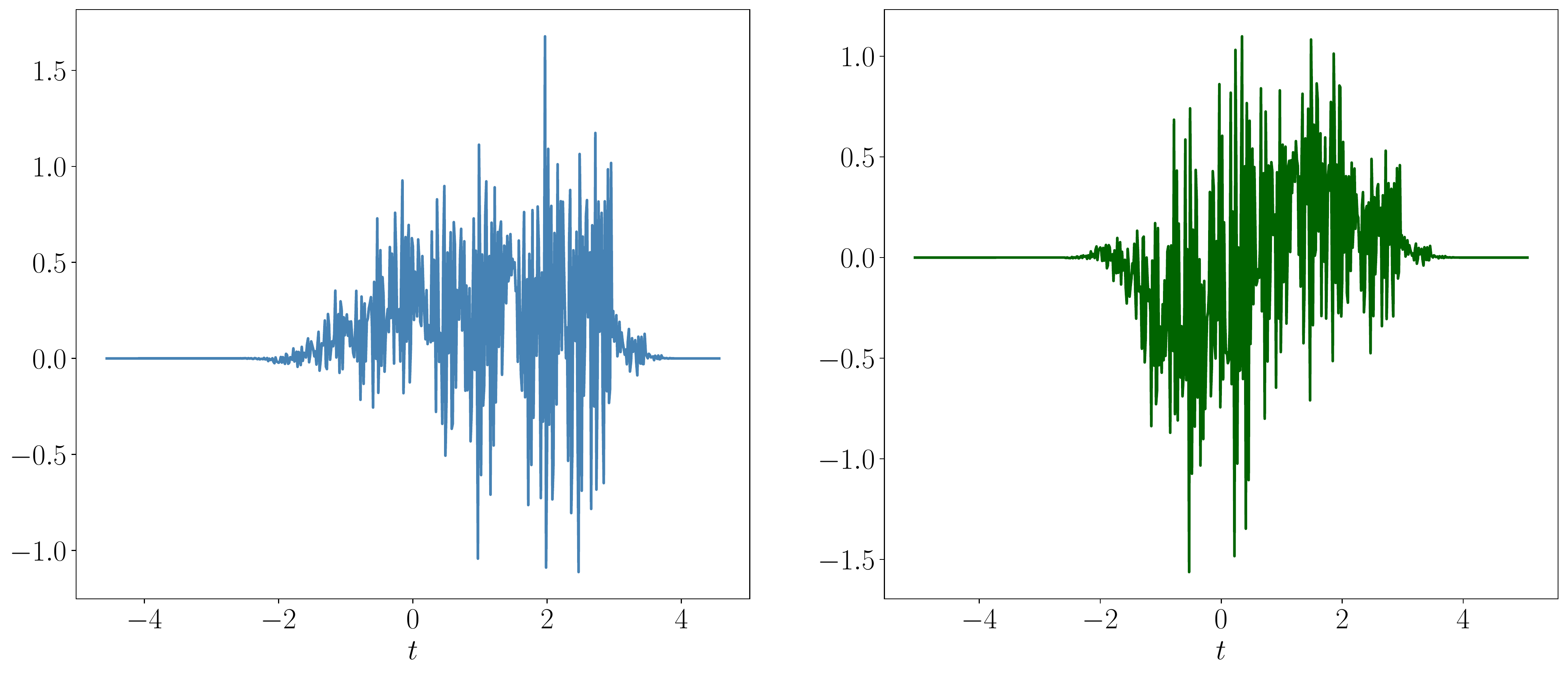} }}
	\subfloat[\centering Order $5$ - task-optimized wavelet ]{{\includegraphics[width=0.44\columnwidth]{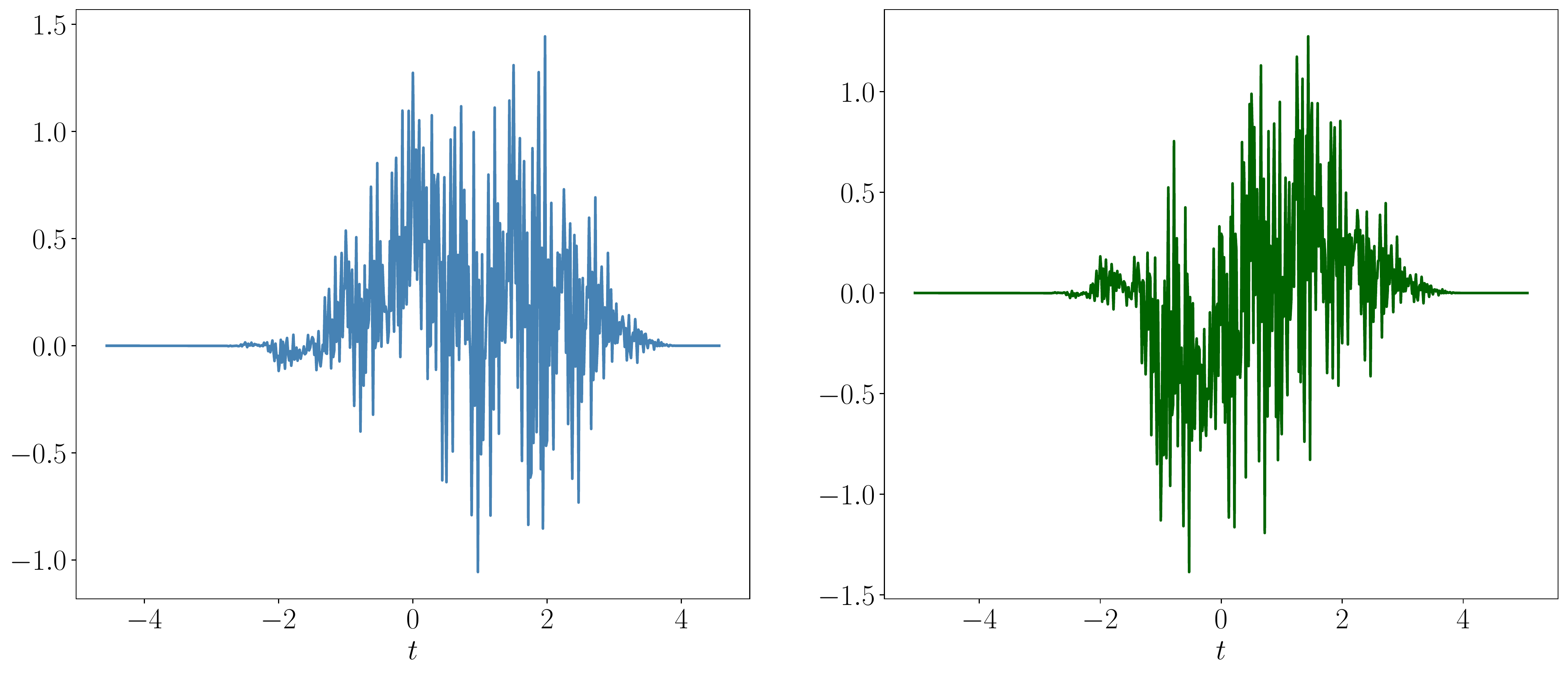} }}
	\\[2ex]    	
	\subfloat[\centering Order $6$ - initial wavelet ]{{\includegraphics[width=0.44\columnwidth]{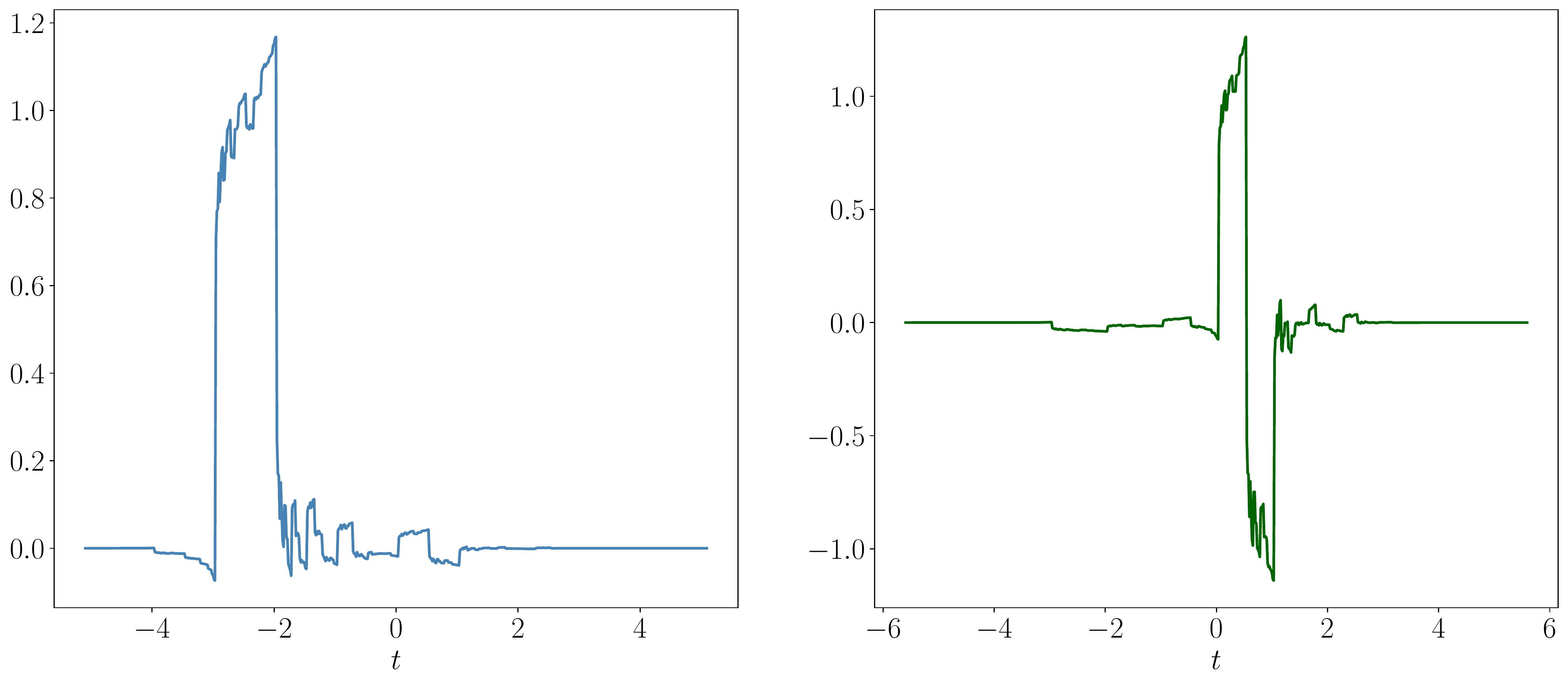} }}
	\subfloat[\centering Order $6$ - task-optimized wavelet ]{{\includegraphics[width=0.44\columnwidth]{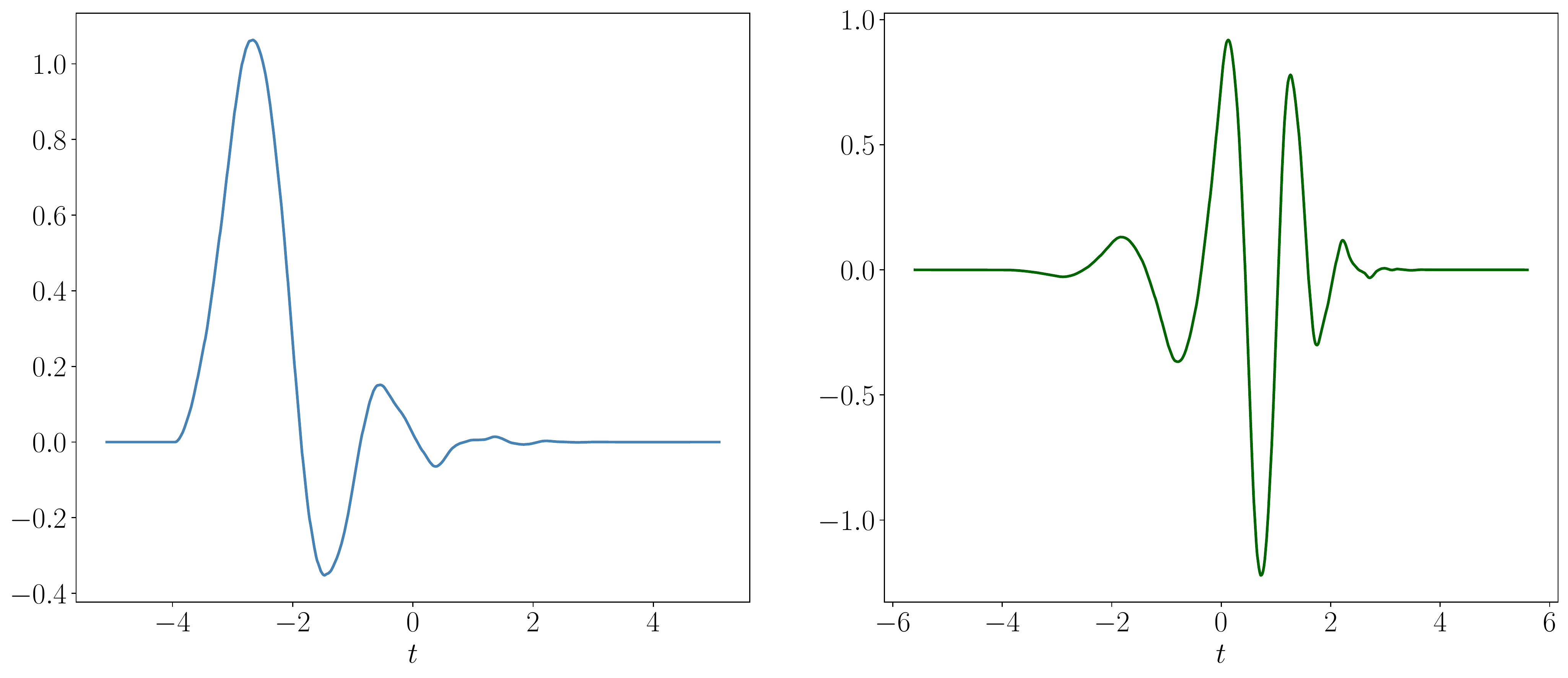} }}
	\\[2ex]    	
	\subfloat[\centering Order $7$ - initial wavelet ]{{\includegraphics[width=0.44\columnwidth]{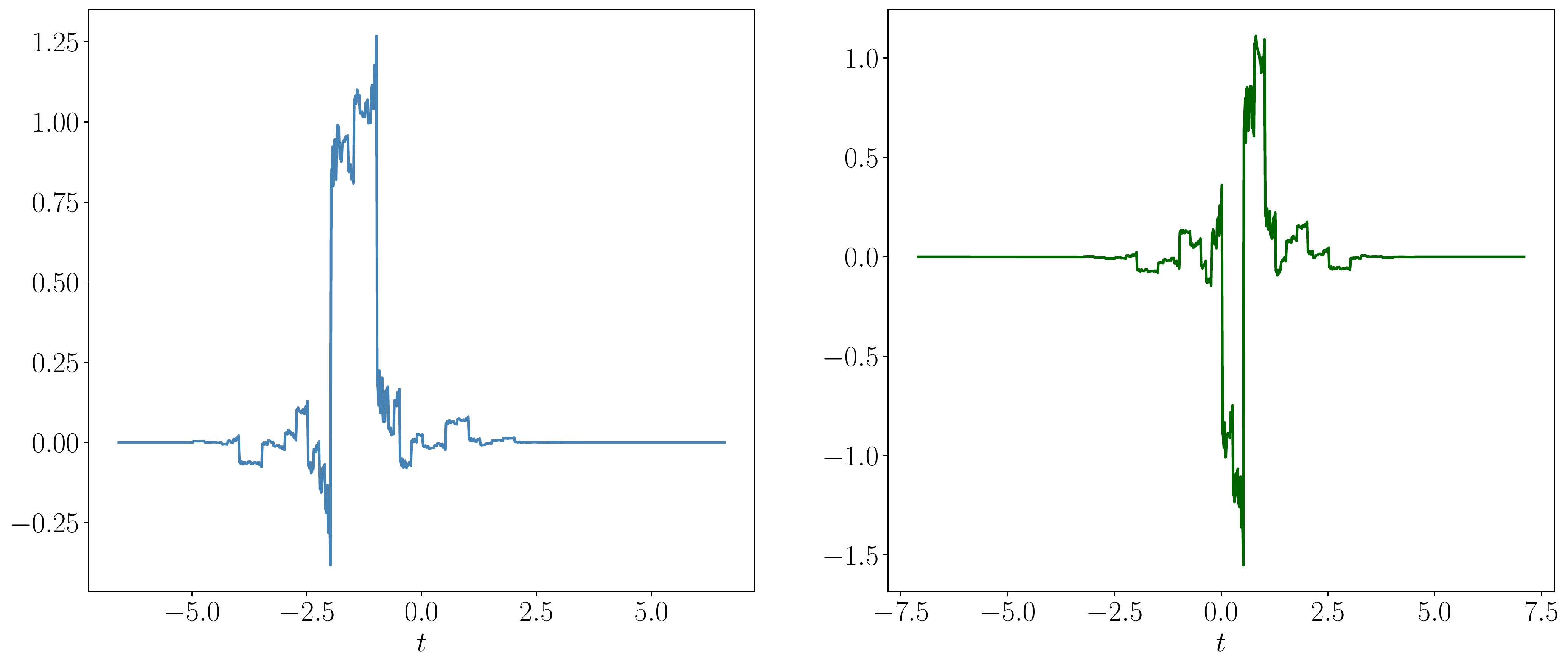} }}
	\subfloat[\centering Order $7$ - task-optimized wavelet ]{{\includegraphics[width=0.44\columnwidth]{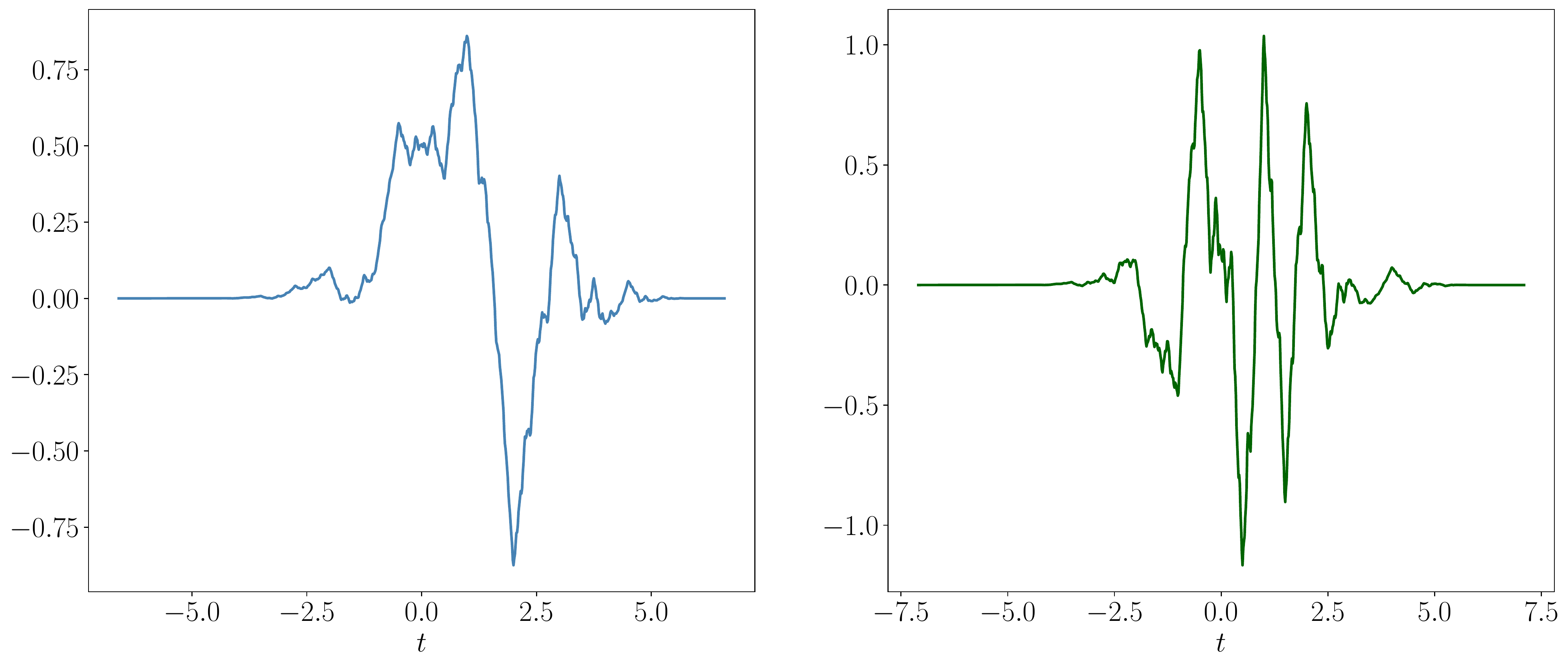} }}
	\\[2ex] 
	\subfloat[\centering Order $8$ - initial wavelet ]{{\includegraphics[width=0.44\columnwidth]{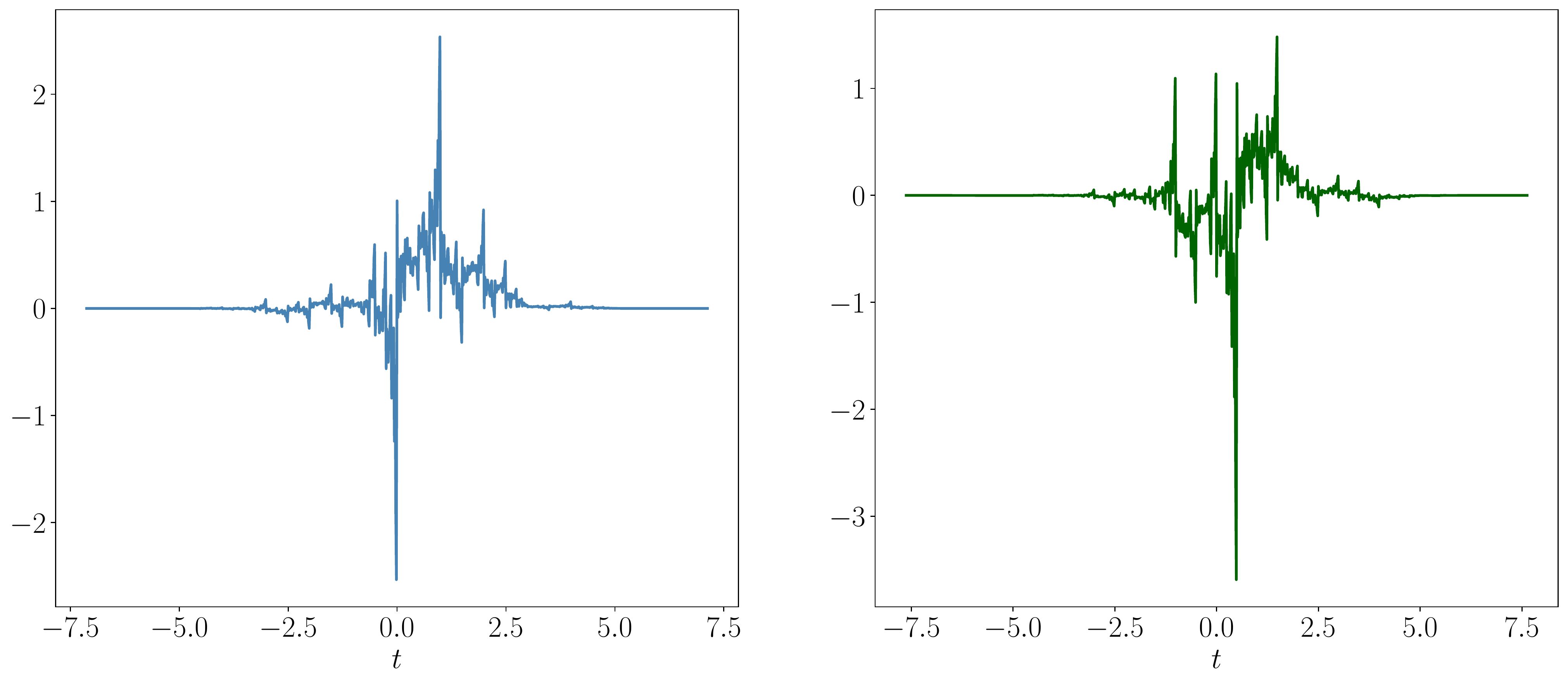} }}
	\subfloat[\centering Order $8$ - task-optimized wavelet ]{{\includegraphics[width=0.44\columnwidth]{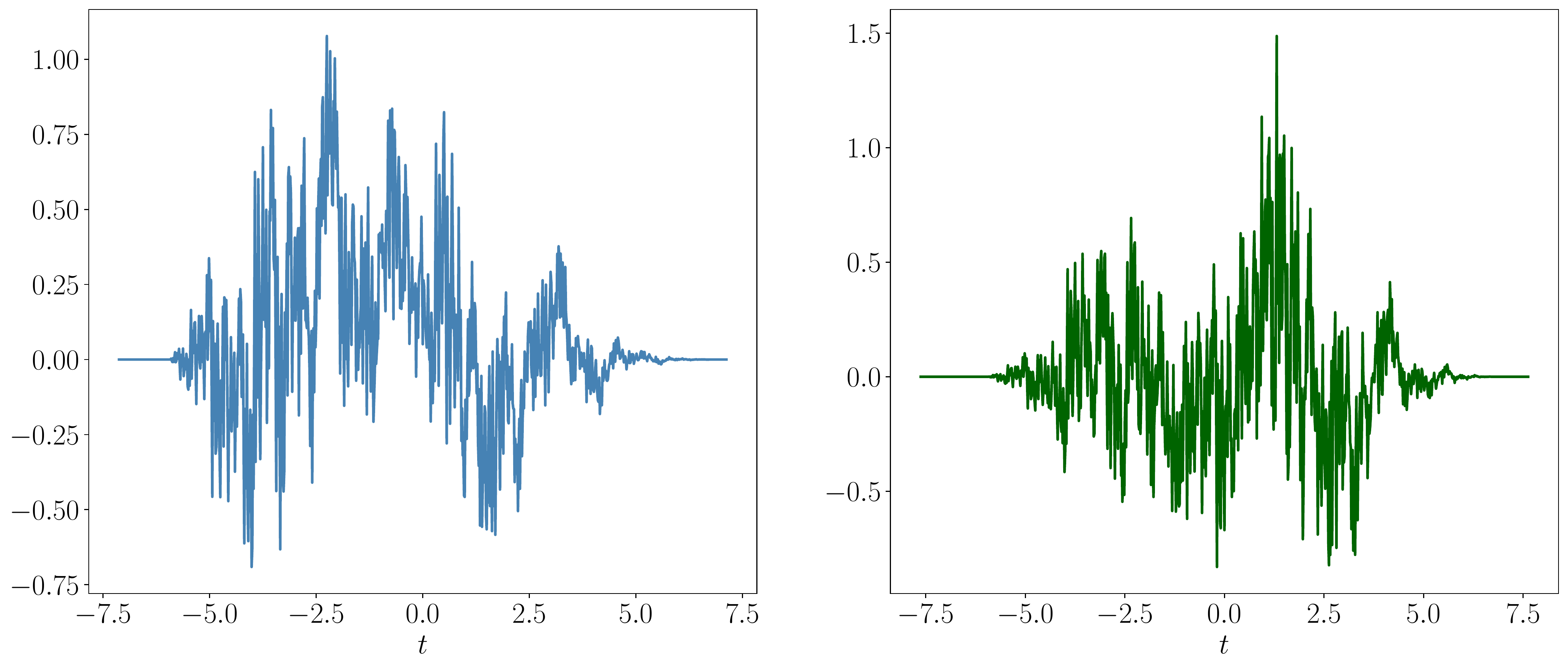} }}
	\\[2ex]    	   	
	\label{fig:prostate_wavelets_comp_1}
\end{figure}
\newpage
\subsection{Refinement masks}
\label{sec:figures_refinement_masks}
In this section we visualize the refinement masks of the initial and task-optimized wavelets shown in the previous section. 
\newpage
\subsubsection{Spleen - first spatial component}
\begin{figure}[!b]
	\centering
	\subfloat[\centering Order $3$ - initial wavelet ]{{\includegraphics[width=0.44\columnwidth]{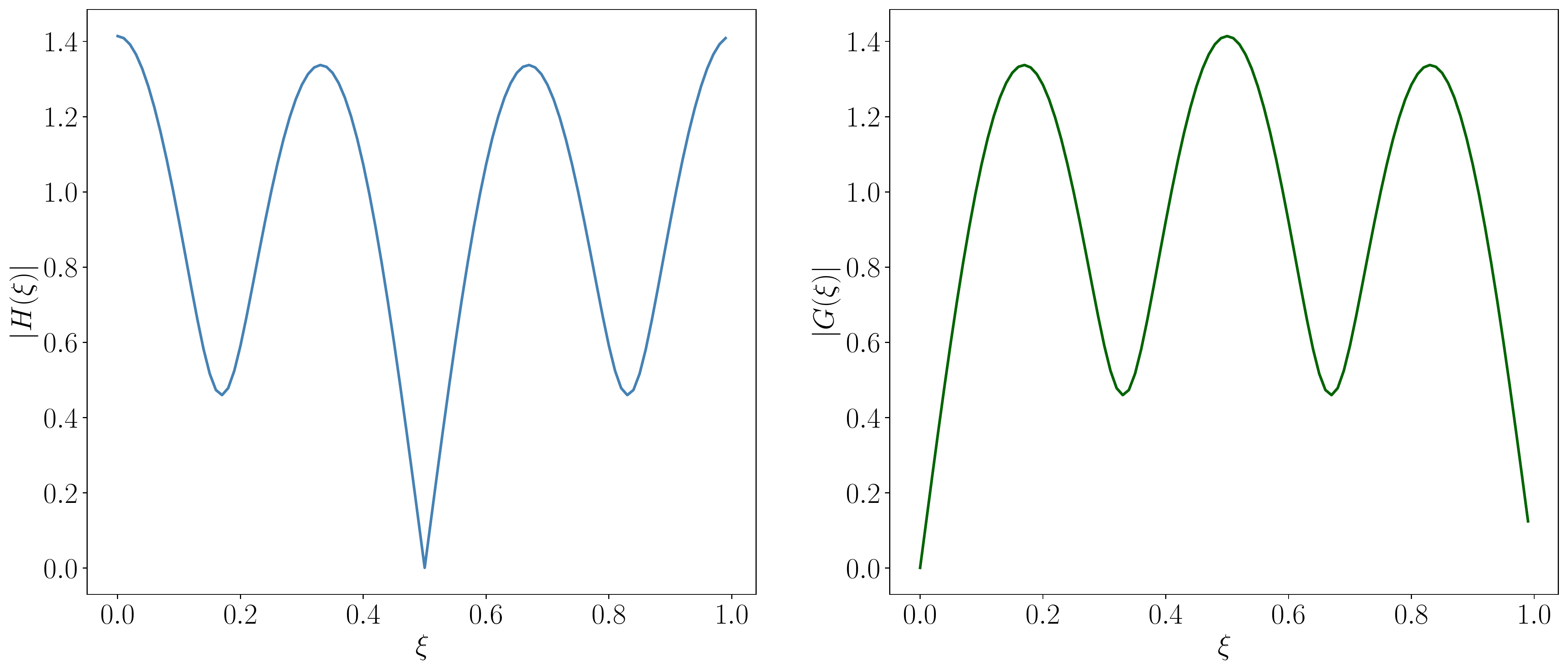} }}
	\subfloat[\centering Order $3$ - task-optimized wavelet ]{{\includegraphics[width=0.44\columnwidth]{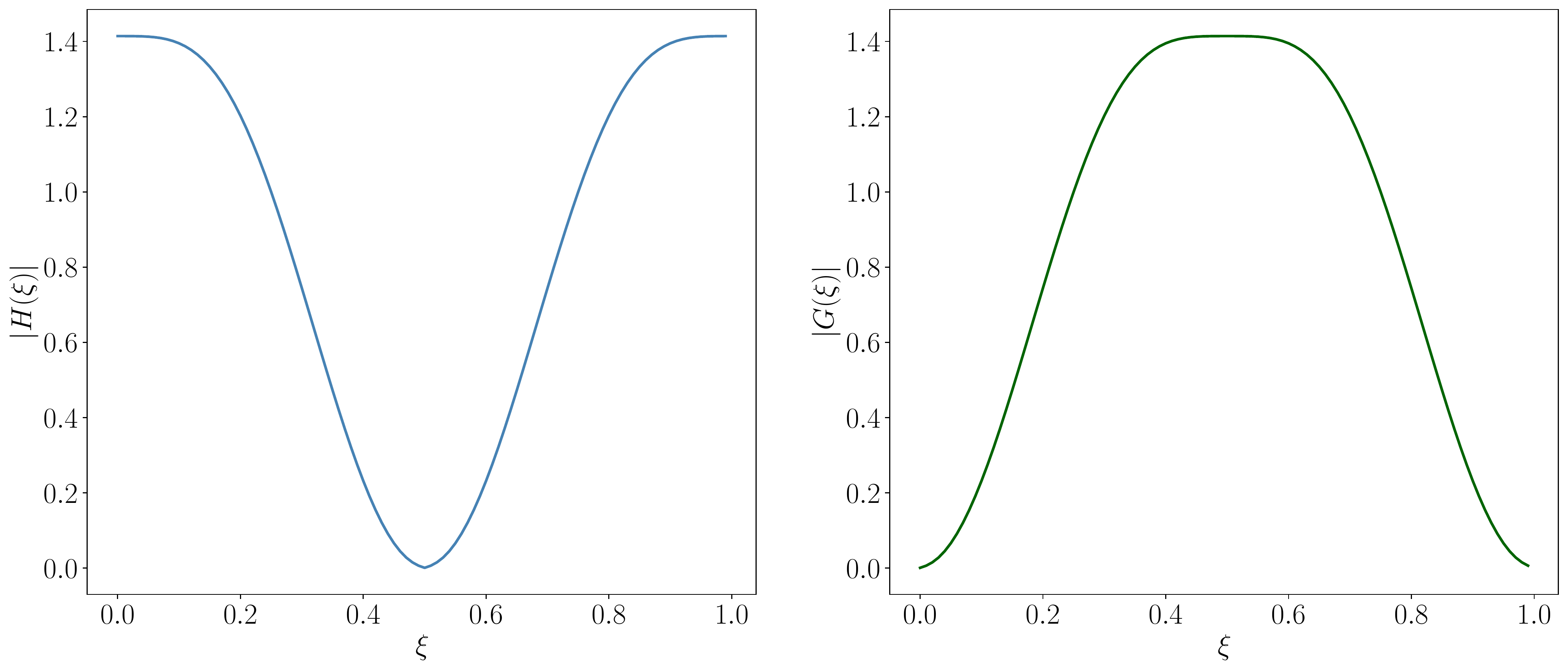} }}
	\\[2ex]    
	\subfloat[\centering Order $4$ - initial wavelet ]{{\includegraphics[width=0.44\columnwidth]{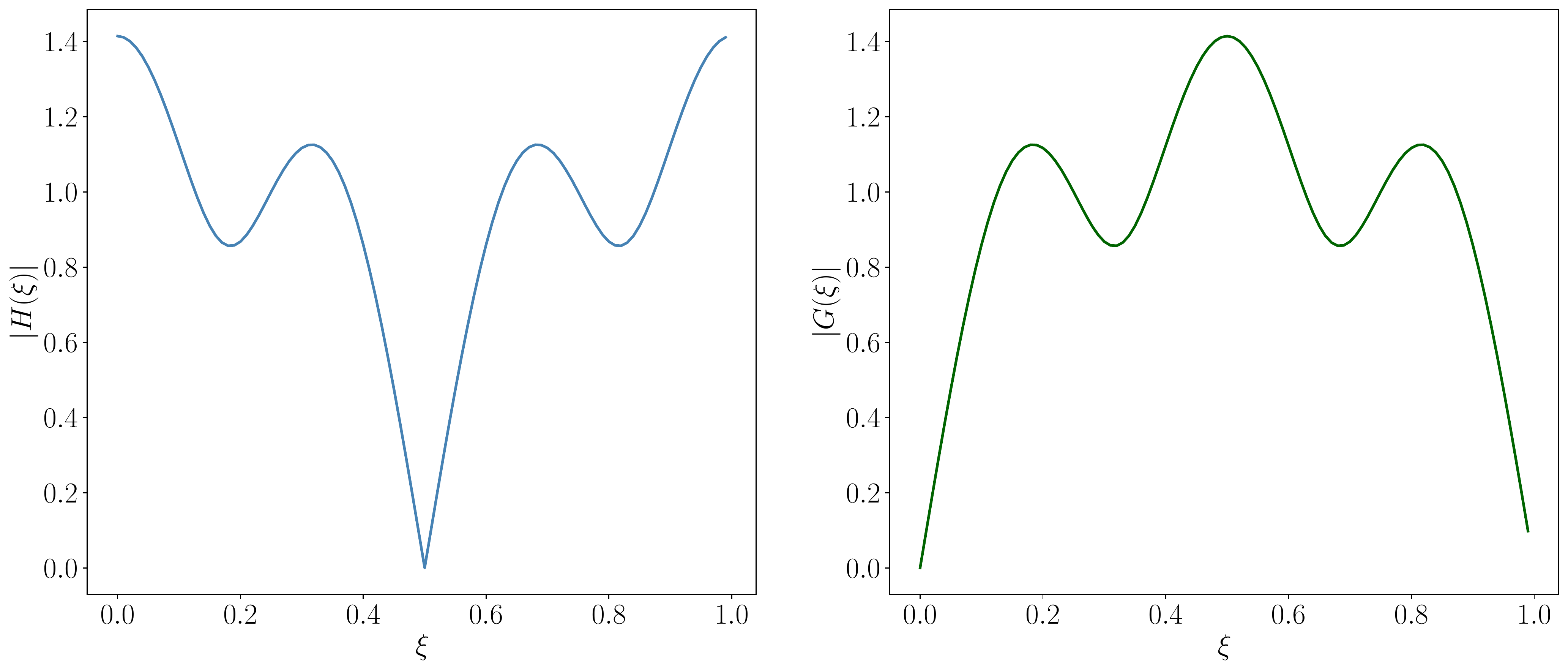} }}
	\subfloat[\centering Order $4$ - task-optimized wavelet ]{{\includegraphics[width=0.44\columnwidth]{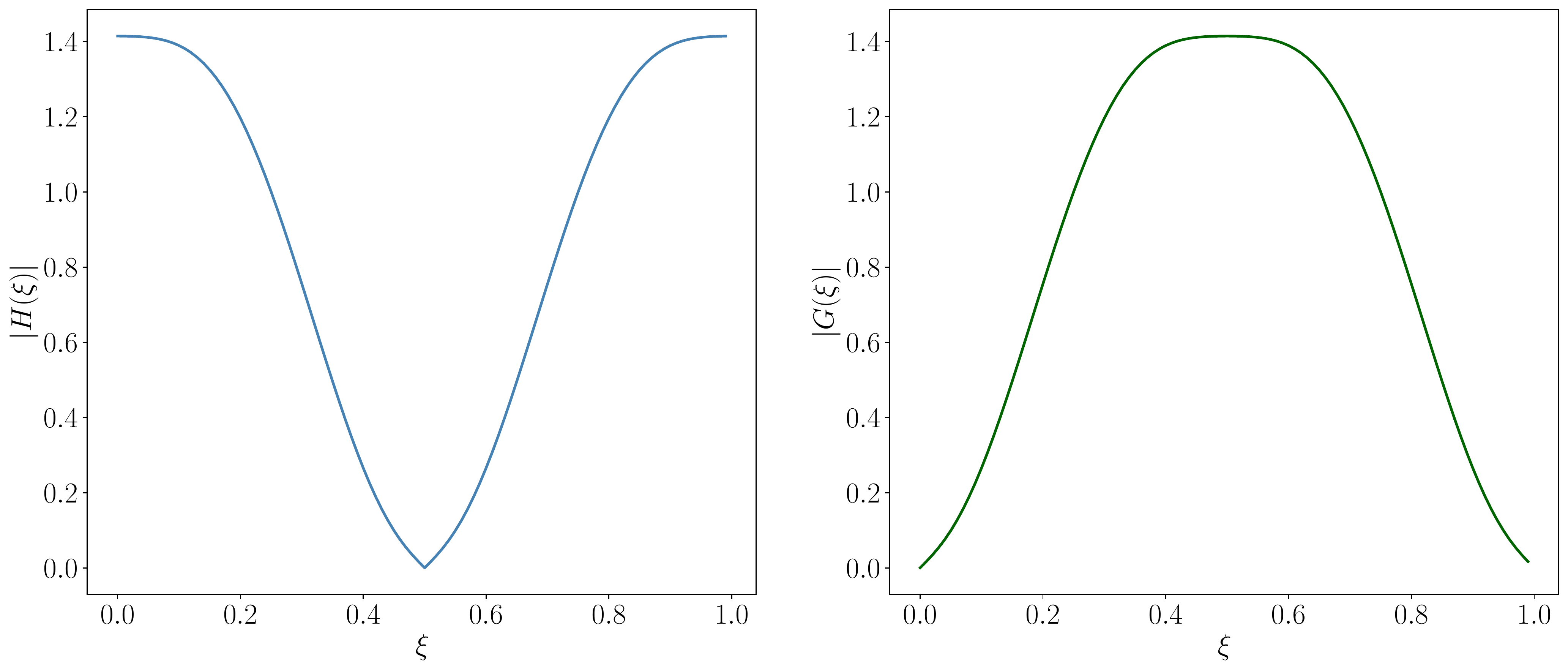} }}
	\\[2ex]    	
	\subfloat[\centering Order $5$ - initial wavelet ]{{\includegraphics[width=0.44\columnwidth]{spleen_refinement_masks/order_5_epoch_0_comp_0} }}
	\subfloat[\centering Order $5$ - task-optimized wavelet ]{{\includegraphics[width=0.44\columnwidth]{spleen_refinement_masks/order_5_epoch_250_comp_0} }}
	\\[2ex]    	
	\subfloat[\centering Order $6$ - initial wavelet ]{{\includegraphics[width=0.44\columnwidth]{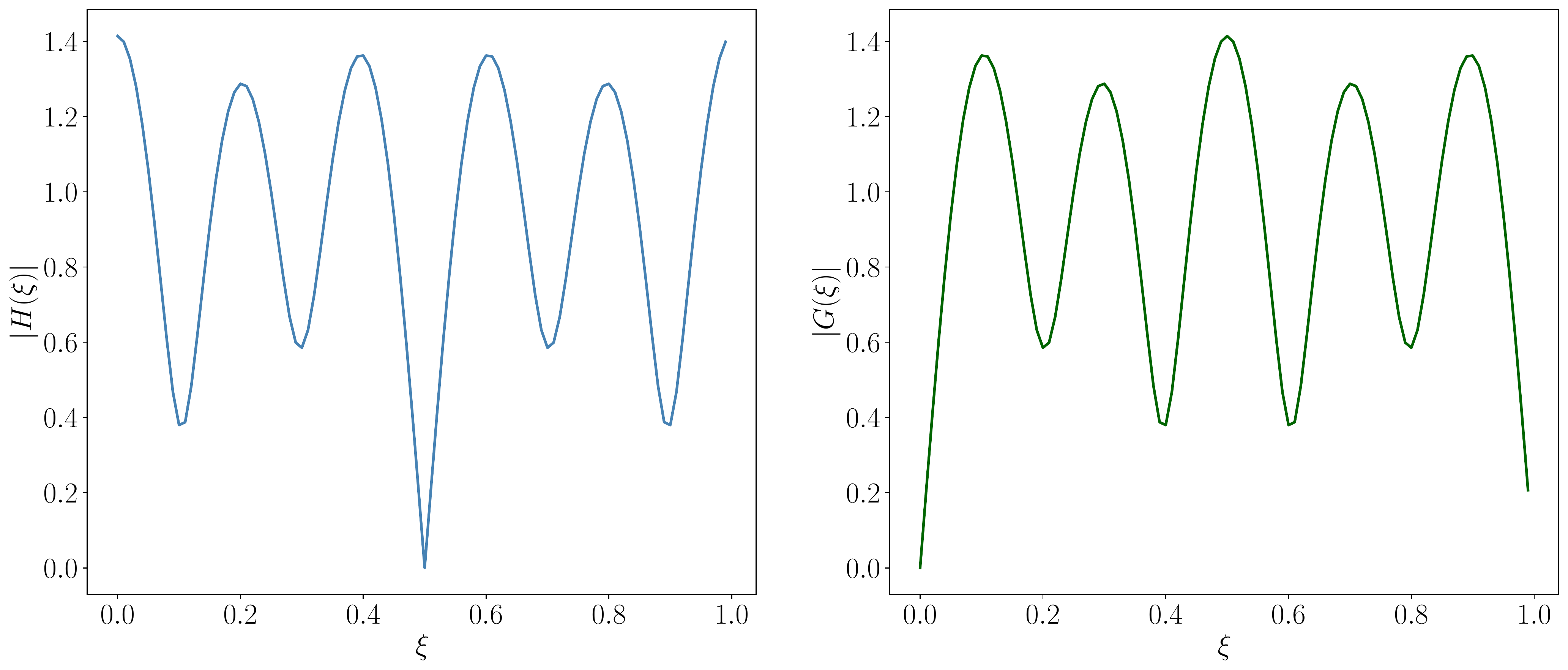} }}
	\subfloat[\centering Order $6$ - task-optimized wavelet ]{{\includegraphics[width=0.44\columnwidth]{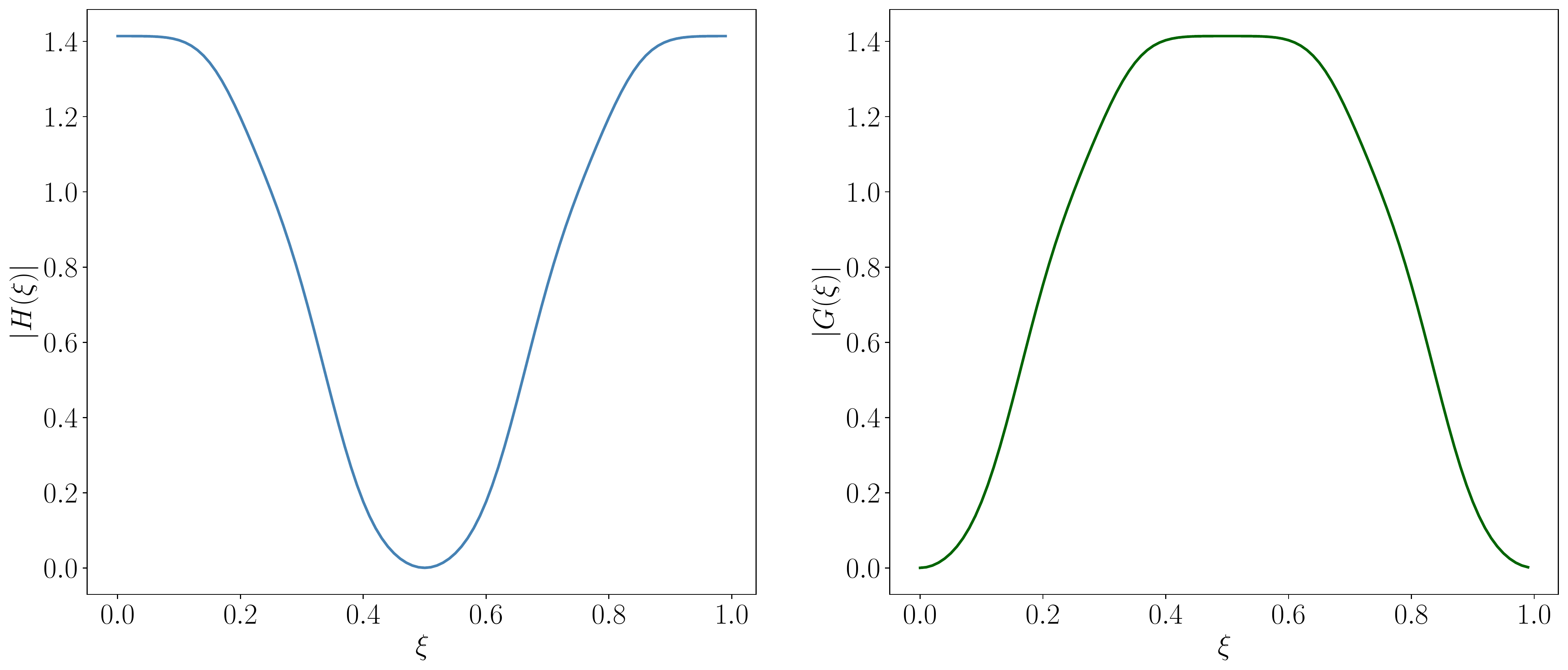} }}
	\\[2ex]    	
	\subfloat[\centering Order $7$ - initial wavelet ]{{\includegraphics[width=0.44\columnwidth]{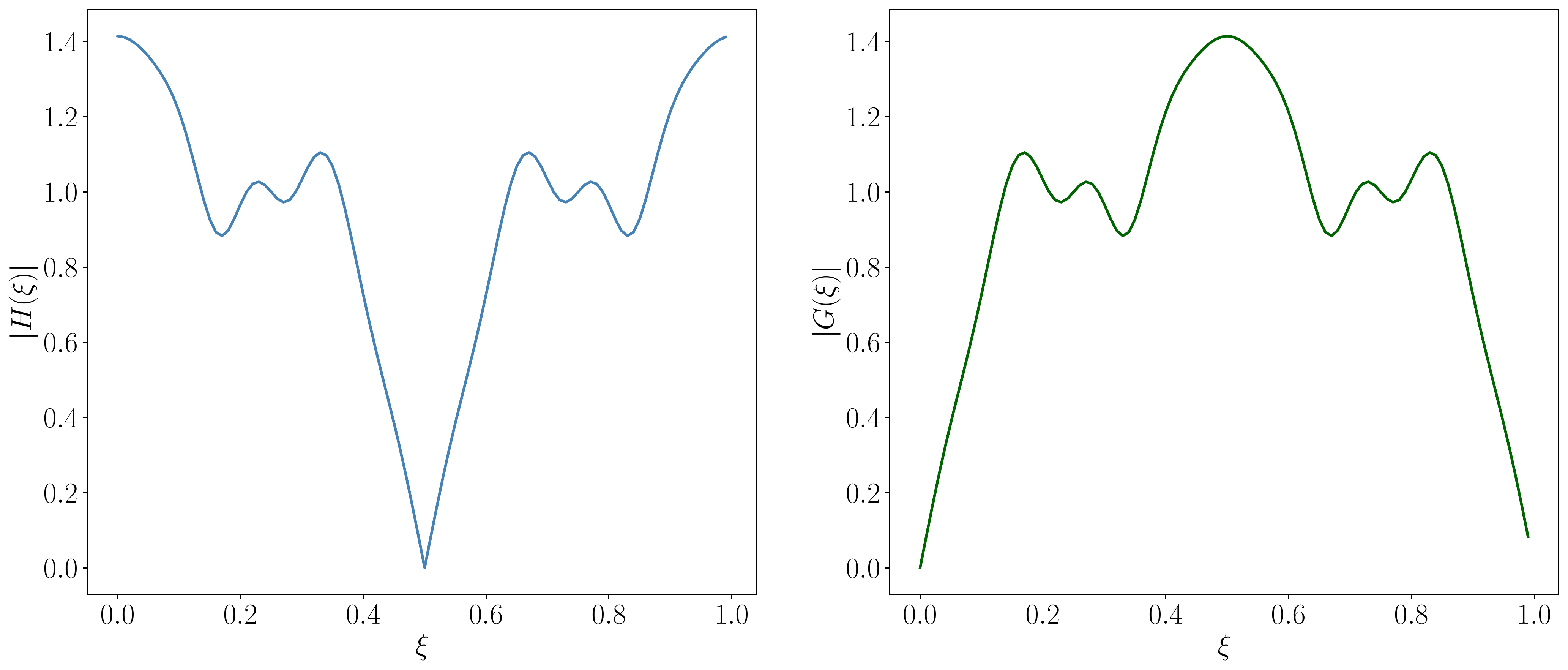} }}
	\subfloat[\centering Order $7$ - task-optimized wavelet ]{{\includegraphics[width=0.44\columnwidth]{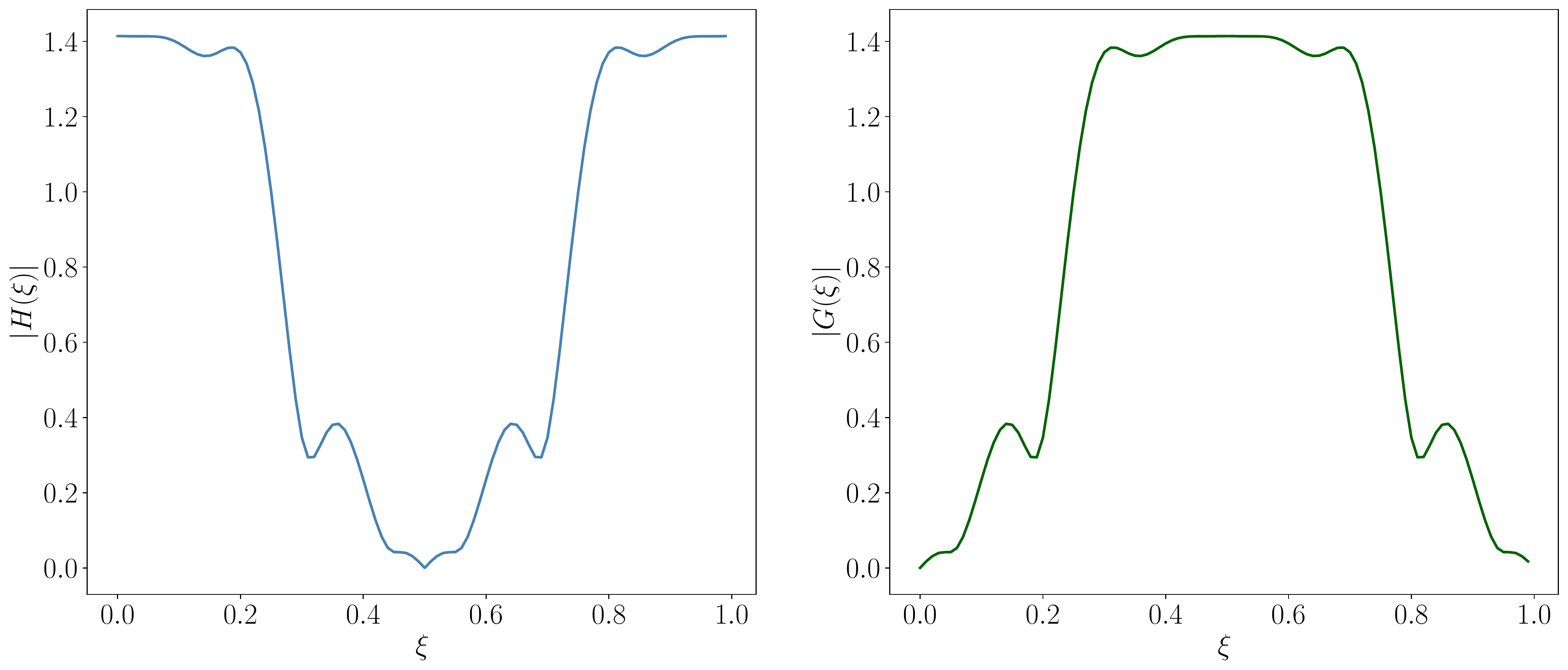} }}
	\\[2ex] 
	\subfloat[\centering Order $8$ - initial wavelet ]{{\includegraphics[width=0.44\columnwidth]{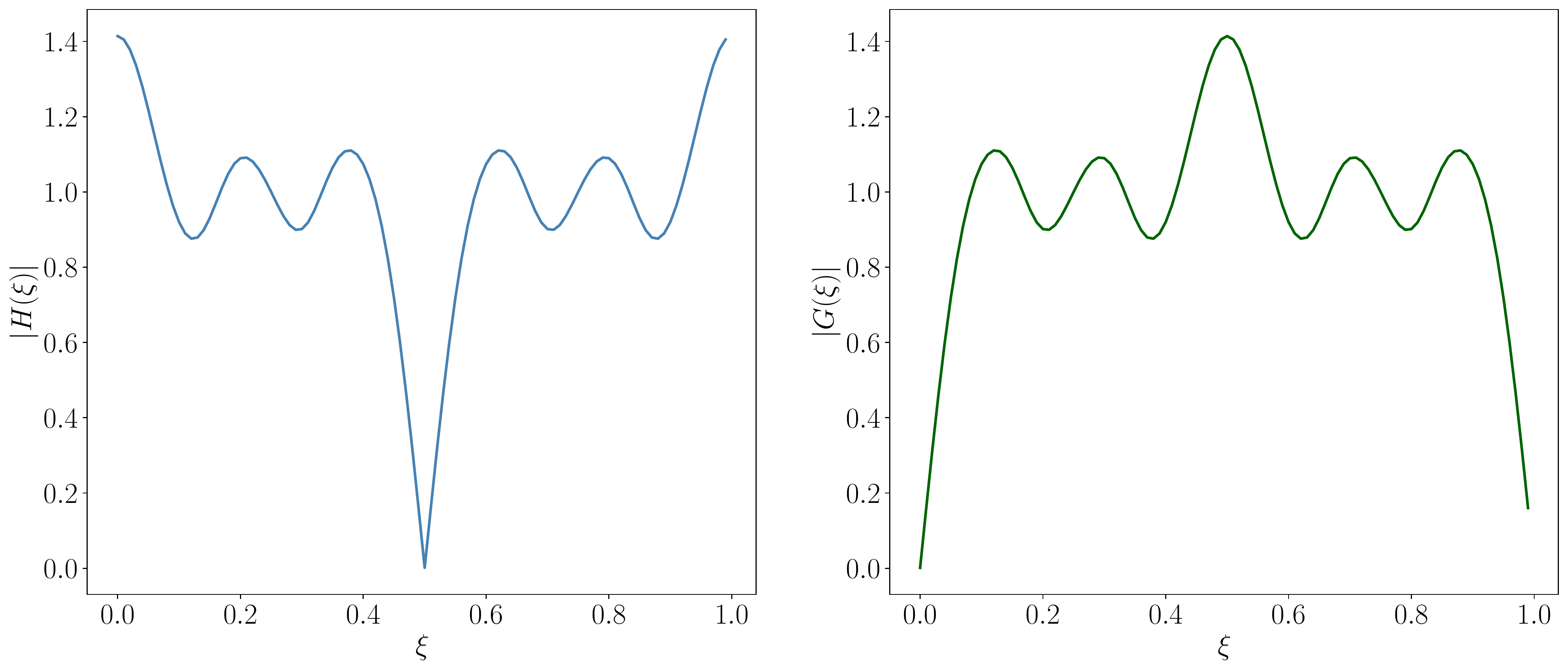} }}
	\subfloat[\centering Order $8$ - task-optimized wavelet ]{{\includegraphics[width=0.44\columnwidth]{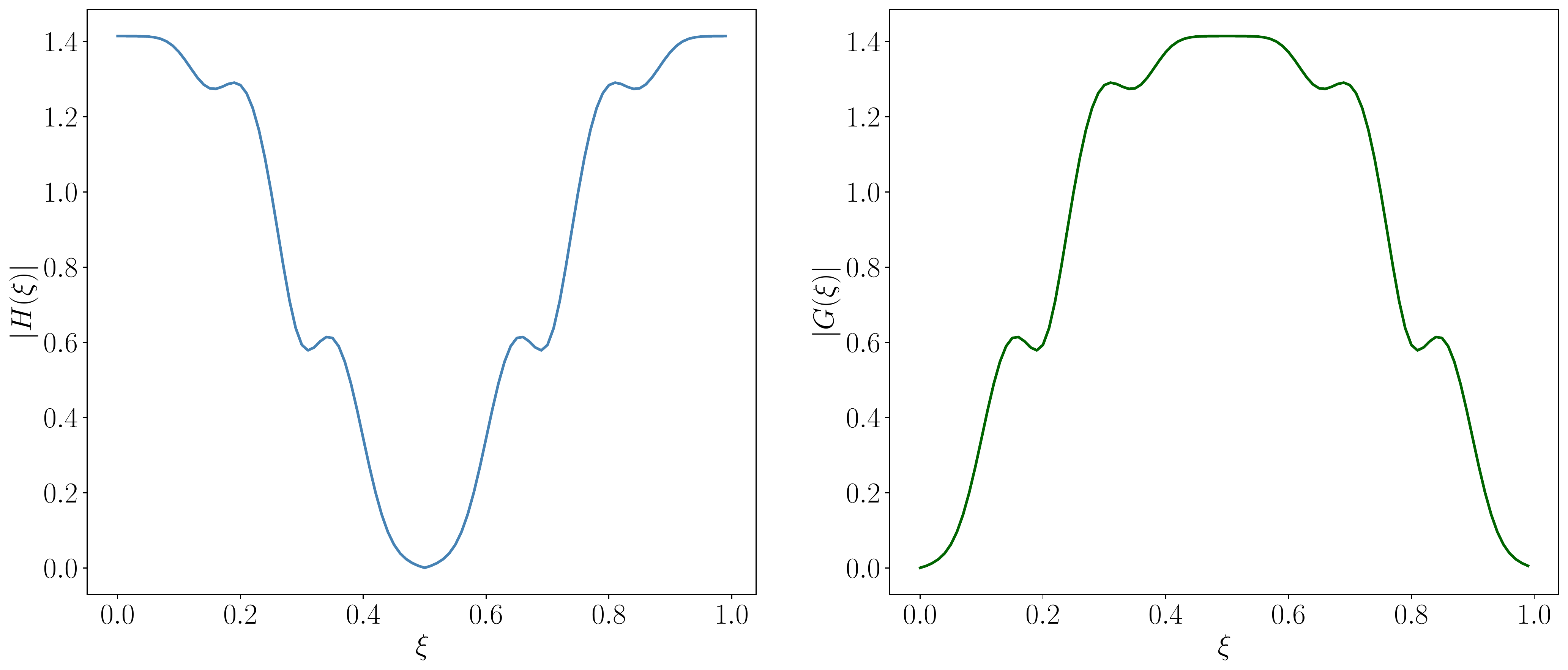} }}
	\\[2ex]    	   	
	\label{fig:spleen_refinement_comp_0}
\end{figure}
\newpage
\subsubsection{Spleen - second spatial component}
\begin{figure}[!b]
	\centering
	\subfloat[\centering Order $3$ - initial wavelet ]{{\includegraphics[width=0.44\columnwidth]{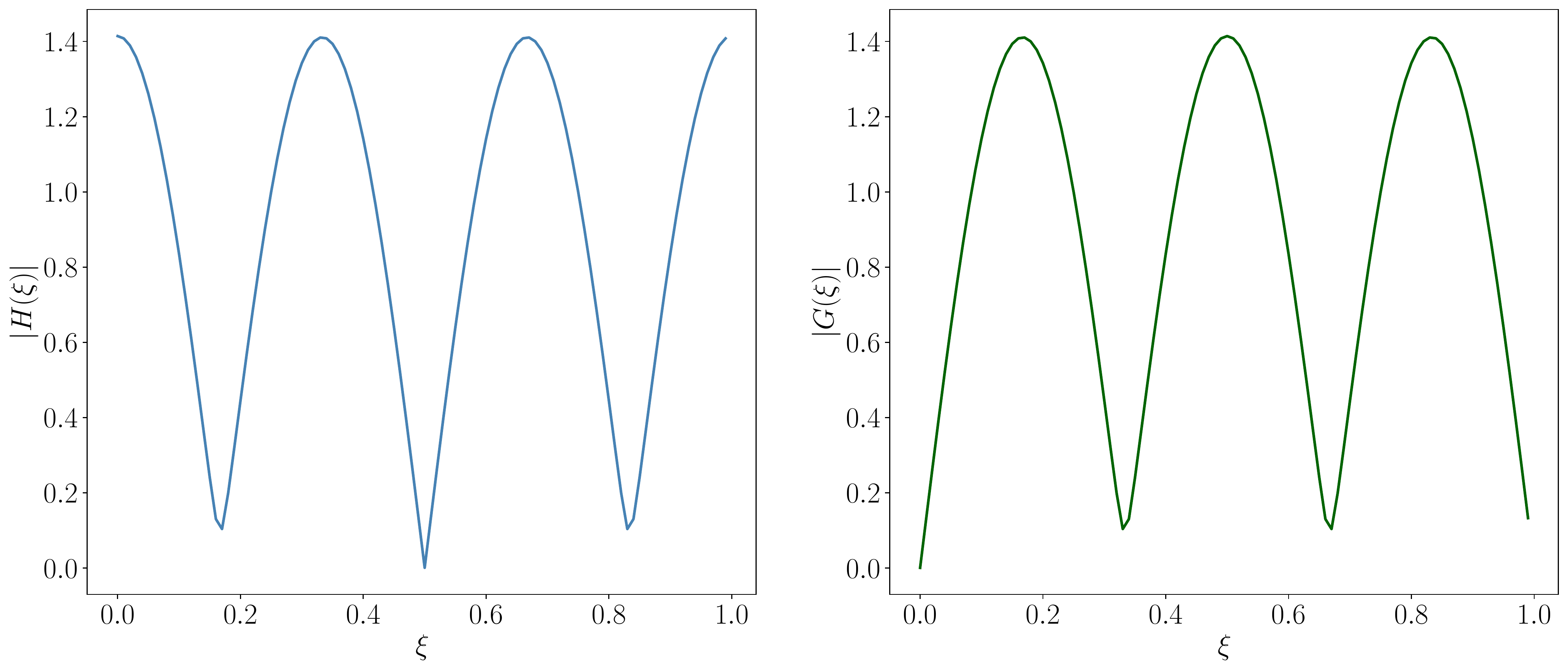} }}
	\subfloat[\centering Order $3$ - task-optimized wavelet ]{{\includegraphics[width=0.44\columnwidth]{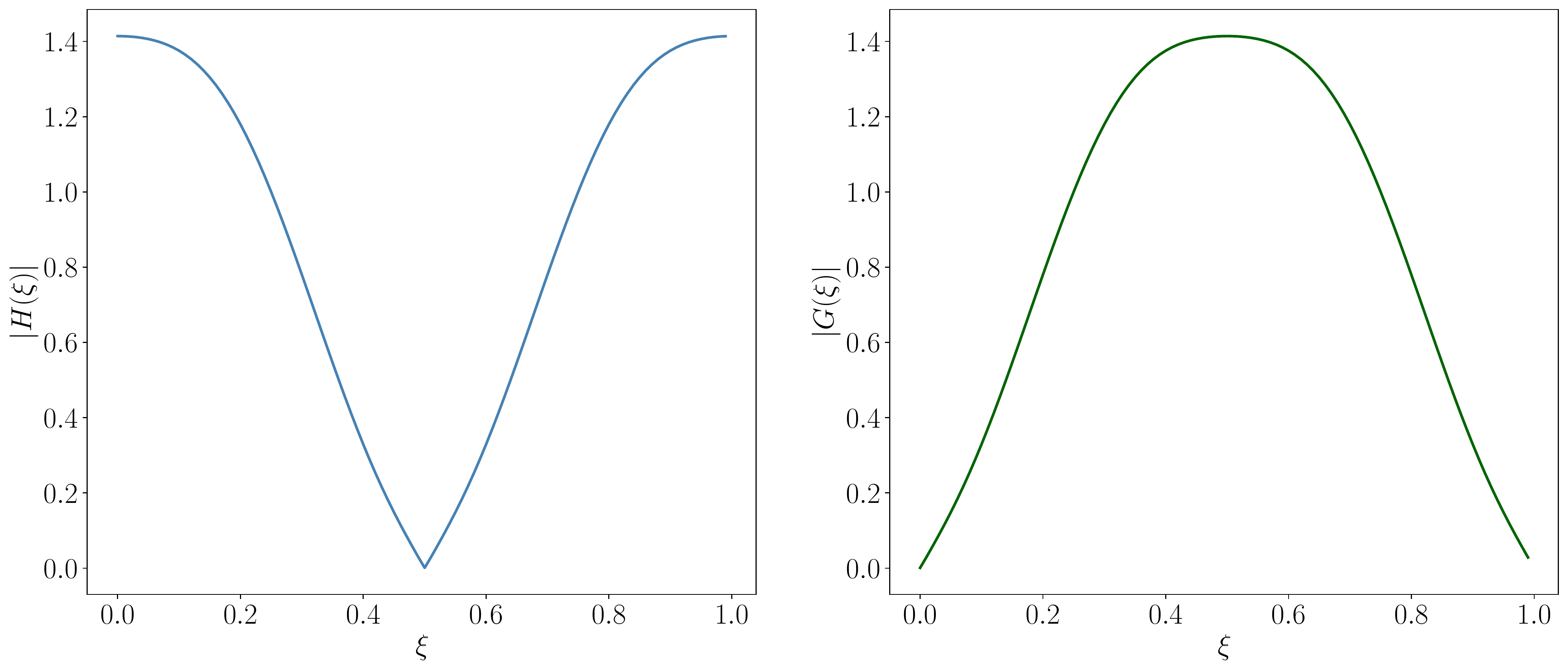} }}
	\\[2ex]    
	\subfloat[\centering Order $4$ - initial wavelet ]{{\includegraphics[width=0.44\columnwidth]{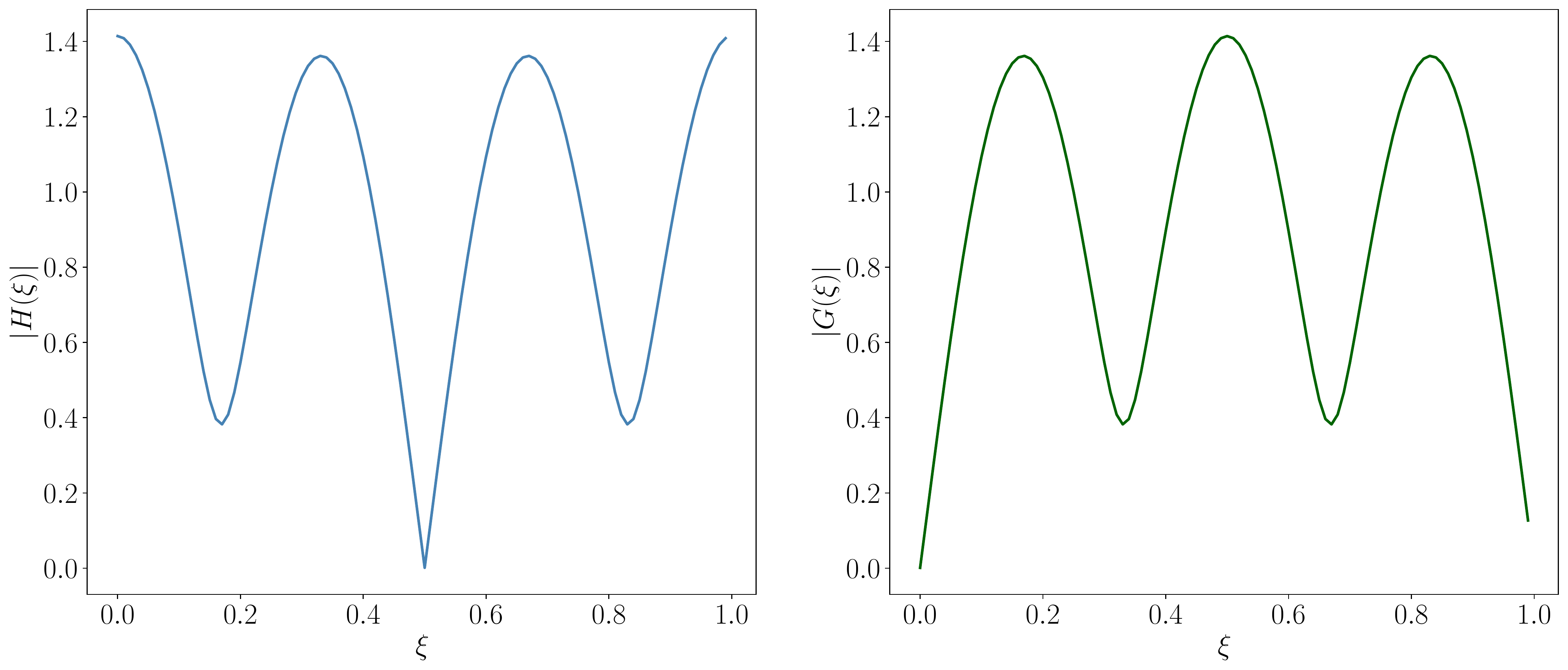} }}
	\subfloat[\centering Order $4$ - task-optimized wavelet ]{{\includegraphics[width=0.44\columnwidth]{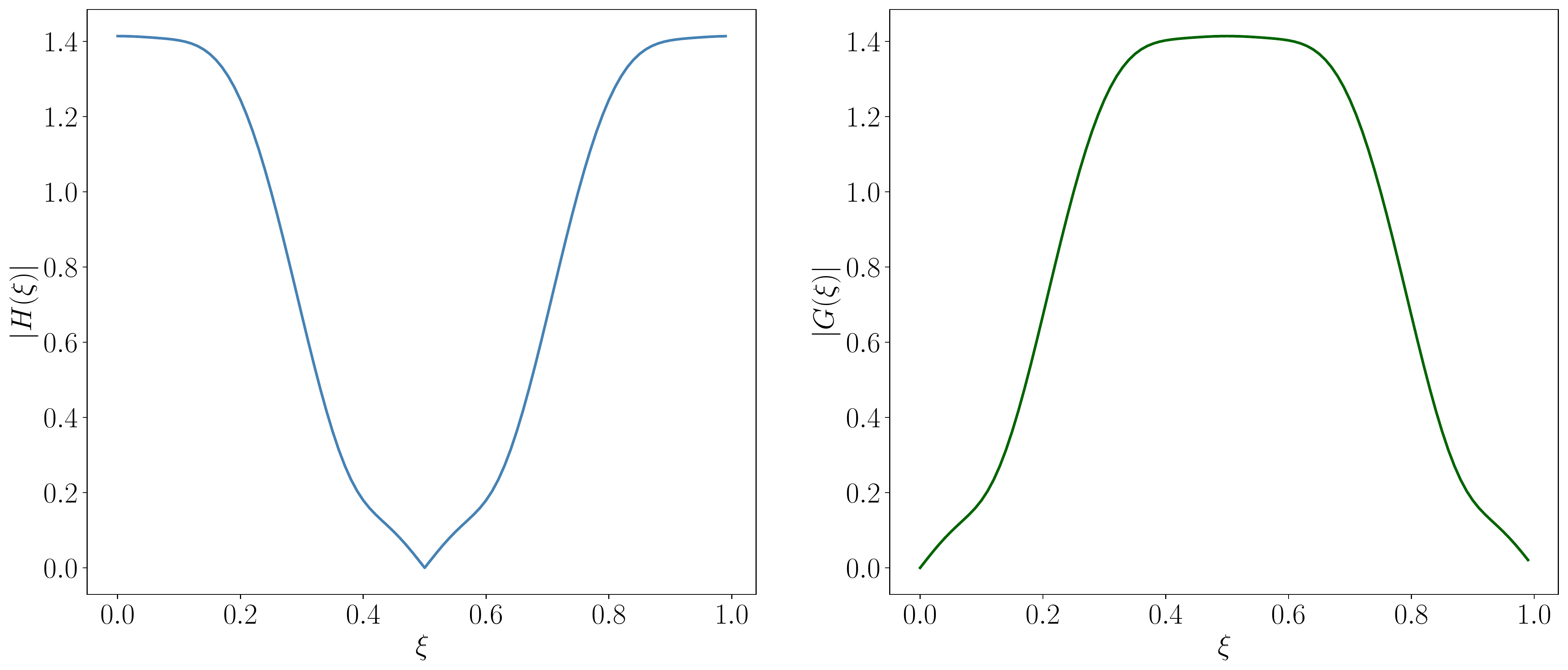} }}
	\\[2ex]    	
	\subfloat[\centering Order $5$ - initial wavelet ]{{\includegraphics[width=0.44\columnwidth]{spleen_refinement_masks/order_5_epoch_0_comp_1} }}
	\subfloat[\centering Order $5$ - task-optimized wavelet ]{{\includegraphics[width=0.44\columnwidth]{spleen_refinement_masks/order_5_epoch_250_comp_1} }}
	\\[2ex]    	
	\subfloat[\centering Order $6$ - initial wavelet ]{{\includegraphics[width=0.44\columnwidth]{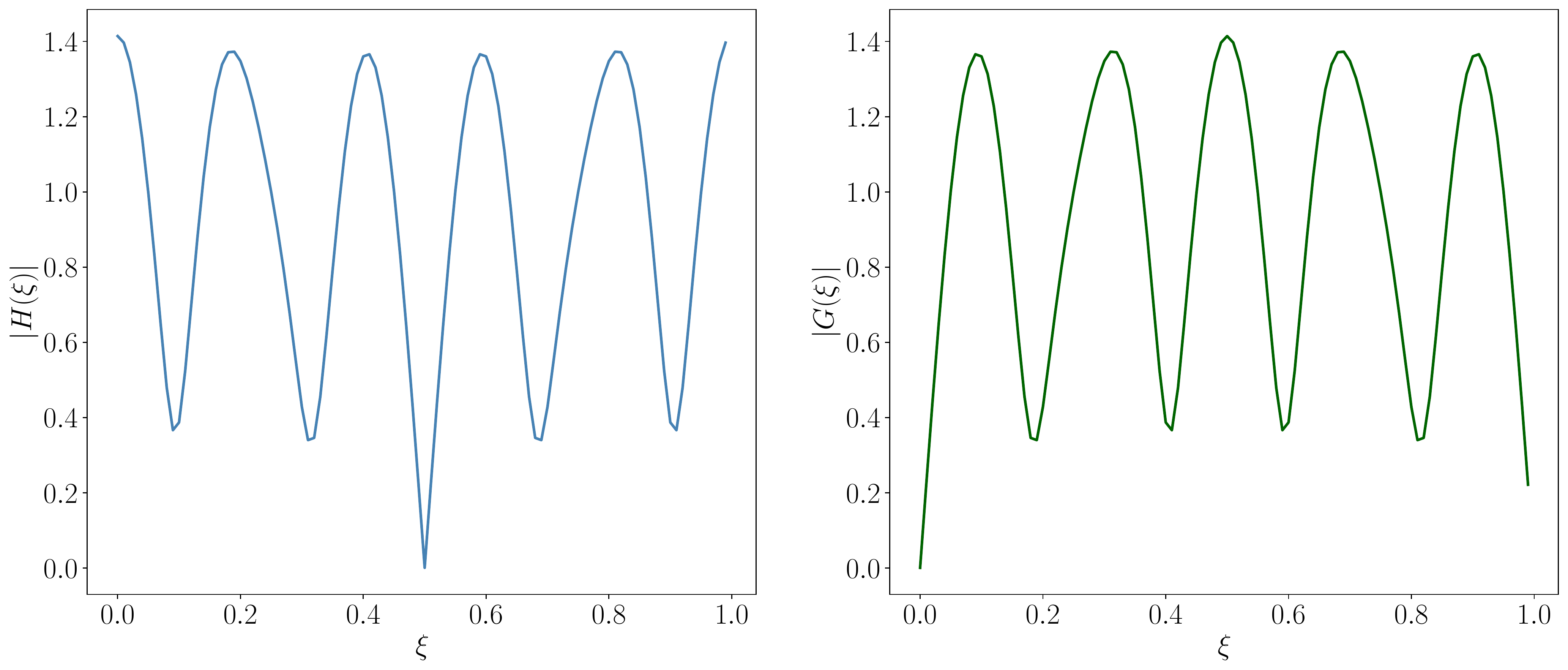} }}
	\subfloat[\centering Order $6$ - task-optimized wavelet ]{{\includegraphics[width=0.44\columnwidth]{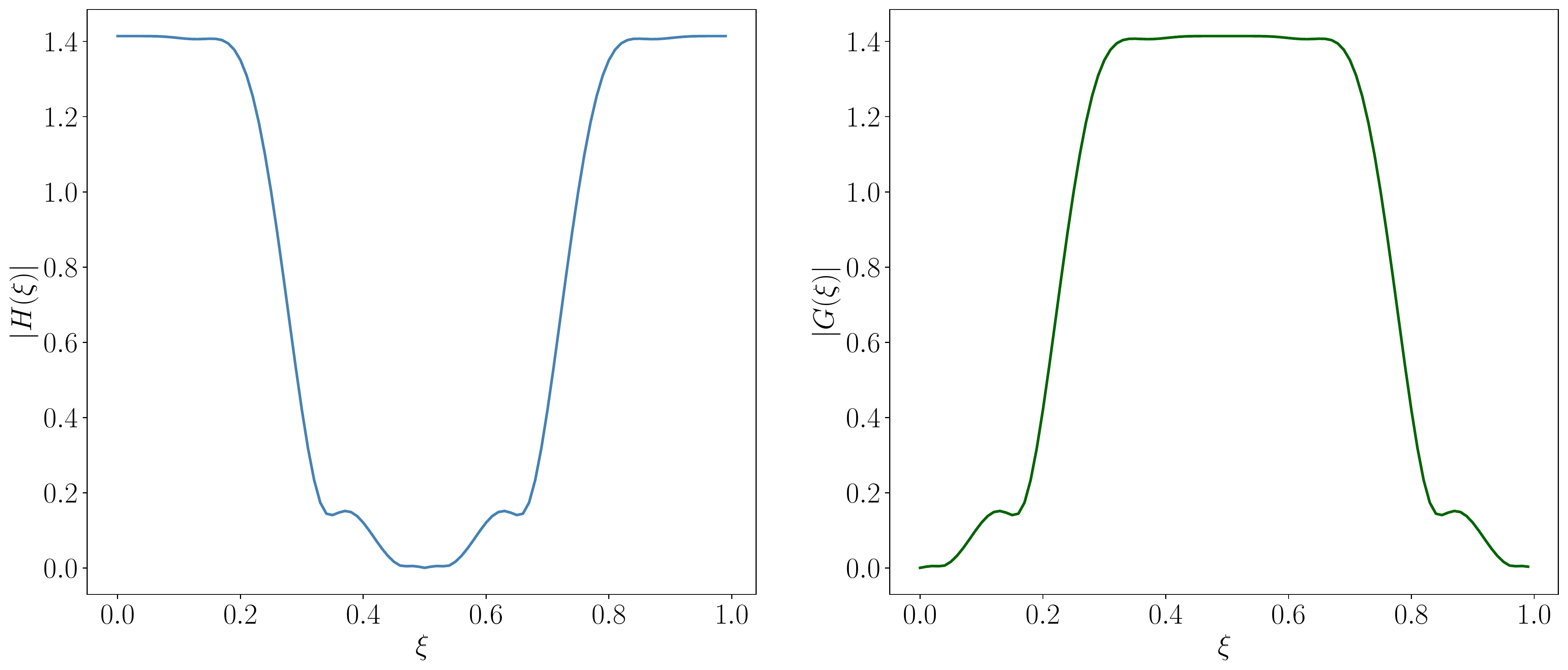} }}
	\\[2ex]    	
	\subfloat[\centering Order $7$ - initial wavelet ]{{\includegraphics[width=0.44\columnwidth]{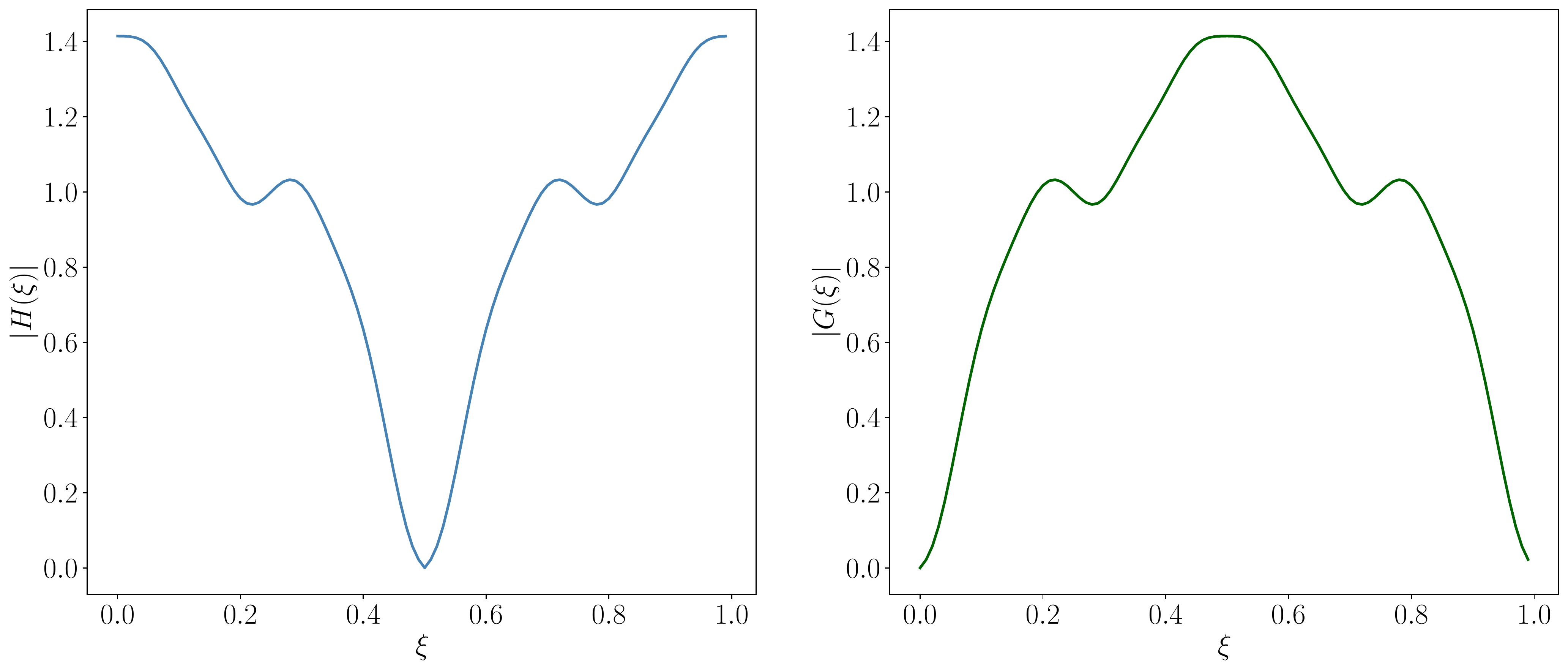} }}
	\subfloat[\centering Order $7$ - task-optimized wavelet ]{{\includegraphics[width=0.44\columnwidth]{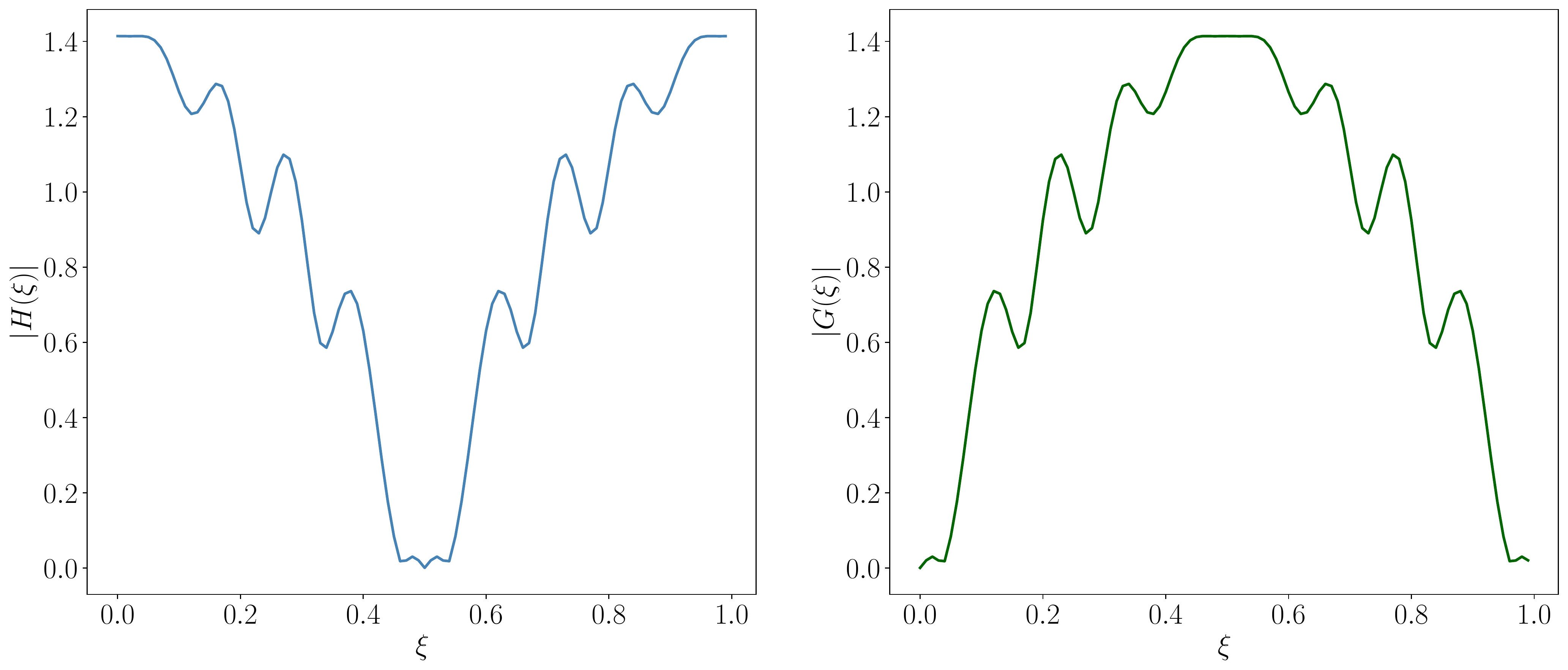} }}
	\\[2ex] 
	\subfloat[\centering Order $8$ - initial wavelet ]{{\includegraphics[width=0.44\columnwidth]{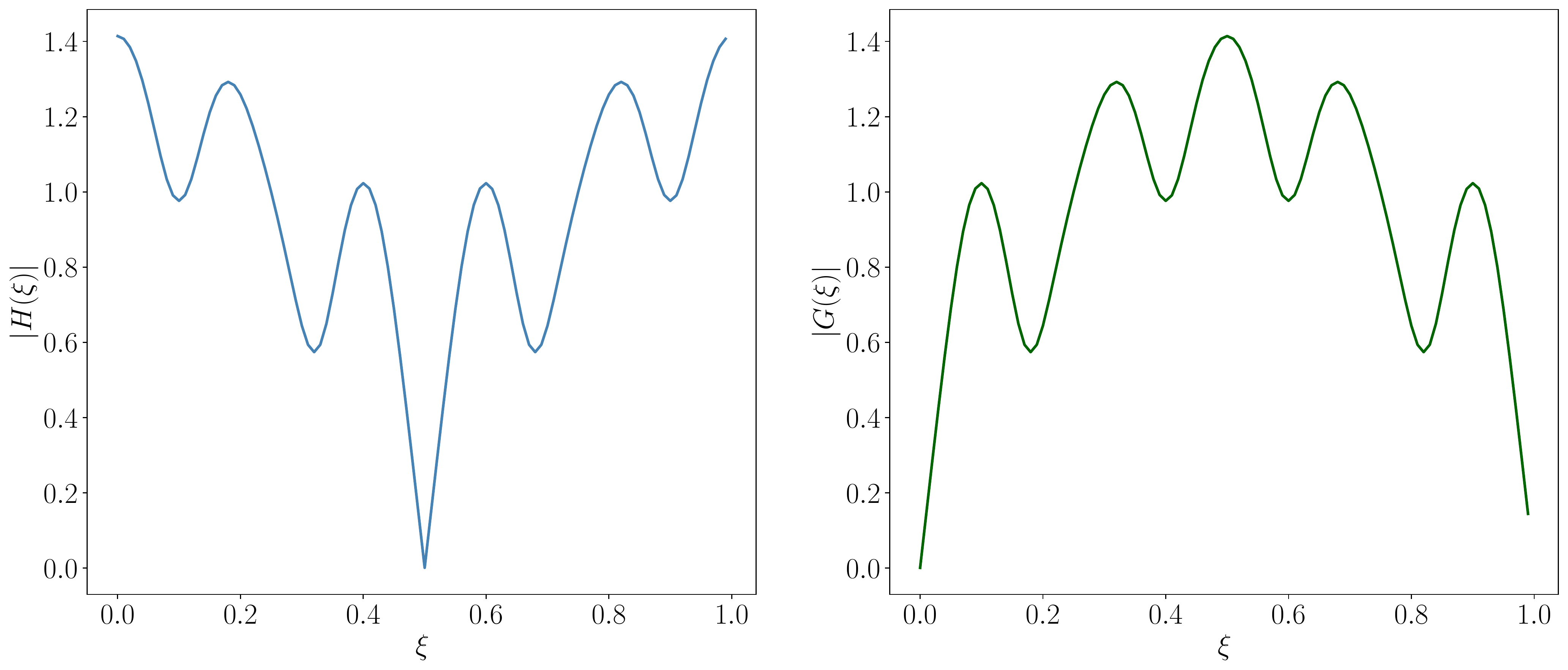} }}
	\subfloat[\centering Order $8$ - task-optimized wavelet ]{{\includegraphics[width=0.44\columnwidth]{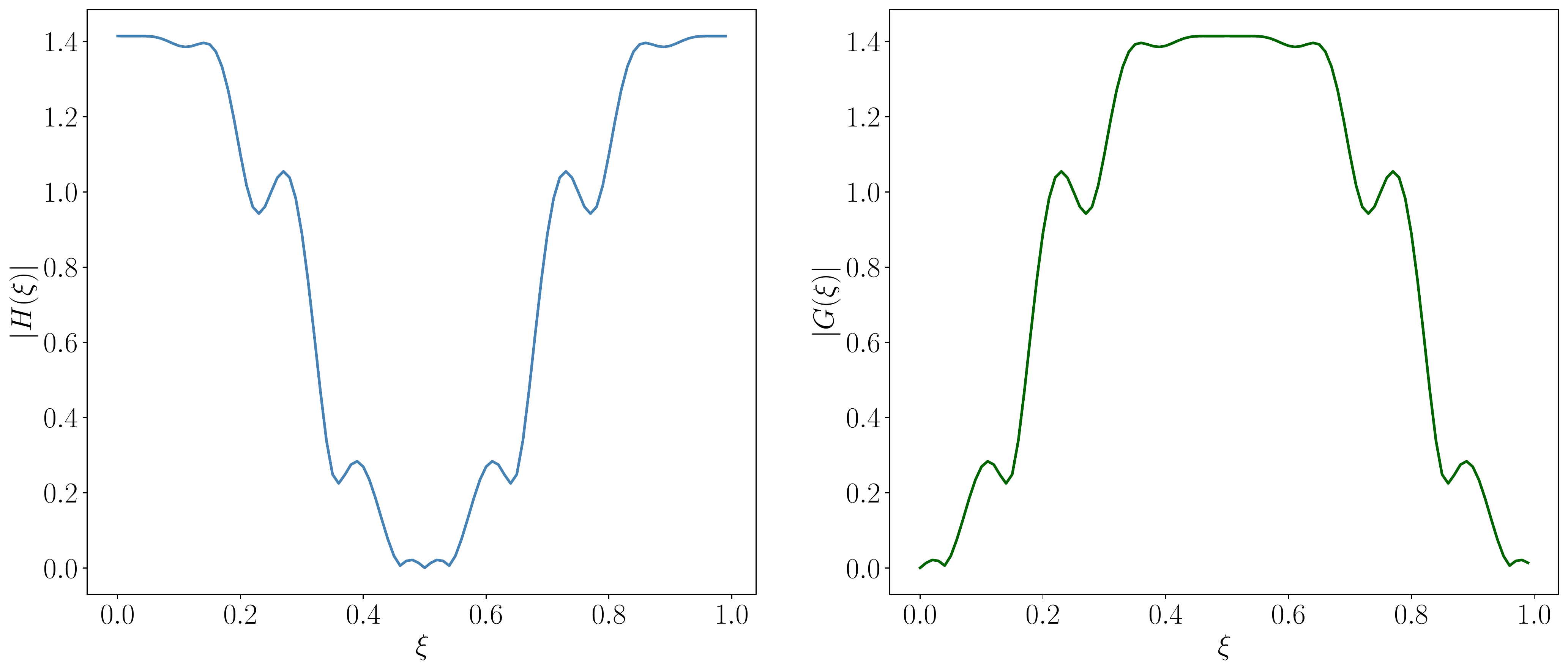} }}
	\\[2ex]    	   	
	\label{fig:spleen_refinement_comp_1}
\end{figure}
\newpage
\subsubsection{Prostate - first spatial component}
\begin{figure}[!b]
	\centering
	\subfloat[\centering Order $3$ - initial wavelet ]{{\includegraphics[width=0.44\columnwidth]{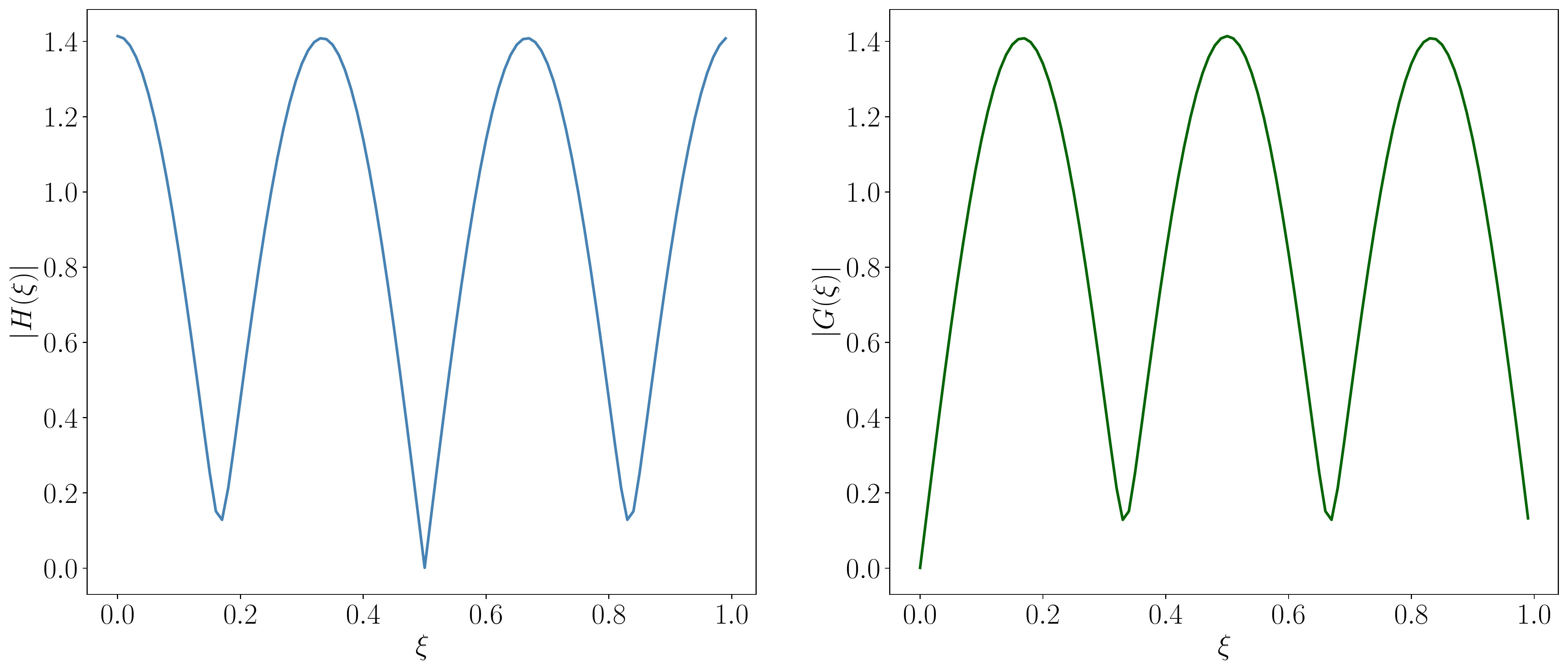} }}
	\subfloat[\centering Order $3$ - task-optimized wavelet ]{{\includegraphics[width=0.44\columnwidth]{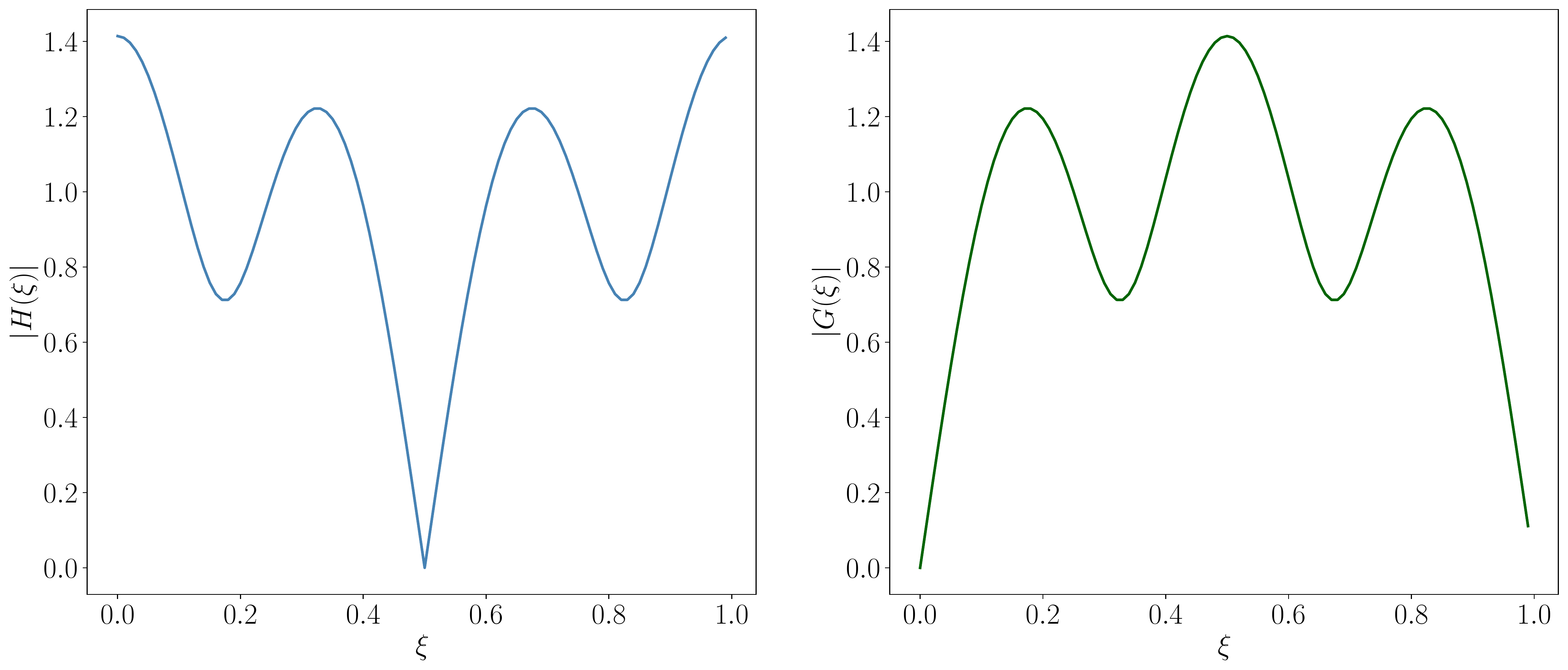} }}
	\\[2ex]    
	\subfloat[\centering Order $4$ - initial wavelet ]{{\includegraphics[width=0.44\columnwidth]{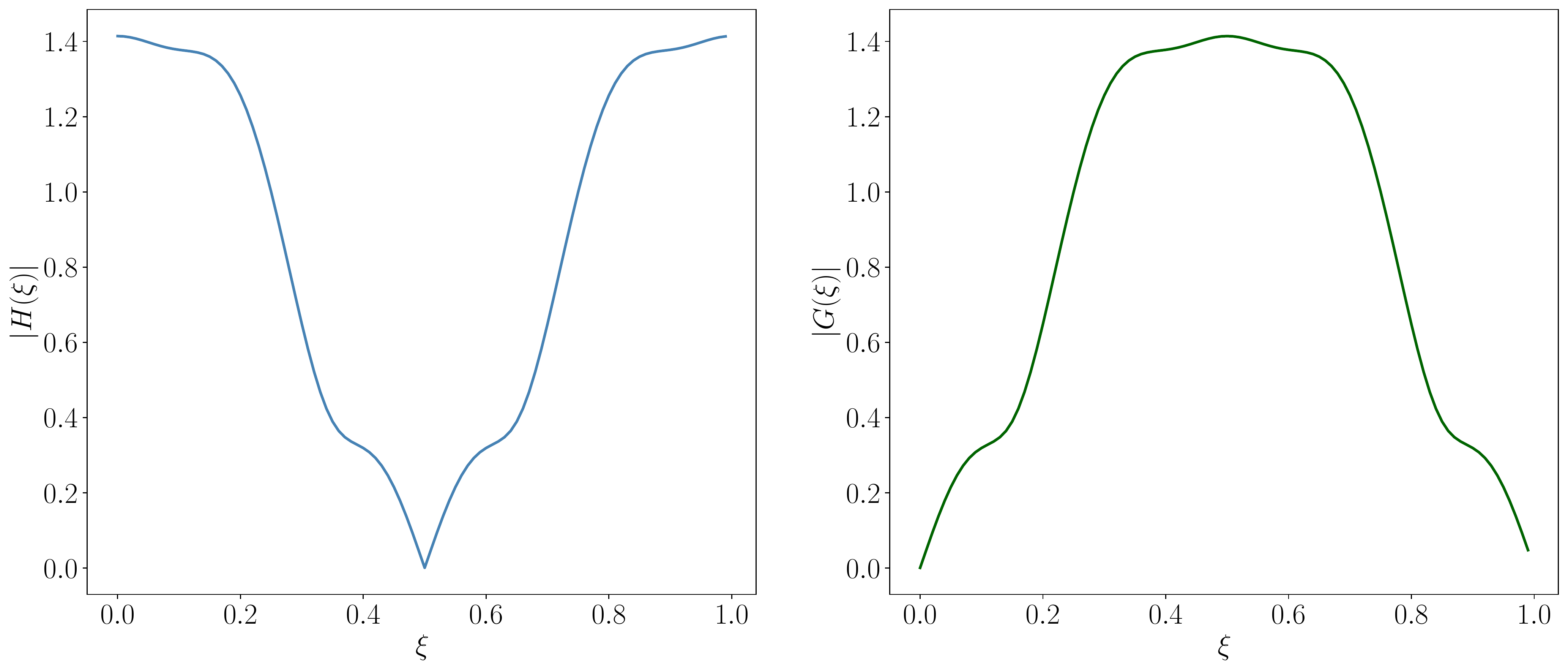} }}
	\subfloat[\centering Order $4$ - task-optimized wavelet ]{{\includegraphics[width=0.44\columnwidth]{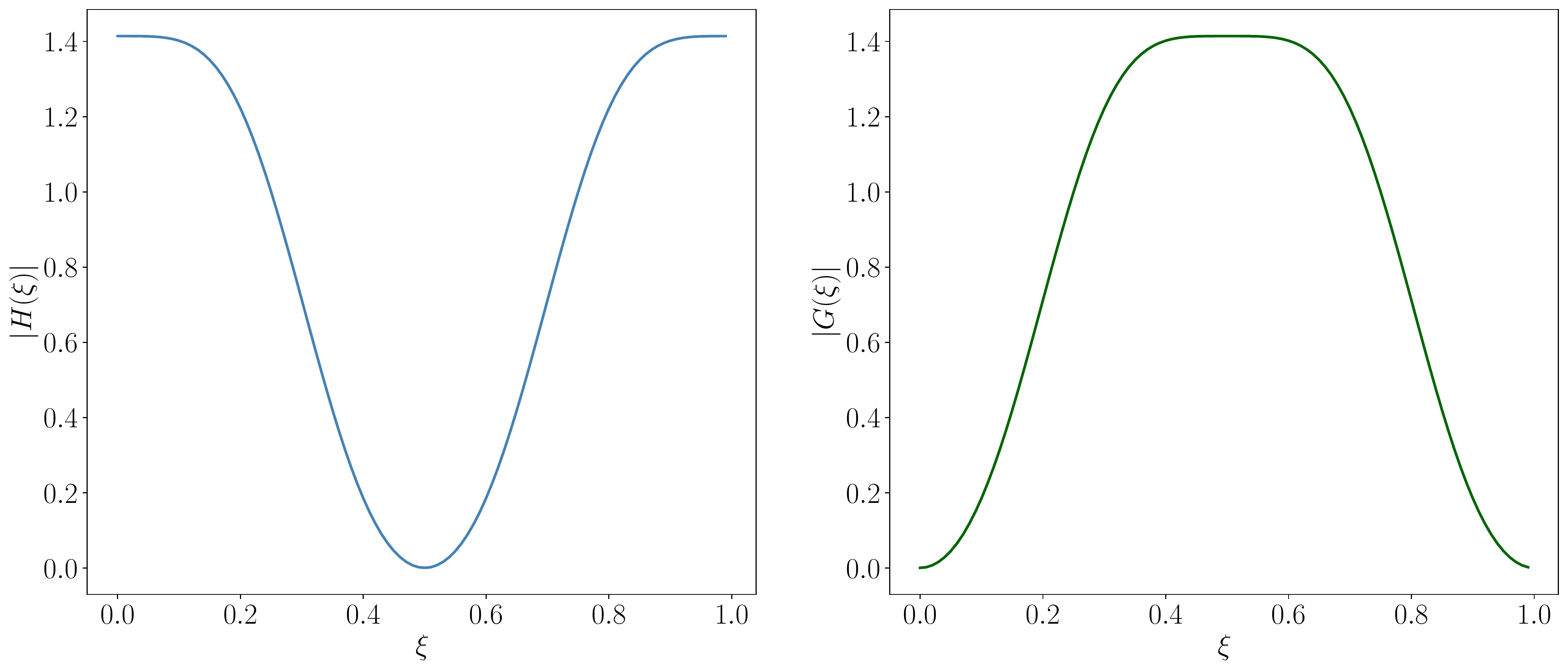} }}
	\\[2ex]    	
	\subfloat[\centering Order $5$ - initial wavelet ]{{\includegraphics[width=0.44\columnwidth]{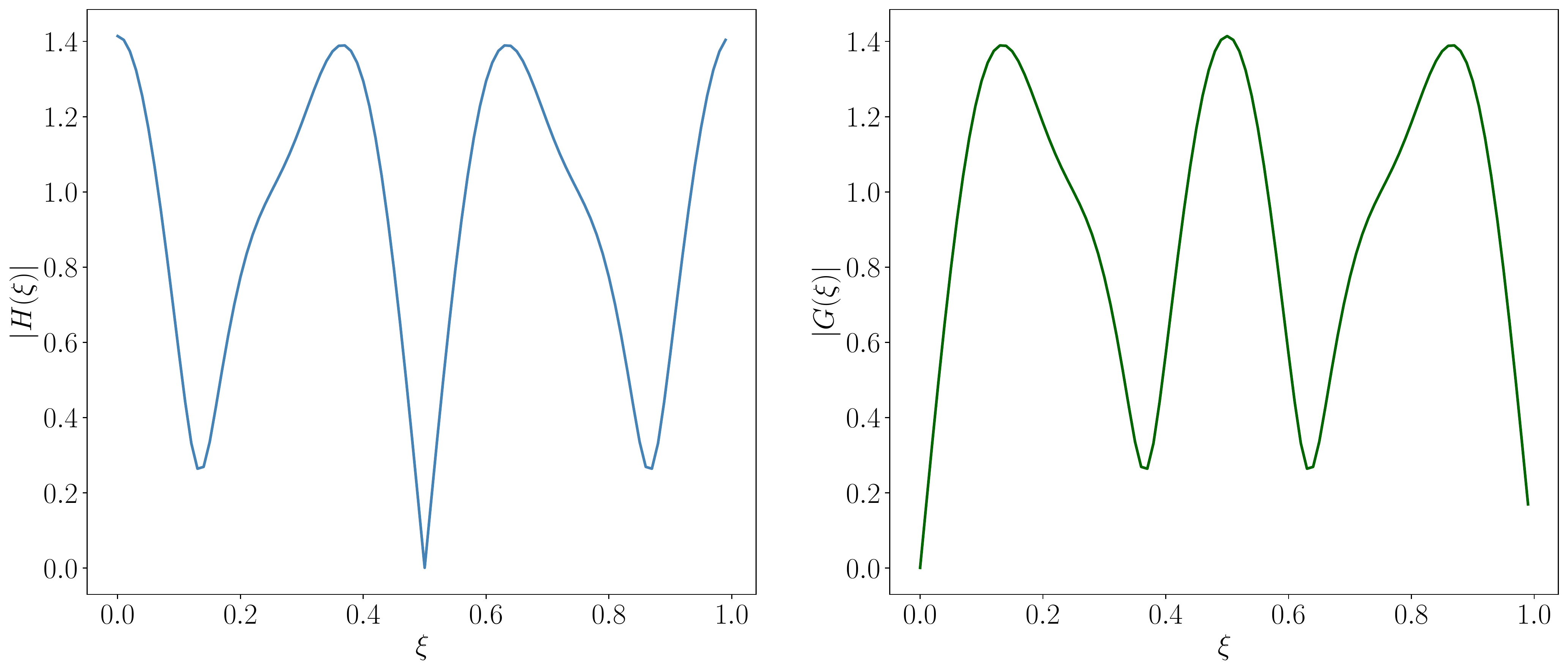} }}
	\subfloat[\centering Order $5$ - task-optimized wavelet ]{{\includegraphics[width=0.44\columnwidth]{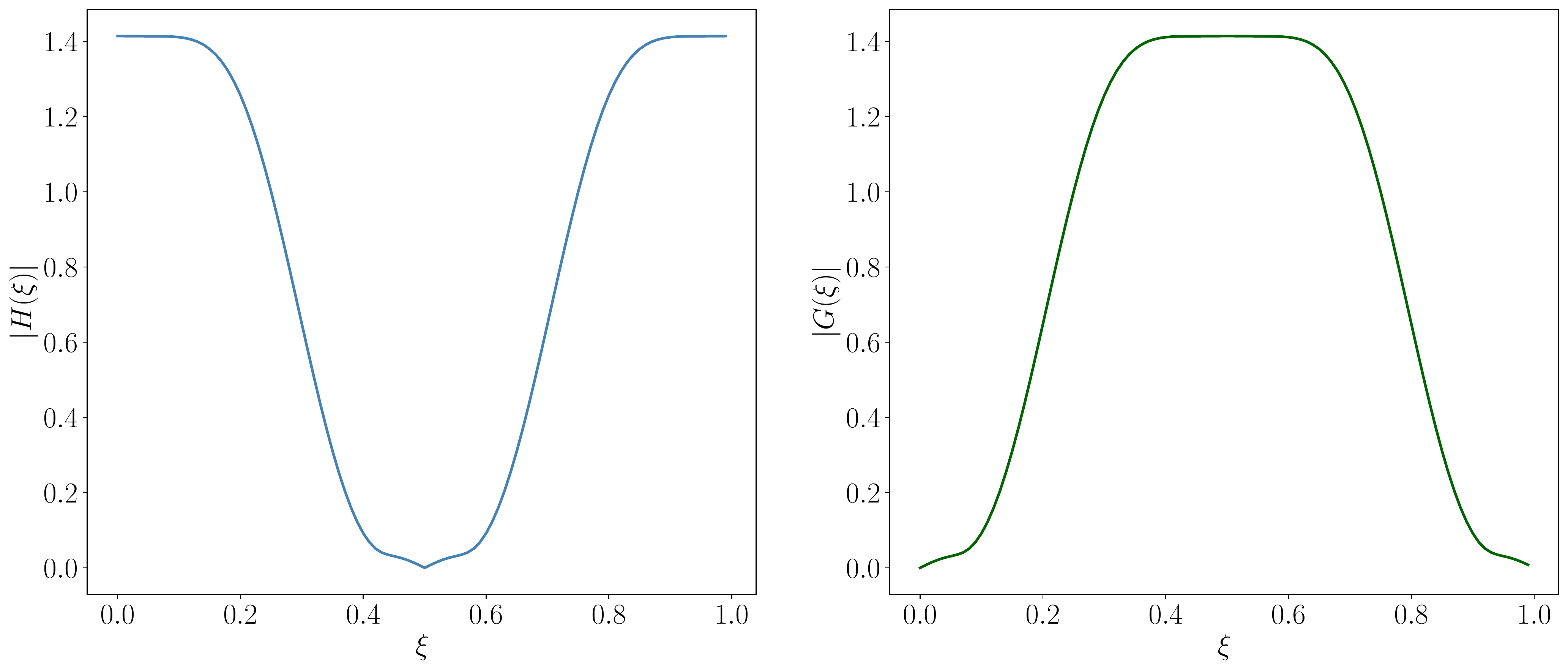} }}
	\\[2ex]    	
	\subfloat[\centering Order $6$ - initial wavelet ]{{\includegraphics[width=0.44\columnwidth]{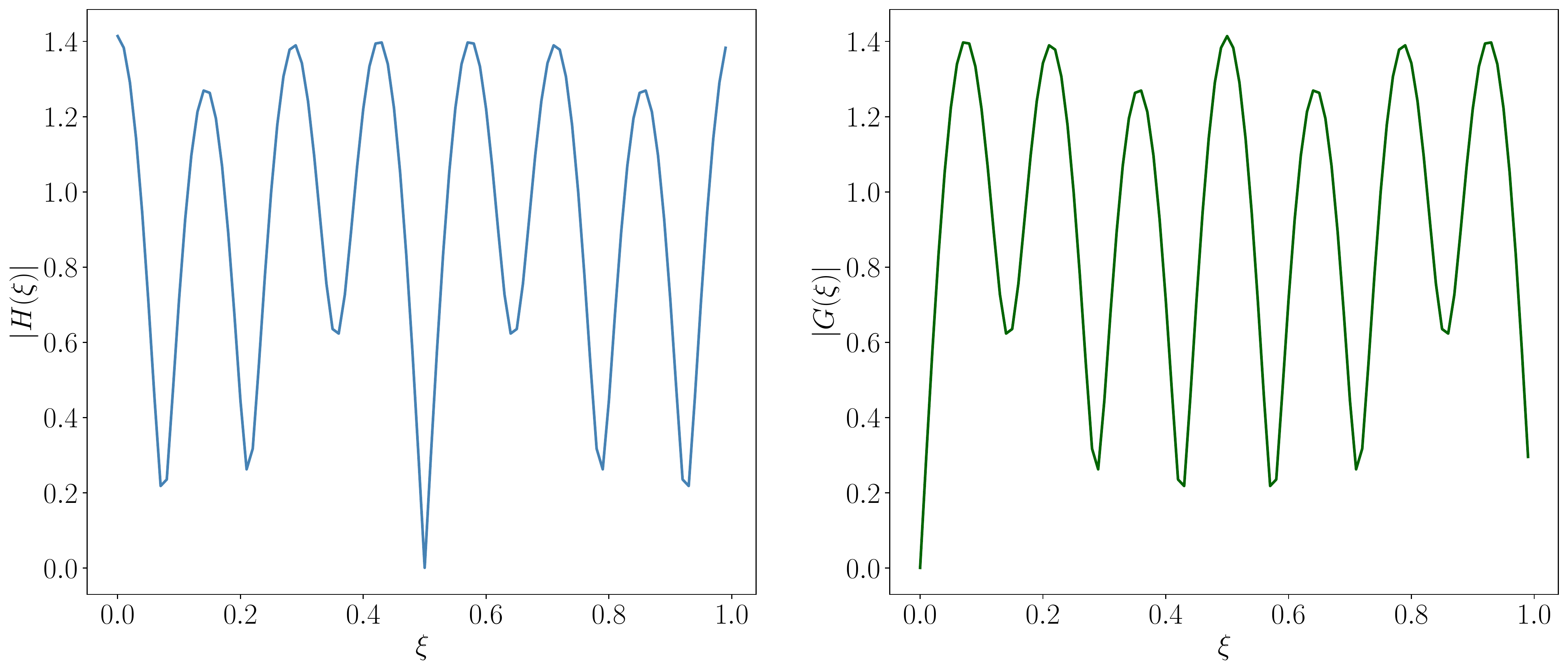} }}
	\subfloat[\centering Order $6$ - task-optimized wavelet ]{{\includegraphics[width=0.44\columnwidth]{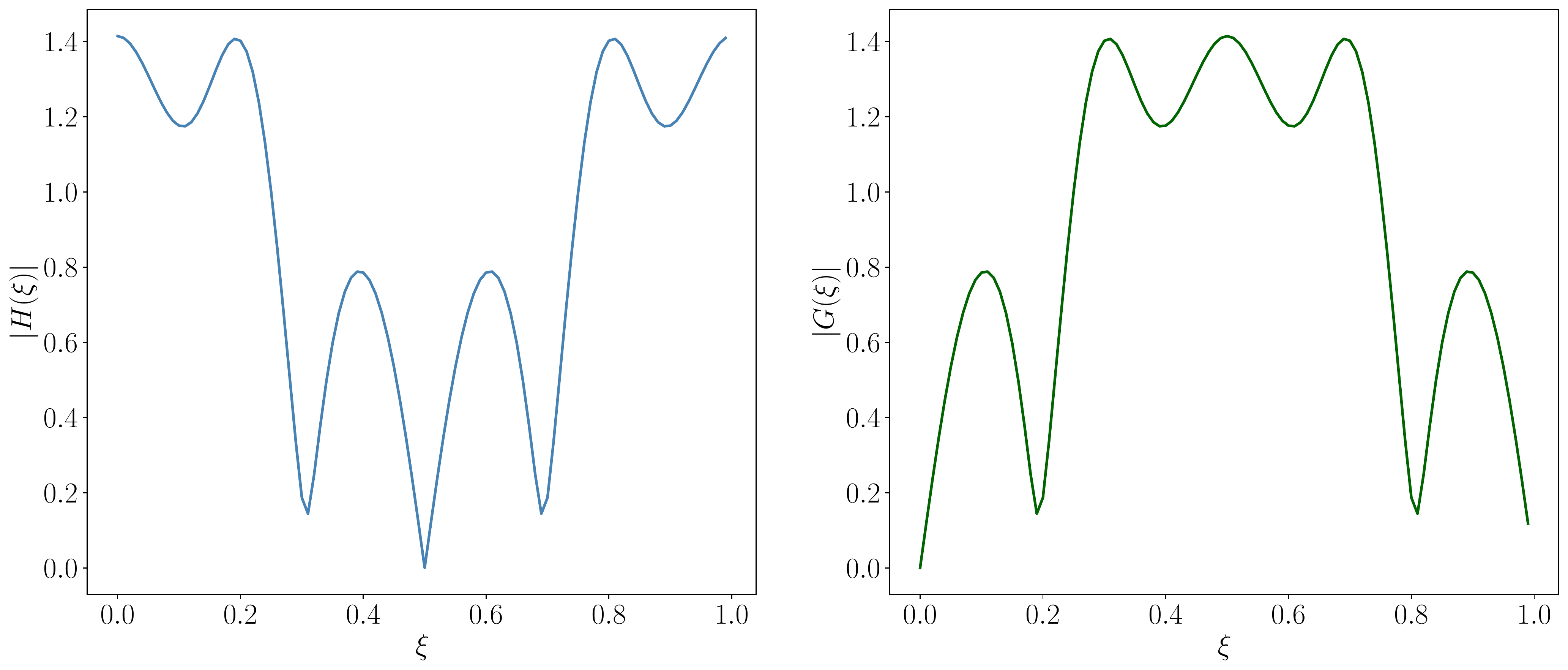} }}
	\\[2ex]    	
	\subfloat[\centering Order $7$ - initial wavelet ]{{\includegraphics[width=0.44\columnwidth]{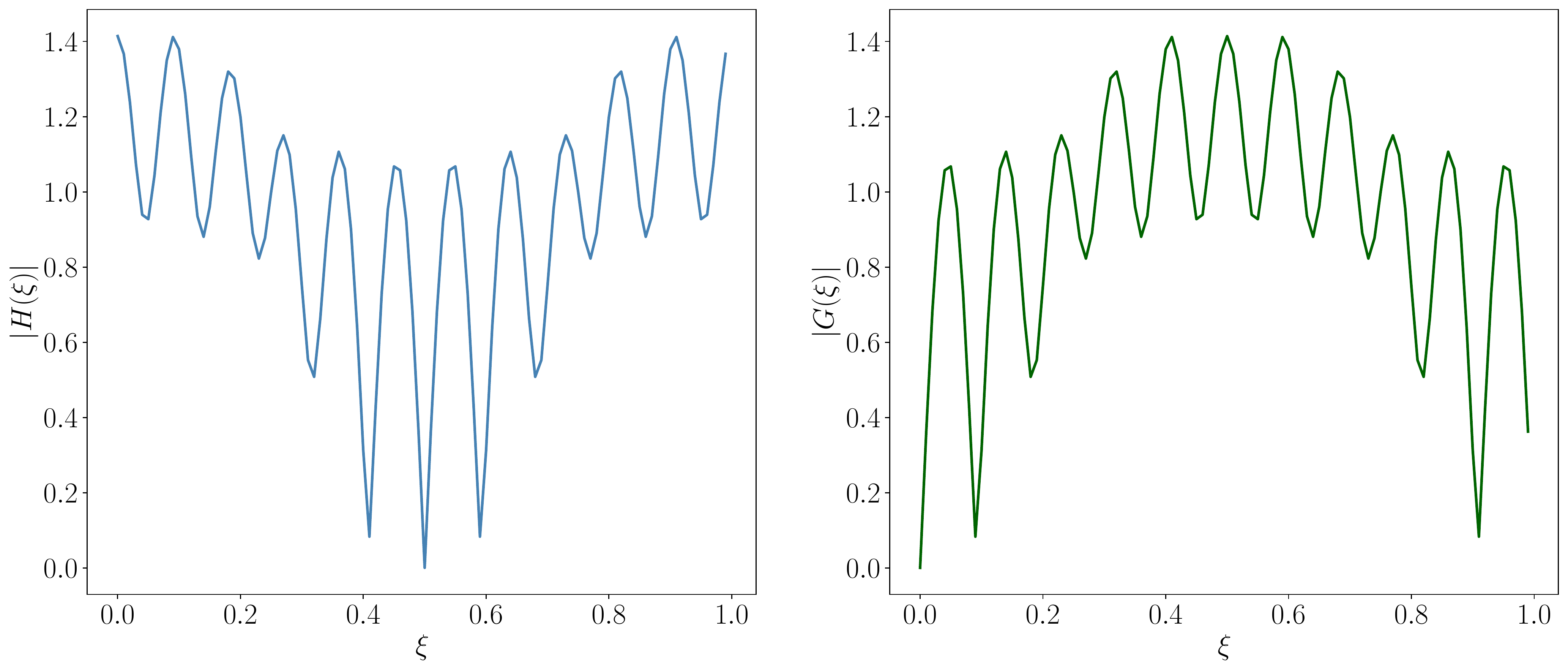} }}
	\subfloat[\centering Order $7$ - task-optimized wavelet ]{{\includegraphics[width=0.44\columnwidth]{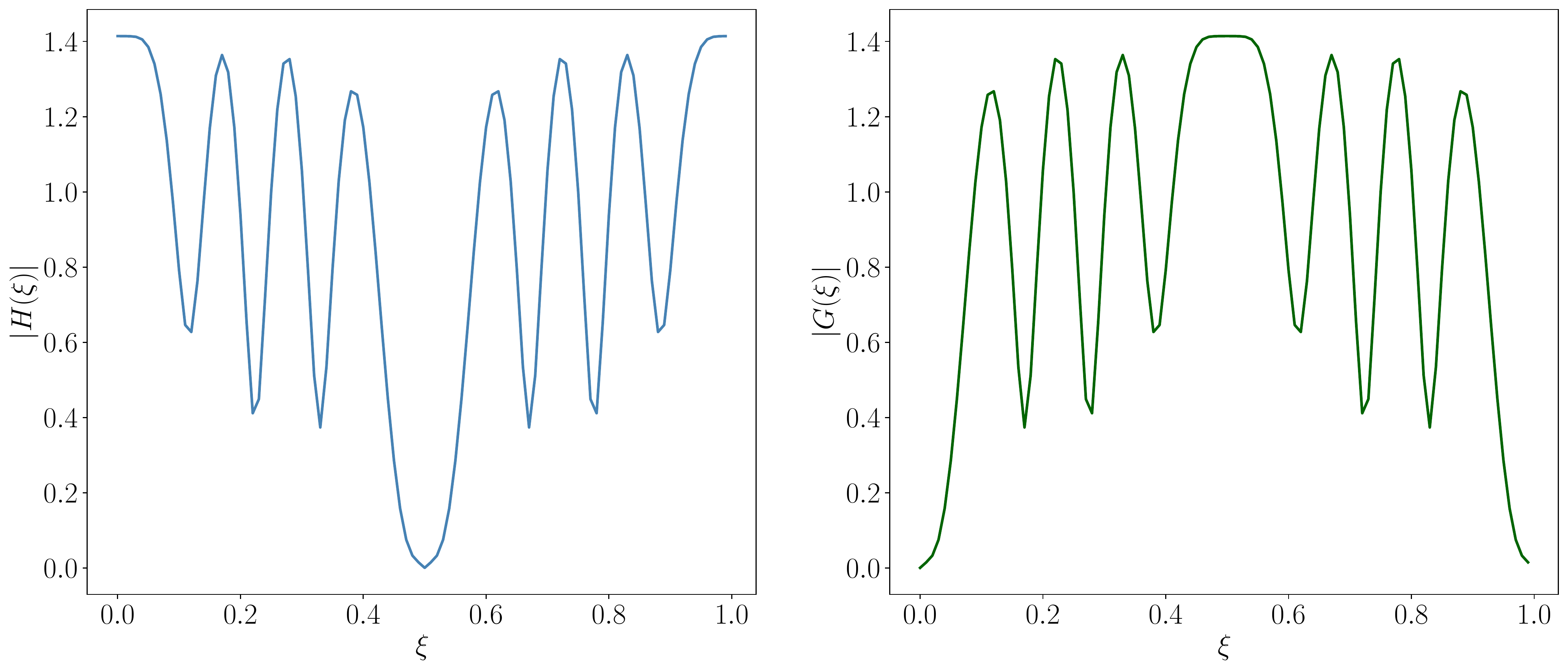} }}
	\\[2ex] 
	\subfloat[\centering Order $8$ - initial wavelet ]{{\includegraphics[width=0.44\columnwidth]{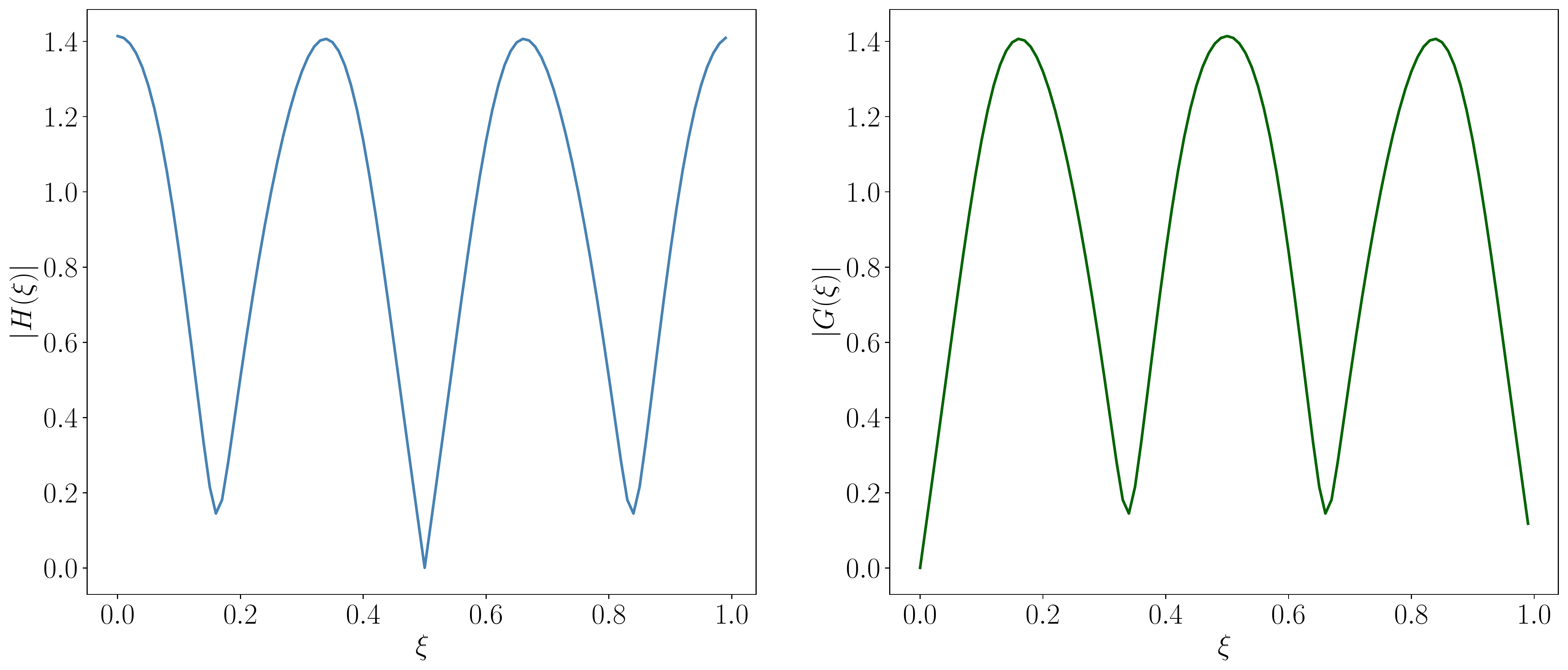} }}
	\subfloat[\centering Order $8$ - task-optimized wavelet ]{{\includegraphics[width=0.44\columnwidth]{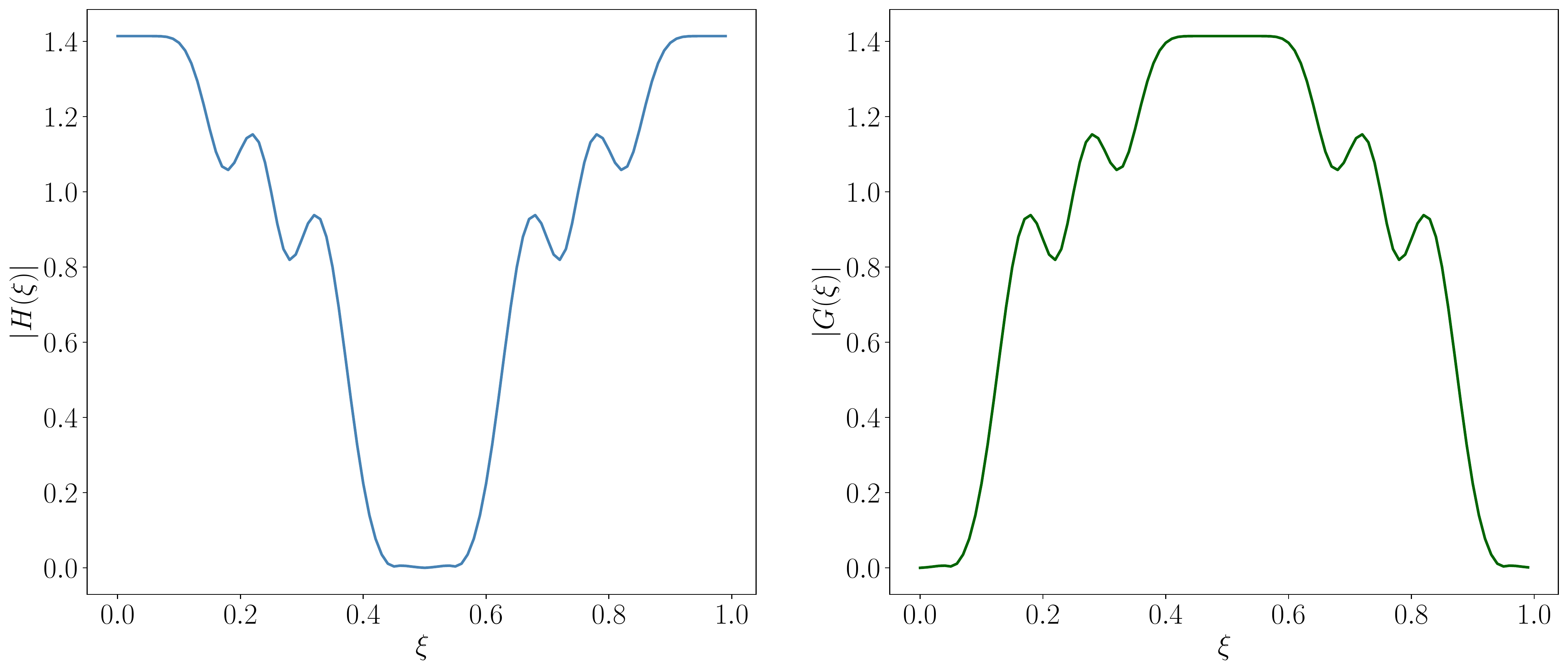} }}
	\\[2ex]    	   	
	\label{fig:prostate_refinement_comp_0}
\end{figure}
\newpage
\subsubsection{Prostate - second spatial component}
\begin{figure}[!b]
	\centering
	\subfloat[\centering Order $3$ - initial wavelet ]{{\includegraphics[width=0.44\columnwidth]{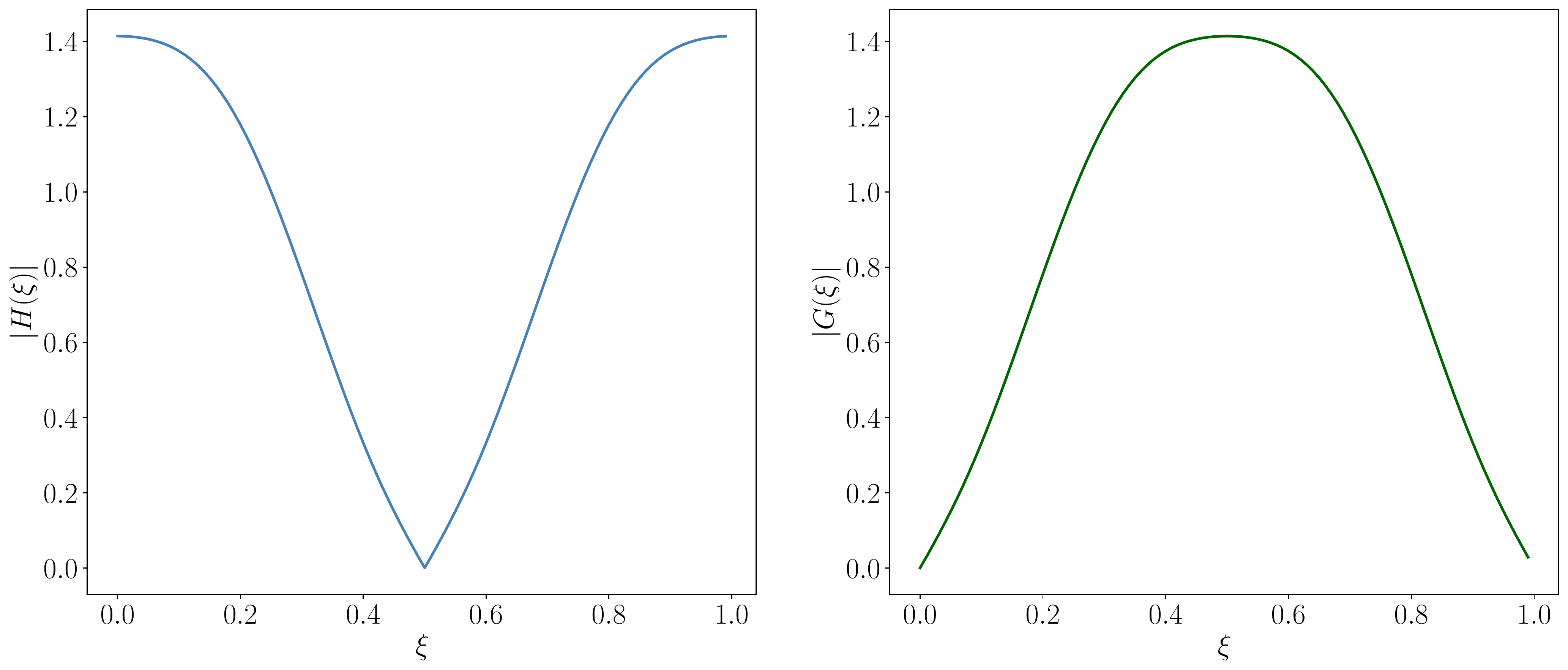} }}
	\subfloat[\centering Order $3$ - task-optimized wavelet ]{{\includegraphics[width=0.44\columnwidth]{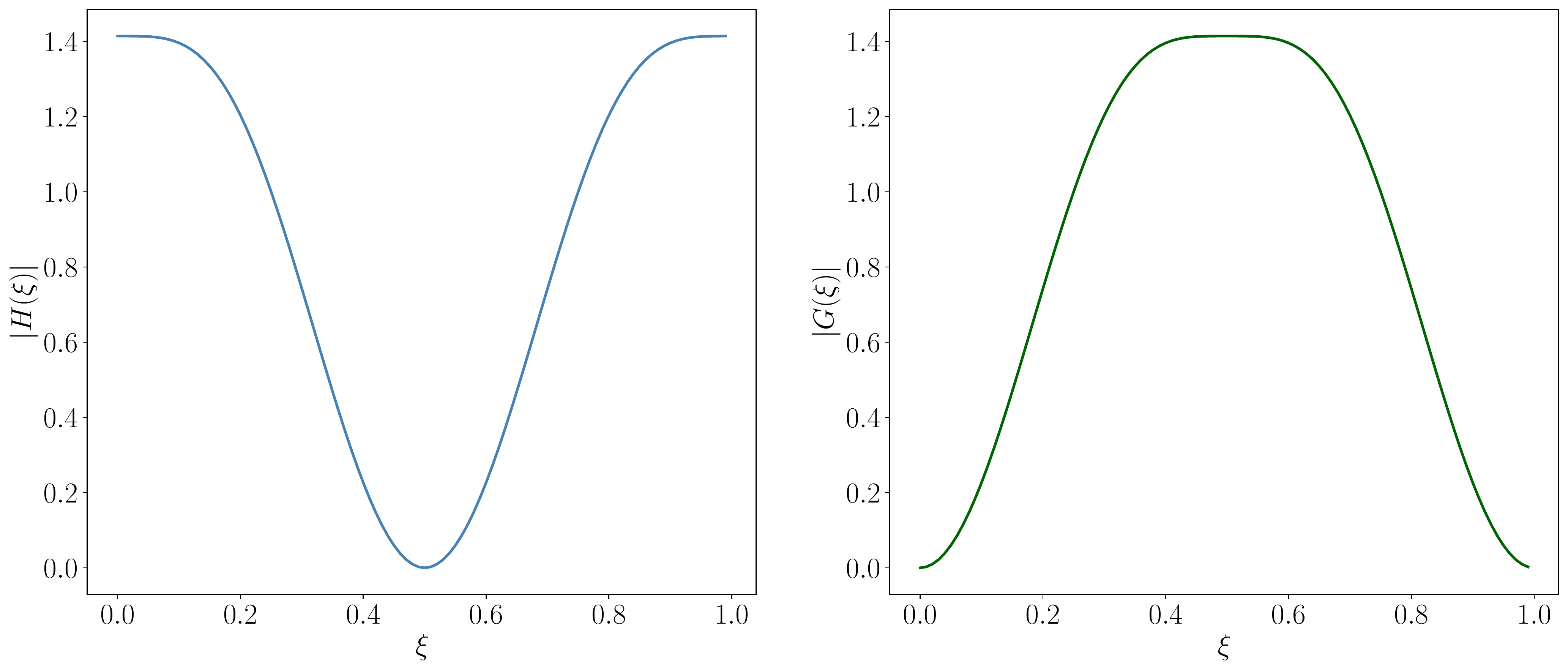} }}
	\\[2ex]    
	\subfloat[\centering Order $4$ - initial wavelet ]{{\includegraphics[width=0.44\columnwidth]{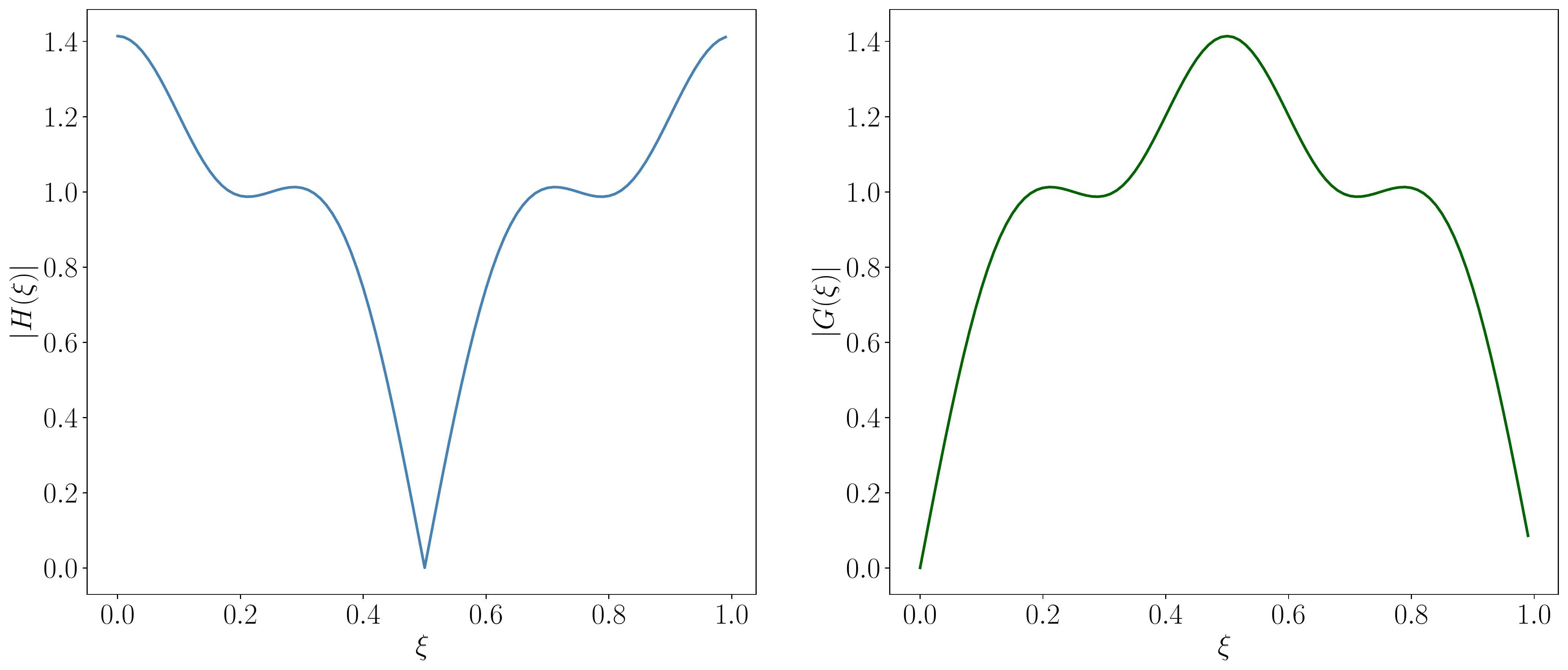} }}
	\subfloat[\centering Order $4$ - task-optimized wavelet ]{{\includegraphics[width=0.44\columnwidth]{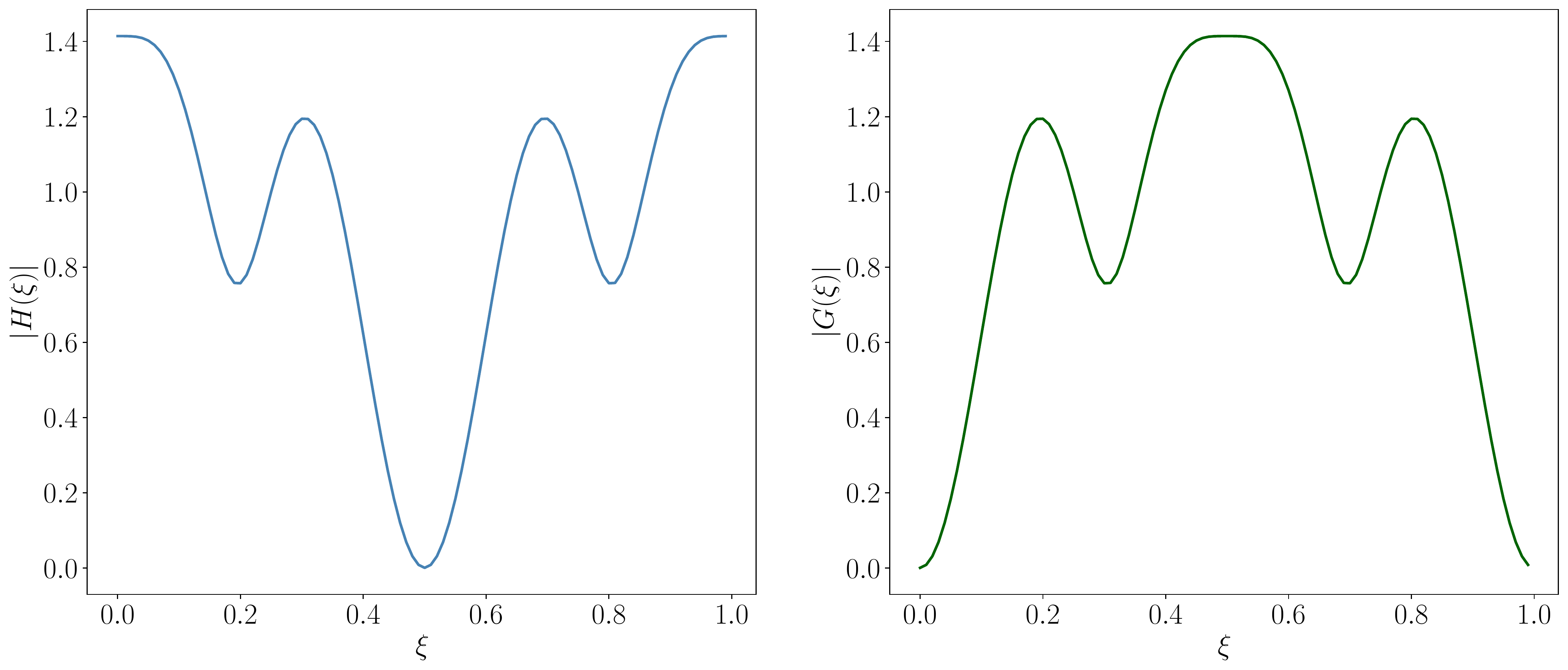} }}
	\\[2ex]    	
	\subfloat[\centering Order $5$ - initial wavelet ]{{\includegraphics[width=0.44\columnwidth]{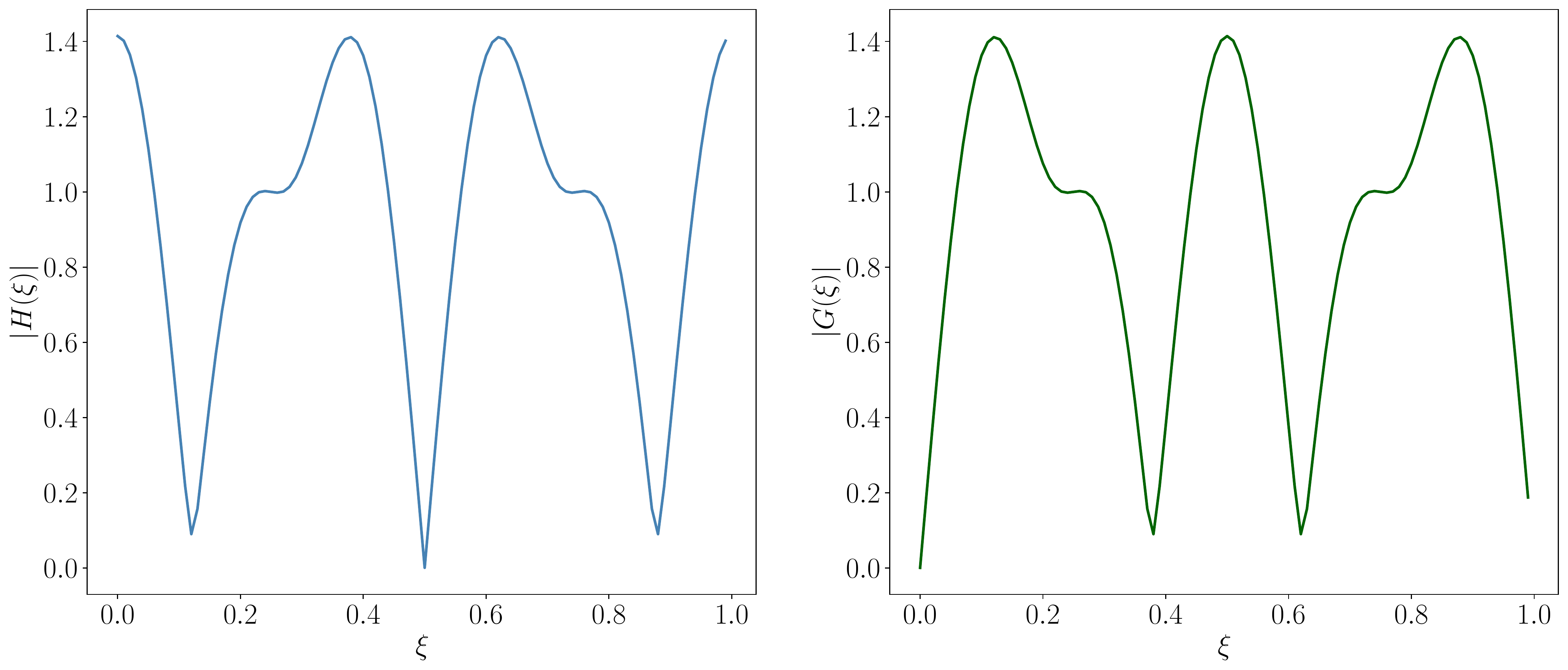} }}
	\subfloat[\centering Order $5$ - task-optimized wavelet ]{{\includegraphics[width=0.44\columnwidth]{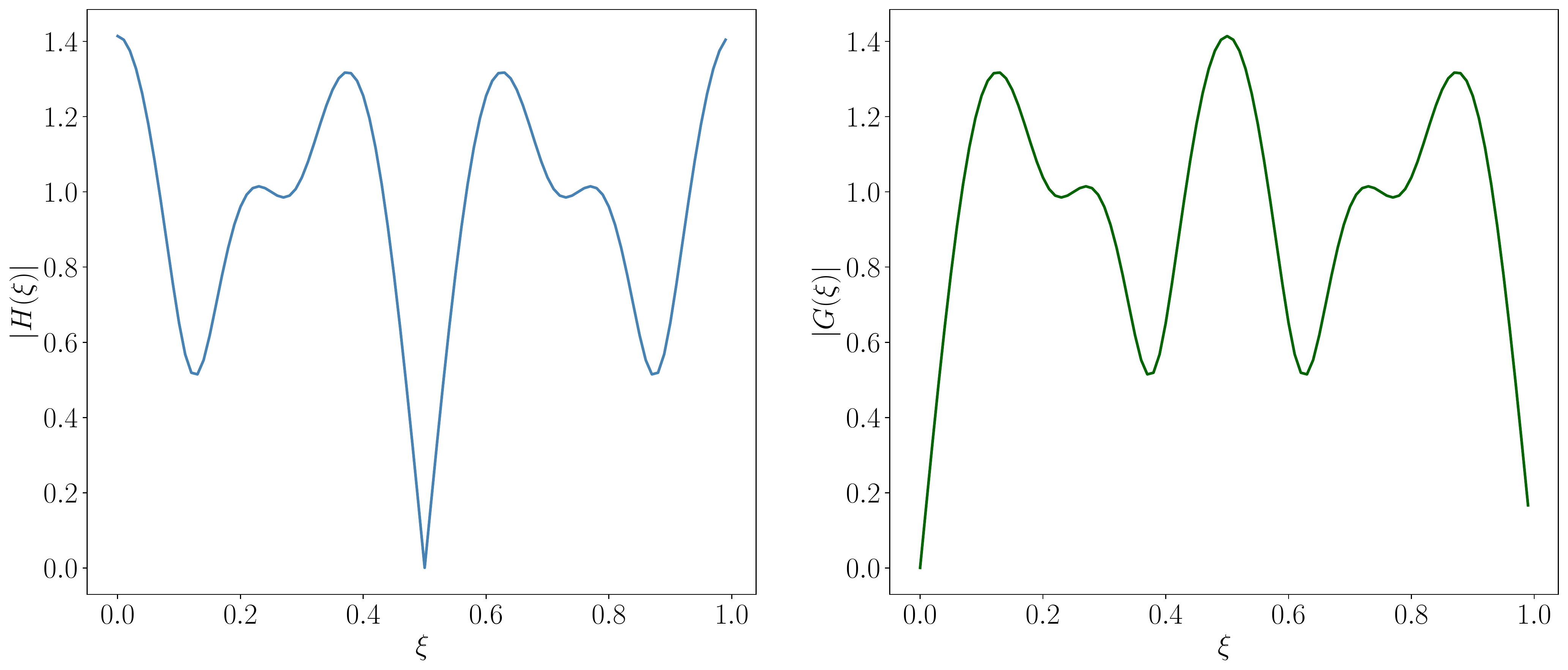} }}
	\\[2ex]    	
	\subfloat[\centering Order $6$ - initial wavelet ]{{\includegraphics[width=0.44\columnwidth]{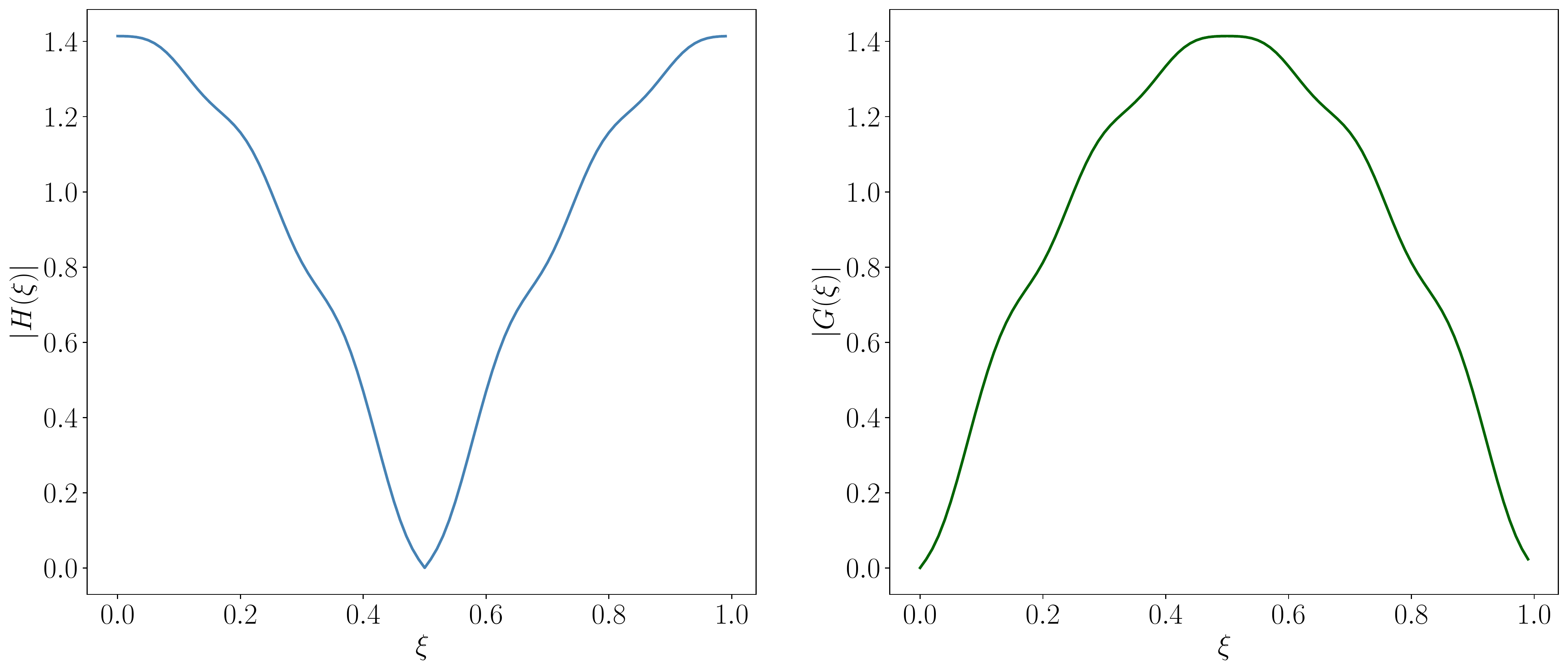} }}
	\subfloat[\centering Order $6$ - task-optimized wavelet ]{{\includegraphics[width=0.44\columnwidth]{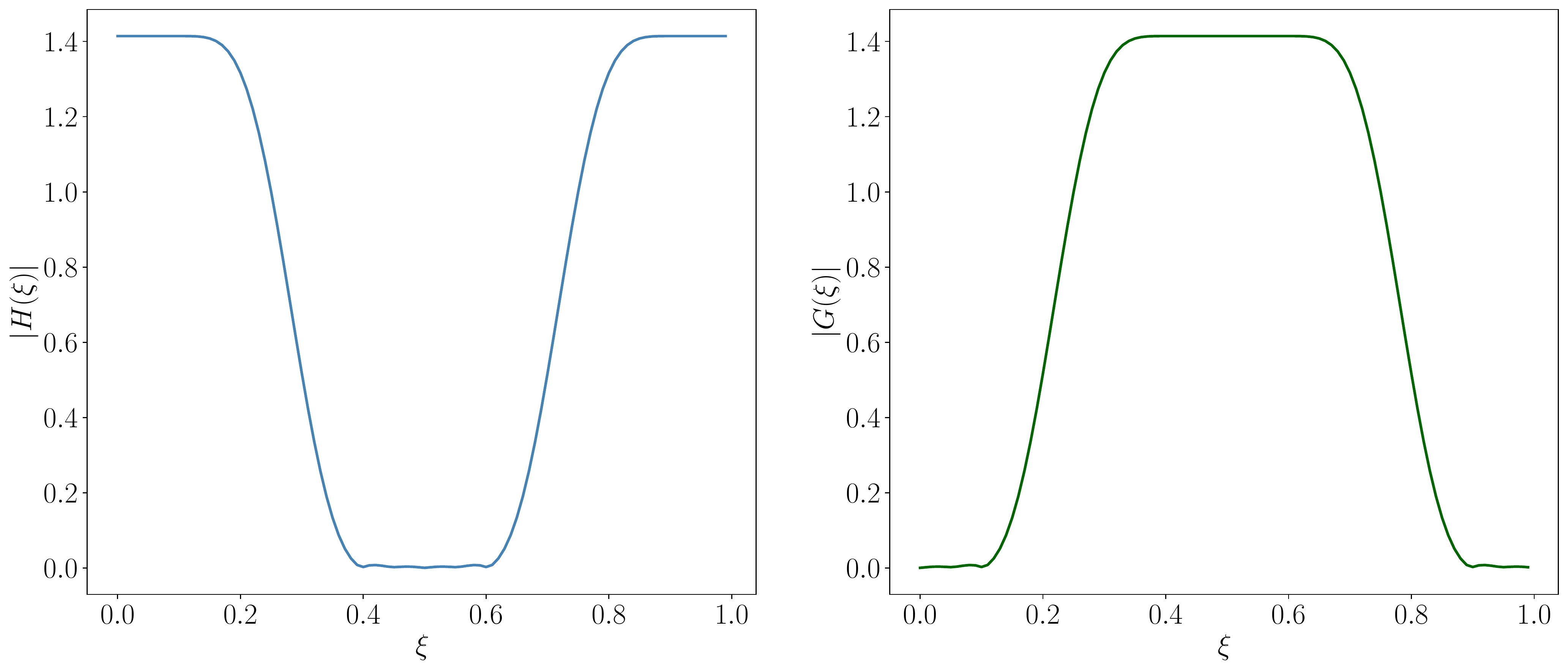} }}
	\\[2ex]    	
	\subfloat[\centering Order $7$ - initial wavelet ]{{\includegraphics[width=0.44\columnwidth]{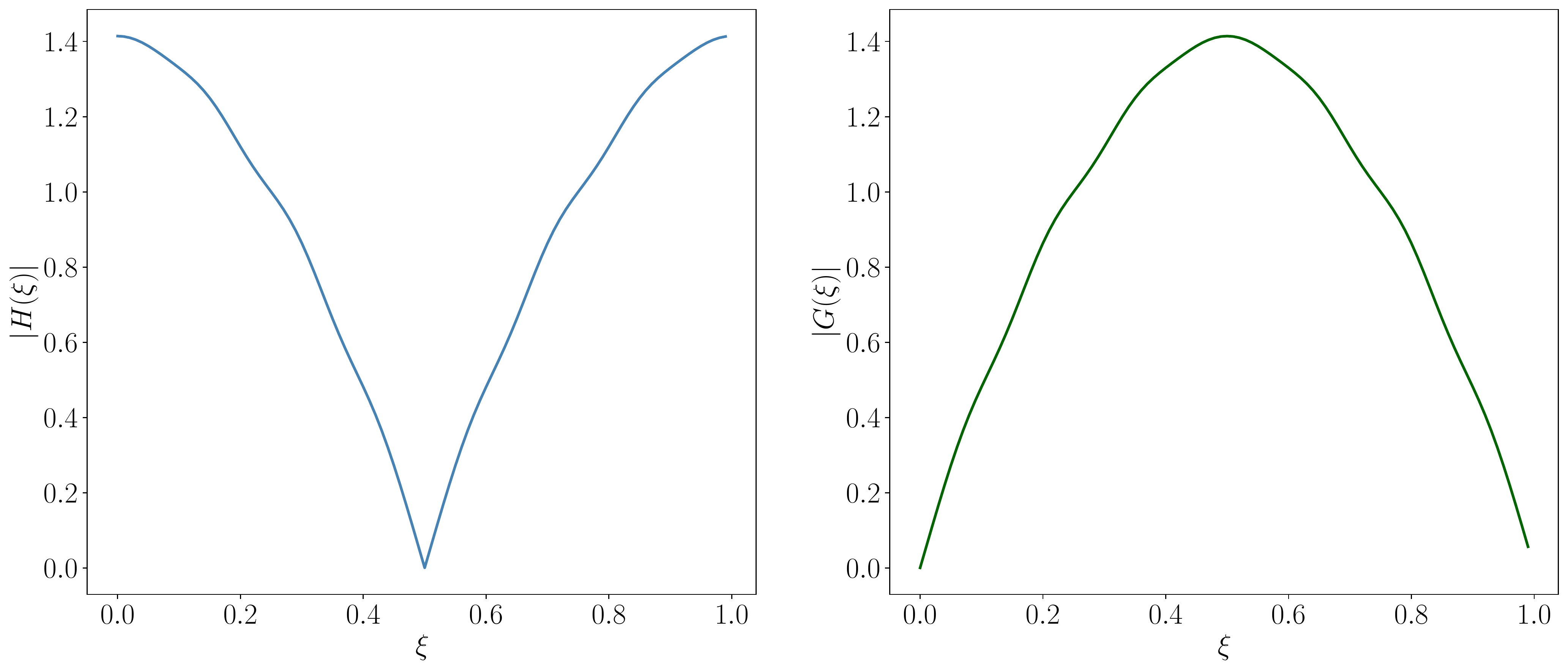} }}
	\subfloat[\centering Order $7$ - task-optimized wavelet ]{{\includegraphics[width=0.44\columnwidth]{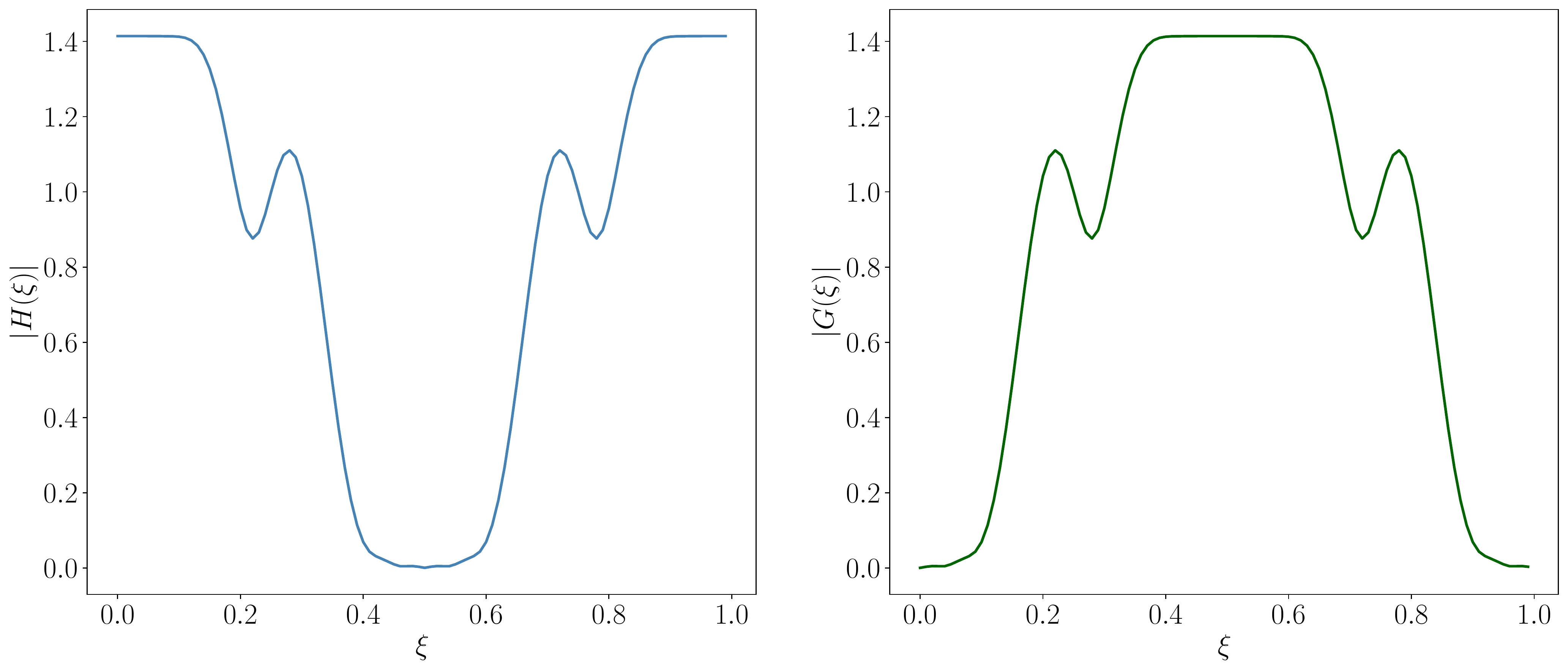} }}
	\\[2ex] 
	\subfloat[\centering Order $8$ - initial wavelet ]{{\includegraphics[width=0.44\columnwidth]{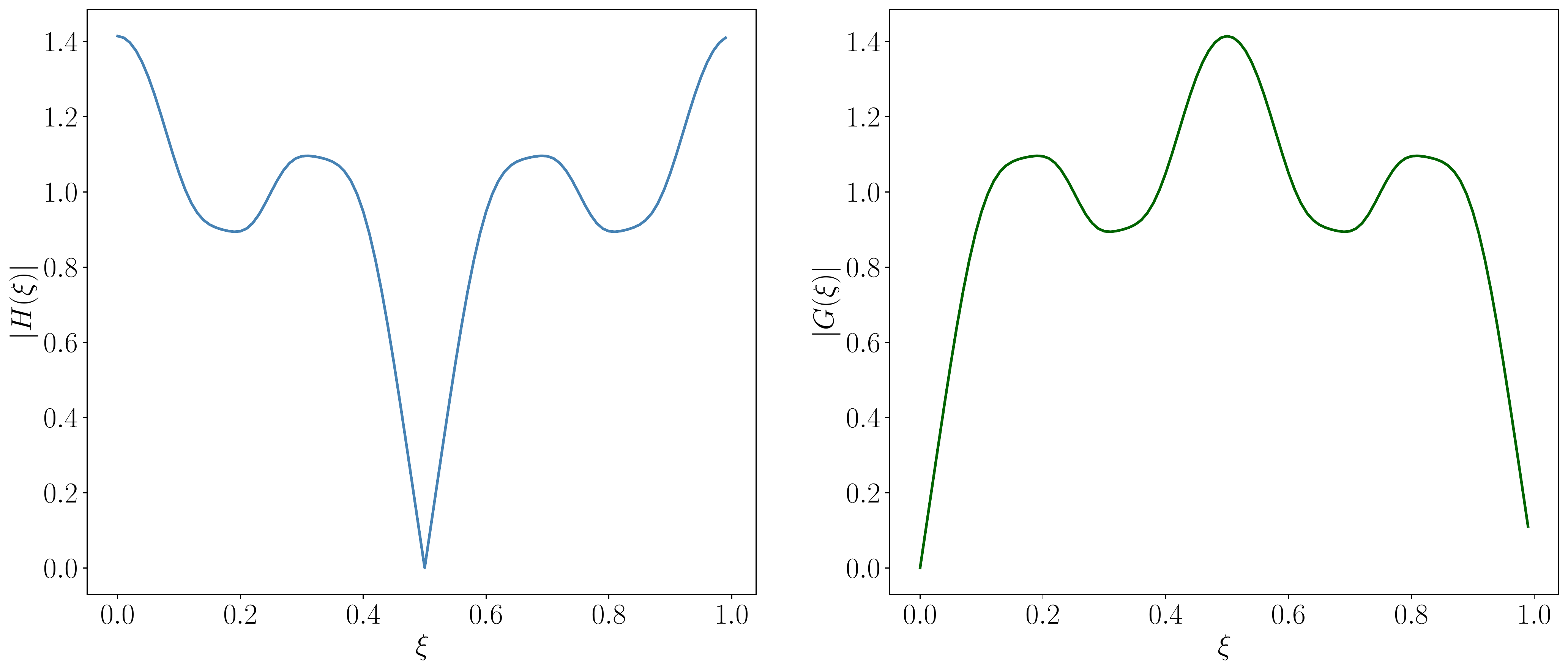} }}
	\subfloat[\centering Order $8$ - task-optimized wavelet ]{{\includegraphics[width=0.44\columnwidth]{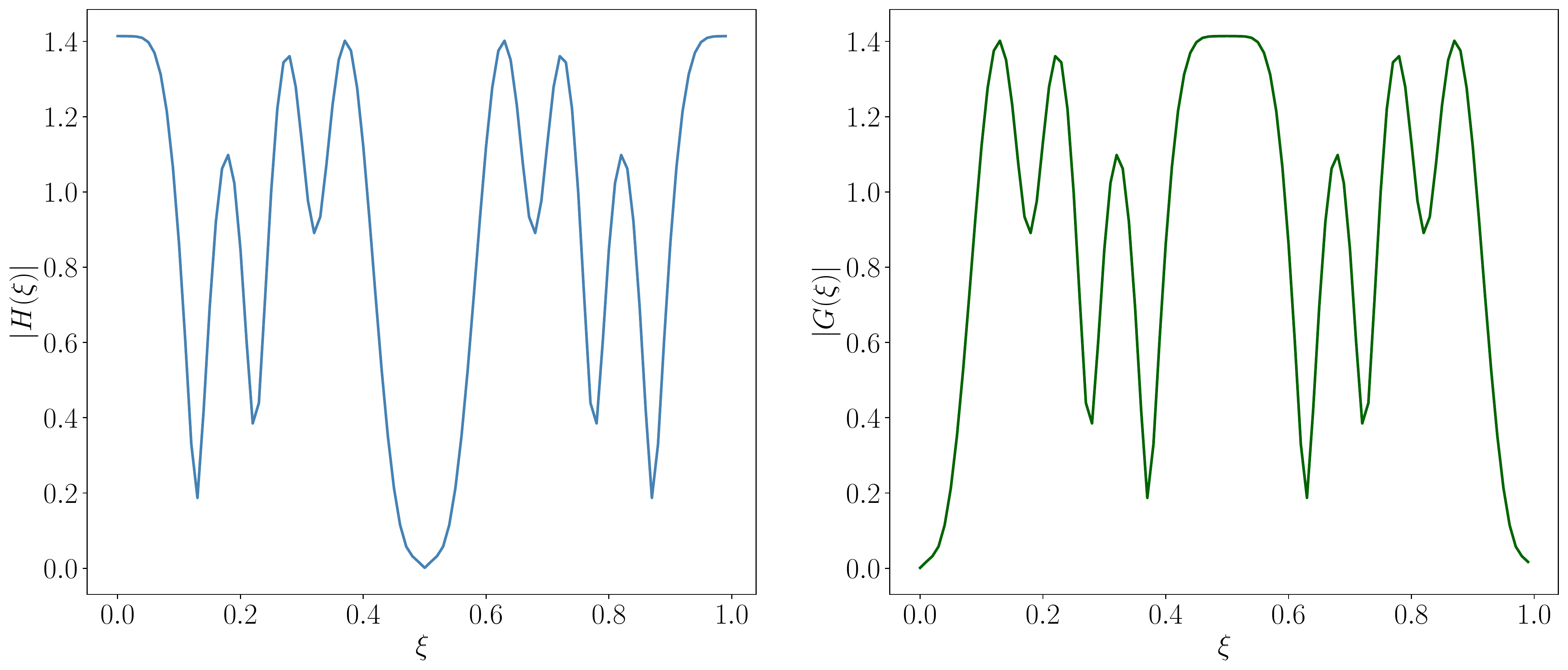} }}
	\\[2ex]    	   	
	\label{fig:prostate_refinement_comp_1}
\end{figure}

\end{document}